\documentclass{article}

    \PassOptionsToPackage{numbers, compress}{natbib}
\usepackage[final]{neurips_2023}

\usepackage[utf8]{inputenc} % allow utf-8 input
\usepackage[T1]{fontenc}    % use 8-bit T1 fonts
\usepackage{hyperref}       % hyperlinks
\usepackage{url}            % simple URL typesetting
\usepackage{booktabs}       % professional-quality tables
\usepackage{amsfonts}       % blackboard math symbols
\usepackage{nicefrac}       % compact symbols for 1/2, etc.
\usepackage{microtype}      % microtypography
\usepackage{xcolor}         % colors

\usepackage{multirow}
\usepackage{adjustbox}

\newif\ifdrafting
\draftingfalse % To hide comments
\ifdrafting
    \newcommand{\ds}[1]{{\color{red}[DS: #1]}}
    \newcommand{\cih}[1]{{\leavevmode\color[rgb]{0.2,0.6,0.2}[Charles: #1]}}
    \newcommand{\df}[1]{{\color{blue}[DF: #1]}}
    \newcommand{\srbs}[1]{{\leavevmode\color[rgb]{1.0,0.4,0}[SS: #1]}}

\else
	\newcommand{\ds} [1] {}
	\newcommand{\cih} [1] {}
    \newcommand{\df}[1]{}
    \newcommand{\srbs}[1]{}
	
\fi

\def\unet{Efficient U-Net}
\def\first{\textbf}
\def\second{\underline}
\def\third{}

\def\ours{DDVM}
\def\oursfull{DDVM (Denoising Diffusion Vision Model)}

\newcommand{\vspaceaftercaption}{\vspace{0.2cm}}

\newcommand{\myparagraph}[1]{\noindent\textbf{#1}}
\usepackage{amsthm,amsmath,amsfonts,bm,xspace}
\usepackage{bbm}
\usepackage{color}
\usepackage{enumitem}

\def\1{\bm{1}}

\def\eps{{\epsilon}}

\def\vx{{\bm{x}}}
\def\veps{{\bm{\epsilon}}}

\def\vy{{\bm{y}}}
\def\vyt{{\bm{y_t}}}

\DeclareMathAlphabet{\mathsfit}{\encodingdefault}{\sfdefault}{m}{sl}
\SetMathAlphabet{\mathsfit}{bold}{\encodingdefault}{\sfdefault}{bx}{n}
\newcommand{\E}{\mathbb{E}}

\makeatletter
\DeclareRobustCommand\onedot{\futurelet\@let@token\@onedot}
\def\@onedot{\ifx\@let@token.\else.\null\fi\xspace}

\def\eg{\emph{e.g}\onedot} 
\def\ie{\emph{i.e}\onedot} 
\def\cf{\emph{cf}\onedot} 
\def\etc{\emph{etc}\onedot} 
\def\wrt{w.r.t\onedot}

\makeatother

\def\unet{Efficient U-Net}

\def\stepunrolled{step-unrolled denoising diffusion}

\usepackage{amsmath,amssymb}
\usepackage{multirow}
\usepackage{tikz}
\usepackage{subcaption}
\usepackage[font=small]{caption}
\usepackage[super]{nth}
\usepackage{algorithm}
\usepackage{algpseudocode}
\usepackage{tabularx}
\usepackage{makecell}

\title{The Surprising Effectiveness of Diffusion Models for Optical Flow and Monocular Depth Estimation}
\author{
   Saurabh Saxena,\:  Charles Herrmann,\:  Junhwa Hur,\: Abhishek Kar, \\[.3em]
    \textbf{Mohammad Norouzi,\: Deqing Sun,\: David J.\ Fleet\thanks{DF is also affiliated with the University of Toronto and the Vector Institute.}} \\[.2cm]
  \texttt{\{srbs,irwinherrmann,junhwahur,abhiskar,deqingsun,davidfleet\}@google.com}\\[.2cm]
Google DeepMind and Google Research
}

\begin{document}
\maketitle

\begin{abstract}
\vspace*{-0.3cm}
Denoising diffusion probabilistic models have transformed image generation with their impressive fidelity and diversity.
We show that they also excel in estimating optical flow and monocular depth, surprisingly, without task-specific architectures and loss functions that are predominant for these tasks. 
Compared to the point estimates of conventional regression-based methods, diffusion models also enable Monte Carlo inference, e.g., capturing uncertainty and ambiguity in flow and depth.
With self-supervised pre-training, the combined use of synthetic and real data for supervised training, and technical innovations (infilling and step-unrolled denoising diffusion training) to handle noisy-incomplete training data, and a simple form of coarse-to-fine refinement, one can train state-of-the-art diffusion models for depth and optical flow estimation. 
Extensive experiments focus on quantitative performance against benchmarks, ablations, and the model's ability to capture uncertainty and multimodality, and impute missing values. Our model, \oursfull, obtains a state-of-the-art relative depth error of 0.074 on the indoor NYU benchmark and an Fl-all outlier rate of 3.26\% on the KITTI  optical flow benchmark, about 25\% better than the best published method. For an overview see \href{https://diffusion-vision.github.io}{diffusion-vision.github.io} 

\end{abstract}
% \vspace*{-0.1cm}
\section{Introduction}
\vspace*{-0.1cm}

Diffusion  models have emerged as powerful generative models for high fidelity image synthesis, capturing rich knowledge about the visual world~\citep{ho2020denoising,ramesh-dalle2,sahariac-imagen,sohl2015deep}. However, at first glance, it is unclear whether these models can be as effective on many classical computer vision tasks. For example, consider two dense vision estimation tasks, namely, optical flow, which estimates frame-to-frame correspondences, and monocular depth perception, which makes depth predictions based on a single image.
% ; both essential vision tasks with myriad potential downstream use cases. 
Both tasks are usually treated as regression problems and addressed with specialized architectures and task-specific loss functions, \eg, cost volumes, feature warps, or suitable losses for depth. Without these specialized components or the regression framework, general generative techniques may be ill-equipped and vulnerable to both generalization and performance issues. 

\begin{figure*}[tp]
    \centering
    \small
    \setlength\tabcolsep{0.2pt} % default value: 6pt
    \newcommand{\Figwidth}{0.19\linewidth}
    \begin{tabular}{c @{\hskip 0.2em} c @{\hskip 0.4em} c @{\hskip 0.4em} ccc}
    &  Input image(s) & Variance heat map & \multicolumn{3}{c}{Multimodal prediction samples} \\
    
    \rotatebox{90}{\hspace{2em}NYU} &
    \includegraphics[trim={0 0 0 0},clip,width=\Figwidth]{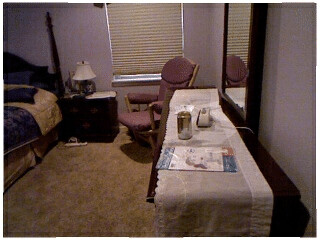} &
    \includegraphics[trim={0 0 0 0},clip,width=\Figwidth]{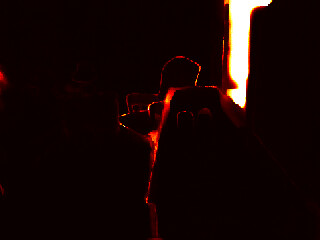} &
    \includegraphics[trim={0 0 0 0},clip,width=\Figwidth]{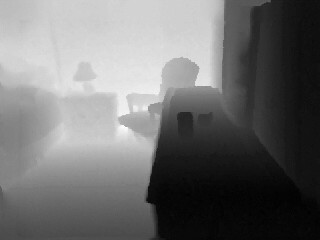} &
    \includegraphics[trim={0 0 0 0},clip,width=\Figwidth]{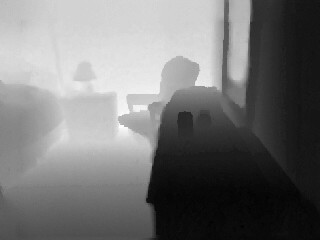} &
    \includegraphics[trim={0 0 0 0},clip,width=\Figwidth]{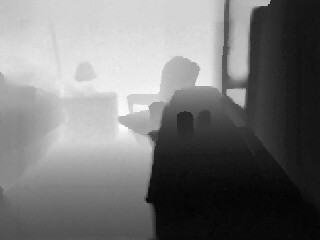}\\
    \rotatebox{90}{\hspace{1.0em}KITTI} &
    \includegraphics[trim={500 0 0 0},clip,width=\Figwidth]{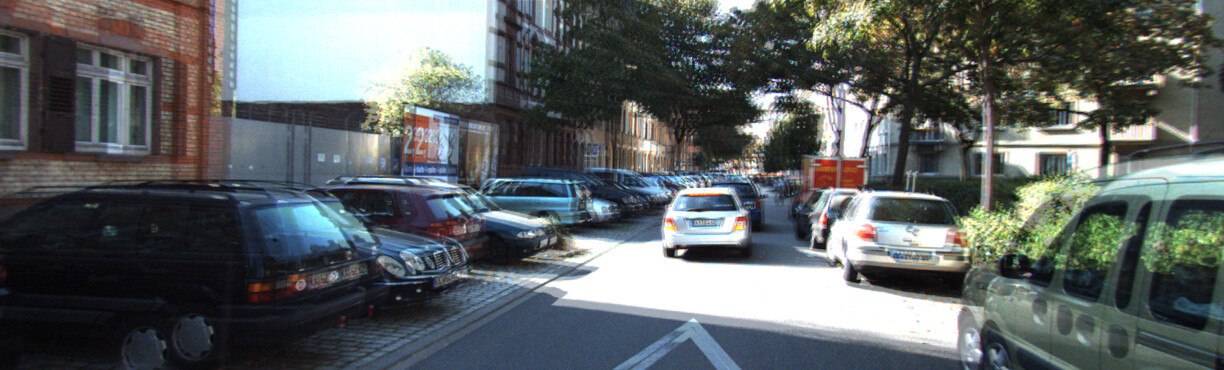} &
    \includegraphics[trim={500 0 0 0},clip,width=\Figwidth]{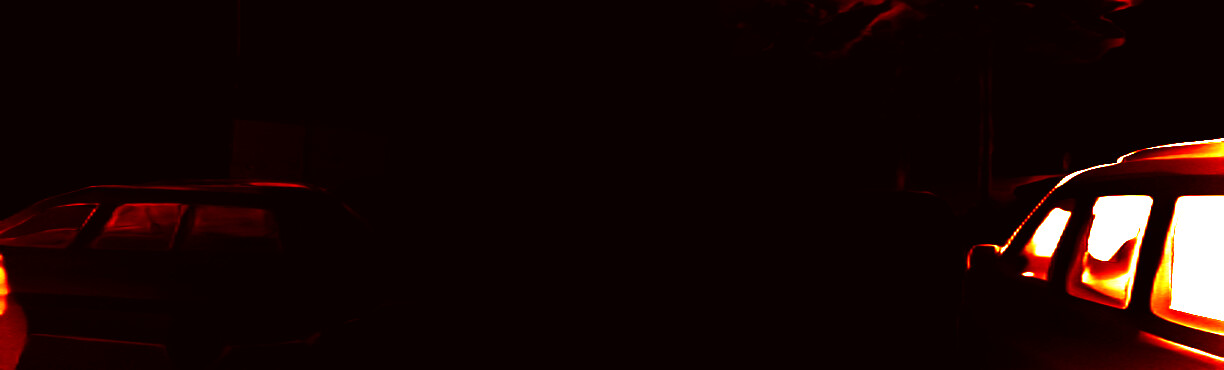} &
    \includegraphics[trim={500 0 0 0},clip,width=\Figwidth]{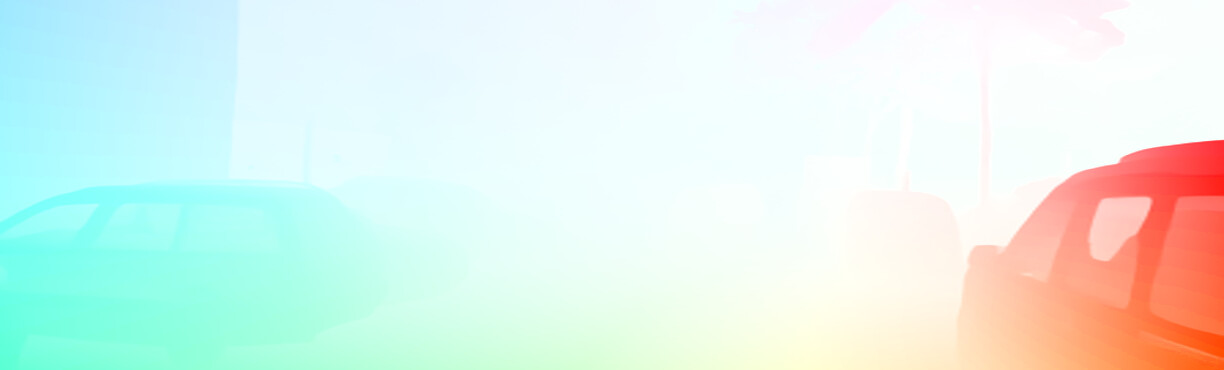} &
    \includegraphics[trim={500 0 0 0},clip,width=\Figwidth]{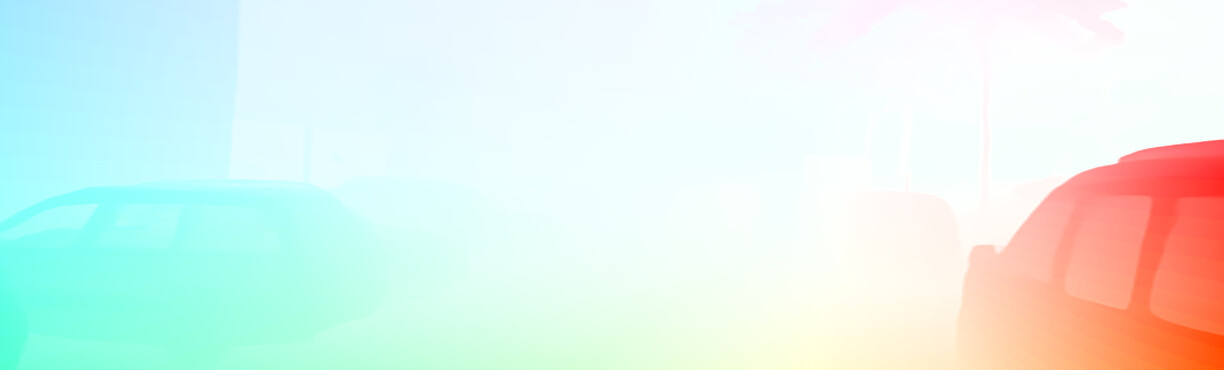} &
    \includegraphics[trim={500 0 0 0},clip,width=\Figwidth]{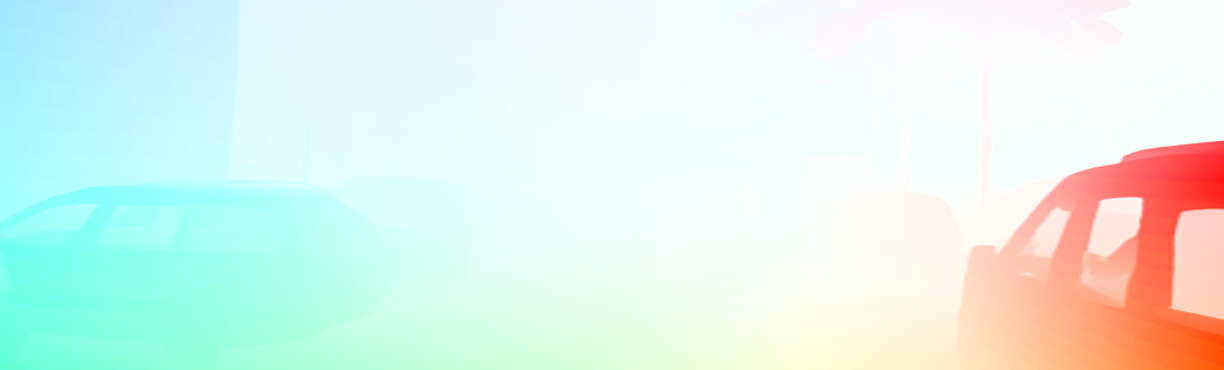} \\
    
    \rotatebox{90}{\hspace{0.5em}Sintel} &
    \includegraphics[trim={0 0 0 0},clip,width=\Figwidth]{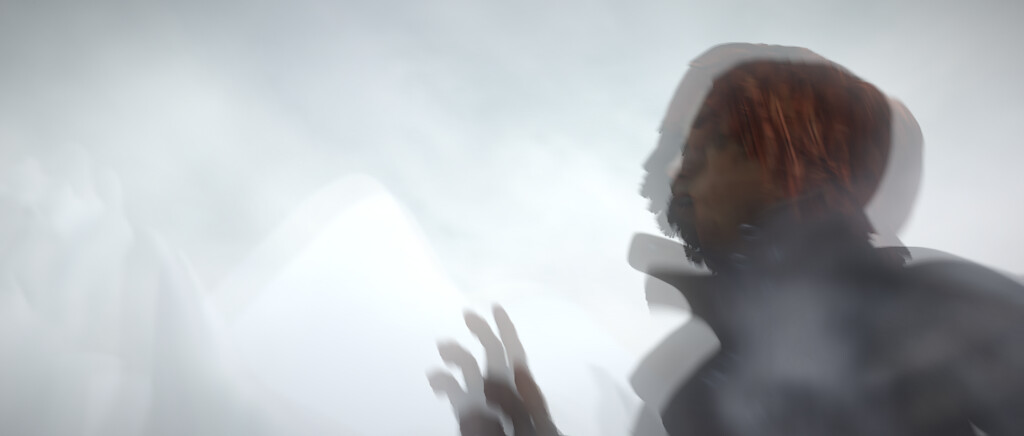} &
    \includegraphics[trim={0 0 0 0},clip,width=\Figwidth]{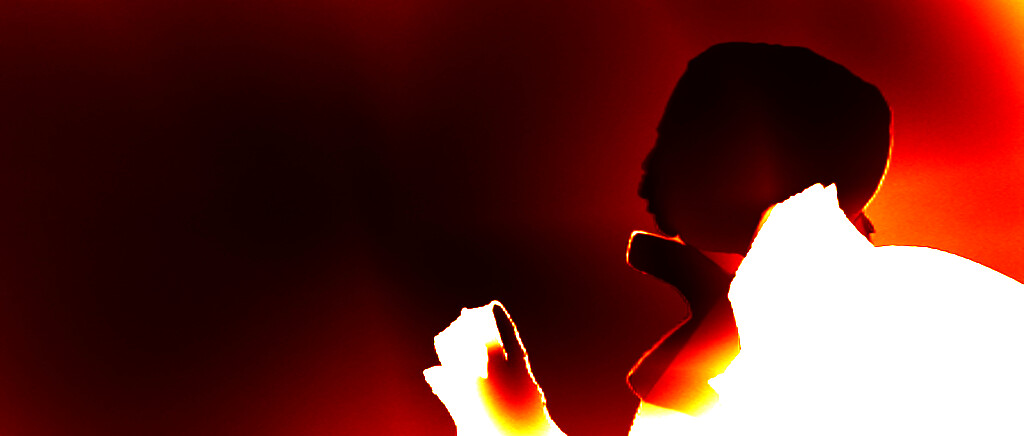} &
    \includegraphics[trim={0 0 0 0},clip,width=\Figwidth]{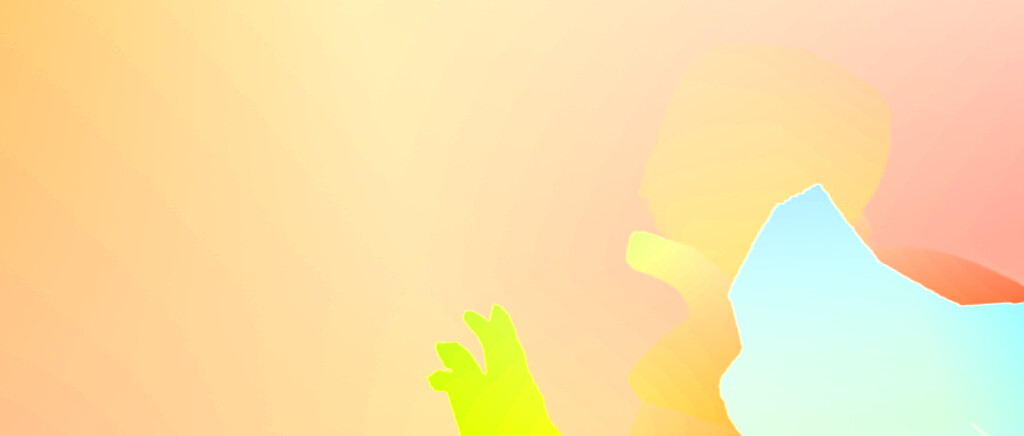} &
    \includegraphics[trim={0 0 0 0},clip,width=\Figwidth]{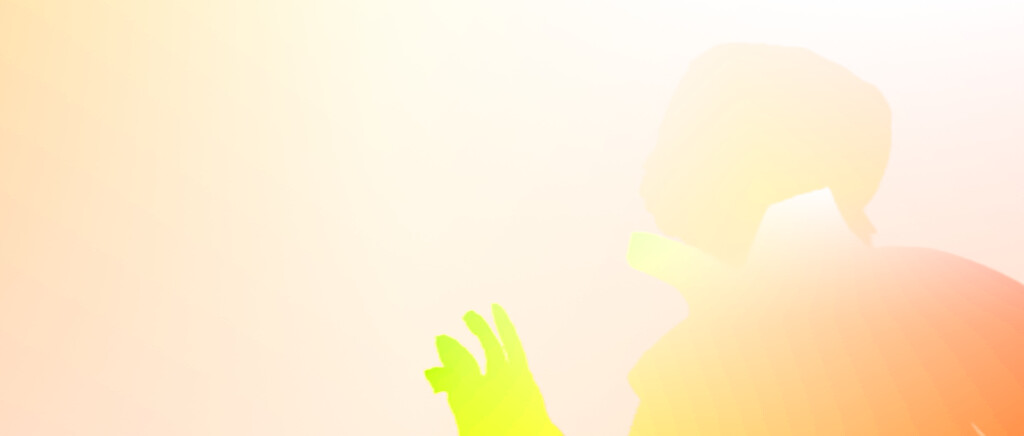} &
    \includegraphics[trim={0 0 0 0},clip,width=\Figwidth]{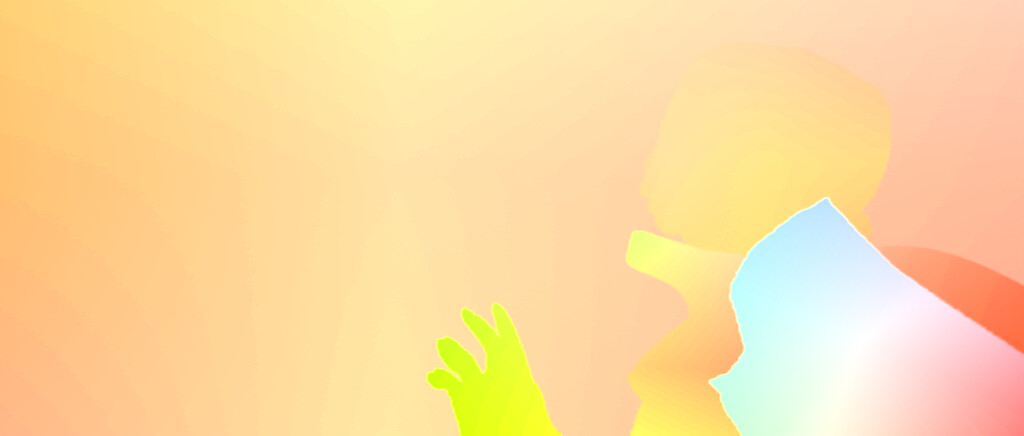} \\
    
    \end{tabular}
        \vspace*{-0.1cm}
	\caption{\textbf{Examples of multi-modal prediction} on depth (NYU) and optical flow (Sintel and KITTI). Each row shows an input image (or overlayed image pair for optical flow), a variance heat map from 8 samples, and 3 individual samples. Our model captures multi-modality in uncertain/ambiguous cases, such as reflective (\eg~mirror on NYU), transparent (\eg~vehicle window on KITTI), and translucent (\eg~fog on Sintel) regions. High variance also occurs at object boundaries, which are often challenging cases for optical flow, and also partially originate from noisy ground truth measurements for depth. See Fig.\  \ref{fig:supp:multi_modal_samples_depth_NYU}, \ref{fig:supp:multi_modal_samples_depth_KITTI}, \ref{fig:supp:multi_modal_samples_flow_KITTI} and \ref{fig:supp:multi_modal_samples_flow_sintel} for more examples.}
    \label{fig:multi_modal_samples}
    \vspace*{-0.2cm}
\end{figure*}

In this paper, we show that these concerns, while valid, can be addressed and that, surprisingly, a generic, conventional diffusion model for image to image translation works impressively well on both tasks, often outperforming the state of the art. In addition, diffusion models provide valuable benefits over networks trained with regression; in particular, diffusion allows for approximate inference with multi-modal distributions, capturing uncertainty and ambiguity (\eg see Figure~\ref{fig:multi_modal_samples}).

One key barrier to training useful diffusion models for monocular depth and optical flow inference concerns the amount and quality of available training data.
Given the limited availability of labelled training data, we propose a training pipeline comprising multi-task self-supervised pre-training followed by supervised pre-training using a combination of real and synthetic data.
Multi-task self-supervised pre-training leverages the strong performance of diffusion models on tasks like colorization and inpainting \citep[e.g.,][]{sahariac-palette}.
We also find that supervised (pre-)training with a combination of real and large-scale synthetic data improves performance significantly.

A further issue concerns the fact that many existing real datasets for depth and optical flow have noisy and incomplete ground truth annotations.
This presents a challenge for the conventional training framework and  
iterative sampling in diffusion models, leading to a problematic distribution shift between training and inference.
To mitigate these issues we propose the use of an $L_1$ loss for robustness, infilling missing depth values during training, and the introduction of {\em \stepunrolled}.
These elements of the model are shown through ablations to be important for both depth and flow estimation.

Our contributions are as follows: 
\begin{enumerate}[nolistsep,leftmargin=0.5cm]
\itemsep 0.05cm
\item We formulate optical flow and monocular depth estimation as image to image translation with generative diffusion models, without specialized loss functions and model architectures.
% We are the first to apply diffusion for optical flow and depth map generation, using a generic image-to-image translation framework, without specialized loss functions and model architectures.

\item We identify and propose solutions to several important issues w.r.t data. For both tasks, to mitigate distribution shift between training and inference with noisy, incomplete data, we propose infilling, step-unrolling, and an $L_1$ loss during training. 
% \item To enable training on noisy, incomplete data, we propose infilling, step-unrolling, and an $L_1$ loss for robustness.
% for approximate infilling via simple interpolation, step-unrolling to reduce latent distribution shift between training and inference and the use of an $L_1$ loss for robustness.
%We demonstrate the value of pre-training on large-scale real and synthetic data. Our pre-training mixture for optical flow, substantially improves the zero-shot perform of the seminal baseline RAFT model on the Sintel and KITTI benchmarks.
For flow, to improve generalization, we introduce a new dataset mixture for  pre-training, yielding a RAFT~\cite{teed2020raft} baseline that outperforms all published methods in zero-shot performance on the Sintel and KITTI training benchmarks.
%which substantially improves the zero-shot performance of the seminal baseline RAFT model \cite{teed2020raft} on the Sintel and KITTI training benchmarks.
\item Our diffusion models is competitive with or surpasses SOTA for both tasks. For monocular depth estimation we achieve a SOTA relative error of 0.074 on the NYU dataset and perform competitively on KITTI. For flow, diffusion surpasses the stronger RAFT baseline by a large margin in pre-training and our fine-tuned model achieves an Fl-all outlier rate of 3.26\% on the public KITTI test benchmark, $\sim$25\% lower 
than the best published method~\cite{sun2022disentangling}. 
\item Our diffusion model is also shown to capture flow and depth uncertainty, and the iterative denoising process enables zero-shot, coarse-to-fine refinement, and imputation.
% Model capture multimodality, and uncertainty through multiple samples from the posterior.
% \item Model also enables imputation, one application of which is NVS in text to 3D.
\end{enumerate}

% \vspace*{-0.1cm}
\section{Related work}

\vspace*{-0.1cm}

\begin{figure}[t]
    \centering
    % \begin{subfigure}
    %   \centering
      \includegraphics[width=1.0\linewidth]{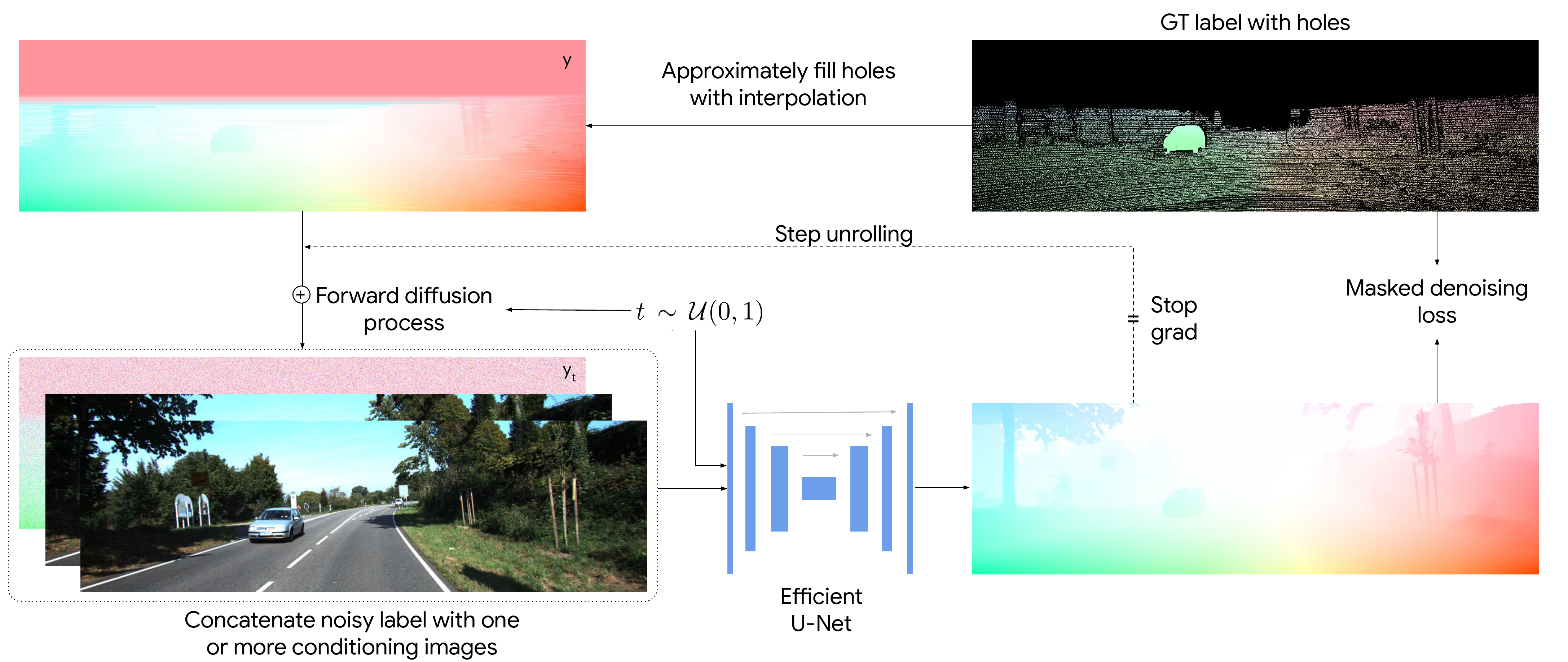}
    % \end{subfigure}
    \vspace*{-0.1cm}
    \caption{
    \label{fig:arch} \textbf{Training architecture}. Given ground truth flow/depth, we first infill missing values using interpolation. Then,  we add noise to the label map and train a neural network to model the conditional distribution of the noise given the RGB image(s), noisy label, and time step. One can optionally unroll the denoising step(s) during training (with stop gradient) to bridge the distribution gap between training and inference for $y_t$.
    }
    \vspace{-1em}
\end{figure}

Optical flow and depth estimation have been extensively studied. Here we briefly review only the most relevant work, and refer the interested readers to the references cited therein. 

% \subsection{Optical flow estimation}
% \vspace*{-0.1cm}

\myparagraph{Optical flow.}
The predominant approach to optical flow is regression-based, with a focus on specialized network architectures to exploit domain knowledge, 
\eg, cost volume construction~\citep{dosovitskiy2015flownet,hosni2012fast,huang2022flowformer,luo2022learning,sun2018pwc,xu2022gmflow,yang2019volumetric,zhang2021separable}, coarse-to-fine estimation~\citep{sun2018pwc,weinzaepfel2013deepflow,xu2011motion}, occlusion handling~\citep{hur2017mirrorflow,ince2008occlusion,sun2012layered}, or iterative refinement~\citep{hur2019iterative,ilg2017flownet,teed2020raft}, as evidenced by public benchmark datasets~\cite{Butler:ECCV:2012,Menze2018JPRS}. Some recent work has also advocated for generic architectures:
Perceiver IO~\citep{jaegleperceiver} introduces a generic transformer-based model that works for any modality, including optical flow and language modeling. 
% FlowFormer~\citep{huang2022flowformer} tokenizes the 4D cost volume, encodes the cost tokens into a cost memory with alternate group transformer, and decodes the
% cost memory via a recurrent transformer decoder.  It sets a new state of the art on Sintel in both the zero-shot and fine-tuning setup. 
Regression-based methods, however, only give a single prediction of the optical flow and do not readily capture uncertainty or ambiguity in the flow.
Our work introduces a surprisingly simple, generic architecture for optical flow using a  denoising diffusion model.

We find that this generic generative model is surprisingly effective for optical flow, recovering fine details on motion boundaries, while capturing multi-modality of the motion distribution.

% \subsection{Monocular depth estimation}
% \vspace*{-0.1cm}

\myparagraph{Monocular depth.}
% Monocular depth estimation is essential for many vision applications~\citep{jing2022depth,zhou2019does}.
Monocular depth estimation has been a long-standing problem in computer vision \cite{NIPS2005_17d8da81,4531745} with recent progress focusing on specialized loss functions and architectures~\cite{pixelformer,cao2016estimating,dorn,laina2016deeper} such as the use of multi-scale networks \cite{eigen2014predicting,eigen2014depth}, adaptive binning \cite{adabins,binsformer} and weighted scale-shift invariant losses \cite{eigen2014depth}.
Large-scale in-domain pre-training has also been effective for depth estimation \citep{ranftl2019robust,dpt,ren2017crossdomain}, which we find to be the case here as well.
We build on this rich literature, but with a simple, generic architecture, leveraging recent advances in generative models.

% \subsection{Diffusion models}
% \vspace*{-0.1cm}

\myparagraph{Diffusion models.}
Diffusion models are latent-variable generative models trained to transform a sample of a Gaussian noise  into a sample from a data distribution \citep{ho2020denoising,sohl2015deep}.
They comprise a {\it forward process} that gradually annihilates data by adding noise, as `time' $t$ increases from 0 to 1, and a learned {\it generative process} that reverses the forward process, starting from a sample of random noise at $t=1$ and incrementally adding structure (attenuating noise) as $t$ decreases to 0. 
A conditional diffusion model conditions the steps of the reverse process (e.g., on labels, text, or an image).
% In the case of depth estimation, our conditioning signal is an RGB image, $\vx$, and the target is a conditional distribution over depth maps, $p(\vy \,|\, \vx)$.

Central to the model is a denoising network $f_{\theta}$ that is trained to take a noisy sample $y_t$
at some time-step $t$, along with a conditioning signal $x$, and predict a less noisy sample.
Using Gaussian noise in the forward process, one can express the training objective over the sequence of transitions (as $t$ slowly decreases) as a sum of non-linear regression objectives, with the L2 loss (here with the $\epsilon$-parameterization):
\begin{equation}
\E_{(\vx,\, \vy)}\, \E_{(t,\, \veps)} \bigg\lVert f_\theta(\vx,\, \underbrace{\sqrt{\gamma_t} \,\vy + \sqrt{1\!-\!\gamma_t}\, \veps}_{\vyt}, \,t) - \veps\, \bigg\rVert^{2}_2
\label{eq:diffusion-loss}
\end{equation}
where $\veps \sim \mathcal{N}(0, I)$, $t \sim \mathcal{U}(0, 1)$, and where $\gamma_t > 0$ is computed with a pre-determined noise schedule.
For inference (i.e., sampling), one draws a random noise sample $\vy_1$, and then iteratively uses $f_{\theta}$
to estimate the noise, from which one can compute the next latent sample $\vy_s$, for $s<t$.

\myparagraph{Self-supervised pre-training.} Prior work has shown that self-supervised tasks such as colorization~\cite{Shakhnarovich2016,ZhangColorization2016} and masked prediction~\cite{xie2022revealing} serve as effective pre-training for downstream vision tasks.
Our work also confirms the benefit of self-supervised pre-training \cite{sahariac-palette} for diffusion-based image-to-image translation, by establishing a new SOTA on optical flow and monocular depth estimation while also representing multi-modality and supporting zero-shot coarse-to-fine refinement and imputation.

% \vspace*{-0.1cm}
\section{Model Framework}
% \vspace*{-0.2cm}

In contrast to the conventional monocular depth and optical flow methods, with rich usage of specialized domain knowledge on their architecture designs, we introduce simple, generic architectures and loss functions.
We replace the inductive bias in state-of-the-art architectures and losses with a powerful generative model along with a combination of self-supervised pre-training and supervised training on both real and synthetic data.

The denoising diffusion model (Figure \ref{fig:arch}) takes a noisy version of the target map (\ie,~a depth or flow) as input, along with the conditioning signal (one RGB image for  depth  and two RGB images for flow).  
The denoiser effectively provides a noise-free estimate of the target map (\ie,~ignoring the specific loss parameterization used).
The training loss penalizes residual error in the denoised map, which is quite distinct from typical image reconstruction losses used in optical flow estimation.

\vspace*{-0.1cm}
\subsection{Synthetic pre-training data and generalization}
\label{sec:pretrain-datasets}
\vspace*{-0.15cm}

Given that we train these models with a generic denoising objective, without task-specific inductive biases in the form of specialized architectures, the choice of training data becomes critical. Below we discuss the datasets used and their contributions in detail.
Because training data with annotated ground truth is limited for many dense vision tasks,
here we make extensive use of synthetic data in the hope that the geometric properties 
acquired from synthetic data during training will transfer to different domains, including 
natural images.

\begin{figure*}[t]
    \centering
    \small
    \setlength\tabcolsep{1pt} % default value: 6pt
    
    \newcommand{\FigwidthK}{0.195\linewidth}
    \begin{tabular}{ccccc}
      First frame & AF+FT & AF+FT+KU & AF+FT+KU+TA & GroundTruth \\
    \includegraphics[trim={600 0 0 0},clip,width=\FigwidthK]{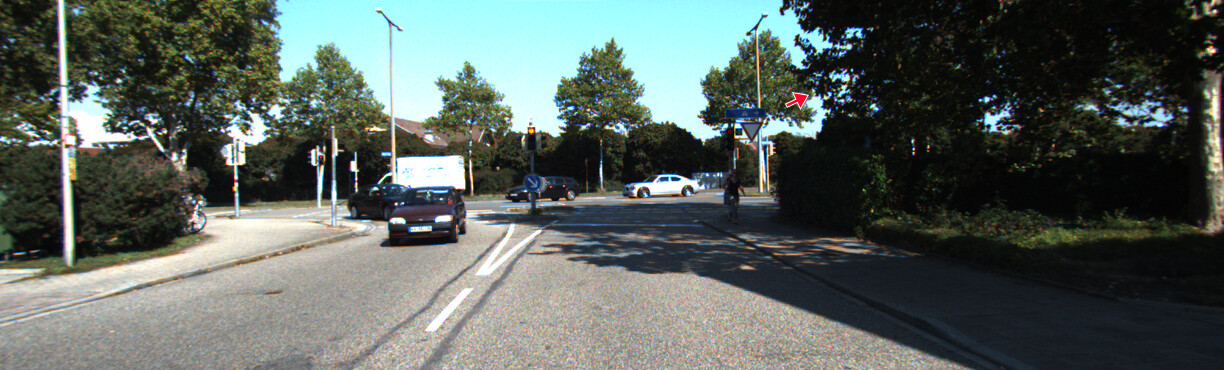} &
    \includegraphics[trim={600 0 0 0},clip,width=\FigwidthK]{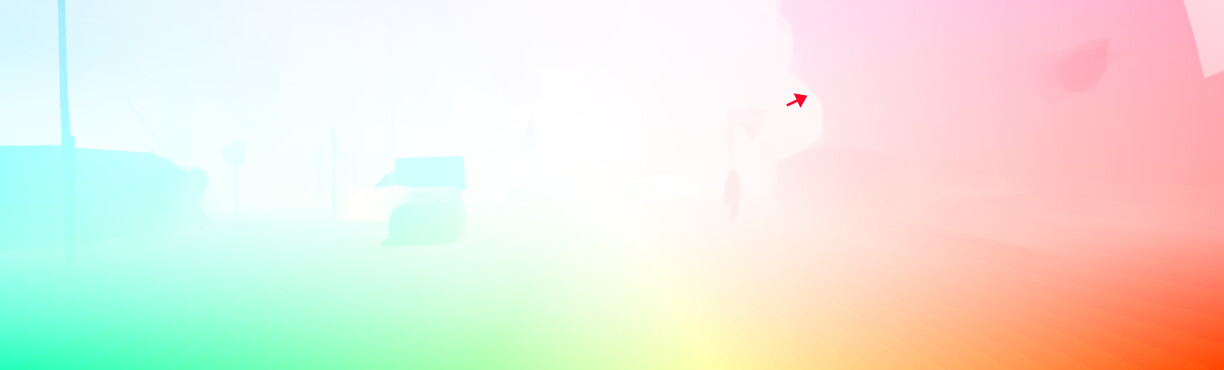} &
    \includegraphics[trim={600 0 0 0},clip,width=\FigwidthK]{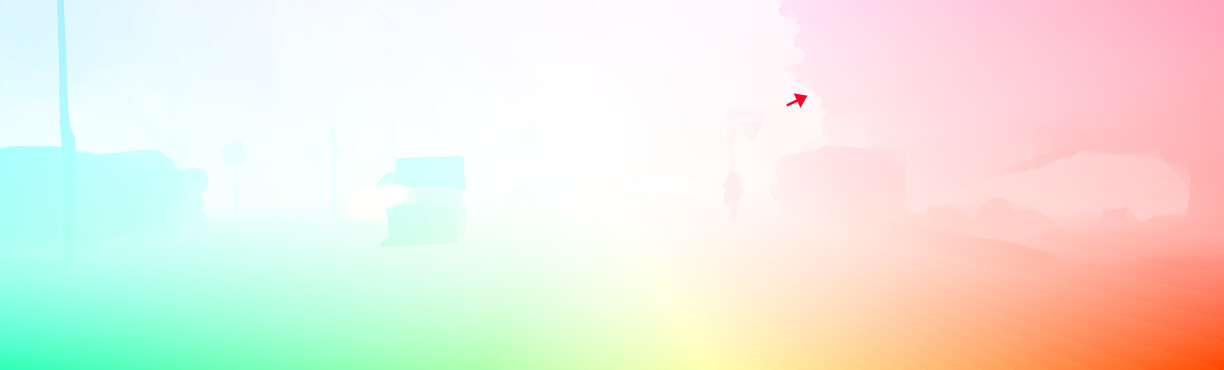} &
    \includegraphics[trim={600 0 0 0},clip,width=\FigwidthK]{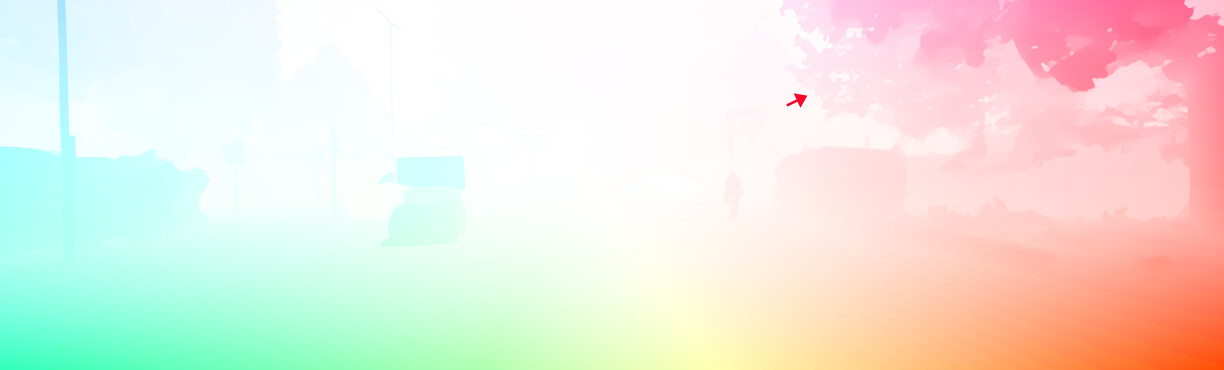} &
    \includegraphics[trim={600 0 0 0},clip,width=\FigwidthK]{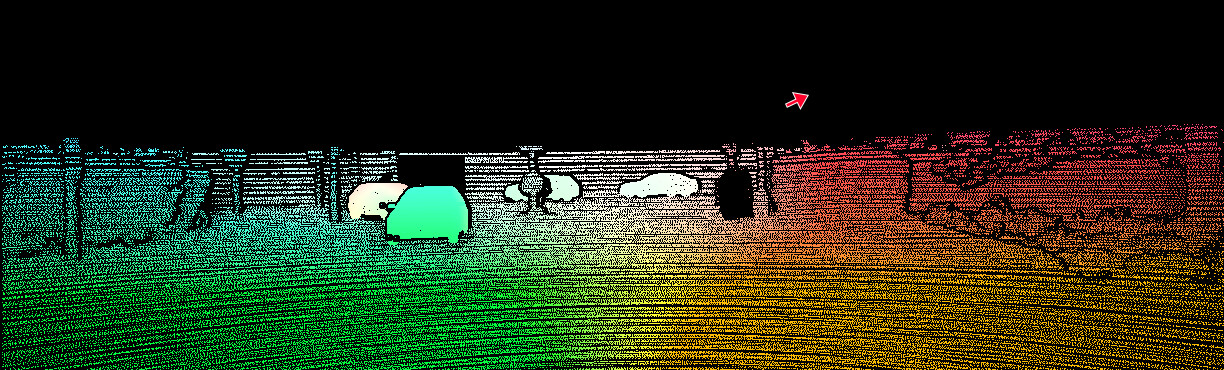}  \\
    \end{tabular}
    
    \newcommand{\FigwidthS}{0.162\linewidth}
    \begin{tabular}{cccccc}
      First frame & AF & AF+FT & AF+FT+KU & AF+FT+KU+TA & GroundTruth \\

    \includegraphics[trim={0 0 0 0},clip,width=\FigwidthS]{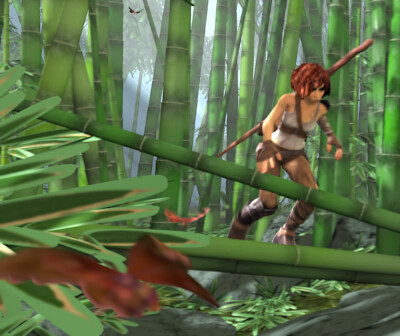} &
    \includegraphics[trim={0 0 0 0},clip,width=\FigwidthS]{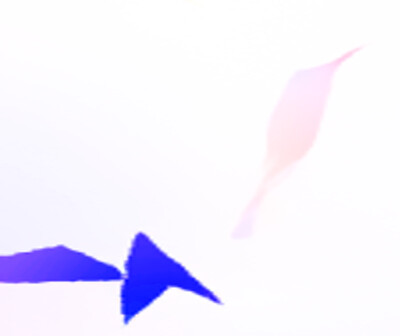} &
    \includegraphics[trim={0 0 0 0},clip,width=\FigwidthS]{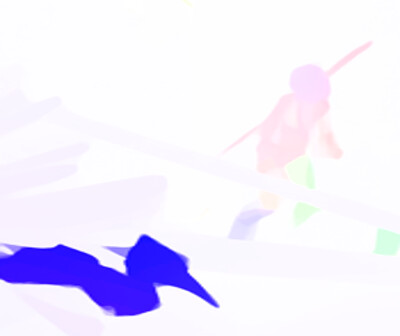} &
    \includegraphics[trim={0 0 0 0},clip,width=\FigwidthS]{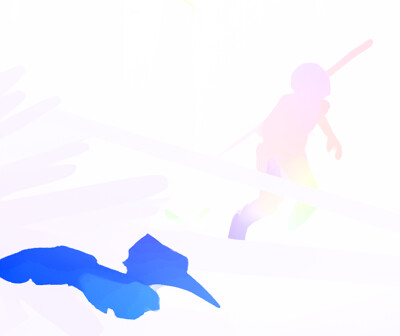} &
    \includegraphics[trim={0 0 0 0},clip,width=\FigwidthS]{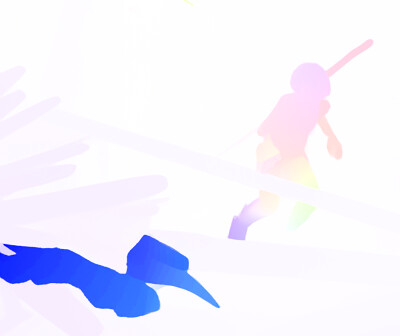} &
    \includegraphics[trim={0 0 0 0},clip,width=\FigwidthS]{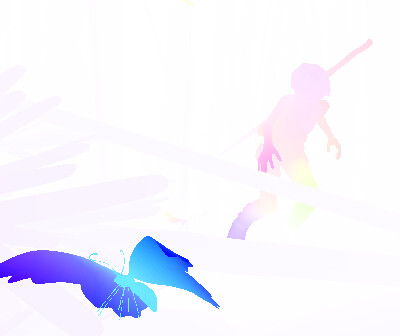}  \\
    \end{tabular}
	\caption{\textbf{Effects of adding synthetic datasets in pretraining}. Diffusion models trained only with AutoFlow (AF) tend to provide very coarse flow estimates and can hallucinate shapes. The addition of FlyingThings (FT), Kubric (KU), and TartanAir (TA) remove the AF-induced bias toward polgonal-shaped regions, and significantly improve flow quality on fine detail, e.g. trees, thin structures, and motion boundaries.}
    \label{fig:flow_dataset_zeroshot}
    \vspace*{-0.1cm}
\end{figure*}

% \paragraph{Optical Flow} 
AutoFlow~\citep{sun2021autoflow} has recently emerged as a powerful synthetic dataset for training flow models. 
We were surprised to find that training on AutoFlow alone is insufficient, as the diffusion model appears to devote a significant fraction of its representation capacity to represent the shapes of AutoFlow regions, rather than solving for correspondence.  
As a result, models trained on AutoFlow alone exhibit a strong bias to generate flow fields with polygonal shaped regions, much like those in AutoFlow, often ignoring the shapes of boundaries in the two-frame RGB inputs (\eg~see Figure~\ref{fig:flow_dataset_zeroshot}).

To mitigate bias induced by AutoFlow in training, we further mix in three synthetic datasets during training, namely,
% In addition to AutoFlow~\citep{sun2021autoflow},  we use the following datasets  for pre-training:
FlyingThings3D~\citep{Mayer_2016_CVPR}, Kubric~\citep{Greff_2022_CVPR} and TartanAir~\citep{tartanair2020iros}.
Given a model pre-trained on AutoFlow, for compute efficiency, we use a greedy mixing strategy where we fix the relative ratio of the previous mixture and tune the proportion of the newly added dataset. 
We leave further exploration of an optimal mixing strategy to future work.
Zero-shot testing of the model on Sintel and KITTI (see  Table~\ref{tab:flow-pretrain-results} and Fig.\ \ref{fig:flow_dataset_zeroshot}) shows substantial performance gains with each additional synthetic dataset.

We find that pre-training is similarly important for 
depth estimation (see Table \ref{tab:nyu-dataset-ablation}). We learn separate indoor and outdoor models.
For the indoor model we pre-train on a mix of ScanNet \citep{scannet} and SceneNet RGB-D \citep{scenenet-rgbd}. 
The outdoor model is pre-trained on the Waymo Open Dataset \citep{waymoOpenDataset}.

\vspace*{-0.1cm}
\subsection{Real data: Challenges with noisy, incomplete ground truth}
\label{sec:handling-missing-depth}
\vspace*{-0.2cm}

Ground truth annotations for real-world depth or flow data are often sparse and noisy, due to highly reflective surfaces, light absorbing surfaces \cite{6664993},  dynamic objects~\cite{Menze2015ISA}, \etc. While regression-based methods can simply compute the loss on pixels with valid ground truth, corruption of the training data is more challenging for diffusion models.
Diffusion models perform inference through iterative refinement of 
the target map $\vy$ conditioned on RGB image data $\vx$. 
It starts with a sample of Gaussian noise $\vy_1$, and terminates with a sample from the predictive distribution $p(\vy_0 \,|\,\vx)$.
A refinement step from time $t$ to $s$, with $s\! <\!t$, proceeds by sampling from the parameterized distribution $p_\theta(\vy_s \,|\, \vy_t, \vx)$; i.e., each step operates on the output from the previous step.
During training, however, the denoising steps are decoupled (see Eqn.\ \ref{eq:diffusion-loss}), where the denoising network operates on a noisy version of the ground truth depth map instead of the output of the previous iteration (reminiscent of teaching forcing in RNN training~\cite{6795228}). 
Thus there is a distribution shift between marginals over
the noisy target maps during training and inference, because the
ground truth maps have missing annotations and heavy-tailed sensor noise while the noisy maps obtained from the previous time step at inference time should not. This distribution shift has a very negative impact on model performance.  Nevertheless, with the following modifications 
during training we find that the problems can be mitigated effectively.

\myparagraph{Infilling.}
One way to reduce the distribution shift is to impute the missing ground truth. 
We explored several ways to do this, including simple interpolation schemes, and inference using our model (trained with nearest neighbor interpolation).
% Empirically, we found that two straightforward steps performed as well as much more sophisticated approaches. \ds{What do ``two straightforward steps'' refer to?}
We find that nearest neighbor interpolation is sufficient to impute missing values in the ground truth maps in the depth and flow field training data.

Despite the imputation of missing ground truth depth and flow values, note that the training loss is only computed and backpropagated from pixels with known (not infilled) ground truth depth.  We refer to this as the masked denoising loss (see Figure \ref{fig:arch}).

% \begin{figure}[t]
\begin{algorithm}[t]
    \footnotesize
   \caption{Denoising diffusion train step with infilling and step unrolling}
   \label{alg:train_step}
\begin{algorithmic}[1]
   \State $x$ $\gets$ conditioning images, $y$ $\gets$ flow or depth map, $mask$ $\gets$ binary mask of known values
   \State $t$ $\sim U(0,1), \eps \sim N(0,1)$
   \State $y$ = fill\_holes\_with\_interpolation($y$)
   \State $y_t = \sqrt{\gamma_t} * y + \sqrt{1-\gamma_t} * \eps$
   \If{unroll\_step}
   \State $\eps_{pred}$ = stop\_gradient($f_{\theta}(x, y_t, t)$)
   \State $y_{pred} = (y_t - \sqrt{1-\gamma_t} * \eps_{pred}) / \sqrt{\gamma_t} $
   \State $y_t = \sqrt{\gamma_t} * y_{pred} + \sqrt{1-\gamma_t} * \eps$
   \State $\eps = (y_t - \sqrt{\gamma_t} * y) / \sqrt{1-\gamma_t}$
   \EndIf
   \State $\eps_{pred} = f_{\theta}(x, y_t, t)$
   \State loss = reduce\_mean($\lvert \eps - \eps_{pred} \lvert$[$mask$])
\end{algorithmic}
% \vspace*{-3em}
\end{algorithm}
% \vspace*{-10em}
% \end{figure}

\myparagraph{Step-unrolled denoising diffusion training.}
\label{step_unrolled}
A second way to mitigate distribution shift in the $y_t$ marginals in training and inference, is to construct $y_t$ from model outputs rather than ground truth maps. One can do this by slightly modifying the training procedure (see Algorithm \ref{alg:train_step})
to run one forward pass of the model and build $y_t$ by adding noise to the model's output rather than the training map. We do not propagate gradients for this forward pass. This process, called {\em \stepunrolled}, slows training only marginally ($\sim$15\% on a TPU v4).
This step-unrolling is akin to the predictor-corrector sampler of \cite{song2021scorebased} which uses an extra Langevin step to improve the target marginal distribution of $y_t$.
Interestingly, the problem of training / inference distribution shift resembles that of \textit{exposure bias} \cite{DBLP:journals/corr/RanzatoCAZ15} in autoregressive models, for which the mismatch is caused by \textit{teacher forcing} during training \cite{6795228}. 
Several solutions have been proposed for this problem in the literature  \cite{Bengio-2015-scheduled-sampling,lamb2016professor,yu2016seqgan}. Step-unrolled denoising diffusion also closely resembles the approach in \cite{savinov2022stepunrolled} for training denoising autoencoders on text.

We only perform \stepunrolled~during model fine-tuning. 
Early in training the denoising predictions are inaccurate, so the latent marginals over the noisy target maps will be closer to the desired \textit{true} marginals than those produced by adding noise to denoiser network outputs. 
% Hence, doing \stepunrolledshort~early in supervised pre-training is not recommended.
One might consider the use of a curriculum for gradually introducing \stepunrolled~in the later stages of supervised pre-training, but this introduces additional hyper-parameters, so we simply invoke \stepunrolled~during fine-tuning, and leave an exploration of curricula to future work.
% might be helpful but it introduces another hyperparameter so we leave that exploration to future work.

\myparagraph{$L_1$ denoiser loss.}
While the $L_2$ loss in Eqn.\ \ref{eq:diffusion-loss} is ideal for Gaussian noise and noise-free ground truth maps, in practice,
real ground truth depth and flow fields are noisy and heavy tailed; \eg,~for distant objects, near object boundaries, and near pixels with missing annotations.
We hypothesize that the robustness afforded by the $L_1$ loss may therefore be useful in training 
the neural denoising network.
(See Tables \ref{tab::loss-ablation-nyu} and \ref{tab::loss-ablation-kitti} in the supplementary material for an ablation of the loss function for monocular depth estimation.)

\subsection{Coarse-to-fine refinement}
\label{sec:tiled-refinement}
\vspace*{-0.1cm}

% \srbs{Camera ready: reworded first two sentences}

Training high resolution diffusion models is often slow and memory intensive but increasing the image resolution of the model has been shown to improve performance on vision tasks \cite{godard2019digging}.
One simple solution, yielding  high-resolution output without increasing the training cost, is to perform inference in a coarse-to-fine manner, first estimating flow over the entire field of view at low resolution, and then refining the estimates in a patch-wise manner. 
For refinement, we first up-sample the low-resolution map to the target resolution using bicubic interpolation.
Patches are cropped from the up-scaled map, denoted $z$, along with the corresponding RGB inputs.
Then we run diffusion model inference  starting at time $t'$ with a noisy map 
$y_{t'} \sim \mathcal{N}(y_{t'};\sqrt{\gamma_{t'}}\,z, (1-\gamma_{t'})I)$.
For simplicity, $t'$ is a fixed hyper-parameter, set based on a 
validation set.
This process is carried out for multiple overlapping patches.
Following  Perceiver IO~\citep{jaegleperceiver}, the patch estimates are merged using weighted masks with lower weight near the patch boundaries since predictions at boundaries are more prone to errors. 
(See Section \ref{sec:supp-inference} for more details.)

% and using our low resolution predicted label $l'$ to initialize $y_{t'} \sim \mathcal{N}(y_{t'};\sqrt{\gamma_{t'}}l', (1-\gamma_{t'})I)$. $t'$ is an inference hyper-parameter that is tuned for each dataset. Patches are merged using weighted masks with lesser weight near the boundaries since predictions at boundaries are more prone to errors. This is similar to the aggregation used in Perceiver IO~\citep{jaegleperceiver}.
% Note that $y_{t'}$ may not be a fair sample from $\mathcal{N}(y_t;\sqrt{\gamma_t}y_0, (1-\gamma_t)I) \forall t$ due to interpolation artifacts which usually necessitates supervised training for super-resolution like in~\cite{saharia2021image} implying that the improvements here are a lower bound for the possible improvements that can be achieved with a model trained for super-resolution. Notwithstanding, we find this approach to be remarkably effective at improving fine-grained detail.

% \vspace*{-0.1cm}
\section{Experiments}
\label{sec:experiments}
% \vspace*{-0.2cm}

\begin{figure*}[t]
    \centering
    \scriptsize
    \setlength\tabcolsep{1pt} % default value: 6pt
    \newcommand{\Figwidth}{0.245\linewidth}
    \begin{tabular}{cccc}
      First frame & Ground Truth & RAFT & \textbf{DDVM (Ours)} \\

    \includegraphics[trim={0 40 0 0},clip,width=\Figwidth]{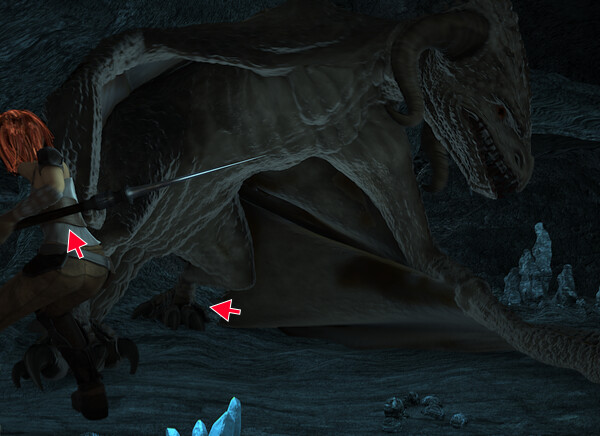} &
    \includegraphics[trim={0 40 0 0},clip,width=\Figwidth]{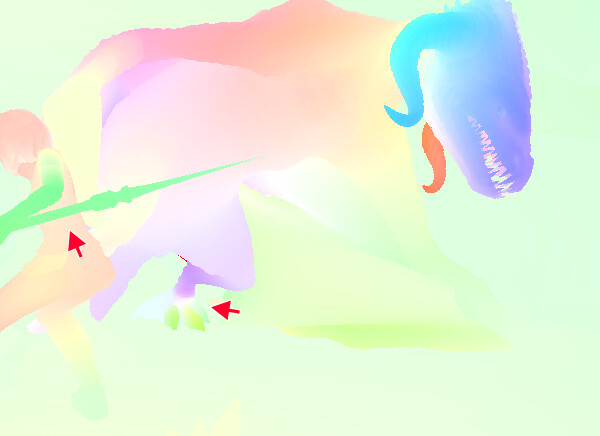} &
    \includegraphics[trim={0 40 0 0},clip,width=\Figwidth]{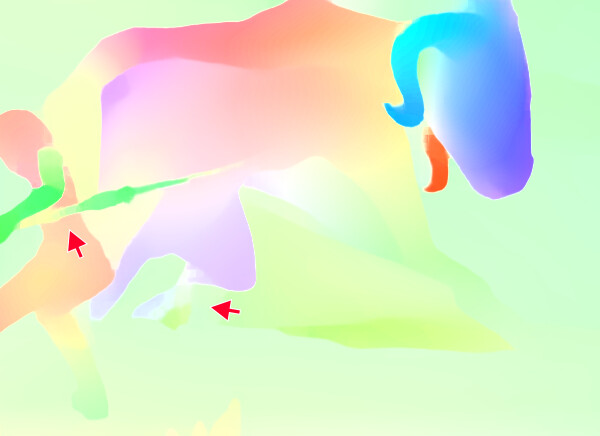} &
    \includegraphics[trim={0 40 0 0},clip,width=\Figwidth]{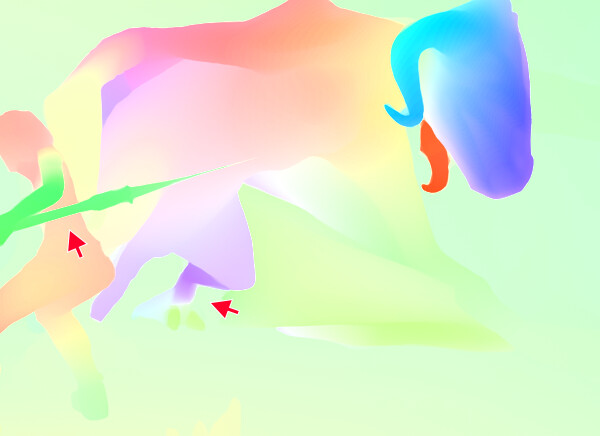} \\
    
        \includegraphics[trim={0 40 0 80},clip,width=\Figwidth]{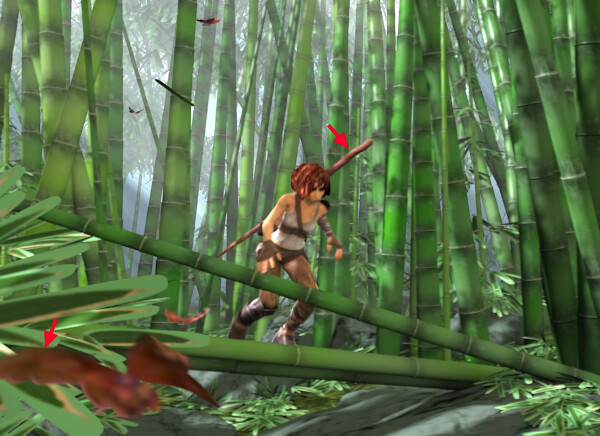} &
    \includegraphics[trim={0 40 0 80},clip,width=\Figwidth]{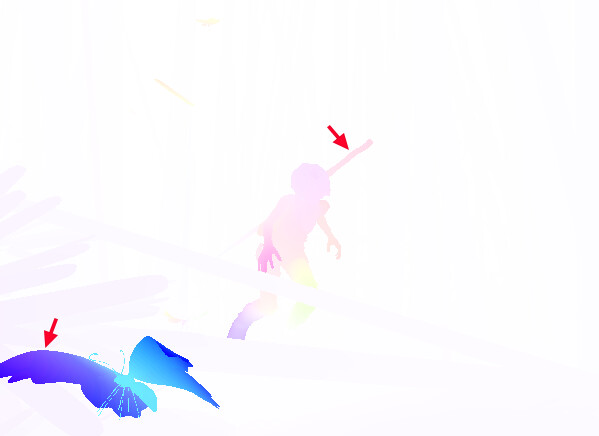} &
    \includegraphics[trim={0 40 0 80},clip,width=\Figwidth]{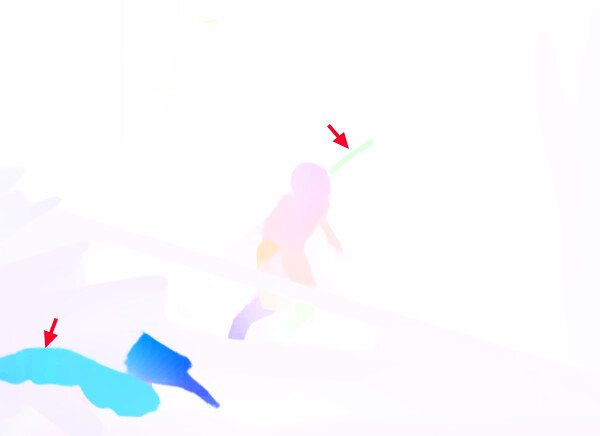} &
    \includegraphics[trim={0 40 0 80},clip,width=\Figwidth]{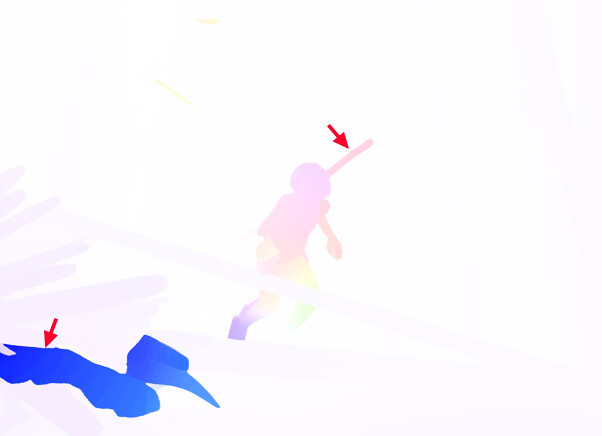} \\
    
    \includegraphics[trim={0 120 0 0},clip,width=\Figwidth]{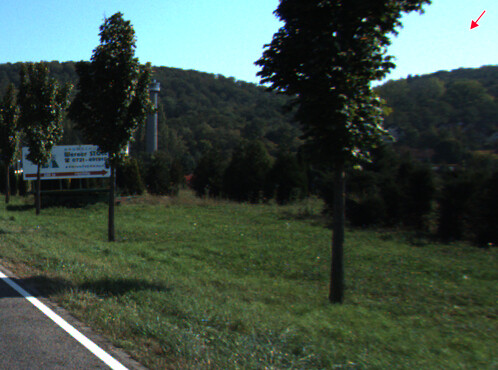} &
    \includegraphics[trim={0 120 0 0},clip,width=\Figwidth]{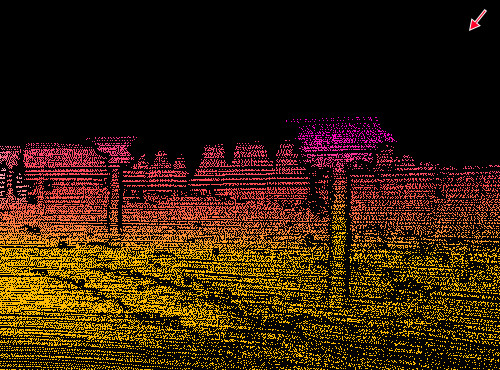} &
    \includegraphics[trim={0 120 0 0},clip,width=\Figwidth]{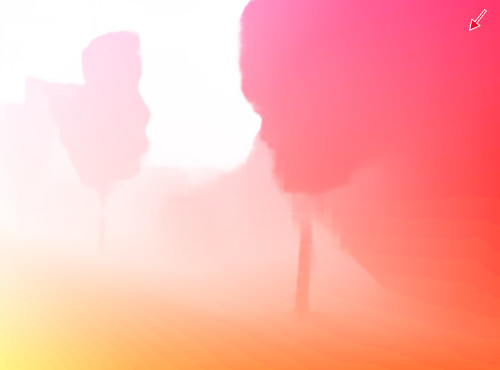} &
    \includegraphics[trim={0 120 0 0},clip,width=\Figwidth]{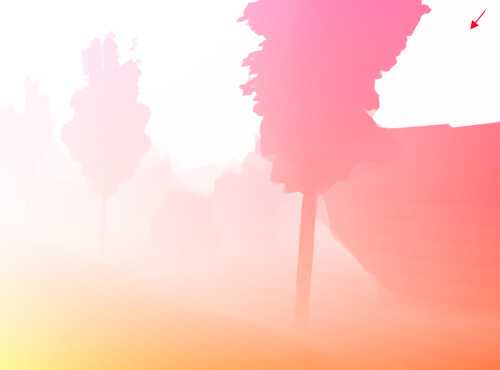} \\
    
    \includegraphics[trim={0 40 0 0},clip,width=\Figwidth]{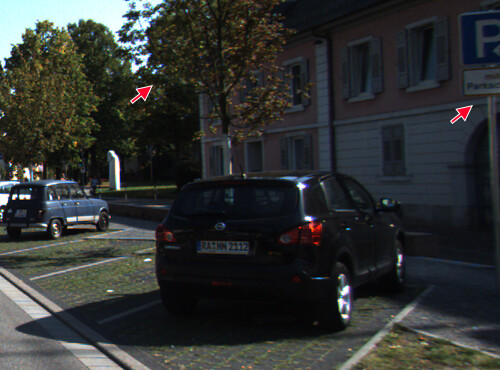} &
    \includegraphics[trim={0 40 0 0},clip,width=\Figwidth]{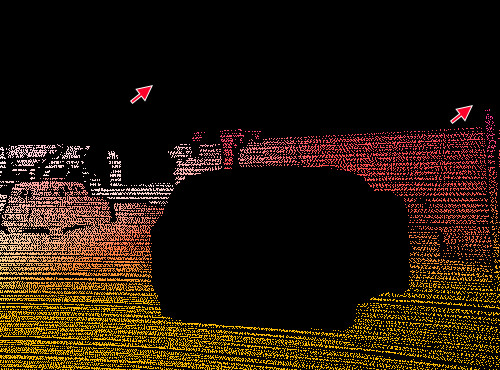} &
    \includegraphics[trim={0 40 0 0},clip,width=\Figwidth]{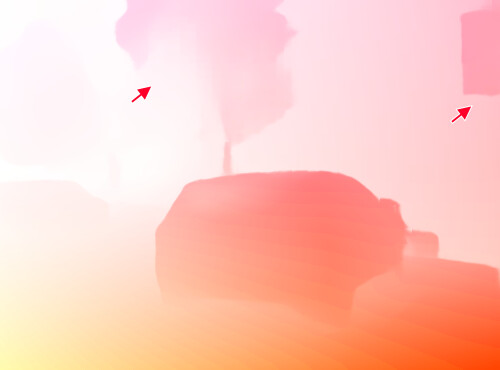} &
    \includegraphics[trim={0 40 0 0},clip,width=\Figwidth]{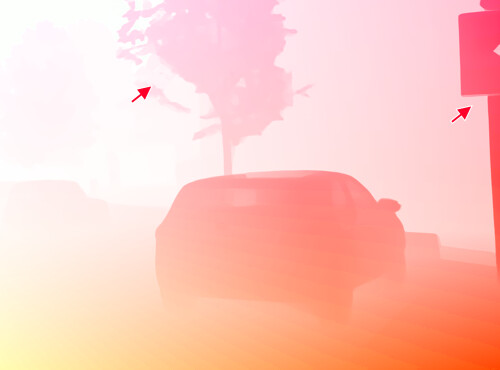} \\

    \end{tabular}
        \vspace*{-0.1cm}
	\caption{\textbf{Visual results comparing RAFT with our method after pretraining}. Note that our method does much better on fine details and ambiguous regions.}
    \label{fig:flow_raft_pretrain_comparison}
    \vspace{-1em}
\end{figure*}

\begin{figure*}[t]
    \centering
    \small
    \setlength\tabcolsep{1pt} % default value: 6pt
    \newcommand{\Figwidth}{0.246\linewidth}
    \begin{tabular}{cccc}
      First frame & Ground Truth & RAFT & \textbf{DDVM (Ours)} \\

    \includegraphics[trim={0 20 0 60},clip,width=\Figwidth]{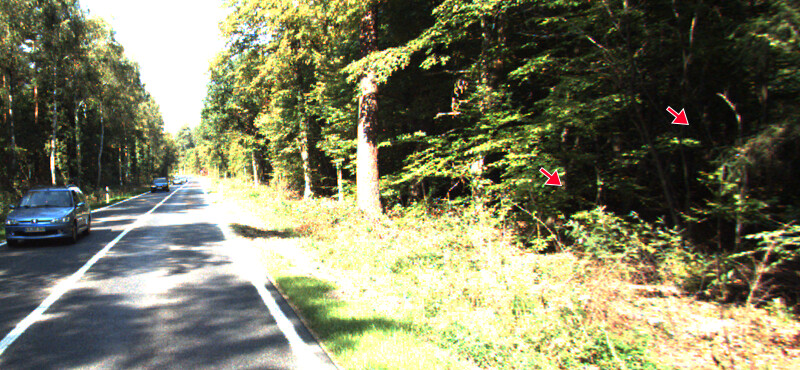} &
    \includegraphics[trim={0 20 0 60},clip,width=\Figwidth]{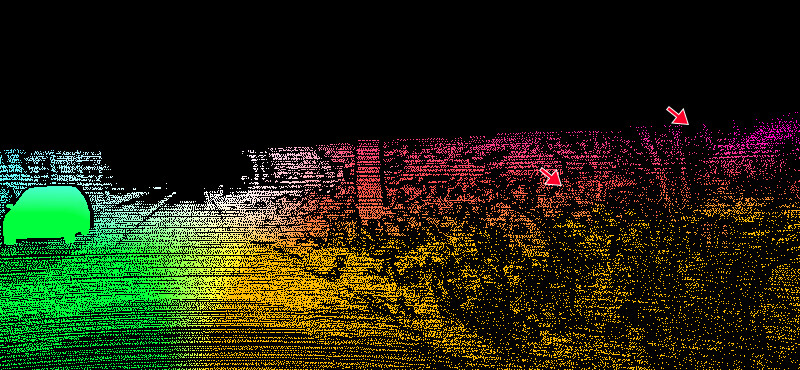} &
    \includegraphics[trim={0 20 0 60},clip,width=\Figwidth]{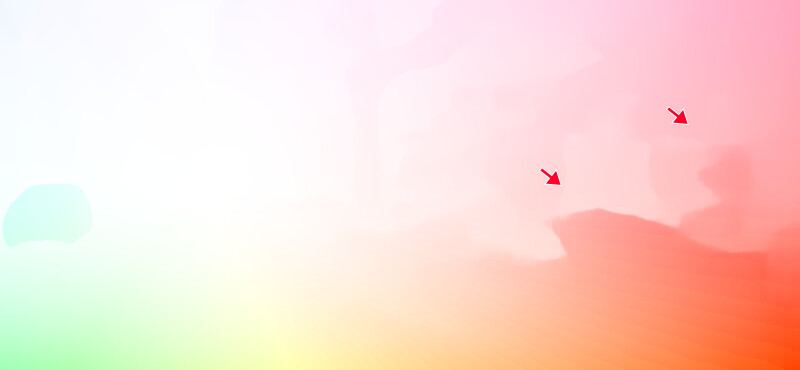} &
    \includegraphics[trim={0 20 0 60},clip,width=\Figwidth]{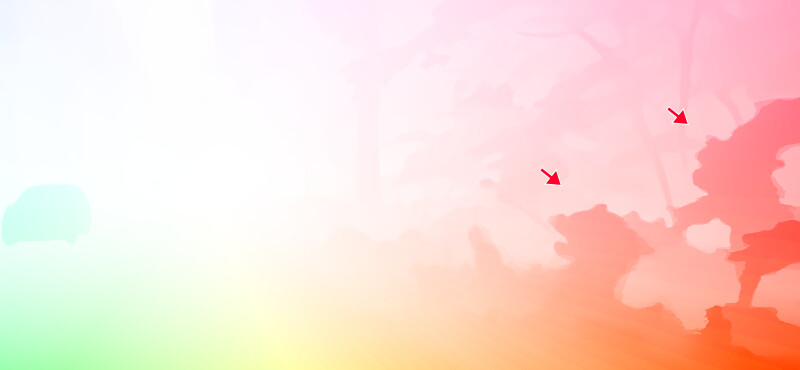} \\
    \includegraphics[trim={0 40 0 0},clip,width=\Figwidth]{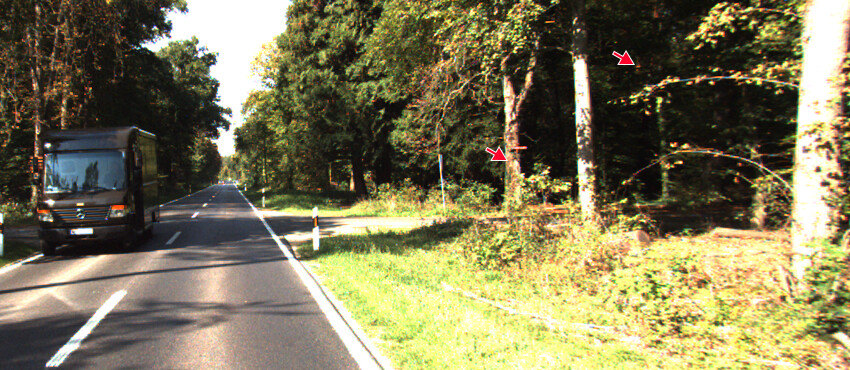} &
    \includegraphics[trim={0 40 0 0},clip,width=\Figwidth]{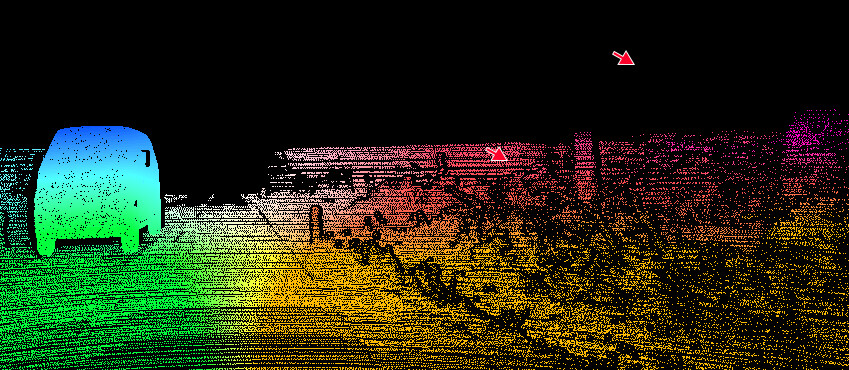} &
    \includegraphics[trim={0 40 0 0},clip,width=\Figwidth]{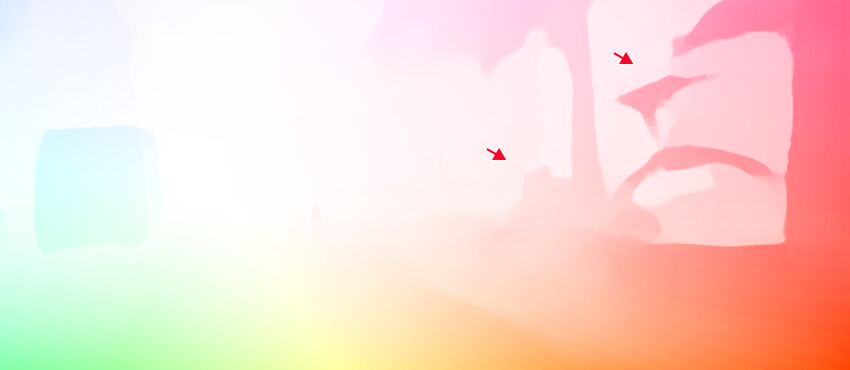} &
    \includegraphics[trim={0 40 0 0},clip,width=\Figwidth]{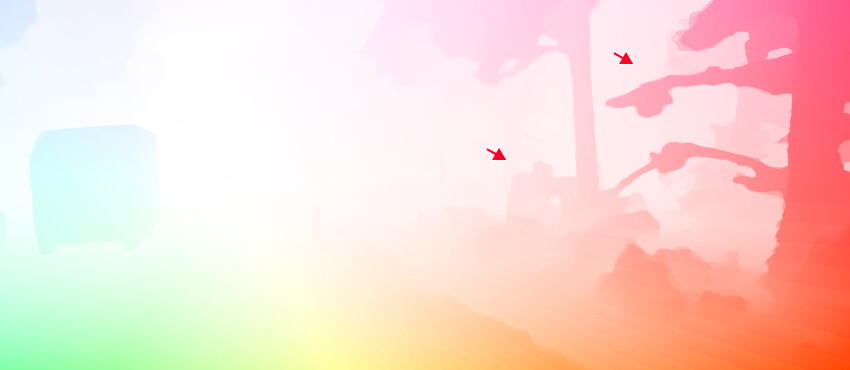} \\
    \includegraphics[trim={10 0 10 40},clip,width=\Figwidth]{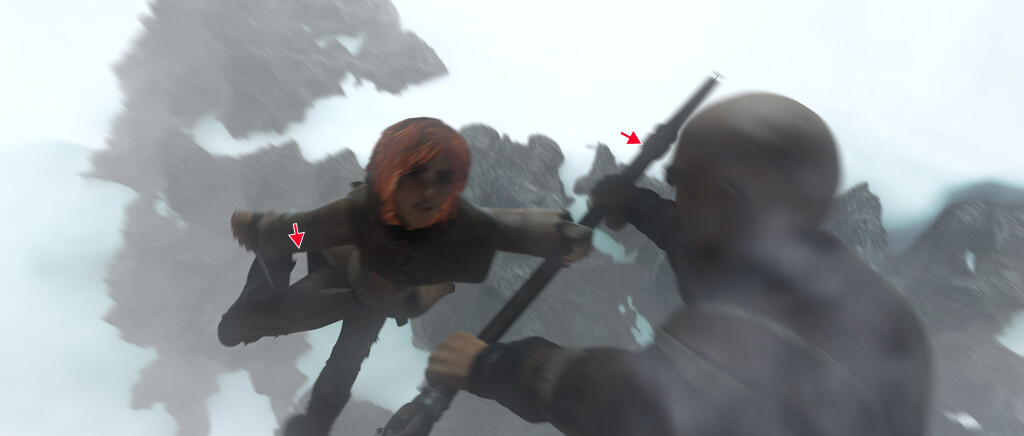} &
    \includegraphics[trim={10 0 10 40},clip,width=\Figwidth]{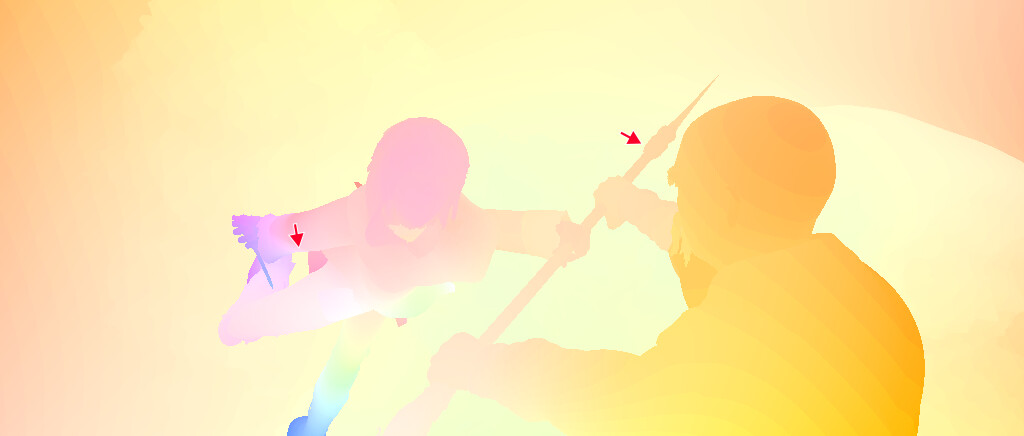} &
    \includegraphics[trim={10 0 10 40},clip,width=\Figwidth]{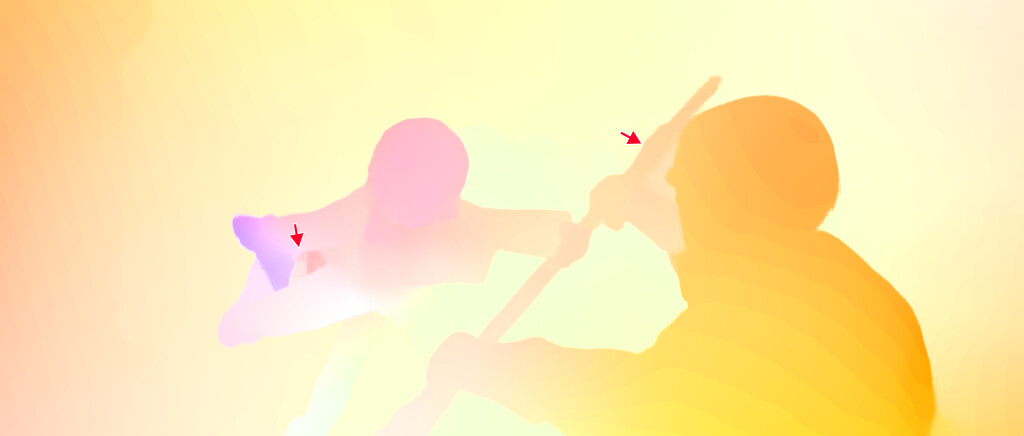} &
    \includegraphics[trim={10 0 10 40},clip,width=\Figwidth]{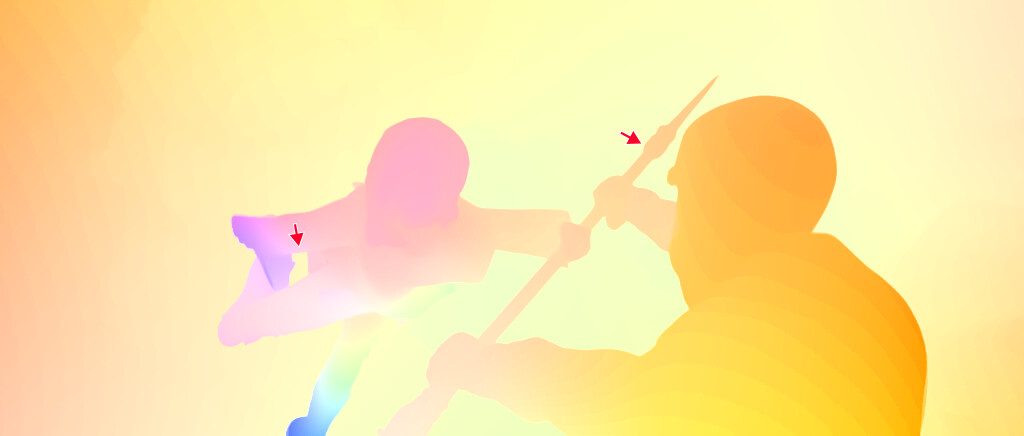} \\
    \includegraphics[trim={0 0 60 60},clip,width=\Figwidth]{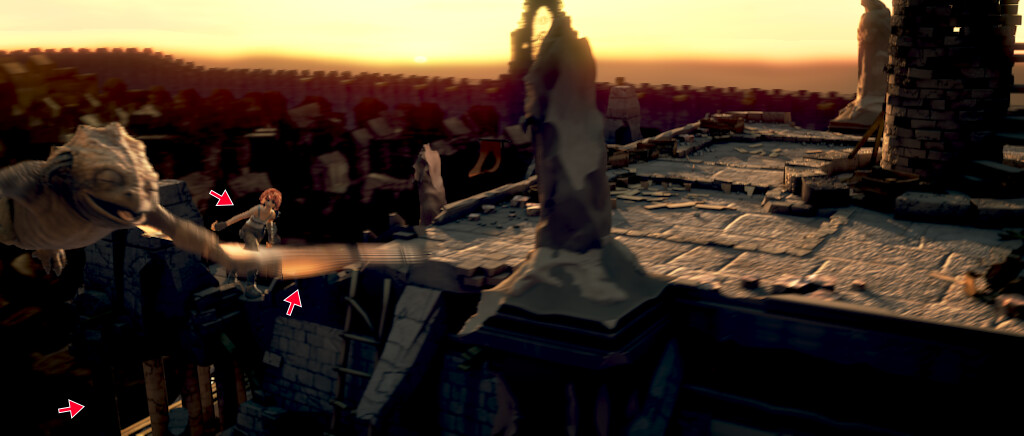} &
    \includegraphics[trim={0 0 60 60},clip,width=\Figwidth]{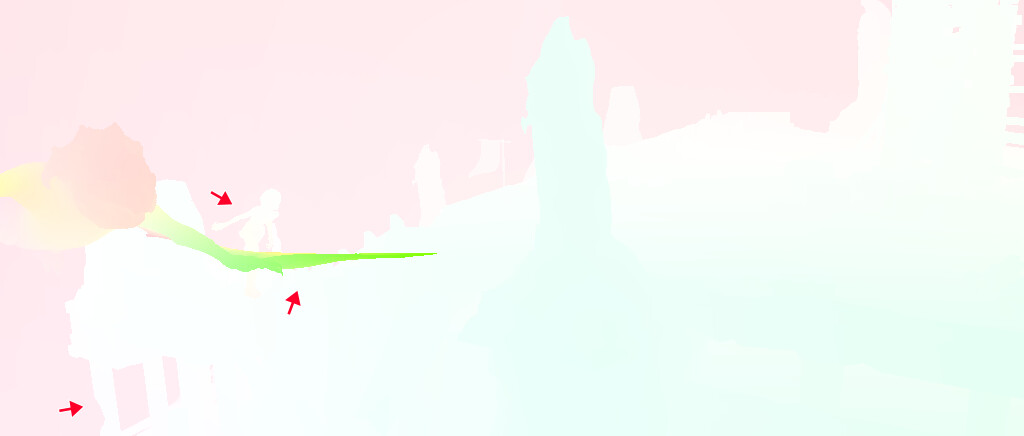} &
    \includegraphics[trim={0 0 60 60},clip,width=\Figwidth]{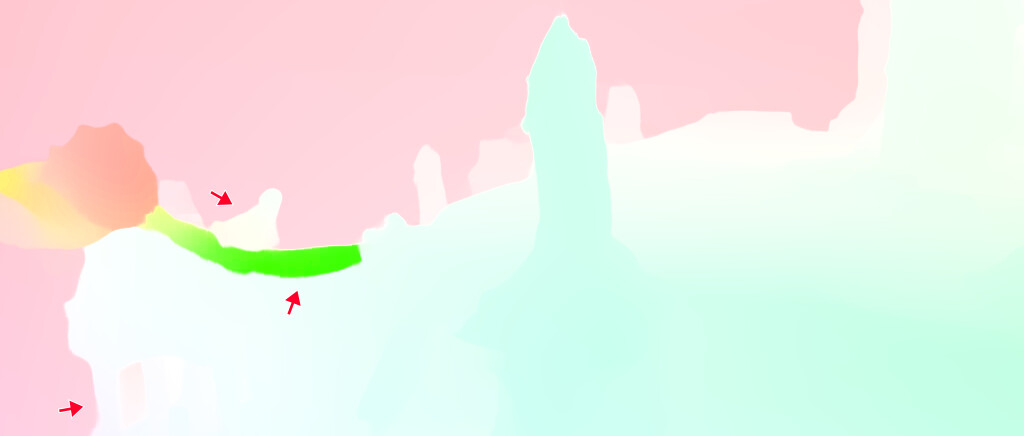} &
    \includegraphics[trim={0 0 60 60},clip,width=\Figwidth]{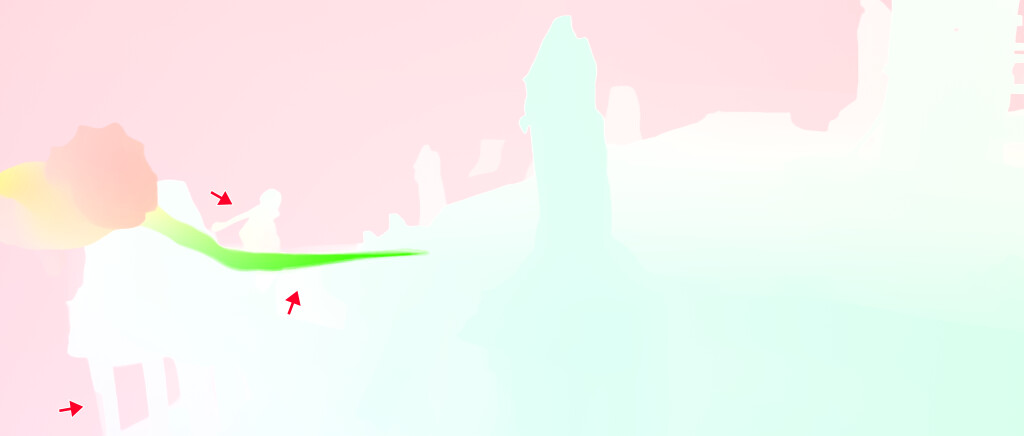} \\

    \end{tabular}
        % \vspace*{-0.05cm}
	\caption{\textbf{Visual results comparing RAFT with our method after finetuning}. Ours does much better on fine details and ambiguous regions.}
    \label{fig:flow_raft_finetune_comparison}
        % \vspace*{-0.1cm}
\end{figure*}

As our denoiser backbone, we adopt the Efficient UNet architecture \cite{sahariac-imagen}, pretrained with Palette \cite{sahariac-palette} style self-supervised pretraining, and slightly modified to have the appropriate input and output channels for each task.
Since diffusion models expect inputs and generate outputs in the range $[-1., 1.]$, we normalize depths using max depth of 10 meters and 80 meters respectively for the indoor and outdoor models. We normalize the flow using the height and width of the ground truth.
Refer to Section \ref{sec:supp-training-inference-details} for more details on the architecture, augmentations and other hyper-parameters.

\myparagraph{Optical flow.} We pre-train on the mixture described in Section \ref{sec:pretrain-datasets} at a resolution of 320$\times$448 and report zero-shot results on the widely used Sintel \cite{Butler:ECCV:2012}
and KITTI \cite{Menze2018JPRS} datasets. We further fine-tune this model on the standard mixture consisting of AutoFlow \cite{sun2021autoflow}, FlyingThings \cite{Mayer_2016_CVPR}, VIPER \cite{richter2017playing}, HD1K \cite{kondermann2016hci}, Sintel and KITTI at a resolution of 320$\times$768 and report results on the test set from the public benchmark. We use a standard average end-point error (AEPE) metric that calculates L2 distance between ground truth and prediction. On KITTI, we additionally use the outlier rate, Fl-all, which reports the outlier ratio in $\%$ among all pixels with valid ground truth, where an estimate is considered as an outlier if its error exceeds 3 pixels and 5$\%$ \wrt the ground truth.

\myparagraph{Depth.} We separately pre-train indoor and outdoor models on the respective pre-training datasets described in Section \ref{sec:pretrain-datasets}. The indoor depth model is then finetuned and evaluated on the NYU depth v2 dataset \cite{nyu} and the outdoor model on the KITTI depth dataset \cite{kitti}. We follow the standard evaluation protocol used in prior work \citep{binsformer}. For both NYU depth v2 and KITTI, we report the absolute relative error (REL), root mean squared error (RMS) and accuracy metrics ($\delta_1 < 1.25$).

\vspace*{-0.1cm}

\subsection{Evaluation on benchmark datasets}
\label{results}
\vspace*{-0.1cm}

\begin{table*}[t]
\centering
\begin{minipage}[t]{0.55\textwidth}
\centering
\scriptsize

\caption{\textbf{Zero-shot optical flow estimation results on Sintel and KITTI}. 
We provide a new RAFT baseline using our proposed pre-training mixture and substantially improve the accuracy over the original. 
Our diffusion model outperforms even this much stronger baseline and  achieves state-of-the-art zero-shot results on Sintel.final and KITTI.}
\label{tab:flow-pretrain-results}
\vspace{-0.3em}
\setlength\tabcolsep{1.15pt}
    \begin{tabular}{l l @{\hskip 0.5em} cccc}
    \toprule
    \multirow{2}{*}{Model} &  \multirow{2}{*}{Dataset} & Sintel.clean & Sintel.final & \multicolumn{2}{c}{KITTI} \\ 
    \cmidrule(lr){3-4} \cmidrule(lr){5-6}
     &    & \multicolumn{2}{c}{AEPE} & AEPE & Fl-all \\ 
    \midrule
    FlowFormer &  Chairs~$\!\rightarrow\!$~Things  & \textbf{1.01} & 2.40 & 4.09 &  14.72\% \\  %
    RAFT & Chairs~$\!\rightarrow\!$~Things & {1.68}	&2.80	&5.92 & -\\
    \hline 
    Perceiver IO & AutoFlow & 1.81 & 2.42 &  4.98 & -\\ 
    RAFT & AutoFlow   & 1.74	&2.41	&4.18  &   13.41\% \\ 
    \hline
    {RAFT (ours)}
    & AF$\rightarrow$AF+FT+KU+TA & 1.27 &	2.28 &	2.71 & 9.16\% \\
    % \multirow{4}{*}{\textbf{Ours}} &AF & 2.04 & 2.55 & 4.47 & 16.59\% \\
    % &AF$\rightarrow$AF+FT & 1.48 & 2.22 & 3.71 & 14.07\% \\
    % &AF$\rightarrow$AF+FT+KU & 1.33 &  2.04 &  2.82 & 9.27\% \\
    % &AF$\rightarrow$AF+FT+KU+TA &  {1.24} & \textbf{2.00} & \textbf{2.19} & \textbf{7.58\%} \\
    \textbf{\ours~(ours)} &AF$\rightarrow$AF+FT+KU+TA &  {1.24} & \textbf{2.00} & \textbf{2.19} & \textbf{7.58\%} \\
    \bottomrule
    \end{tabular}
\end{minipage} \hspace{1.5em}
\begin{minipage}[t]{0.4\textwidth}
    % \vspace{-0.5em}
    \centering
    \scriptsize
    \caption{{\bf Optical flow finetuning evaluation} on public benchmark datasets (AEPE$\downarrow$ for Sintel and Fl-all$\downarrow$ for KITTI). 
    \textbf{Bold} indicates the best and \underline{underline} the \nth{2}-best.
    $^{\S}$ uses extra datasets (AutoFlow and VIPER) on top of defaults (FlyingThings, HD1K, KITTI, and Sintel). $^*$uses warm start on Sintel.
    }
    \label{tab:flow_benchmark}
    \setlength\tabcolsep{2.15pt}
    \begin{tabular}{l @{\hskip 1em} cccc}
    \toprule
    Method & Sintel.clean & Sintel.final & KITTI \\ \midrule
    SKFlow \citep{sunskflow}$^*$ & \underline{1.30} & \underline{2.26} & 4.84\% \\ 
    CRAFT \citep{sui2022craft}$^*$ & {1.44} & 2.42 & 4.79\% \\
    FlowFormer \citep{huang2022flowformer} & \textbf{1.14} & \textbf{2.18} & 4.68\% \\ 
    RAFT-OCTC~\citep{jeong2022imposing}$^{*}$ & 1.51 &2.57 & 4.33\% \\
    RAFT-it~\citep{sun2022disentangling}$^{\S}$ & 1.55 & 2.90 & \underline{4.31\%} \\
    \textbf{\ours~(ours)}$^{\S}$ & 1.75 & 2.48 & \textbf{3.26\%} \\ 
    \bottomrule
    \end{tabular}
\end{minipage}
\vspace{-0.5em}
\end{table*}

\begin{table}[t]
\centering
\scriptsize

\caption{\textbf{Performance comparison on the NYU-Depth-v2 and KITTI datasets}. $\top$ indicates method uses unsupervised pretraining, \dag indicates supervised pretraining and \ddag~indicates use of auxilliary supervised depth data. \textbf{Best} / \underline{second best} results are bolded / underlined respectively. $\downarrow$: lower is better $\uparrow$: higher is better.}
\label{tab::results-depth-merged}
\vspaceaftercaption

\setlength\tabcolsep{3.5pt}
    \begin{tabularx}{0.77\linewidth}{l @{\hskip 1em} l @{\hskip 1.5em} c @{\hskip 3em} c c c c c c c}
    \toprule
    \multicolumn{2}{l}{\multirow{2}{*}{Method}} & \multirow{2}{*}{Architecture} & \multicolumn{3}{c}{NYU-Depth-v2} & & \multicolumn{3}{c}{KITTI} \\
    \cmidrule(r){4-6} \cmidrule(lr){8-10}
    & & & $\delta_1\uparrow$ & REL$\downarrow$ & RMS$\downarrow$ & & $\delta_1\uparrow$ & REL$\downarrow$ & RMS$\downarrow$ \\
    \midrule
    \multicolumn{2}{l}{TransDepth \citep{transdepth}} & Res-50+ViT-B$^\dag$ & 0.900 & 0.106 & 0.365 & & 0.956 & 0.064 & 2.755 \\
    \multicolumn{2}{l}{DPT \citep{dpt}} & Res-50+ViT-B$^{\dag\ddag}$ & 0.904 & 0.110 & 0.357  & & 0.959 & 0.062 & 2.573 \\
    \multicolumn{2}{l}{BTS \citep{bts}} & DenseNet-161$^\dag$ &0.885 & 0.110 & 0.392 & & 0.956 & 0.059 & 2.756 \\
    \multicolumn{2}{l}{AdaBins \citep{adabins}} & E-B5+Mini-ViT$^{\dag}$ & 0.903 & 0.103 & 0.364 & & 0.964 & 0.058 & 2.360  \\
    \multicolumn{2}{l}{BinsFormer \citep{binsformer}} & Swin-Large$^\dag$ & 0.925 & 0.094 & 0.330  & & 0.974 & 0.052 & \second{2.098}  \\
    \multicolumn{2}{l}{PixelFormer \citep{pixelformer}} & Swin-Large$^\dag$ & 0.929 & 0.090 & 0.322 & & \second{0.976} & \second{0.051} & \second{2.081} \\
    \multicolumn{2}{l}{MIM \citep{xie2022revealing}} & SwinV2-L$^\top$ & \underline{0.949} & 0.083 & \second{0.287} & & \first{0.977} & \first{0.050} & \first{1.966} \\
    \multicolumn{2}{l}{AiT-P \citep{AllInTokens}} & SwinV2-L$^\top$ &  \first{0.953} & 0.076 & \first{0.279} & & -- & -- & --  \\
    \midrule
    \multirow{3}{*}{\textbf{\ours}} & samples=1 & \unet$^{\top\ddag}$ & 0.944 & \second{0.075} & 0.324 & & 0.964 & 0.056 & 2.700 \\
    & samples=2 & \unet$^{\top\ddag}$ & 0.944 & \first{0.074} & 0.319 & & 0.965 & 0.055 & 2.660 \\
    & samples=4 & \unet$^{\top\ddag}$ & 0.946 & \first{0.074} & 0.315 & & 0.965 & 0.055 & 2.613 \\
    \bottomrule
\end{tabularx} 

\vspace{-1.0em}
\end{table}

\myparagraph{Depth.}
Table \ref{tab::results-depth-merged} reports the results on NYU depth v2 and KITTI (see Section \ref{sec:supp-depth-results} for more detailed results and Section \ref{sec:supp-dpt} for qualitative comparison with DPT on NYU). We achieve a state-of-the-art absolute relative error of 0.074 on NYU depth v2.
On KITTI, our method performs competitively with prior work.
We report results with averaging depth maps from one or more samples. Note that most prior works use post processing that averages two samples, one from the input image, and the other based on its reflection about the vertical axis.

\myparagraph{Flow.} 
Table \ref{tab:flow-pretrain-results} reports the zero-shot  results of our model on Sintel and KITTI Train datasets where ground truth are provided. The model is trained on our newly proposed pre-training mixtures (AutoFlow (AF), FlyingThings (FT), Kubric (KU), and TartanAir (TA)). We report results by averaging 8 samples at a coarse resolution and then refining them to the full resolution as described in Section \ref{sec:tiled-refinement}.
For a fair comparison, we re-train RAFT on this pre-training mixture; this new RAFT model significantly outperforms the original RAFT model.
And our diffusion model outperforms the stronger RAFT baseline.
It achieves the state-of-the-art zero-shot results on both the challenging Sintel Final and KITTI datasets.
% When gradually adding up Kubric and TartanAir, we observe a substantially accuracy boost. \srbs{Should this last point be moved to dataset ablation?}

Figure~\ref{fig:flow_raft_pretrain_comparison} provides a qualitative comparison of pre-trained models.
Our method demonstrates finer details on both object and motion boundaries.
Especially on KITTI, our model  recovers fine details remarkably well, \eg~on trees and its layered motion between tree and background.

We further finetune our model on the mixture of the following datasets, AutoFlow, FlyingThings, HD1K, KITTI, Sintel, and VIPER.
Table \ref{tab:flow_benchmark} reports the comparison to state-of-the-art optical flow methods on public benchmark datasets, Sintel and KITTI.
On KITTI, our method outperforms all existing optical flow methods by a substantial margin (even most scene flow methods that use stereo inputs), and sets the new state of the art. 
On the challenging Sintel final, our method is competitive with other state of the art models.
Except for methods using warm-start strategies, our method is only behind FlowFormer which adopts strong domain knowledge on optical flow (\eg~cost volume, iterative refinement, or attention layers for larger context) unlike our generic model.
Interestingly, we find that our model outperforms FlowFormer on 11/12 Sintel test sequences and our overall worse performance can be attributed to a much higher AEPE on a single (possibly out-of-distribution) test sequence. We discuss this in more detail in Section \ref{sec:limitations}.
On KITTI, our diffusion model outperforms FlowFormer by a large margin (30.34$\%$).

\vspace*{-0.1cm}
\subsection{Ablation study}
\vspace*{-0.1cm}

\begin{table*}[t]
\centering
\scriptsize
\begin{minipage}[t]{0.56\textwidth}
    \centering
    \scriptsize
    \caption{\textbf{Ablation on infilling and step-unrolling}. Without either one, performance deteriorates. Without both, optical flow models fail to train on KITTI. 
    % (failing to train on optical flow estimation with KITTI). 
    % Using both infilling and step-unrolling gives the best performance.
    }
    \label{tab::infilling_step_unrolling_ablation}
    \vspace{-0.8em}
    \setlength\tabcolsep{4.6pt}
    \begin{tabular}{l @{\hskip 2em} cccc|cc}
        \toprule
        & \multicolumn{2}{c}{NYU val} & \multicolumn{2}{c|}{KITTI val} & \multicolumn{2}{c}{KITTI val} \\
        \cmidrule(lr){2-3} \cmidrule(lr){4-5} \cmidrule(lr){6-7}
        & REL & RMS & REL & RMS & AEPE & Fl-all \\
        \midrule
        Baseline & 0.079 & 0.331 & 0.222 & 3.770 & - & -\\
        Step-unroll & 0.076 & \textbf{0.324} & 0.085 & 2.844 & 1.84 & 6.16\% \\
        Infill & 0.077 & 0.338 & 0.057 & 2.744 & 1.53 & 5.24\% \\
        \textbf{Step-unroll $\&$ infill} & \textbf{0.075} & \textbf{0.324} & \textbf{0.056} & \textbf{2.700} & \textbf{1.47} & \textbf{4.74\%} \\
        \bottomrule
    \end{tabular} 
\end{minipage} \hspace{1.5em}
\begin{minipage}[t]{0.40\textwidth}
    \centering
    \scriptsize
    \setlength\tabcolsep{1.15pt}
    \caption{\textbf{Coarse-to-fine refinement} improves zero-shot optical flow estimation results on Sintel and KITTI, along with the qualitative improvements shown in Figure \ref{fig:flow_tiled_ablation}.}
    \label{tab::tiled_refinement_ablation}
    \begin{tabular}{l @{\hskip 1em} cccc}
        \toprule
        \multirow{2}{*}{\makecell[l]{Coarse-to-fine \\ refinement}}
        & Sintel.clean & Sintel.final
        &\multicolumn{2}{c}{KITTI} \\
        \cmidrule(lr){2-3} \cmidrule(lr){4-5}
         & AEPE & AEPE & AEPE & Fl-all \\
        \midrule
        Without & 1.42 & 2.12 & 2.35 & 8.65\% \\
        \textbf{With} & \textbf{1.24} & \textbf{2.00} & \textbf{2.19} & \textbf{7.58\%} \\
        \bottomrule
    \end{tabular}
\end{minipage}
\vspace{-1em}
\end{table*}

\begin{table*}[t]
\centering

\begin{minipage}[t]{0.56\textwidth}
\centering
\scriptsize
\caption{\textbf{The addition of optical flow synthetic datasets} substantially improves the zero-shot results on Sintel and KITTI.}
\label{tab:flow-dataset-ablation}
\setlength\tabcolsep{4.5pt}
\begin{tabular}{lcccc}
\toprule
Dataset & Sintel.clean & Sintel.final & KITTI & KITTI Fl-all\\
\midrule
AF pretraining & 2.04 & 2.55 & 4.47 & 16.59\% \\
AF$\rightarrow$AF+FT & 1.48 & 2.22 & 3.71 & 14.07\% \\
AF$\rightarrow$AF+FT+KU & 1.33 &  2.04 &  2.82 & 9.27\% \\
AF$\rightarrow$AF+FT+KU+TA &  \textbf{1.24} & \textbf{2.00} & \textbf{2.19} & \textbf{7.58\%} \\
\bottomrule
\end{tabular}
\end{minipage} \hspace{1.5em}
\begin{minipage}[t]{0.39\textwidth}
    \caption{\textbf{The addition of synthetic depth data in pre-training} substantially improves fine-tuning performance on NYU.}
    \label{tab:nyu-dataset-ablation}
    \vspace{-0.2em}
    \centering
    \scriptsize
    \begin{tabular}{l @{\hskip 2.3em} cc}
        \toprule
        Dataset & REL & RMS \\ 
        \midrule
        SceneNet RGB-D & 0.089 & 0.362 \\
        ScanNet & 0.081 & 0.346 \\
        SceneNet RGB-D + ScanNet & \textbf{0.075} & \textbf{0.324} \\
        \bottomrule
    \end{tabular}
\end{minipage}
\vspace{-0.5em}
\end{table*}

\myparagraph{Infilling and step-unrolling.}
We study the effect of infilling and step-unrolling in Table \ref{tab::infilling_step_unrolling_ablation}. For depth, we report results for fine-tuning our pre-trained model on the NYU and KITTI datasets with the same resolution and augmentations as our best results. For flow, we fine-tune on the KITTI train set alone (with nearest neighbor resizing to the target resolution being the only augmentation) at a resolution of 320$\times$448 and report metrics on the KITTI val set \citep{lv2018learning}.
We report results with a single sample and no coarse-to-fine refinement.
We find that training on raw sparse data without infilling and step unrolling leads to poor results, especially on KITTI where the ground truth is quite sparse.
Step-unrolling helps to stabilize training without requiring any extra data pre-processing. 
However, we find that most gains come from interpolating missing values in the sparse labels. Infilling and step-unrolling compose well as our best results use both; infilling (being an approximation) does not completely bridge the training-inference distribution shift of the noisy latent.

\begin{figure*}[t]
    \centering
    \scriptsize
    \setlength\tabcolsep{1pt} % default value: 6pt
    \newcommand{\Figwidth}{0.245\linewidth}
    \begin{tabular}{cccc}
      First frame & Ground Truth & Base & Refined \\
    \includegraphics[trim={400 0 100 40},clip,width=\Figwidth]{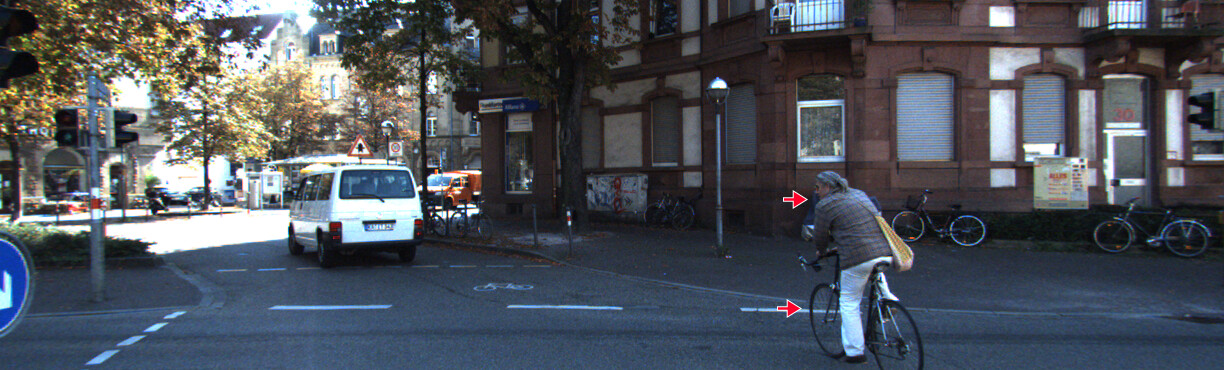} &
    \includegraphics[trim={400 0 100 40},clip,width=\Figwidth]{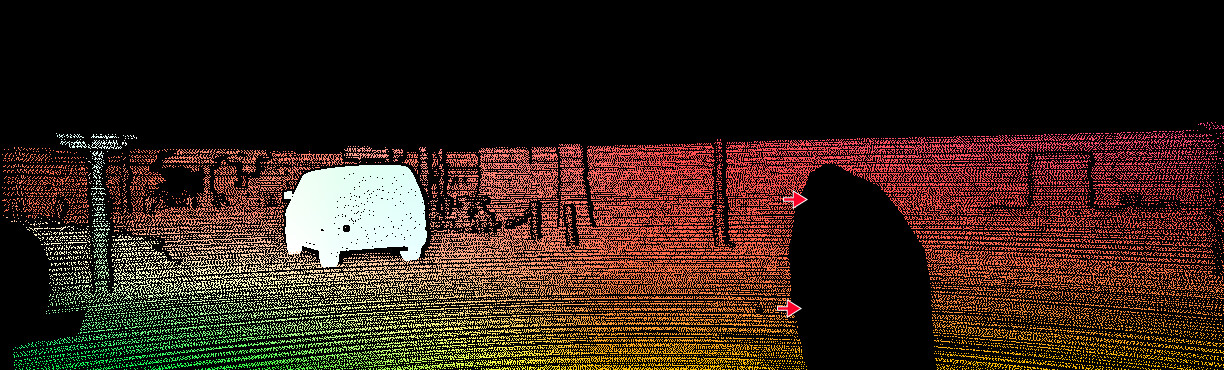} &
    \includegraphics[trim={400 0 100 40},clip,width=\Figwidth]{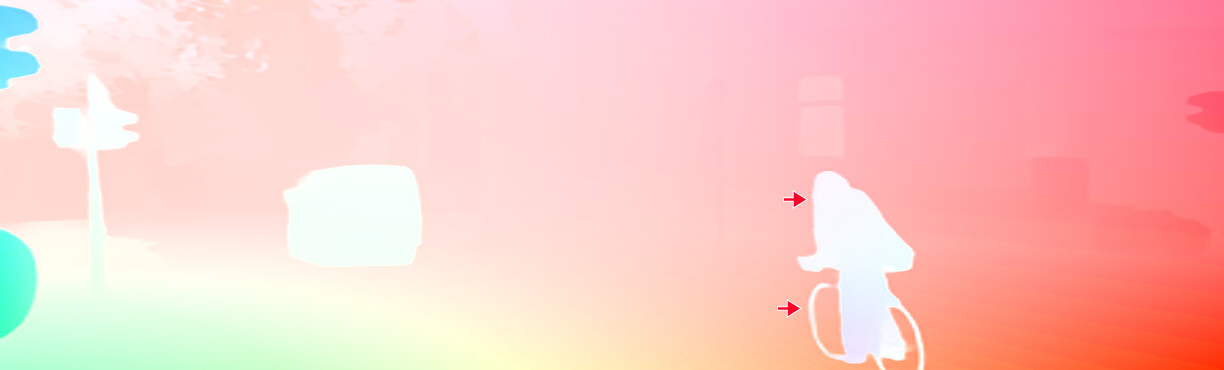} &
    \includegraphics[trim={400 0 100 40},clip,width=\Figwidth]{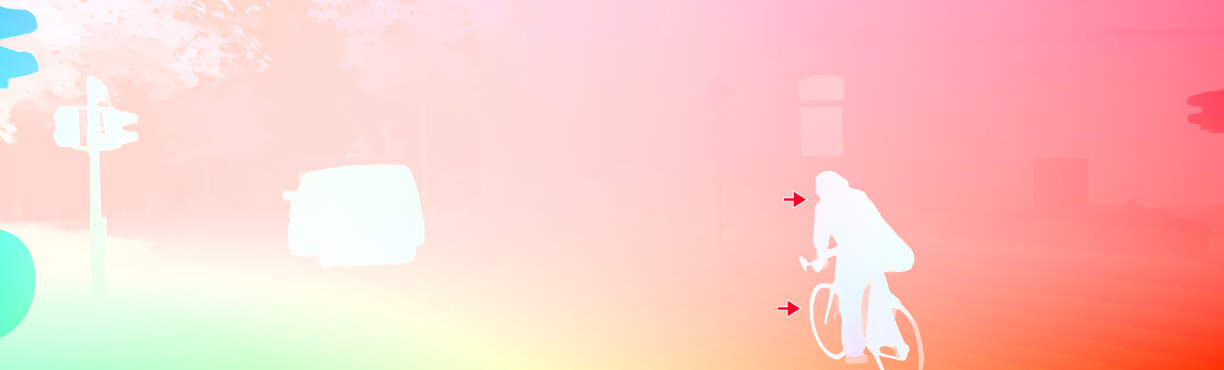} \\
    \includegraphics[trim={200 0 100 100},clip,width=\Figwidth]{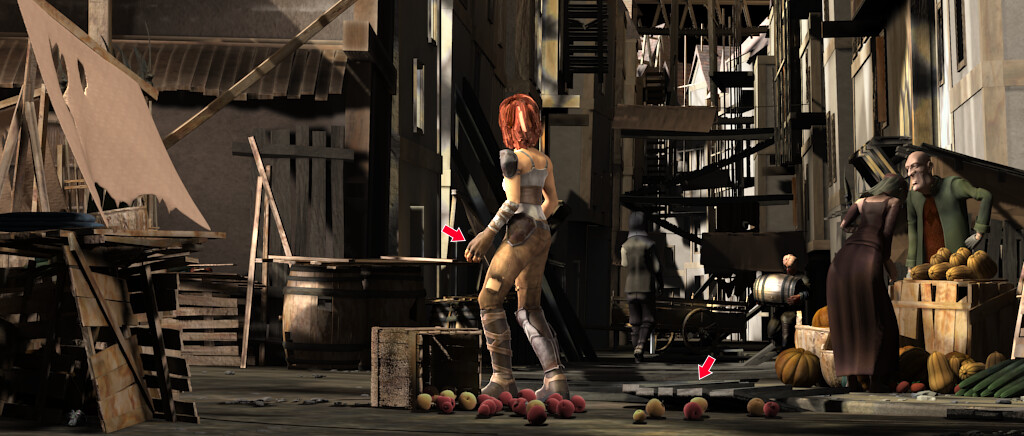} &
    \includegraphics[trim={200 0 100 100},clip,width=\Figwidth]{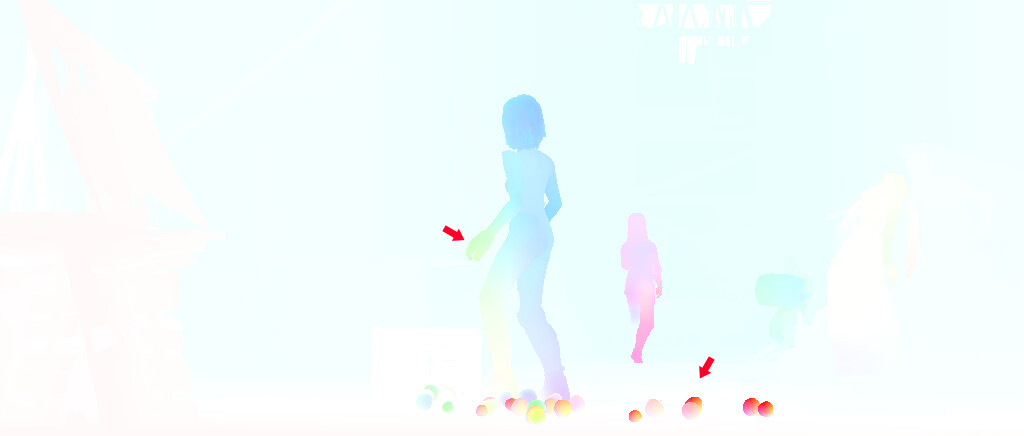} &
    \includegraphics[trim={200 0 100 100},clip,width=\Figwidth]{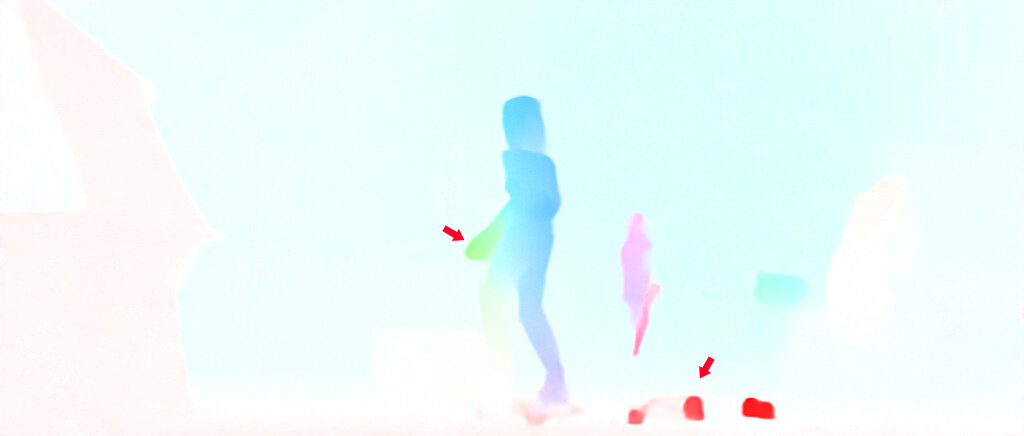} &
    \includegraphics[trim={200 0 100 100},clip,width=\Figwidth]{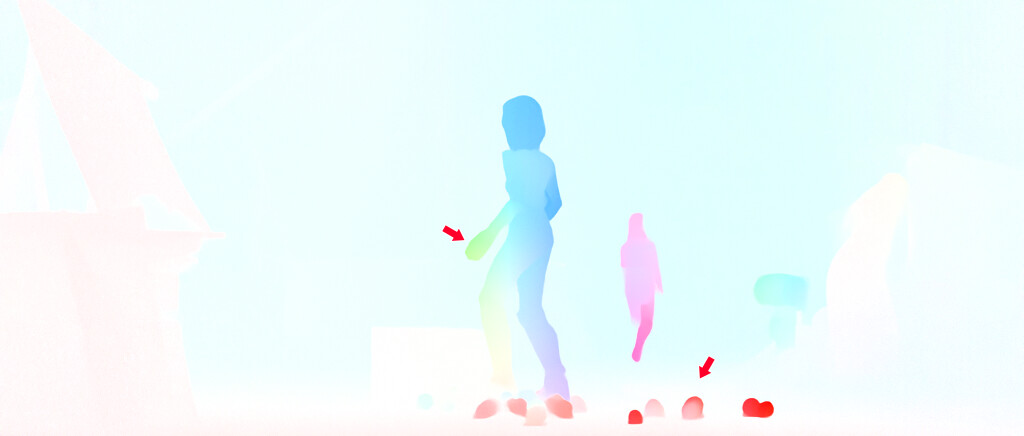} \\
    \end{tabular}
        % \vspace*{-0.1cm}
	\caption{\textbf{Visual results with and without coarse-to-fine refinement}. For our pretrained model, refinement helps correct wrong flow and adds details to correct flow.}
    \label{fig:flow_tiled_ablation}
        % \vspace*{-0.1cm}
\end{figure*}

\myparagraph{Coarse-to-fine refinement.}
Figure \ref{fig:flow_tiled_ablation} shows that coarse-to-fine refinement (Section \ref{sec:tiled-refinement}) substantially improves fine-grained details in estimated optical flow fields.
It also improves the metrics for zero-shot optical flow estimation on both KITTI and Sintel, as shown in Table \ref{tab::tiled_refinement_ablation}.

\myparagraph{Datasets.}
When using different mixtures of datasets for pretraining, we find that diffusion models sometimes capture
region boundaries and shape at the expense of local textural variation (eg see Figure~\ref{fig:flow_dataset_zeroshot}).
% When using a different mixture of datasets for pretraining, we find that the interesting property of the diffusion model -- which relies heavily on shape-based features instead of textures -- also holds for optical flow, as shown in Figure~\ref{fig:flow_dataset_zeroshot}.
The model trained solely on AutoFlow tends to provide very coarse flow, and mimics the object shapes found in AutoFlow.
The addition of FlyingThings, Kubric, and TartanAir removes this hallucination and significantly improves the fine details in the flow estimates (eg, shadows, trees, thin structure, and motion boundaries) together with a substantial boost in  accuracy (\cf Table \ref{tab:flow-dataset-ablation}).
Similarly, we find that mixing SceneNet RGB-D \cite{scenenet-rgbd}, a synthetic dataset, along with ScanNet \cite{scannet} provides a performance boost for fine-tuning results on NYU depth v2, shown in Table \ref{tab:nyu-dataset-ablation}.

\vspace*{-0.1cm}
\subsection{Interesting properties of diffusion models}
\vspace*{-0.1cm}

\myparagraph{Multimodality.}
One strength of diffusion models is their ability to capture complex multimodal distributions. This can be effective in representing uncertainty, especially where there may exist natural ambiguities and thus multiple predictions, \eg~in cases of transparent, translucent, or reflective cases.
Figure~\ref{fig:multi_modal_samples} presents multiple samples on the NYU, KITTI, and Sintel datasets, showing that our model captures multimodality and provides plausible samples when ambiguities exist. More details and examples are available in Section \ref{sec:supp-multimodal}.

\begin{figure}[t]
    \centering
    %   \vspace*{0.1cm}
    \includegraphics[width=0.994\textwidth]{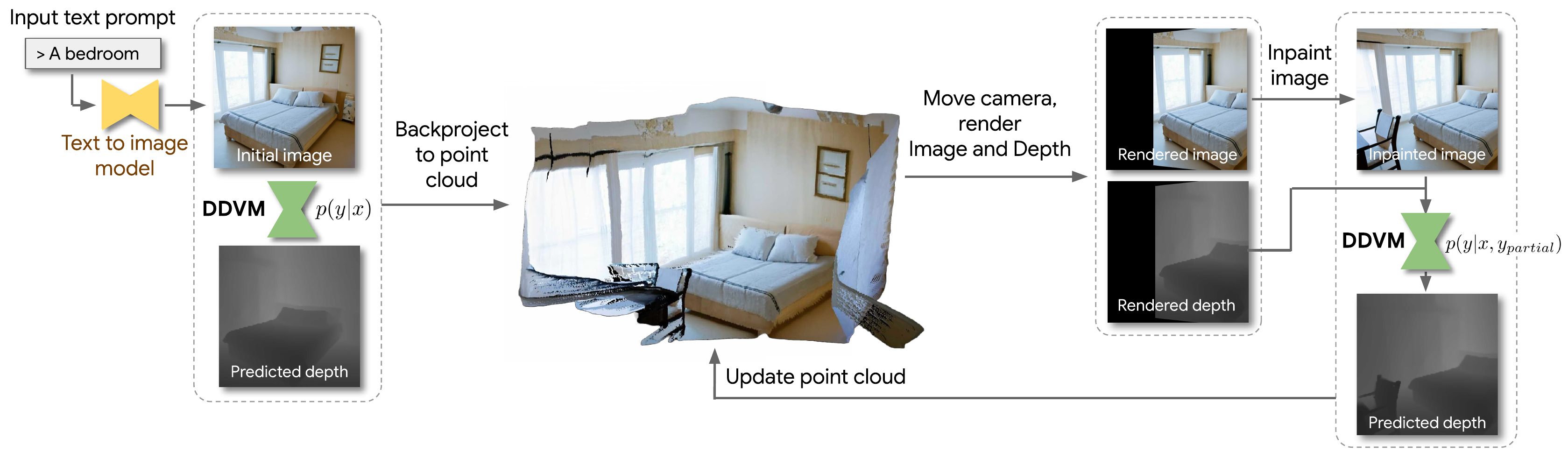}
    \vspace*{-0.05cm}
    \caption{\textbf{Application of zero-shot depth completion} with our model by incorporating it into an iterative 3D scene generation pipeline. Starting with a initial image (optionally generated from a text-to-image model), we sample an image-only conditioned depth map using our model. The image-depth pair is added to a point cloud. We then iteratively render images and depth maps (with holes) from this point cloud by moving the camera.
    We then fill image holes using an existing image inpainter (optionally text conditioned), and then use our model with replacement guidance to impute missing depths (conditioned on the filled RGB image and known depth).}
    \label{fig:text_to_3d}
    \vspace*{-0.1cm}
\end{figure}

\myparagraph{Imputation of missing labels.}
A diffusion model trained to model the conditional distribution $p(y|x)$ can be zero-shot leveraged to sample from $p(y|x,y_{partial})$ where $y_{partial}$ is the partially known label. One approach for doing this, known as the \textit{replacement method} for conditional inference \cite{song2021scorebased}, is to replace the known portion the latent $y_t$ at each inference step with the noisy latent built by applying the forward process to the known label. We qualitatively study the results of leveraging replacement guidance for depth completion and find it to be surprisingly effective. We illustrate this by building a pipeline for iteratively generating 3D scenes (conditioned on a text prompt) as shown in Figure \ref{fig:text_to_3d} by leveraging existing models for text-to-image generation and text-conditional image inpainting. While a more thorough evaluation of depth completion and novel view synthesis against existing methods is warranted, we leave that exploration to future work. (See Section \ref{sec:supp-imputation} for more details and examples.)

\section{Limitations}

\label{sec:limitations}
\myparagraph{Latency.} We adopt standard practices from image-generation models, leading to larger models and slower running times than RAFT. However, we are excited by the recent progress on progressive distillation~\cite{meng2022on,salimans2022progressive} and consistency models~\cite{song2023consistency} to improve inference speed in diffusion models. 

\myparagraph{Sintel fine-tuning.} Under the zero-shot setting, our method achieves state-of-the-art results on both Sintel Final and KITTI. Under the fine-tuning setting, ours is state-of-the-art on KITTI but is behind FlowFormer \cite{huang2022flowformer} on Sintel Final. We discuss several possible reasons for why this may be the case.

\begin{itemize}[leftmargin=*]
    \item We follow the fine-tuning procedure in~\cite{sun2022disentangling}. While their zero-shot RAFT results are comparable to FlowFormer on Sintel and KITTI, the fine-tuned RAFT-it is significantly better on KITTI but less accurate on Sintel than FlowFormer. It is possible that the fine-tuning procedure (\eg dataset mixture or augmentations) developed in~\cite{sun2022disentangling} is more suited for KITTI than Sintel.
    \item Another possible reason is that there is substantial domain gap between the training and test data on Sintel than KITTI. On Sintel test, there is a particular sequence ``Ambush 1'', where the girl's right arm moves out of the image boundary. Our method has an AEPE close to 30 while FlowFormer has lower than 10. It is likely that the attention on the cost volume mechanism by FlowFormer can better reason about the motion globally and handles this particular sequence well.  This particular sequence may account for the major difference in the overall results; among 12 available results on the Sintel website, ours has lower AEPE on 11 sequences but a higher AEPE on the  ``Ambush 1'' sequence, as shown in Table~\ref{tab:sintel:finetuning}. Figure~\ref{fig:supp:sintel_finetuning} in the appendix further provides visualization.
\end{itemize}

\vspace{-0.4cm}

\begin{table}[h!]
    \centering
    \scriptsize
    \caption{\textbf{Average end-point error (AEPE) on 12 Sintel test sequences} available from the public website.}
    \vspace*{0.1cm}
    \begin{tabular}{l @{\hskip 3em} c c}
        \toprule
        Sequence & Ours & FlowFormer \cite{huang2022flowformer}\\
        \midrule
Perturbed Market 3	&\textbf{0.787}	& 0.869\\
Perturbed Shaman 1	&\textbf{0.219}	& 0.252\\
Ambush 1	& {\color{red} 29.33}	& \textbf{8.141}\\
Ambush 3	& \textbf{2.855}	& 2.973\\
Bamboo 3	& \textbf{0.415}	& 0.577\\
Cave 3	& \textbf{2.042}	& 2.352\\
Market 1	& \textbf{0.719}	& 1.174\\
Market 4	& \textbf{5.517}	& 8.768\\
Mountain 2	& \textbf{0.176}	& 0.518\\
Temple 1	& \textbf{0.452}	& 0.612\\
Tiger	& \textbf{0.413}	& 0.596\\
Wall	& \textbf{1.639}	& 1.723\\
        \bottomrule
    \end{tabular} 
    % \vspace*{-0.5cm}
    \label{tab:sintel:finetuning}
\end{table}

\vspace*{-0.15cm}
\section{Conclusion}
\vspace*{-0.25cm}

We introduced a simple denoising diffusion model for monocular depth and optical flow estimation using an image-to-image translation framework. Our generative approach obtains state-of-the-art results without task-specific architectures or loss functions. In particular, our model achieves an Fl-all score of 3.26\% on KITTI, about 25\% better than the best published method~\cite{sun2022disentangling}.
Further, our model captures the multi-modality and uncertainty through multiple samples from the posterior. It also allows imputation of missing values, which enables iterative generation of 3D scenes conditioned on a text prompt. Our work suggests that diffusion models could be a simple and generic framework for dense vision tasks, and we hope to see more work in this direction.

\subsection*{Acknowledgements} 
We thank Ting Chen, Daniel Watson, Hugo Larochelle and the rest of Google DeepMind  for feedback on this work.
Thanks to Klaus Greff and Andrea Tagliasacchi for their help with the Kubric generator, and to Chitwan Saharia for help training the Palette model. 
\vspace*{0.1cm}

\nocite{chen2023generalist}

\bibliographystyle{plainnat}
\bibliography{main}

\begin{thebibliography}{88}
\providecommand{\natexlab}[1]{#1}
\providecommand{\url}[1]{\texttt{#1}}
\expandafter\ifx\csname urlstyle\endcsname\relax
  \providecommand{\doi}[1]{doi: #1}\else
  \providecommand{\doi}{doi: \begingroup \urlstyle{rm}\Url}\fi

\bibitem[Agarwal and Arora(2023)]{pixelformer}
Ashutosh Agarwal and Chetan Arora.
\newblock Attention {A}ttention {E}verywhere: {M}onocular depth prediction with
  skip attention.
\newblock In \emph{WACV}, 2023.

\bibitem[Bengio et~al.(2015)Bengio, Vinyals, Jaitly, and
  Shazeer]{Bengio-2015-scheduled-sampling}
Samy Bengio, Oriol Vinyals, Navdeep Jaitly, and Noam Shazeer.
\newblock Scheduled sampling for sequence prediction with recurrent neural
  networks.
\newblock \emph{NIPS}, 2015.

\bibitem[Bhat et~al.(2021)Bhat, Alhashim, and Wonka]{adabins}
Shariq~Farooq Bhat, Ibraheem Alhashim, and Peter Wonka.
\newblock Ada{B}ins: {D}epth estimation using adaptive bins.
\newblock In \emph{CVPR}, pages 4009--4018, 2021.

\bibitem[Butler et~al.(2012)Butler, Wulff, Stanley, and
  Black]{Butler:ECCV:2012}
Daniel~J. Butler, Jonas Wulff, Garrett~B. Stanley, and Michael~J. Black.
\newblock A naturalistic open source movie for optical flow evaluation.
\newblock In \emph{ECCV}, pages 611--625, 2012.

\bibitem[Cao et~al.(2017)Cao, Wu, and Shen]{cao2016estimating}
Yuanzhouhan Cao, Zifeng Wu, and Chunhua Shen.
\newblock Estimating depth from monocular images as classification using deep
  fully convolutional residual networks.
\newblock \emph{IEEE T-CSVT}, 28\penalty0 (11):\penalty0 3174--3182, 2017.

\bibitem[Chang et~al.(2015)Chang, Funkhouser, Guibas, Hanrahan, Huang, Li,
  Savarese, Savva, Song, Su, Xiao, Yi, and Yu]{shapenet}
Angel~X. Chang, Thomas Funkhouser, Leonidas Guibas, Pat Hanrahan, Qixing Huang,
  Zimo Li, Silvio Savarese, Manolis Savva, Shuran Song, Hao Su, Jianxiong Xiao,
  Li~Yi, and Fisher Yu.
\newblock Shape{N}et: {A}n information-rich 3{D} model repository.
\newblock \emph{arXiv:1512.03012}, 2015.

\bibitem[Chen et~al.(2023{\natexlab{a}})Chen, Li, Saxena, Hinton, and
  Fleet]{chen2023generalist}
Ting Chen, Lala Li, Saurabh Saxena, Geoffrey Hinton, and David~J. Fleet.
\newblock A generalist framework for panoptic segmentation of images and
  videos.
\newblock In \emph{ICCV}, 2023{\natexlab{a}}.

\bibitem[Chen et~al.(2023{\natexlab{b}})Chen, Zhang, and
  Hinton]{chen2023analog}
Ting Chen, Ruixiang Zhang, and Geoffrey Hinton.
\newblock Analog bits: Generating discrete data using diffusion models with
  self-conditioning.
\newblock In \emph{ICLR}, 2023{\natexlab{b}}.

\bibitem[Dai et~al.(2017)Dai, Chang, Savva, Halber, Funkhouser, and
  Nie{\ss}ner]{scannet}
Angela Dai, Angel~X. Chang, Manolis Savva, Maciej Halber, Thomas Funkhouser,
  and Matthias Nie{\ss}ner.
\newblock Scan{N}et: {R}ichly-annotated 3{D} reconstructions of indoor scenes.
\newblock In \emph{CVPR}, 2017.

\bibitem[Deng et~al.(2009)Deng, Dong, Socher, Li, Li, and
  Fei-Fei]{imagenet_cvpr09}
Jia Deng, Wei Dong, Richard Socher, Li-Jia Li, Kai Li, and Li~Fei-Fei.
\newblock Image{N}et: {A} large-scale hierarchical image database.
\newblock In \emph{CVPR}, pages 248--255, 2009.

\bibitem[Dhariwal and Nichol(2022)]{dhariwal2021diffusion}
Prafulla Dhariwal and Alex Nichol.
\newblock Diffusion models beat {GAN}s on image synthesis.
\newblock In \emph{{NeurIPS}}, 2022.

\bibitem[Eigen and Fergus(2015)]{eigen2014predicting}
David Eigen and Rob Fergus.
\newblock Predicting depth, surface normals and semantic labels with a common
  multi-scale convolutional architecture.
\newblock In \emph{ICCV}, pages 2650--2658, 2015.

\bibitem[Eigen et~al.(2014)Eigen, Puhrsch, and Fergus]{eigen2014depth}
David Eigen, Christian Puhrsch, and Rob Fergus.
\newblock Depth map prediction from a single image using a multi-scale deep
  network.
\newblock In \emph{NIPS}, volume~27, 2014.

\bibitem[Fischer et~al.(2015)Fischer, Dosovitskiy, Ilg, H{\"{a}}usser,
  Hazirbas, Golkov, van~der Smagt, Cremers, and Brox]{dosovitskiy2015flownet}
Philipp Fischer, Alexey Dosovitskiy, Eddy Ilg, Philip H{\"{a}}usser, Caner
  Hazirbas, Vladimir Golkov, Patrick van~der Smagt, Daniel Cremers, and Thomas
  Brox.
\newblock Flow{N}et: {L}earning optical flow with convolutional networks.
\newblock In \emph{ICCV}, pages 2758--2766, 2015.

\bibitem[Fu et~al.(2018)Fu, Gong, Wang, Batmanghelich, and Tao]{dorn}
Huan Fu, Mingming Gong, Chaohui Wang, Kayhan Batmanghelich, and Dacheng Tao.
\newblock Deep ordinal regression network for monocular depth estimation.
\newblock In \emph{CVPR}, pages 2002--2011, 2018.

\bibitem[Garg et~al.(2016)Garg, Bg, Carneiro, and Reid]{garg2016unsupervised}
Ravi Garg, Vijay~Kumar Bg, Gustavo Carneiro, and Ian Reid.
\newblock Unsupervised {CNN} for single view depth estimation: {G}eometry to
  the rescue.
\newblock In \emph{ECCV}, pages 740--756, 2016.

\bibitem[Geiger et~al.(2013)Geiger, Lenz, Stiller, and Urtasun]{kitti}
Andreas Geiger, Philip Lenz, Christoph Stiller, and Raquel Urtasun.
\newblock Vision meets {R}obotics: The {KITTI} dataset.
\newblock \emph{The International Journal of Robotics Research}, 32\penalty0
  (11):\penalty0 1231--1237, 2013.

\bibitem[Godard et~al.(2019)Godard, Mac~Aodha, Firman, and
  Brostow]{godard2019digging}
Cl{\'e}ment Godard, Oisin Mac~Aodha, Michael Firman, and Gabriel~J. Brostow.
\newblock Digging into self-supervised monocular depth estimation.
\newblock In \emph{ICCV}, pages 3828--3838, 2019.

\bibitem[Greff et~al.(2022)Greff, Belletti, Beyer, Doersch, Du, Duckworth,
  Fleet, Gnanapragasam, Golemo, Herrmann, Kipf, Kundu, Lagun, Laradji, Liu,
  Meyer, Miao, Nowrouzezahrai, Oztireli, Pot, Radwan, Rebain, Sabour, Sajjadi,
  Sela, Sitzmann, Stone, Sun, Vora, Wang, Wu, Yi, Zhong, and
  Tagliasacchi]{Greff_2022_CVPR}
Klaus Greff, Francois Belletti, Lucas Beyer, Carl Doersch, Yilun Du, Daniel
  Duckworth, David~J. Fleet, Dan Gnanapragasam, Florian Golemo, Charles
  Herrmann, Thomas Kipf, Abhijit Kundu, Dmitry Lagun, Issam Laradji,
  Hsueh-Ti~(Derek) Liu, Henning Meyer, Yishu Miao, Derek Nowrouzezahrai, Cengiz
  Oztireli, Etienne Pot, Noha Radwan, Daniel Rebain, Sara Sabour, Mehdi S.~M.
  Sajjadi, Matan Sela, Vincent Sitzmann, Austin Stone, Deqing Sun, Suhani Vora,
  Ziyu Wang, Tianhao Wu, Kwang~Moo Yi, Fangcheng Zhong, and Andrea
  Tagliasacchi.
\newblock Kubric: A scalable dataset generator.
\newblock In \emph{CVPR}, pages 3749--3761, June 2022.

\bibitem[Handa et~al.(2016)Handa, Patraucean, Badrinarayanan, Stent, and
  Cipolla]{https://doi.org/10.48550/arxiv.1511.07041}
Ankur Handa, Viorica Patraucean, Vijay Badrinarayanan, Simon Stent, and Roberto
  Cipolla.
\newblock Understanding real world indoor scenes with synthetic data.
\newblock In \emph{CVPR}, pages 4077--4085, 2016.

\bibitem[Ho et~al.(2020)Ho, Jain, and Abbeel]{ho2020denoising}
Jonathan Ho, Ajay Jain, and Pieter Abbeel.
\newblock {Denoising Diffusion Probabilistic Models}.
\newblock \emph{{NeurIPS}}, 2020.

\bibitem[Hosni et~al.(2012)Hosni, Rhemann, Bleyer, Rother, and
  Gelautz]{hosni2012fast}
Asmaa Hosni, Christoph Rhemann, Michael Bleyer, Carsten Rother, and Margrit
  Gelautz.
\newblock Fast cost-volume filtering for visual correspondence and beyond.
\newblock \emph{IEEE T-PAMI}, 35\penalty0 (2):\penalty0 504--511, 2012.

\bibitem[Huang et~al.(2022)Huang, Shi, Zhang, Wang, Cheung, Qin, Dai, and
  Li]{huang2022flowformer}
Zhaoyang Huang, Xiaoyu Shi, Chao Zhang, Qiang Wang, Ka~Chun Cheung, Hongwei
  Qin, Jifeng Dai, and Hongsheng Li.
\newblock {FlowFormer}: {A} transformer architecture for optical flow.
\newblock In \emph{ECCV}, pages 668--685, 2022.

\bibitem[Hur and Roth(2017)]{hur2017mirrorflow}
Junhwa Hur and Stefan Roth.
\newblock Mirror{F}low: {E}xploiting symmetries in joint optical flow and
  occlusion estimation.
\newblock In \emph{ICCV}, pages 312--321, 2017.

\bibitem[Hur and Roth(2019)]{hur2019iterative}
Junhwa Hur and Stefan Roth.
\newblock Iterative residual refinement for joint optical flow and occlusion
  estimation.
\newblock In \emph{CVPR}, pages 5754--5763, 2019.

\bibitem[Ilg et~al.(2017)Ilg, Mayer, Saikia, Keuper, Dosovitskiy, and
  Brox]{ilg2017flownet}
Eddy Ilg, Nikolaus Mayer, Tonmoy Saikia, Margret Keuper, Alexey Dosovitskiy,
  and Thomas Brox.
\newblock Flow{N}et 2.0: {E}volution of optical flow estimation with deep
  networks.
\newblock In \emph{CVPR}, pages 2462--2470, 2017.

\bibitem[Ince and Konrad(2008)]{ince2008occlusion}
Serdar Ince and Janusz Konrad.
\newblock Occlusion-aware optical flow estimation.
\newblock \emph{IEEE T-IP}, 17\penalty0 (8):\penalty0 1443--1451, 2008.

\bibitem[Jaegle et~al.(2022)Jaegle, Borgeaud, Alayrac, Doersch, Ionescu, Ding,
  Koppula, Zoran, Brock, Shelhamer, et~al.]{jaegleperceiver}
Andrew Jaegle, Sebastian Borgeaud, Jean-Baptiste Alayrac, Carl Doersch, Catalin
  Ionescu, David Ding, Skanda Koppula, Daniel Zoran, Andrew Brock, Evan
  Shelhamer, et~al.
\newblock Perceiver {IO}: {A} general architecture for structured inputs \&
  outputs.
\newblock In \emph{ICLR}, 2022.

\bibitem[Jeong et~al.(2022)Jeong, Lin, Porikli, and Kwak]{jeong2022imposing}
Jisoo Jeong, Jamie Lin, Fatih Porikli, and Nojun Kwak.
\newblock Imposing consistency for optical flow estimation.
\newblock In \emph{CVPR}, 2022.

\bibitem[Kondermann et~al.(2016)Kondermann, Nair, Honauer, Krispin, Andrulis,
  Brock, Gussefeld, Rahimimoghaddam, Hofmann, Brenner,
  et~al.]{kondermann2016hci}
Daniel Kondermann, Rahul Nair, Katrin Honauer, Karsten Krispin, Jonas Andrulis,
  Alexander Brock, Burkhard Gussefeld, Mohsen Rahimimoghaddam, Sabine Hofmann,
  Claus Brenner, et~al.
\newblock The {HCI} {B}enchmark {S}uite: {S}tereo and flow ground truth with
  uncertainties for urban autonomous driving.
\newblock In \emph{CVPR Workshops}, pages 19--28, 2016.

\bibitem[Laina et~al.(2016)Laina, Rupprecht, Belagiannis, Tombari, and
  Navab]{laina2016deeper}
Iro Laina, Christian Rupprecht, Vasileios Belagiannis, Federico Tombari, and
  Nassir Navab.
\newblock Deeper depth prediction with fully convolutional residual networks.
\newblock In \emph{3DV}, pages 239--248, 2016.

\bibitem[Lamb et~al.(2016)Lamb, Goyal, Zhang, Zhang, Courville, and
  Bengio]{lamb2016professor}
Alex Lamb, Anirudh Goyal, Ying Zhang, Saizheng Zhang, Aaron Courville, and
  Yoshua Bengio.
\newblock Professor {F}orcing: {A} new algorithm for training recurrent
  networks.
\newblock \emph{NIPS}, 29, 2016.

\bibitem[Larsson et~al.(2016)Larsson, Maire, and
  Shakhnarovich]{Shakhnarovich2016}
Gustav Larsson, Michael Maire, and Gregory Shakhnarovich.
\newblock Learning representations for automatic colorization.
\newblock In \emph{ECCV}, pages 577--593, 2016.

\bibitem[Lee et~al.(2019)Lee, Han, Ko, and Suh]{bts}
Jin~Han Lee, Myung-Kyu Han, Dong~Wook Ko, and Il~Hong Suh.
\newblock From big to small: Multi-scale local planar guidance for monocular
  depth estimation.
\newblock \emph{arXiv:1907.10326}, 2019.

\bibitem[Li et~al.(2022)Li, Wang, Liu, and Jiang]{binsformer}
Zhenyu Li, Xuyang Wang, Xianming Liu, and Junjun Jiang.
\newblock {BinsFormer}: {R}evisiting adaptive bins for monocular depth
  estimation.
\newblock \emph{arxiv.2204.00987}, 2022.

\bibitem[Liba et~al.(2020)Liba, Cai, Tsai, Eban, Movshovitz-Attias, Pritch,
  Chen, and Barron]{Liba_2020_CVPR_Workshops}
Orly Liba, Longqi Cai, Yun-Ta Tsai, Elad Eban, Yair Movshovitz-Attias, Yael
  Pritch, Huizhong Chen, and Jonathan~T. Barron.
\newblock Sky {O}ptimization: Semantically aware image processing of skies in
  low-light photography.
\newblock In \emph{CVPR Workshops}, June 2020.

\bibitem[Liu et~al.(2021)Liu, Tucker, Jampani, Makadia, Snavely, and
  Kanazawa]{infinite_nature_2020}
Andrew Liu, Richard Tucker, Varun Jampani, Ameesh Makadia, Noah Snavely, and
  Angjoo Kanazawa.
\newblock Infinite {N}ature: {P}erpetual view generation of natural scenes from
  a single image.
\newblock In \emph{ICCV}, 2021.

\bibitem[Luo et~al.(2022)Luo, Yang, Luo, Li, Fan, and Liu]{luo2022learning}
Ao~Luo, Fan Yang, Kunming Luo, Xin Li, Haoqiang Fan, and Shuaicheng Liu.
\newblock Learning optical flow with adaptive graph reasoning.
\newblock In \emph{AAAI}, pages 1890--1898, 2022.

\bibitem[Lv et~al.(2018)Lv, Kim, Troccoli, Sun, Rehg, and
  Kautz]{lv2018learning}
Zhaoyang Lv, Kihwan Kim, Alejandro Troccoli, Deqing Sun, James~M. Rehg, and Jan
  Kautz.
\newblock Learning rigidity in dynamic scenes with a moving camera for 3{D}
  motion field estimation.
\newblock In \emph{ECCV}, pages 468--484, 2018.

\bibitem[Mayer et~al.(2016)Mayer, Ilg, Hausser, Fischer, Cremers, Dosovitskiy,
  and Brox]{Mayer_2016_CVPR}
Nikolaus Mayer, Eddy Ilg, Philip Hausser, Philipp Fischer, Daniel Cremers,
  Alexey Dosovitskiy, and Thomas Brox.
\newblock A large dataset to train convolutional networks for disparity,
  optical flow, and scene flow estimation.
\newblock In \emph{CVPR}, 2016.

\bibitem[McCormac et~al.(2017)McCormac, Handa, Leutenegger, and
  Davison]{scenenet-rgbd}
John McCormac, Ankur Handa, Stefan Leutenegger, and Andrew~J. Davison.
\newblock {SceneNet RGB-D}: Can 5{M} synthetic images beat generic imagenet
  pre-training on indoor segmentation?
\newblock In \emph{ICCV}, 2017.

\bibitem[Meng et~al.(2022)Meng, Gao, Kingma, Ermon, Ho, and
  Salimans]{meng2022on}
Chenlin Meng, Ruiqi Gao, Diederik~P. Kingma, Stefano Ermon, Jonathan Ho, and
  Tim Salimans.
\newblock On distillation of guided diffusion models.
\newblock In \emph{NeurIPS 2022 Workshop on Score-Based Methods}, 2022.

\bibitem[Menze et~al.(2015)Menze, Heipke, and Geiger]{Menze2015ISA}
Moritz Menze, Christian Heipke, and Andreas Geiger.
\newblock Joint 3{D} estimation of vehicles and scene flow.
\newblock In \emph{ISPRS Workshop on Image Sequence Analysis (ISA)}, 2015.

\bibitem[Menze et~al.(2018)Menze, Heipke, and Geiger]{Menze2018JPRS}
Moritz Menze, Christian Heipke, and Andreas Geiger.
\newblock Object scene flow.
\newblock \emph{ISPRS Journal of Photogrammetry and Remote Sensing (JPRS)},
  2018.

\bibitem[Nichol and Dhariwal(2021)]{nichol2021improved}
Alexander~Quinn Nichol and Prafulla Dhariwal.
\newblock Improved denoising diffusion probabilistic models.
\newblock In \emph{ICML}, pages 8162--8171, 2021.

\bibitem[Ning et~al.(2023)Ning, Li, Zhang, Geng, Dai, He, and Hu]{AllInTokens}
Jia Ning, Chen Li, Zheng Zhang, Zigang Geng, Qi~Dai, Kun He, and Han Hu.
\newblock {All in Tokens}: {U}nifying output space of visual tasks via soft
  token.
\newblock In \emph{ICCV}, 2023.

\bibitem[Perez et~al.(2018)Perez, Strub, de~Vries, Dumoulin, and
  Courville]{perez2018film}
Ethan Perez, Florian Strub, Harm de~Vries, Vincent Dumoulin, and Aaron~C.
  Courville.
\newblock Fi{L}m: {V}isual reasoning with a general conditioning layer.
\newblock In \emph{AAAI}, 2018.

\bibitem[Ramesh et~al.(2022)Ramesh, Dhariwal, Nichol, Chu, and
  Chen]{ramesh-dalle2}
Aditya Ramesh, Prafulla Dhariwal, Alex Nichol, Casey Chu, and Mark Chen.
\newblock Hierarchical text-conditional image generation with clip latents.
\newblock \emph{arXiv:2204.06125}, 2022.

\bibitem[Ranftl et~al.(2020)Ranftl, Lasinger, Hafner, Schindler, and
  Koltun]{ranftl2019robust}
Ren{\'e} Ranftl, Katrin Lasinger, David Hafner, Konrad Schindler, and Vladlen
  Koltun.
\newblock Towards robust monocular depth estimation: Mixing datasets for
  zero-shot cross-dataset transfer.
\newblock \emph{IEEE T-PAMI}, 44\penalty0 (3):\penalty0 1623--1637, 2020.

\bibitem[Ranftl et~al.(2021)Ranftl, Bochkovskiy, and Koltun]{dpt}
Ren\'e Ranftl, Alexey Bochkovskiy, and Vladlen Koltun.
\newblock Vision transformers for dense prediction.
\newblock In \emph{ICCV}, pages 12179--12188, 2021.

\bibitem[Ranzato et~al.(2016)Ranzato, Chopra, Auli, and
  Zaremba]{DBLP:journals/corr/RanzatoCAZ15}
Marc'Aurelio Ranzato, Sumit Chopra, Michael Auli, and Wojciech Zaremba.
\newblock Sequence level training with recurrent neural networks.
\newblock In \emph{ICLR}, 2016.

\bibitem[Ren and Lee(2018)]{ren2017crossdomain}
Zhongzheng Ren and Yong~Jae Lee.
\newblock Cross-{D}omain self-supervised multi-task feature learning using
  synthetic imagery.
\newblock In \emph{CVPR}, 2018.

\bibitem[Richter et~al.(2017)Richter, Hayder, and Koltun]{richter2017playing}
Stephan~R. Richter, Zeeshan Hayder, and Vladlen Koltun.
\newblock Playing for benchmarks.
\newblock In \emph{ICCV}, pages 2213--2222, 2017.

\bibitem[Saharia et~al.(2022{\natexlab{a}})Saharia, Chan, Chang, Lee, Ho,
  Salimans, Fleet, and Norouzi]{sahariac-palette}
Chitwan Saharia, William Chan, Huiwen Chang, Chris~A. Lee, Jonathan Ho, Tim
  Salimans, David~J. Fleet, and Mohammad Norouzi.
\newblock {Palette: Image-to-Image Diffusion Models}.
\newblock In \emph{SIGGRAPH}, 2022{\natexlab{a}}.

\bibitem[Saharia et~al.(2022{\natexlab{b}})Saharia, Chan, Saxena, Li, Whang,
  Denton, Ghasemipour, Ayan, Mahdavi, Lopes, Salimans, Ho, Fleet, and
  Norouzi]{sahariac-imagen}
Chitwan Saharia, William Chan, Saurabh Saxena, Lala Li, Jay Whang, Emily
  Denton, Seyed Kamyar~Seyed Ghasemipour, Burcu~Karagol Ayan, S.~Sara Mahdavi,
  Rapha~Gontijo Lopes, Tim Salimans, Jonathan Ho, David~J. Fleet, and Mohammad
  Norouzi.
\newblock {Photorealistic Text-to-Image Diffusion Models with Deep Language
  Understanding}.
\newblock In \emph{NeurIPS}, 2022{\natexlab{b}}.

\bibitem[Salimans and Ho(2022)]{salimans2022progressive}
Tim Salimans and Jonathan Ho.
\newblock Progressive distillation for fast sampling of diffusion models.
\newblock In \emph{ICLR}, 2022.

\bibitem[Savinov et~al.(2022)Savinov, Chung, Binkowski, Elsen, and van~den
  Oord]{savinov2022stepunrolled}
Nikolay Savinov, Junyoung Chung, Mikolaj Binkowski, Erich Elsen, and Aaron
  van~den Oord.
\newblock Step-unrolled denoising autoencoders for text generation.
\newblock In \emph{ICLR}, 2022.

\bibitem[Saxena et~al.(2005)Saxena, Chung, and Ng]{NIPS2005_17d8da81}
Ashutosh Saxena, Sung Chung, and Andrew Ng.
\newblock Learning depth from single monocular images.
\newblock \emph{NIPS}, 2005.

\bibitem[Saxena et~al.(2009)Saxena, Sun, and Ng]{4531745}
Ashutosh Saxena, Min Sun, and Andrew~Y. Ng.
\newblock Make3{D}: {L}earning 3{D} scene structure from a single still image.
\newblock \emph{IEEE T-PAMI}, 31\penalty0 (5):\penalty0 824--840, 2009.

\bibitem[Shih et~al.(2020)Shih, Su, Kopf, and Huang]{3dphotos}
Meng-Li Shih, Shih-Yang Su, Johannes Kopf, and Jia-Bin Huang.
\newblock 3{D} photography using context-aware layered depth inpainting.
\newblock In \emph{CVPR}, 2020.

\bibitem[Silberman et~al.(2012)Silberman, Hoiem, Kohli, and Fergus]{nyu}
Nathan Silberman, Derek Hoiem, Pushmeet Kohli, and Rob Fergus.
\newblock Indoor segmentation and support inference from {RGBD} images.
\newblock In \emph{ECCV}, pages 746--760, 2012.

\bibitem[Sohl-Dickstein et~al.(2015)Sohl-Dickstein, Weiss, Maheswaranathan, and
  Ganguli]{sohl2015deep}
Jascha Sohl-Dickstein, Eric Weiss, Niru Maheswaranathan, and Surya Ganguli.
\newblock Deep unsupervised learning using nonequilibrium thermodynamics.
\newblock In \emph{ICML}, pages 2256--2265, 2015.

\bibitem[Song et~al.(2021)Song, Sohl-Dickstein, Kingma, Kumar, Ermon, and
  Poole]{song2021scorebased}
Yang Song, Jascha Sohl-Dickstein, Diederik~P. Kingma, Abhishek Kumar, Stefano
  Ermon, and Ben Poole.
\newblock Score-based generative modeling through stochastic differential
  equations.
\newblock In \emph{ICLR}, 2021.

\bibitem[Song et~al.(2023)Song, Dhariwal, Chen, and
  Sutskever]{song2023consistency}
Yang Song, Prafulla Dhariwal, Mark Chen, and Ilya Sutskever.
\newblock Consistency models.
\newblock In \emph{ICML}, 2023.

\bibitem[Stommel et~al.(2014)Stommel, Beetz, and Xu]{6664993}
Martin Stommel, Michael Beetz, and Weiliang Xu.
\newblock Inpainting of missing values in the kinect sensor's depth maps based
  on background estimates.
\newblock \emph{IEEE Sensors Journal}, 14\penalty0 (4):\penalty0 1107--1116,
  2014.

\bibitem[Sui et~al.(2022)Sui, Li, Geng, Wu, Xu, Liu, Goh, and
  Zhu]{sui2022craft}
Xiuchao Sui, Shaohua Li, Xue Geng, Yan Wu, Xinxing Xu, Yong Liu, Rick Goh, and
  Hongyuan Zhu.
\newblock {CRAFT}: {C}ross-attentional flow transformer for robust optical
  flow.
\newblock In \emph{CVPR}, pages 17602--17611, 2022.

\bibitem[Sun et~al.(2012)Sun, Sudderth, and Black]{sun2012layered}
Deqing Sun, Erik~B. Sudderth, and Michael~J. Black.
\newblock Layered segmentation and optical flow estimation over time.
\newblock In \emph{CVPR}, pages 1768--1775, 2012.

\bibitem[Sun et~al.(2018)Sun, Yang, Liu, and Kautz]{sun2018pwc}
Deqing Sun, Xiaodong Yang, Ming-Yu Liu, and Jan Kautz.
\newblock {PWC}-{N}et: {CNN}s for optical flow using pyramid, warping, and cost
  volume.
\newblock In \emph{CVPR}, pages 8934--8943, 2018.

\bibitem[Sun et~al.(2021)Sun, Vlasic, Herrmann, Jampani, Krainin, Chang, Zabih,
  Freeman, and Liu]{sun2021autoflow}
Deqing Sun, Daniel Vlasic, Charles Herrmann, Varun Jampani, Michael Krainin,
  Huiwen Chang, Ramin Zabih, William~T. Freeman, and Ce~Liu.
\newblock Auto{F}low: {L}earning a better training set for optical flow.
\newblock In \emph{CVPR}, pages 10093--10102, 2021.

\bibitem[Sun et~al.(2022{\natexlab{a}})Sun, Herrmann, Reda, Rubinstein, Fleet,
  and Freeman]{sun2022disentangling}
Deqing Sun, Charles Herrmann, Fitsum Reda, Michael Rubinstein, David~J. Fleet,
  and William~T. Freeman.
\newblock Disentangling architecture and training for optical flow.
\newblock In \emph{ECCV}, pages 165--182, 2022{\natexlab{a}}.

\bibitem[Sun et~al.(2020)Sun, Kretzschmar, Dotiwalla, Chouard, Patnaik, Tsui,
  Guo, Zhou, Chai, Caine, Vasudevan, Han, Ngiam, Zhao, Timofeev, Ettinger,
  Krivokon, Gao, Joshi, Zhao, Cheng, Zhang, Shlens, Chen, and
  Anguelov]{waymoOpenDataset}
Pei Sun, Henrik Kretzschmar, Xerxes Dotiwalla, Aurelien Chouard, Vijaysai
  Patnaik, Paul Tsui, James Guo, Yin Zhou, Yuning Chai, Benjamin Caine, Vijay
  Vasudevan, Wei Han, Jiquan Ngiam, Hang Zhao, Aleksei Timofeev, Scott
  Ettinger, Maxim Krivokon, Amy Gao, Aditya Joshi, Sheng Zhao, Shuyang Cheng,
  Yu~Zhang, Jonathon Shlens, Zhifeng Chen, and Dragomir Anguelov.
\newblock Scalability in perception for autonomous driving: Waymo open dataset.
\newblock In \emph{CVPR}, 2020.

\bibitem[Sun et~al.(2022{\natexlab{b}})Sun, Chen, Zhu, Guo, and Li]{sunskflow}
Shangkun Sun, Yuanqi Chen, Yu~Zhu, Guodong Guo, and Ge~Li.
\newblock {SKFlow}: {L}earning optical flow with super kernels.
\newblock In \emph{NeurIPS}, 2022{\natexlab{b}}.

\bibitem[Tan and Le(2019)]{tan2019efficientnet}
Mingxing Tan and Quoc Le.
\newblock Efficient{N}et: {R}ethinking model scaling for convolutional neural
  networks.
\newblock In \emph{ICML}, pages 6105--6114, 2019.

\bibitem[Teed and Deng(2020)]{teed2020raft}
Zachary Teed and Jia Deng.
\newblock {RAFT}: {R}ecurrent all-pairs field transforms for optical flow.
\newblock In \emph{ECCV}, pages 402--419, 2020.

\bibitem[Wang et~al.(2023)Wang, Saharia, Montgomery, Pont-Tuset, Noy,
  Pellegrini, Onoe, Laszlo, Fleet, Soricut, Baldridge, Norouzi, Anderson, and
  Chan]{ImagenEditor}
Su~Wang, Chitwan Saharia, Ceslee Montgomery, Jordi Pont-Tuset, Shai Noy,
  Stefano Pellegrini, Yasumasa Onoe, Sarah Laszlo, David~J. Fleet, Radu
  Soricut, Jason Baldridge, Mohammad Norouzi, Peter Anderson, and William Chan.
\newblock {Imagen Editor and EditBench}: {A}dvancing and evaluating text-guided
  image inpainting.
\newblock In \emph{CVPR}, 2023.

\bibitem[Wang et~al.(2020)Wang, Zhu, Wang, Hu, Qiu, Wang, Hu, Kapoor, and
  Scherer]{tartanair2020iros}
Wenshan Wang, Delong Zhu, Xiangwei Wang, Yaoyu Hu, Yuheng Qiu, Chen Wang, Yafei
  Hu, Ashish Kapoor, and Sebastian Scherer.
\newblock {TartanAir}: {A} dataset to push the limits of visual {SLAM}.
\newblock In \emph{IROS}, 2020.

\bibitem[Weinzaepfel et~al.(2013)Weinzaepfel, Revaud, Harchaoui, and
  Schmid]{weinzaepfel2013deepflow}
Philippe Weinzaepfel, Jerome Revaud, Zaid Harchaoui, and Cordelia Schmid.
\newblock Deep{F}low: {L}arge displacement optical flow with deep matching.
\newblock In \emph{ICCV}, pages 1385--1392, 2013.

\bibitem[Wiles et~al.(2020)Wiles, Gkioxari, Szeliski, and
  Johnson]{wiles2020synsin}
Olivia Wiles, Georgia Gkioxari, Richard Szeliski, and Justin Johnson.
\newblock {SynSin}: {E}nd-to-end view synthesis from a single image.
\newblock In \emph{CVPR}, 2020.

\bibitem[Williams and Zipser(1989)]{6795228}
Ronald~J. Williams and David Zipser.
\newblock A learning algorithm for continually running fully recurrent neural
  networks.
\newblock \emph{Neural Computation}, 1\penalty0 (2):\penalty0 270--280, 1989.

\bibitem[Xie et~al.(2023)Xie, Geng, Hu, Zhang, Hu, and Cao]{xie2022revealing}
Zhenda Xie, Zigang Geng, Jingcheng Hu, Zheng Zhang, Han Hu, and Yue Cao.
\newblock Revealing the dark secrets of masked image modeling.
\newblock In \emph{CVPR}, 2023.

\bibitem[Xu et~al.(2022)Xu, Zhang, Cai, Rezatofighi, and Tao]{xu2022gmflow}
Haofei Xu, Jing Zhang, Jianfei Cai, Hamid Rezatofighi, and Dacheng Tao.
\newblock {GMFlow}: {L}earning optical flow via global matching.
\newblock In \emph{CVPR}, pages 8121--8130, 2022.

\bibitem[Xu et~al.(2011)Xu, Jia, and Matsushita]{xu2011motion}
Li~Xu, Jiaya Jia, and Yasuyuki Matsushita.
\newblock Motion detail preserving optical flow estimation.
\newblock \emph{IEEE T-PAMI}, 34\penalty0 (9):\penalty0 1744--1757, 2011.

\bibitem[Yang and Ramanan(2019)]{yang2019volumetric}
Gengshan Yang and Deva Ramanan.
\newblock Volumetric correspondence networks for optical flow.
\newblock \emph{NeurIPS}, 2019.

\bibitem[Yu et~al.(2017)Yu, Zhang, Wang, and Yu]{yu2016seqgan}
Lantao Yu, Weinan Zhang, Jun Wang, and Yong Yu.
\newblock Seq{GAN}: {S}equence generative adversarial nets with policy
  gradient.
\newblock In \emph{AAAI}, 2017.

\bibitem[Zhang et~al.(2021)Zhang, Woodford, Prisacariu, and
  Torr]{zhang2021separable}
Feihu Zhang, Oliver~J. Woodford, Victor~Adrian Prisacariu, and Philip~H.S.
  Torr.
\newblock Separable flow: Learning motion cost volumes for optical flow
  estimation.
\newblock In \emph{ICCV}, pages 10807--10817, 2021.

\bibitem[Zhang et~al.(2016)Zhang, Isola, and Efros]{ZhangColorization2016}
Richard Zhang, Phillip Isola, and Alexei~A. Efros.
\newblock Colorful image colorization.
\newblock In \emph{ECCV}, pages 649--666, 2016.

\bibitem[Zhao et~al.(2021)Zhao, Yan, Zhao, Guo, Huang, and Li]{transdepth}
Jiawei Zhao, Ke~Yan, Yifan Zhao, Xiaowei Guo, Feiyue Huang, and Jia Li.
\newblock Transformer-based dual relation graph for multi-label image
  recognition.
\newblock In \emph{ICCV}, pages 163--172, 2021.

\bibitem[Zhou et~al.(2017)Zhou, Lapedriza, Khosla, Oliva, and
  Torralba]{zhou2017places}
Bolei Zhou, Agata Lapedriza, Aditya Khosla, Aude Oliva, and Antonio Torralba.
\newblock Places: {A} 10 million image database for scene recognition.
\newblock \emph{IEEE T-PAMI}, 2017.

\end{thebibliography}

\clearpage
\appendix

\section{Multimodal prediction}
\label{sec:supp-multimodal}
\begin{figure*}[h]
    \centering
    \footnotesize
    \setlength\tabcolsep{0.8pt}
    \renewcommand{\arraystretch}{0.0}
    
    \newcommand{\Figwidth}{0.195\linewidth}
    \begin{tabular}{c @{\hskip 0.3em} c @{\hskip 0.3em} ccc}
    Input image & Variance heat map & \multicolumn{3}{c}{Multimodal prediction samples} \\

    \includegraphics[trim={0 0 0 0},clip,width=\Figwidth]{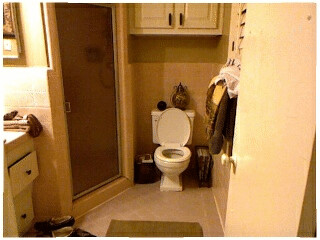} &
    \includegraphics[trim={0 0 0 0},clip,width=\Figwidth]{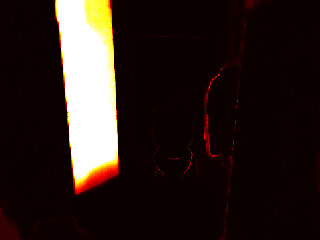} & 
    \includegraphics[trim={0 0 0 0},clip,width=\Figwidth]{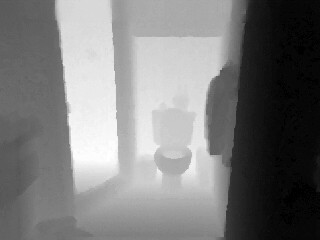} &  
    \includegraphics[trim={0 0 0 0},clip,width=\Figwidth]{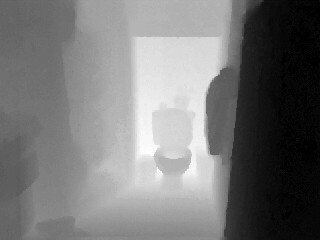} &  
    \includegraphics[trim={0 0 0 0},clip,width=\Figwidth]{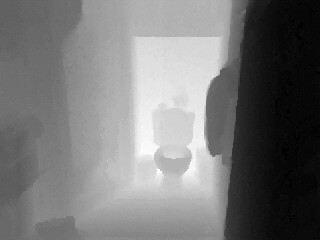}\\
    [0.2em] \\
    
    \includegraphics[trim={0 0 0 0},clip,width=\Figwidth]{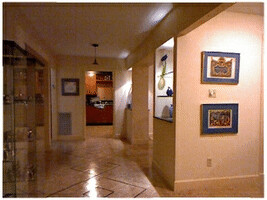} &
    \includegraphics[trim={0 0 0 0},clip,width=\Figwidth]{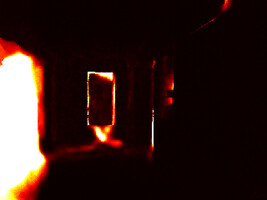} &
    \includegraphics[trim={0 0 0 0},clip,width=\Figwidth]{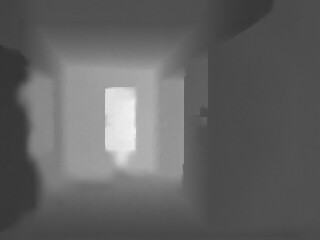} &  
    \includegraphics[trim={0 0 0 0},clip,width=\Figwidth]{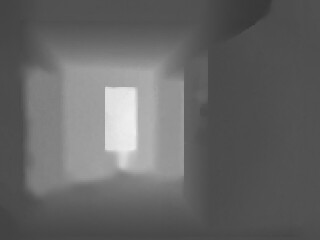} & 
    \includegraphics[trim={0 0 0 0},clip,width=\Figwidth]{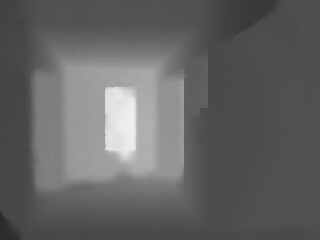}\\
    [0.2em] \\
    
    \includegraphics[trim={0 0 0 0},clip,width=\Figwidth]{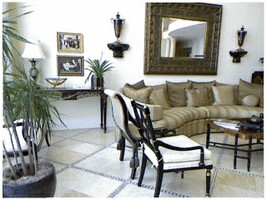} &
    \includegraphics[trim={0 0 0 0},clip,width=\Figwidth]{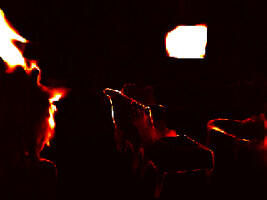} &
    \includegraphics[trim={0 0 0 0},clip,width=\Figwidth]{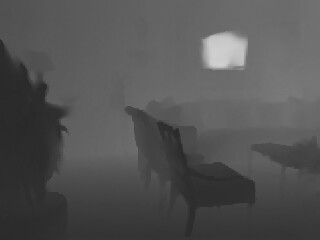} &  
    \includegraphics[trim={0 0 0 0},clip,width=\Figwidth]{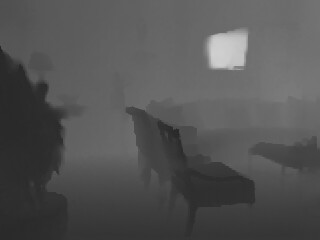} & 
    \includegraphics[trim={0 0 0 0},clip,width=\Figwidth]{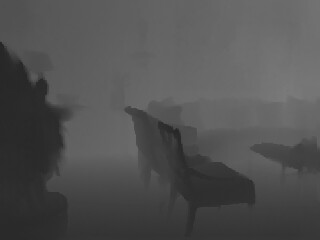} \\ [0.2em] \\
    
    \includegraphics[trim={0 0 0 0},clip,width=\Figwidth]{images/multi_modal_samples/depth_nyu/nyu_6_img.jpg} &
    \includegraphics[trim={0 0 0 0},clip,width=\Figwidth]{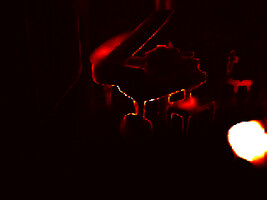} &
    \includegraphics[trim={0 0 0 0},clip,width=\Figwidth]{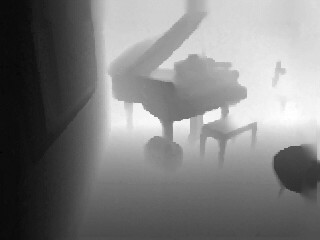} & 
    \includegraphics[trim={0 0 0 0},clip,width=\Figwidth]{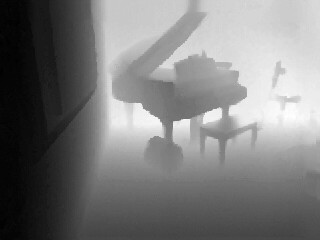} & 
    \includegraphics[trim={0 0 0 0},clip,width=\Figwidth]{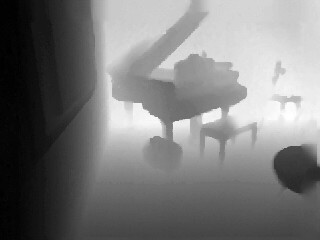}\\ [0.2em] \\
    
    \includegraphics[trim={0 0 0 0},clip,width=\Figwidth]{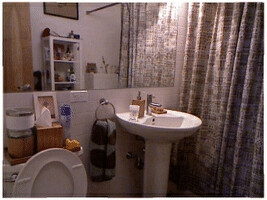} &
    \includegraphics[trim={0 0 0 0},clip,width=\Figwidth]{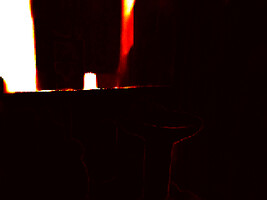} &
    \includegraphics[trim={0 0 0 0},clip,width=\Figwidth]{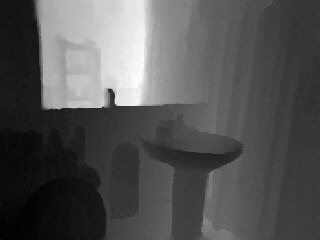} & 
    \includegraphics[trim={0 0 0 0},clip,width=\Figwidth]{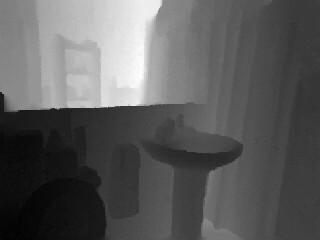} & 
    \includegraphics[trim={0 0 0 0},clip,width=\Figwidth]{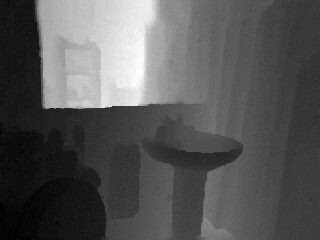}\\ [0.2em] \\

    \includegraphics[trim={0 0 0 0},clip,width=\Figwidth]{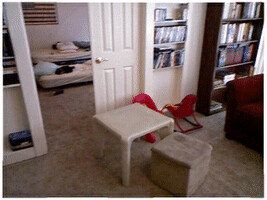} &
    \includegraphics[trim={0 0 0 0},clip,width=\Figwidth]{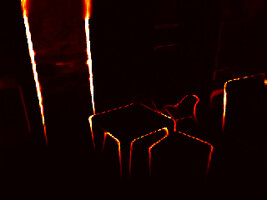} &
    \includegraphics[trim={0 0 0 0},clip,width=\Figwidth]{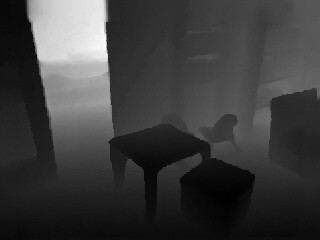} & 
    \includegraphics[trim={0 0 0 0},clip,width=\Figwidth]{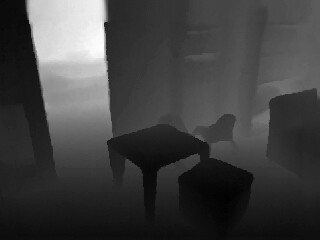} & 
    \includegraphics[trim={0 0 0 0},clip,width=\Figwidth]{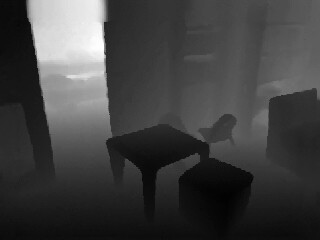}\\
    
    \end{tabular}
    \vspace*{0.2cm}
	\caption{\textbf{Qualitative examples of multimodal estimation on the NYU depth dataset}. Our model is able to output multiple plausible depth maps where ambiguity exists. Rows \emph{1 to 5} show transparent or reflective surfaces where two answers exist. In all samples (specially \emph{the last row}) we observe the model's ability to capture uncertainty in depth near object boundaries (see the areas of high variance).}
	   %\vspace*{0.4cm}
	   %\vspace{-1cm}
    \label{fig:supp:multi_modal_samples_depth_NYU}
\end{figure*}

\begin{figure*}[h]
    \centering
    \vspace*{0.25cm}
    \footnotesize
    \setlength\tabcolsep{1pt} % default value: 6pt
    \newcommand{\Figwidth}{0.196\linewidth}
    \begin{tabular}{c @{\hskip 0.2em} c @{\hskip 0.2em} ccc}
    Input image & Variance heat map & \multicolumn{3}{c}{Multimodal prediction samples} \\

    \includegraphics[trim={450 0 0 50},clip,width=\Figwidth]{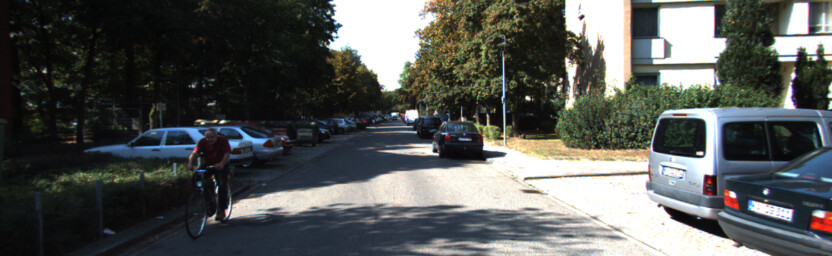} &
    \includegraphics[trim={450 0 0 50},clip,width=\Figwidth]{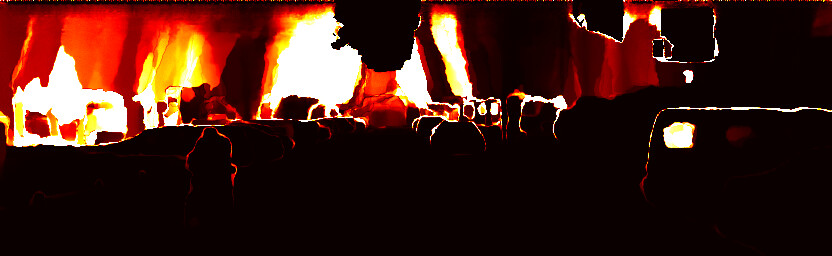} &
    \includegraphics[trim={450 0 0 50},clip,width=\Figwidth]{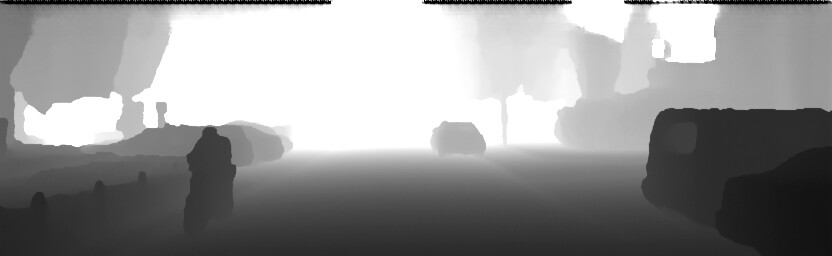} &
    \includegraphics[trim={450 0 0 50},clip,width=\Figwidth]{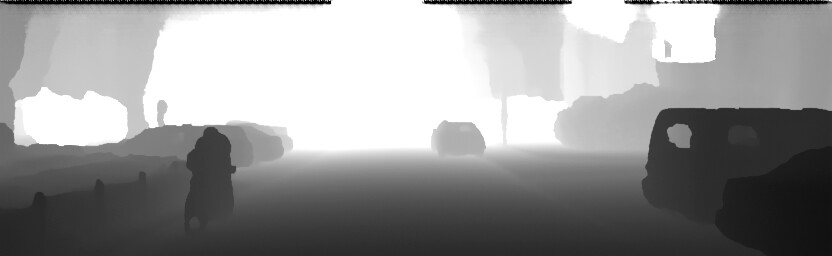} &
    \includegraphics[trim={450 0 0 50},clip,width=\Figwidth]{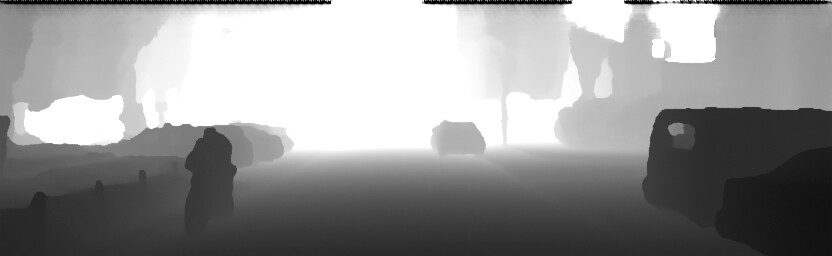}\\
    
    \includegraphics[trim={450 0 0 50},clip,width=\Figwidth]{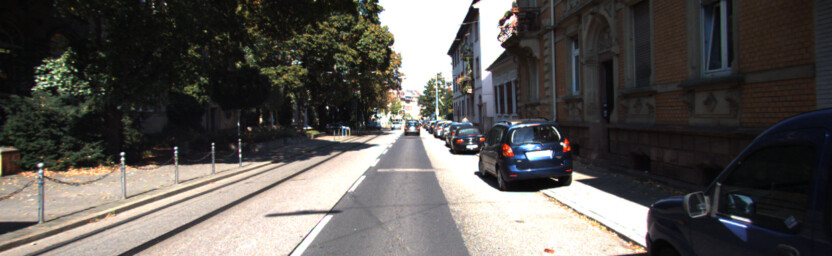} &
    \includegraphics[trim={450 0 0 50},clip,width=\Figwidth]{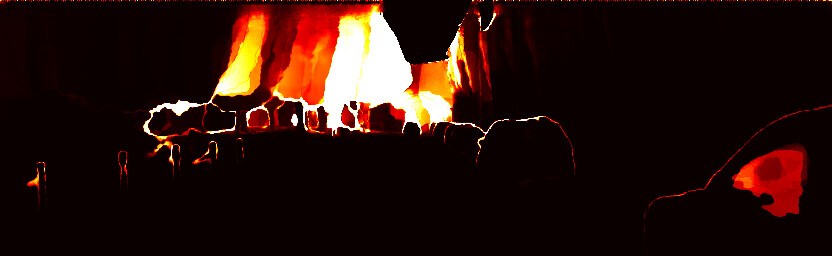} &
    \includegraphics[trim={450 0 0 50},clip,width=\Figwidth]{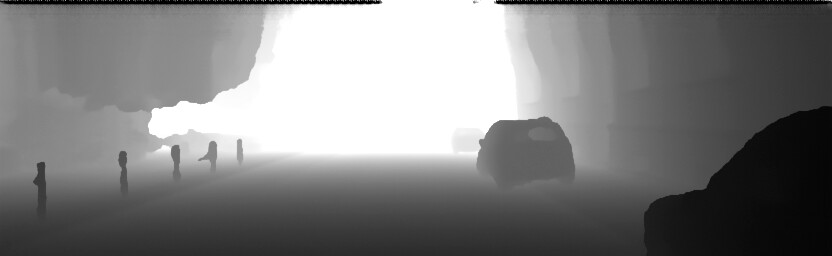} &
    \includegraphics[trim={450 0 0 50},clip,width=\Figwidth]{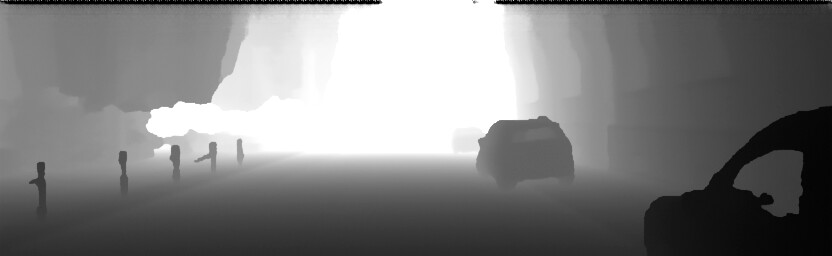} &
    \includegraphics[trim={450 0 0 50},clip,width=\Figwidth]{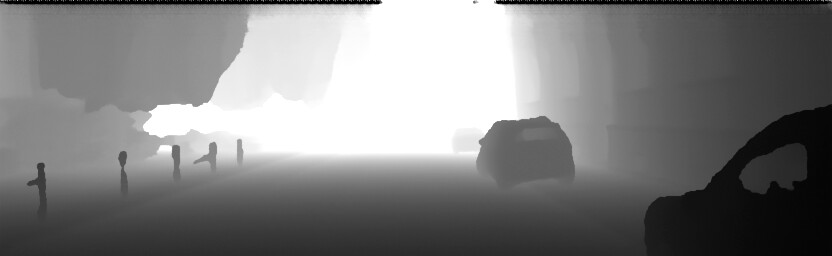}\\
    
    \includegraphics[trim={0 0 500 50},clip,width=\Figwidth]{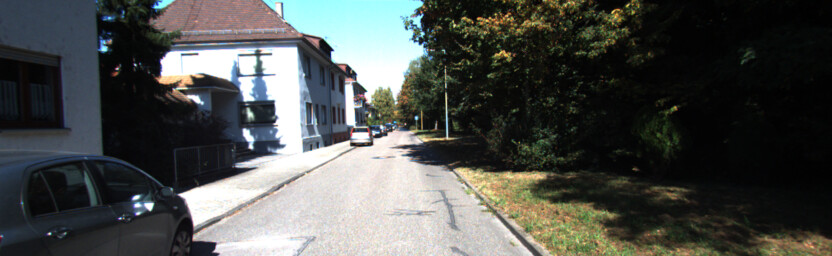} &
    \includegraphics[trim={0 0 500 50},clip,width=\Figwidth]{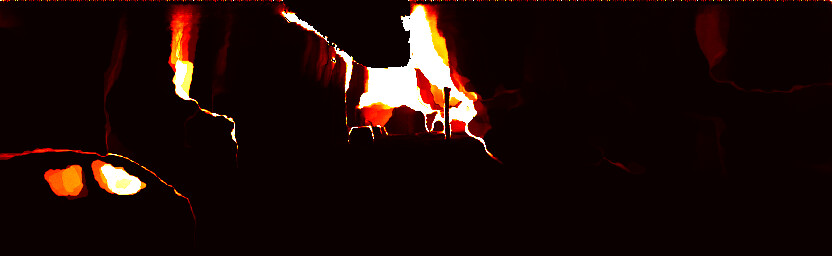} &
    \includegraphics[trim={0 0 500 50},clip,width=\Figwidth]{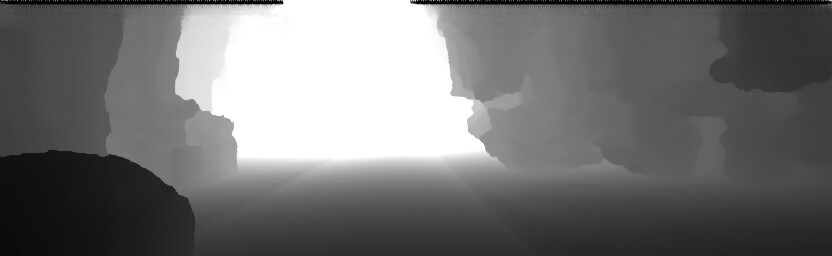} &
    \includegraphics[trim={0 0 500 50},clip,width=\Figwidth]{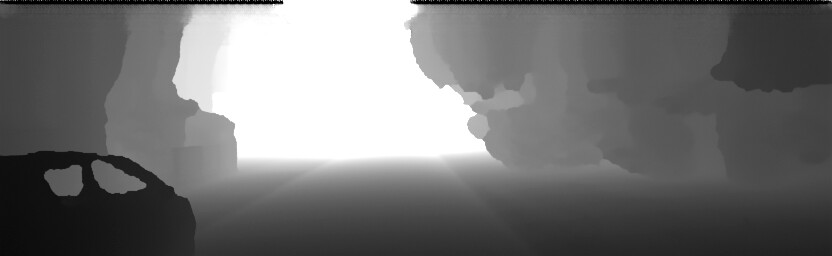} &
    \includegraphics[trim={0 0 500 50},clip,width=\Figwidth]{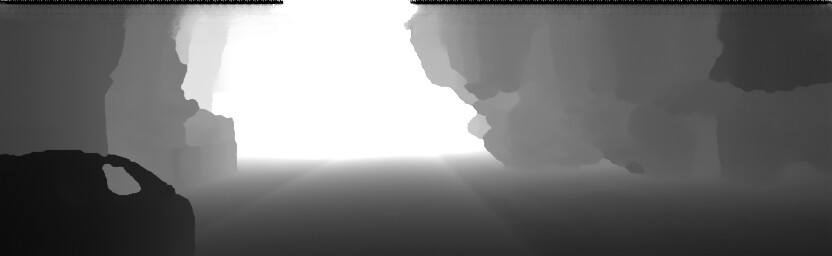}\\

    \includegraphics[trim={450 0 0 50},clip,width=\Figwidth]{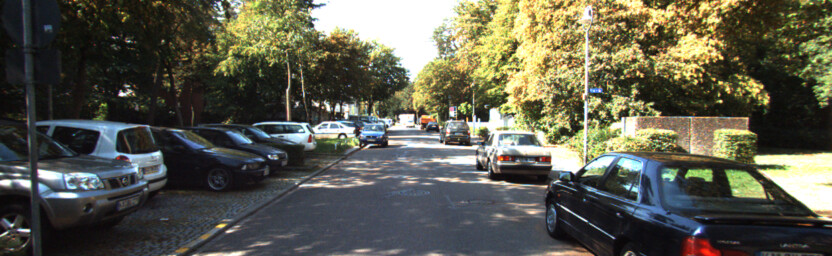} &
    \includegraphics[trim={450 0 0 50},clip,width=\Figwidth]{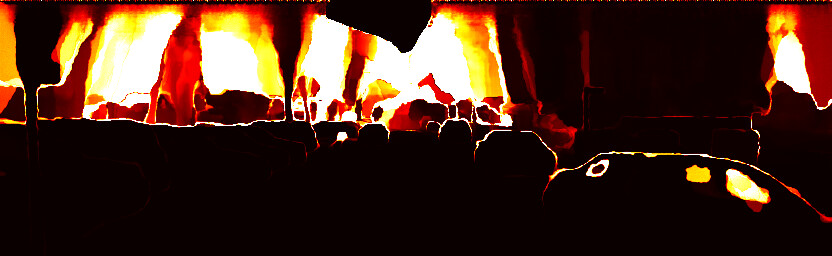} &
    \includegraphics[trim={450 0 0 50},clip,width=\Figwidth]{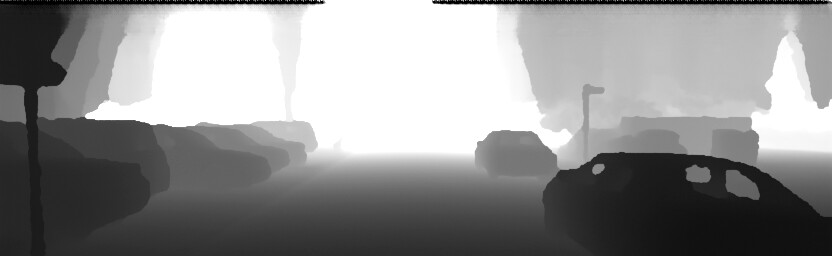} &
    \includegraphics[trim={450 0 0 50},clip,width=\Figwidth]{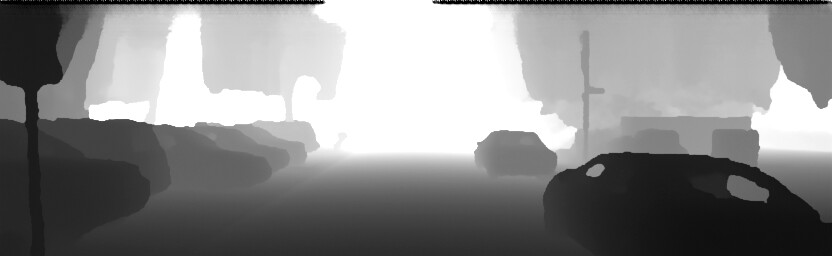} &
    \includegraphics[trim={450 0 0 50},clip,width=\Figwidth]{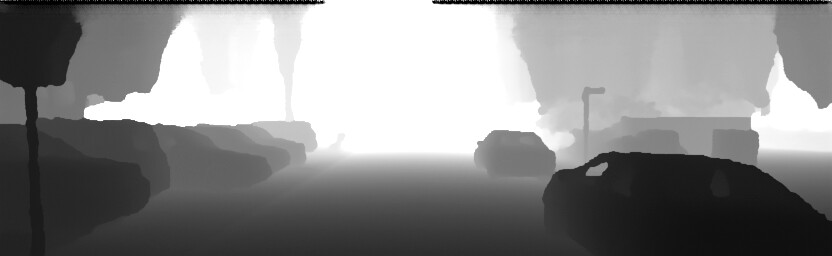}\\

    \end{tabular}
	\caption{
	\textbf{Qualitative examples of multimodal depth estimation on KITTI}. Our model is able to predict multimodal samples, especially windows of cars and object boundaries. Please refer to the areas with high variance.}
    \label{fig:supp:multi_modal_samples_depth_KITTI}
\end{figure*}

We provide more qualitative examples for multimodal prediction.
Figures \ref{fig:supp:multi_modal_samples_depth_NYU} and \ref{fig:supp:multi_modal_samples_depth_KITTI} illustrate multimodal depth predictions on NYU and KITTI respectively.
Multimodality of the posterior distribution exists in regions where there are multiple plausible predictions.
For example, this includes reflective and transparent surfaces (mirrors and glass surfaces in rows 1 to 5 of Figure  \ref{fig:supp:multi_modal_samples_depth_NYU} and windows of cars in Figure \ref{fig:supp:multi_modal_samples_depth_KITTI}).
We further find that the model captures uncertainty in depth estimates in the vicinity of object boundaries, some of which arise due to noise in ground truth measurements in the training data. This can be observed at the boundaries of cars in Figure \ref{fig:supp:multi_modal_samples_depth_KITTI} and around edges of objects in Figure \ref{fig:supp:multi_modal_samples_depth_NYU} (most clearly visible in the last row).

\begin{figure*}[tp]
    \centering
        %   \vspace*{0.35cm}
    \footnotesize
    \setlength\tabcolsep{0.8pt} % default value: 6pt
    \newcommand{\Figwidth}{0.195\linewidth}
    \begin{tabular}{c @{\hskip 0.3em} c @{\hskip 0.3em} ccc}
    Overlayed inputs & Variance heat map & \multicolumn{3}{c}{Multimodal prediction samples} \\

    \includegraphics[trim={0 0 700 100},clip,width=\Figwidth]{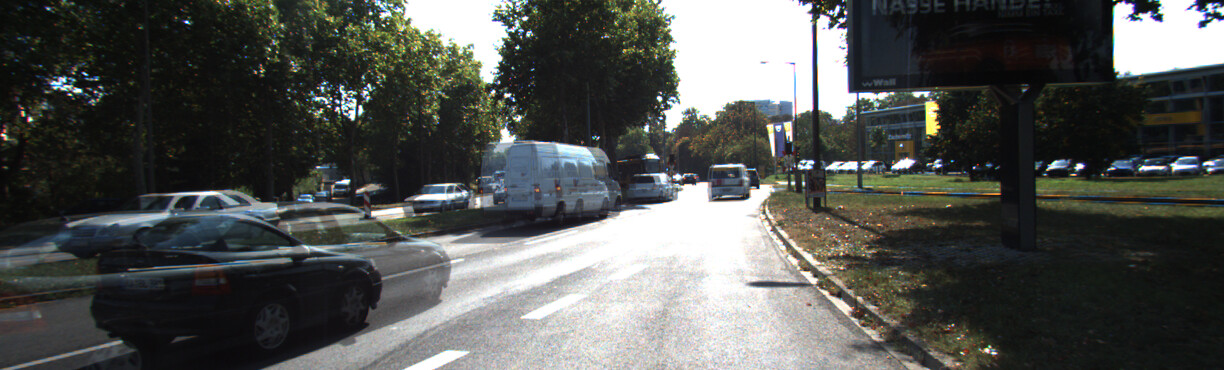} &
    \includegraphics[trim={0 0 700 100},clip,width=\Figwidth]{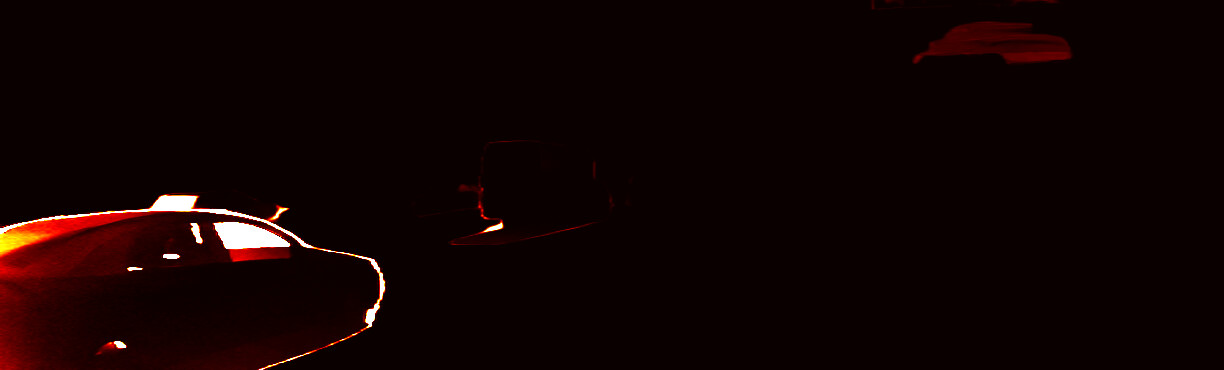} &
    \includegraphics[trim={0 0 700 100},clip,width=\Figwidth]{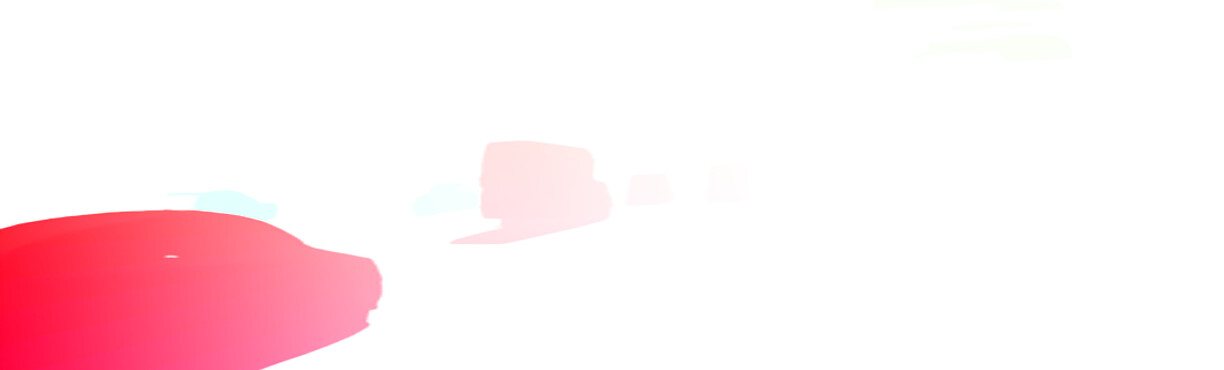} &
    \includegraphics[trim={0 0 700 100},clip,width=\Figwidth]{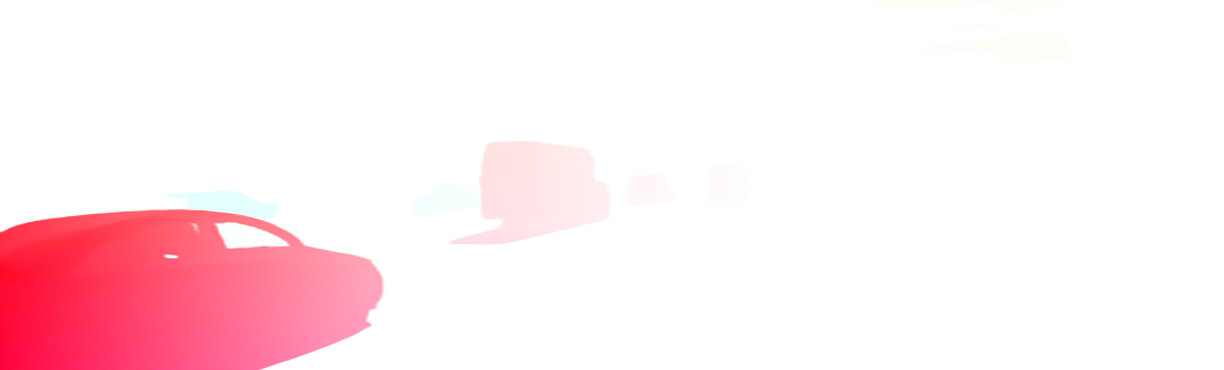} &
    \includegraphics[trim={0 0 700 100},clip,width=\Figwidth]{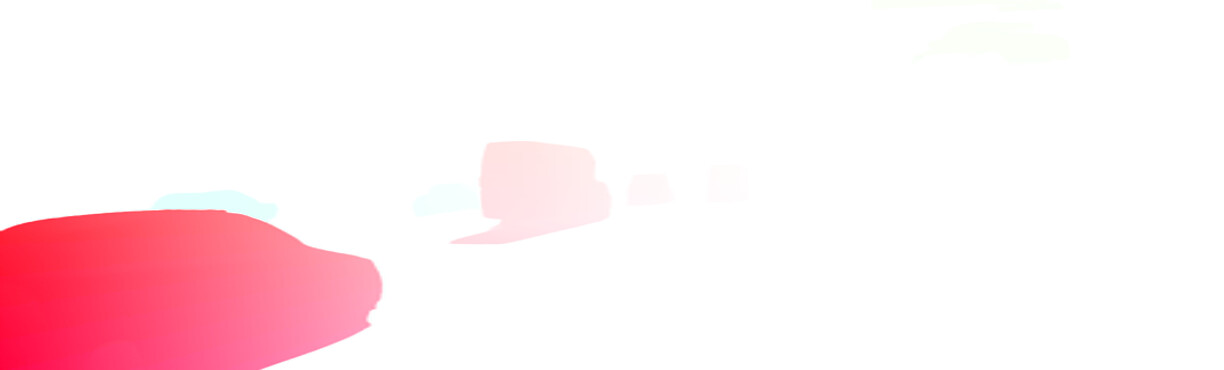}\\
    
    \includegraphics[trim={250 0 250 0},clip,width=\Figwidth]{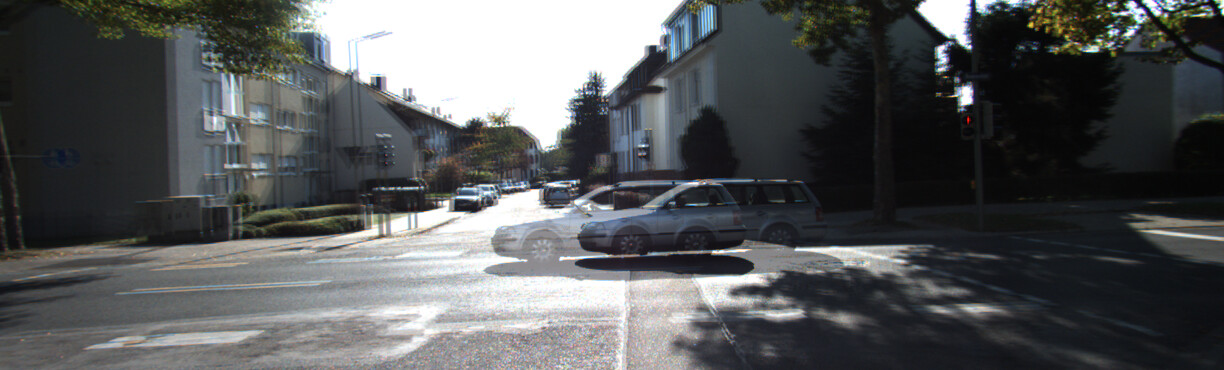} &
    \includegraphics[trim={250 0 250 0},clip,width=\Figwidth]{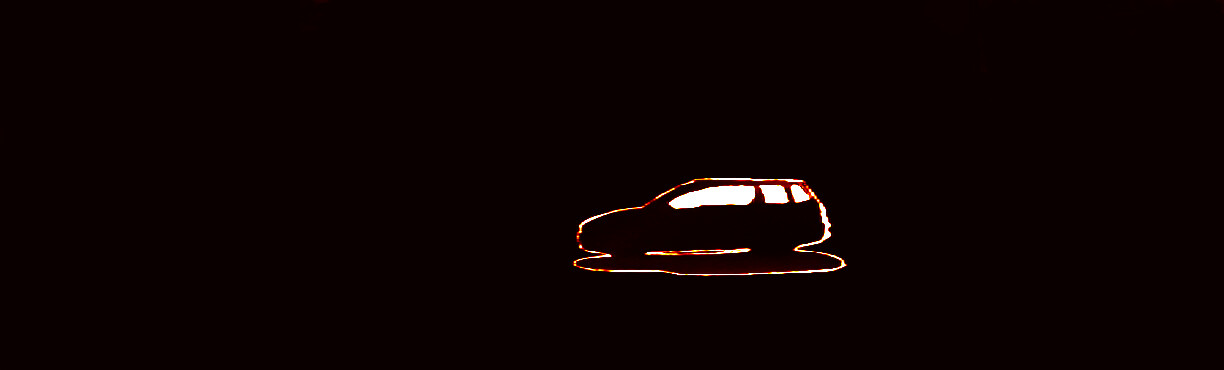} &
    \includegraphics[trim={250 0 250 0},clip,width=\Figwidth]{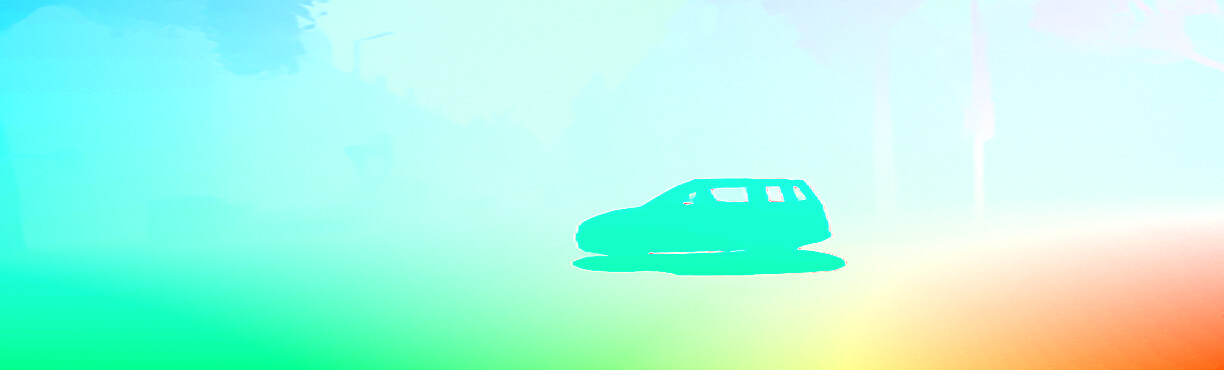} &
    \includegraphics[trim={250 0 250 0},clip,width=\Figwidth]{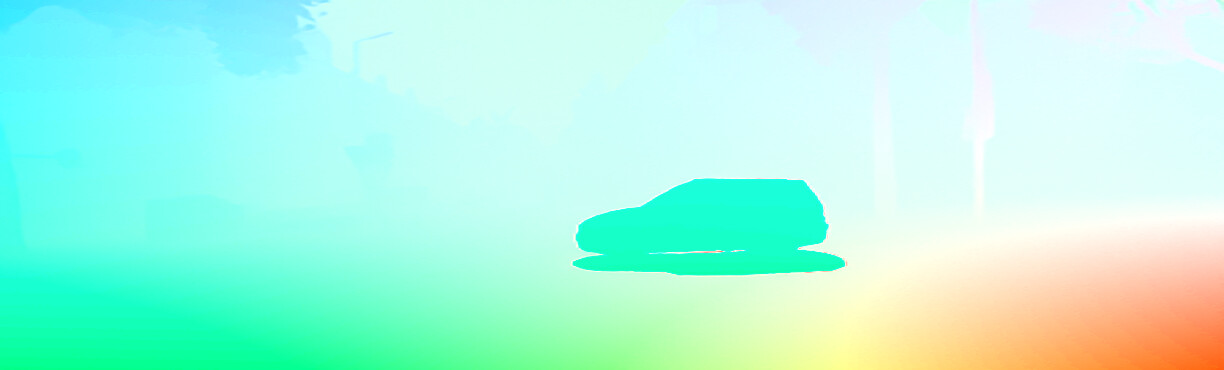} &
    \includegraphics[trim={250 0 250 0},clip,width=\Figwidth]{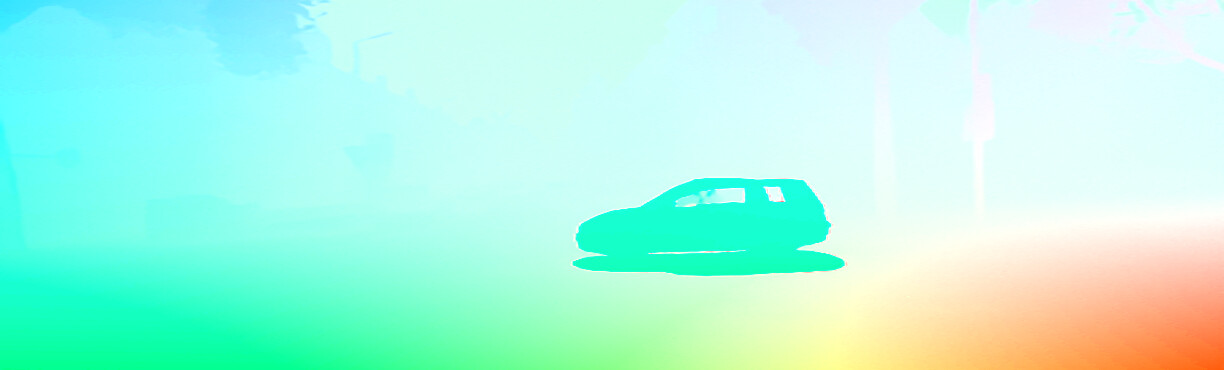}\\

    \includegraphics[trim={400 0 0 0},clip,width=\Figwidth]{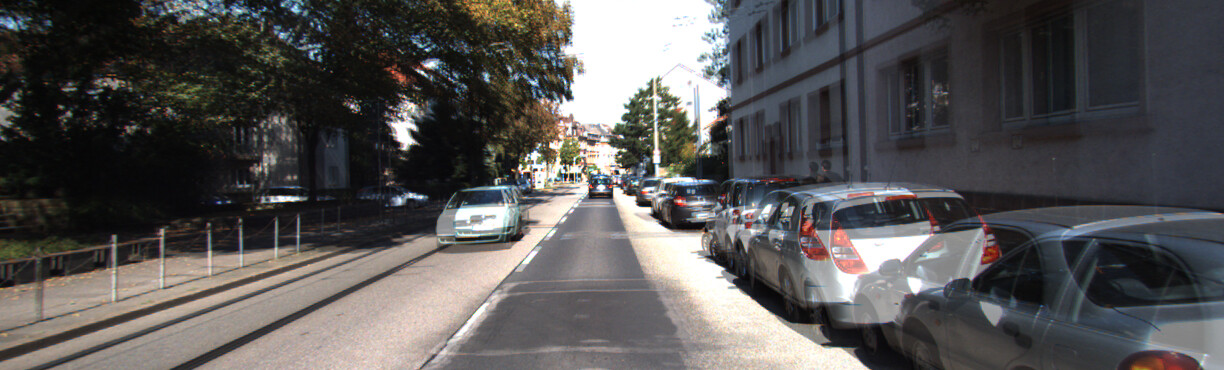} &
    \includegraphics[trim={400 0 0 0},clip,width=\Figwidth]{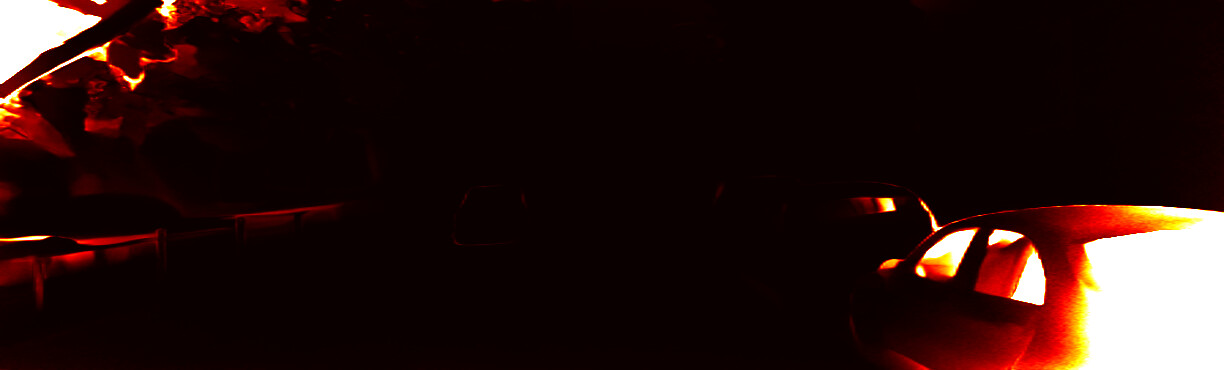} &
    \includegraphics[trim={400 0 0 0},clip,width=\Figwidth]{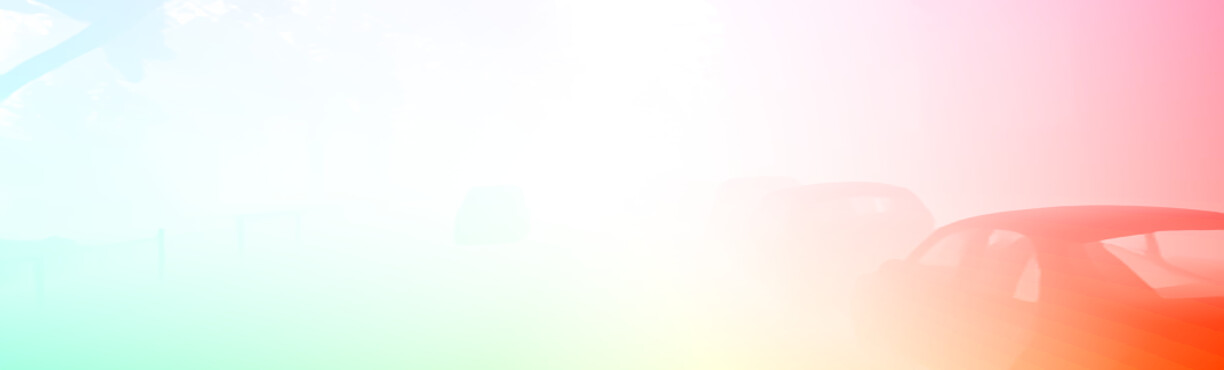} &
    \includegraphics[trim={400 0 0 0},clip,width=\Figwidth]{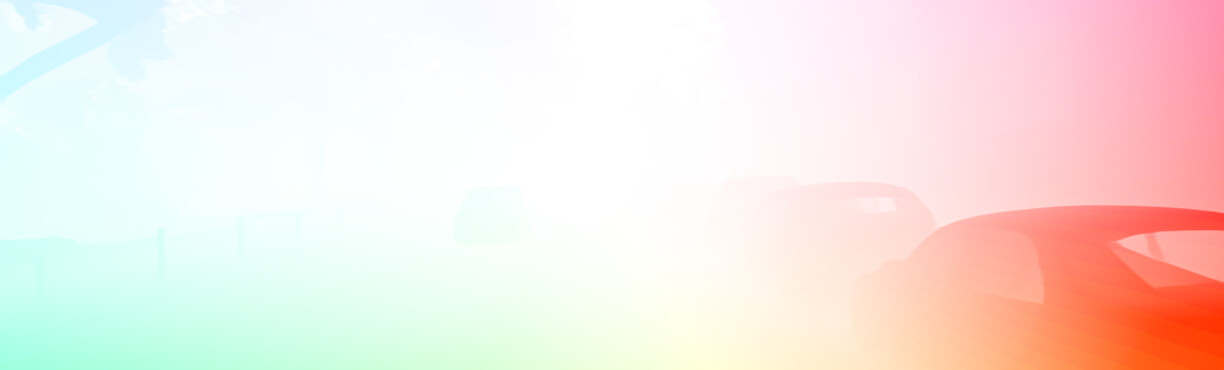} &
    \includegraphics[trim={400 0 0 0},clip,width=\Figwidth]{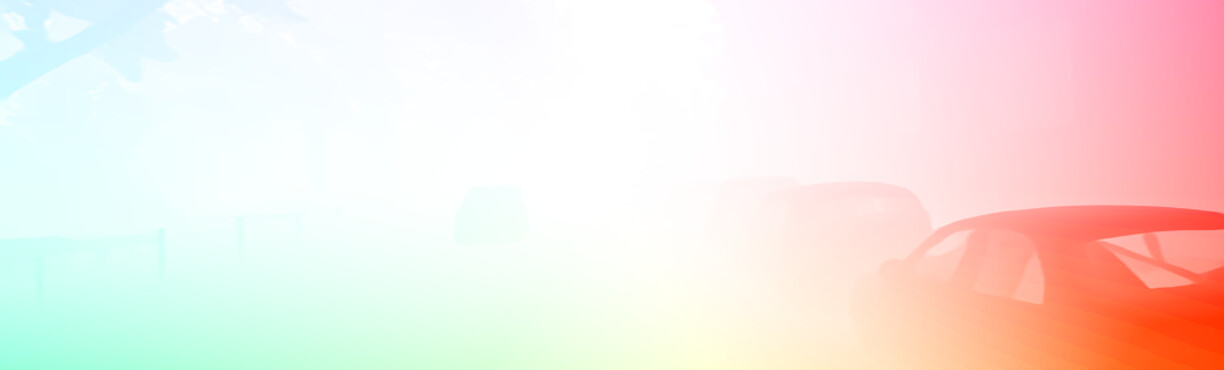}\\
    
    \includegraphics[trim={700 50 0 50},clip,width=\Figwidth]{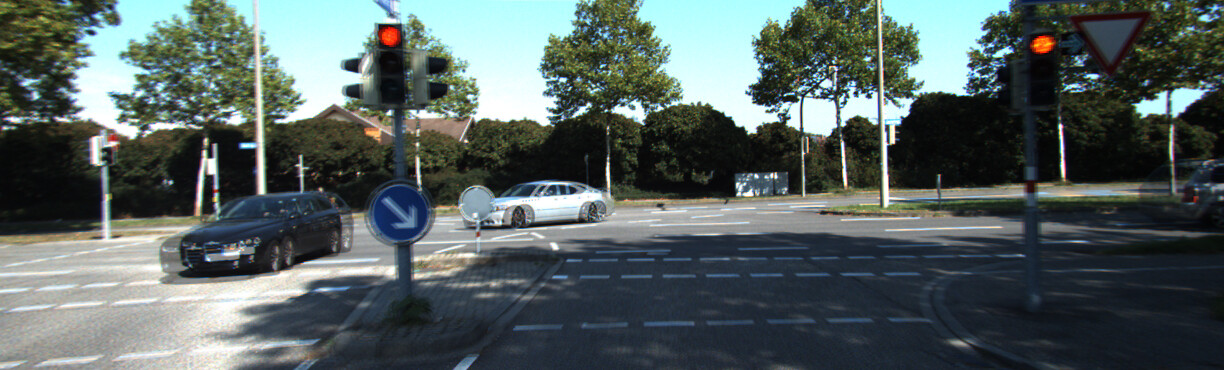} &
    \includegraphics[trim={700 50 0 50},clip,width=\Figwidth]{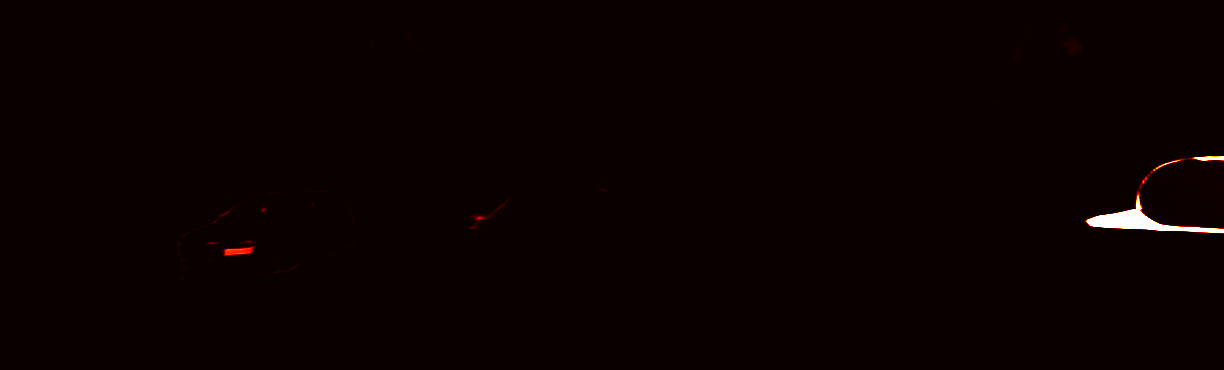} &
    \includegraphics[trim={700 50 0 50},clip,width=\Figwidth]{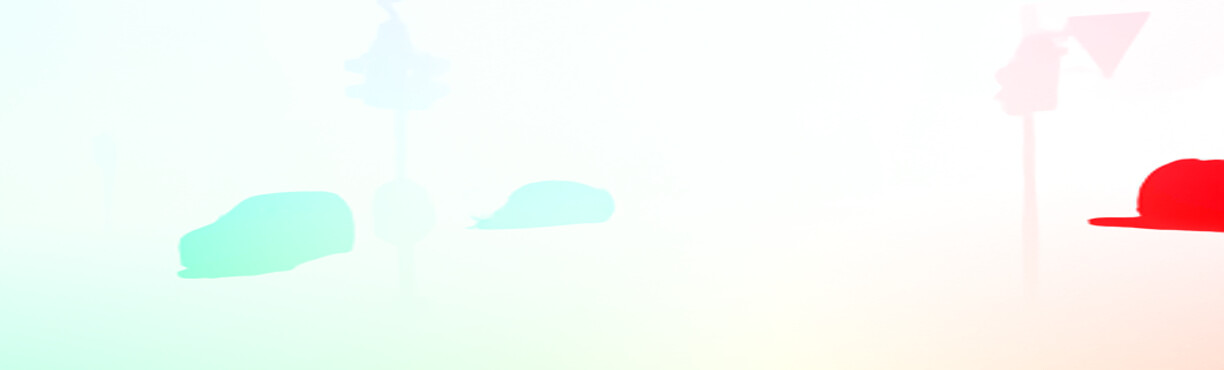} &
    \includegraphics[trim={700 50 0 50},clip,width=\Figwidth]{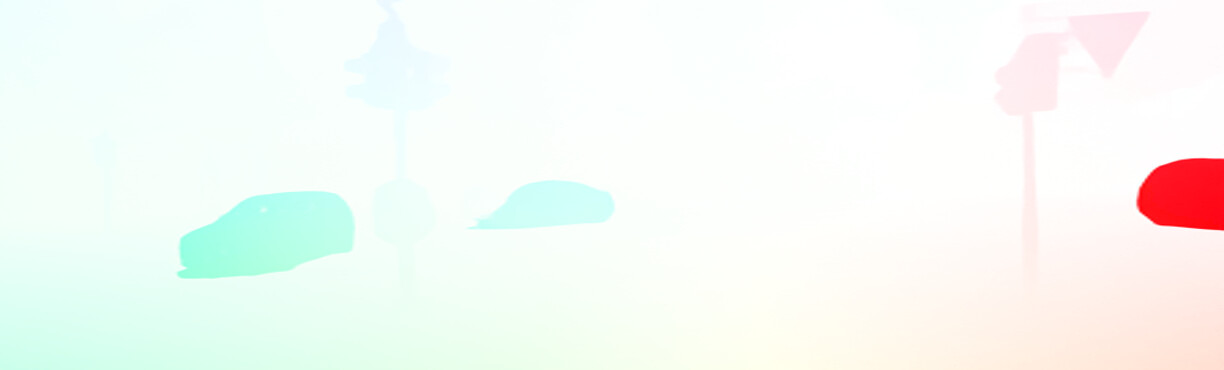} &
    \includegraphics[trim={700 50 0 50},clip,width=\Figwidth]{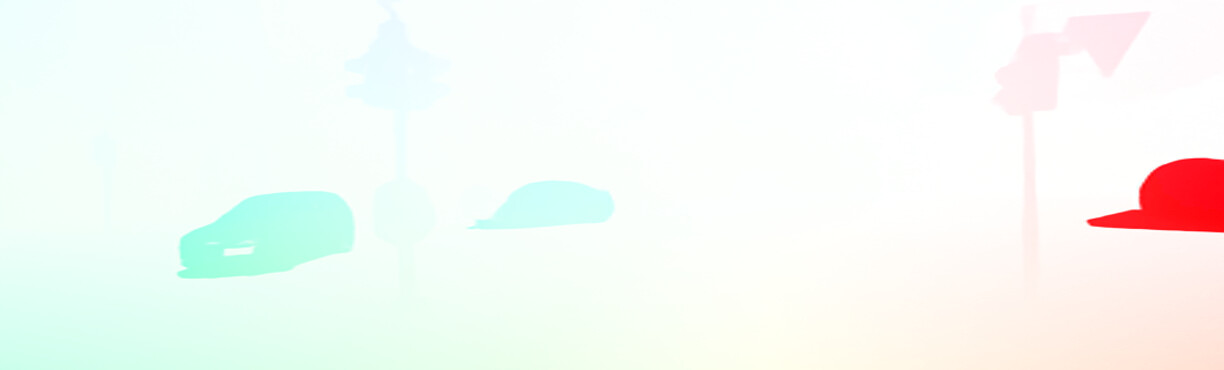}\\

    \end{tabular}
	\caption{\textbf{Qualitative examples of multimodal optical flow estimation on KITTI}. Multimodality exists on transparent surfaces (\eg, windows of cars) and shadows where our model estimates layered motion in different samples (\emph{see the last row}).}
    \label{fig:supp:multi_modal_samples_flow_KITTI}
    % \vspace*{0.2cm}
\end{figure*}

\begin{figure*}[h]
    \centering
    %   \vspace*{0.15cm}
    \footnotesize
    \setlength\tabcolsep{0.8pt}
    \renewcommand{\arraystretch}{0.0}
    \newcommand{\Figwidth}{0.247\linewidth}
    \begin{tabular}{c @{\hskip 0.4em} ccc}
    Input images & \multicolumn{3}{c}{Variance heat map $\&$ multimodal samples} \\
    \includegraphics[trim={0 0 0 0},clip,width=\Figwidth]{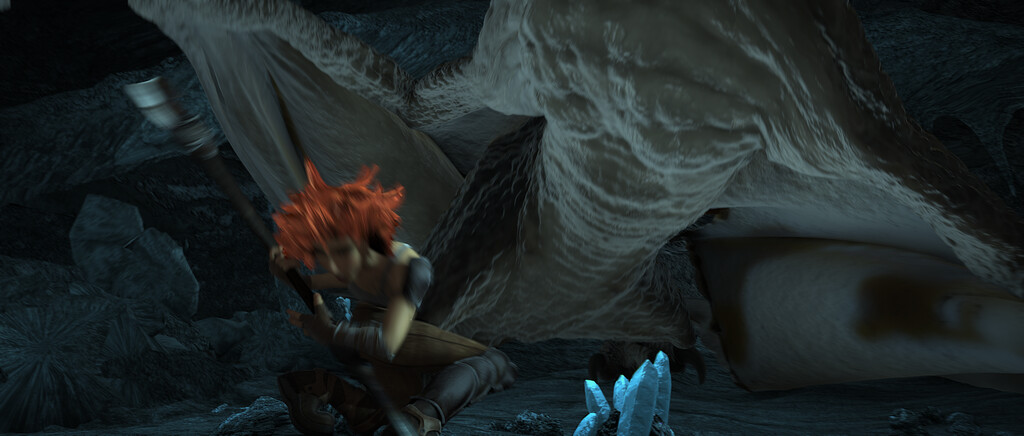} &
    \includegraphics[trim={0 0 0 0},clip,width=\Figwidth]{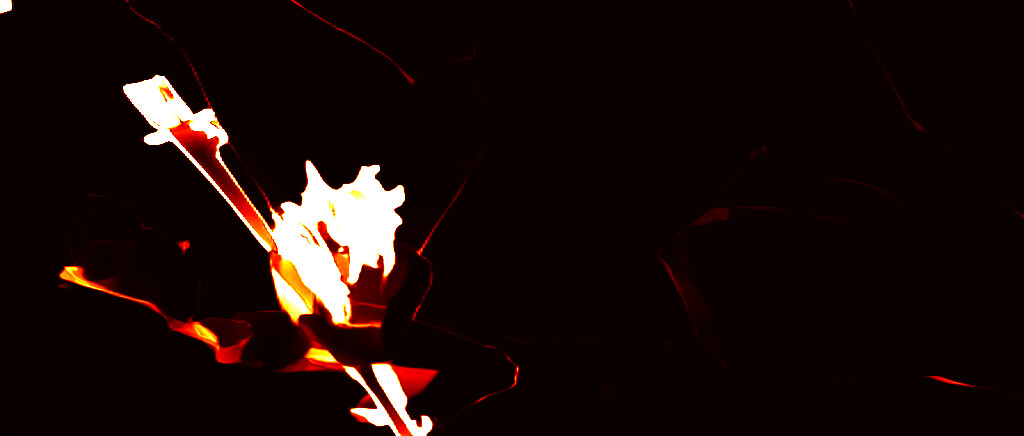} &
    \includegraphics[trim={0 0 0 0},clip,width=\Figwidth]{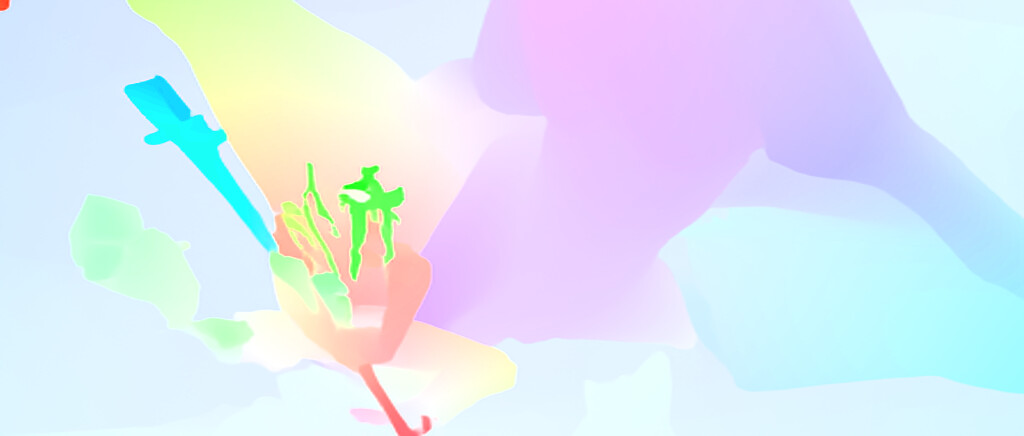} &
    \includegraphics[trim={0 0 0 0},clip,width=\Figwidth]{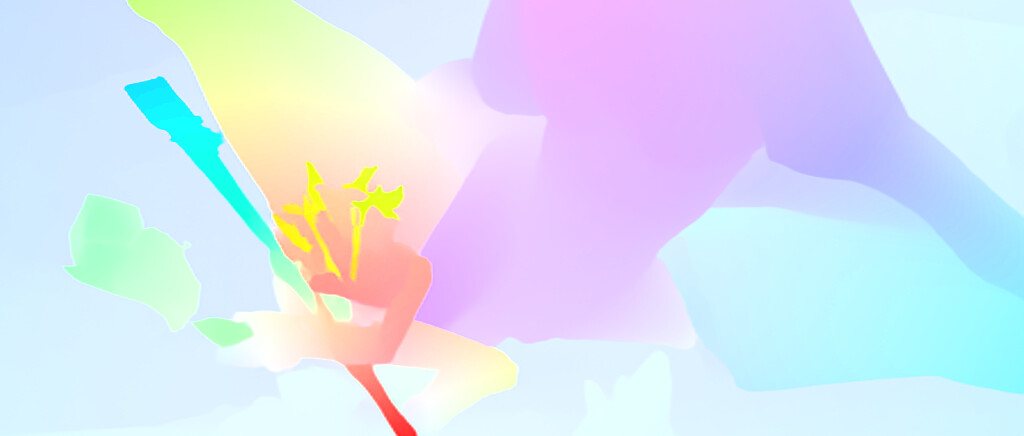} \\ 
    [0.1em] \\
    \includegraphics[trim={0 0 0 0},clip,width=\Figwidth]{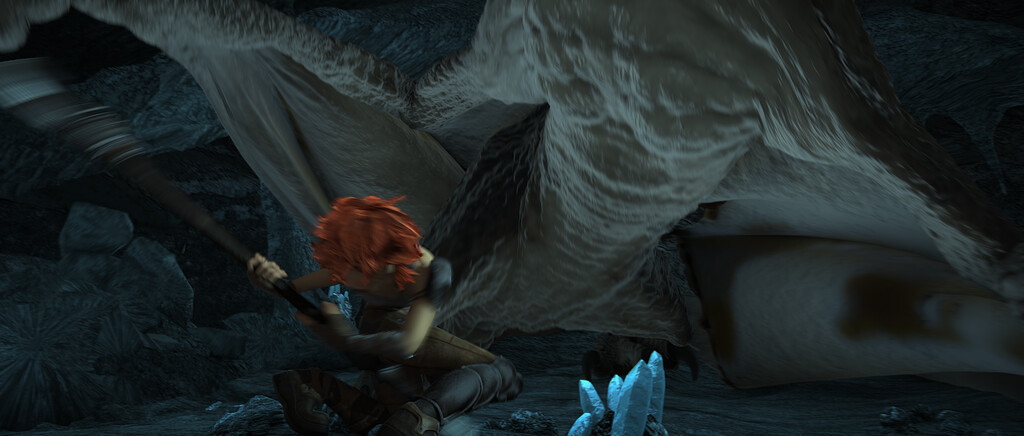} &
    \includegraphics[trim={0 0 0 0},clip,width=\Figwidth]{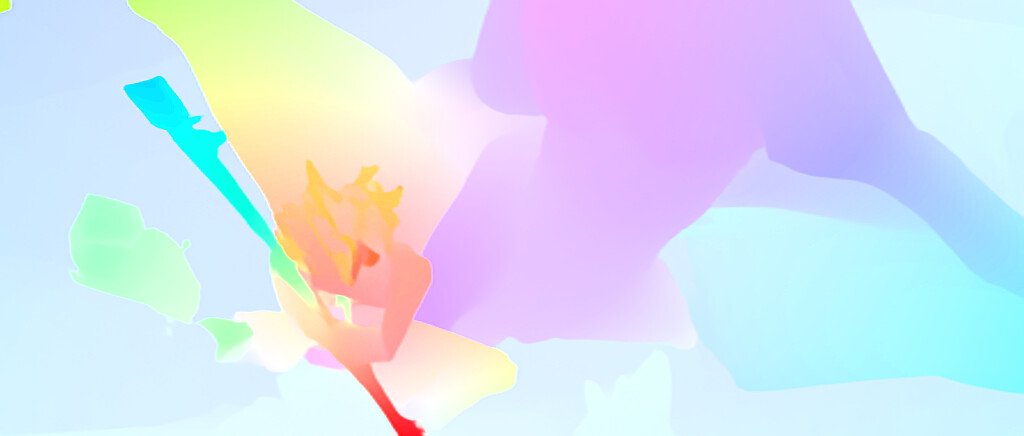} &
    \includegraphics[trim={0 0 0 0},clip,width=\Figwidth]{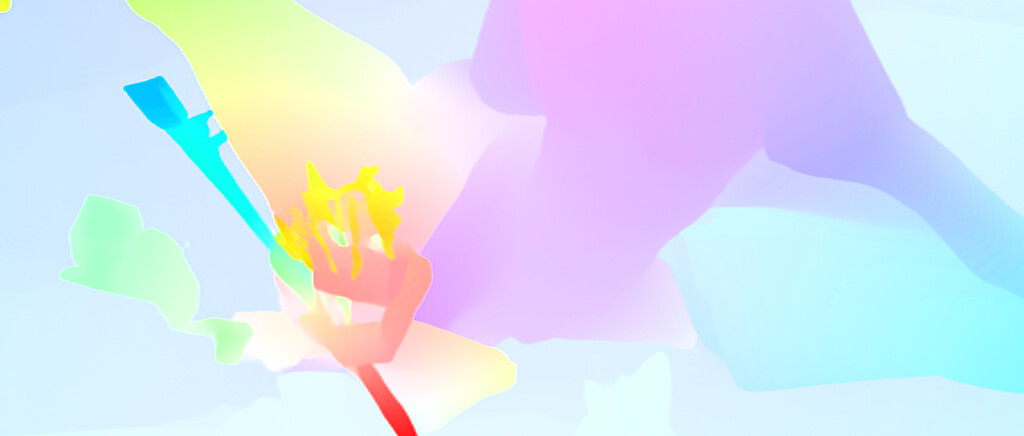} &
    \includegraphics[trim={0 0 0 0},clip,width=\Figwidth]{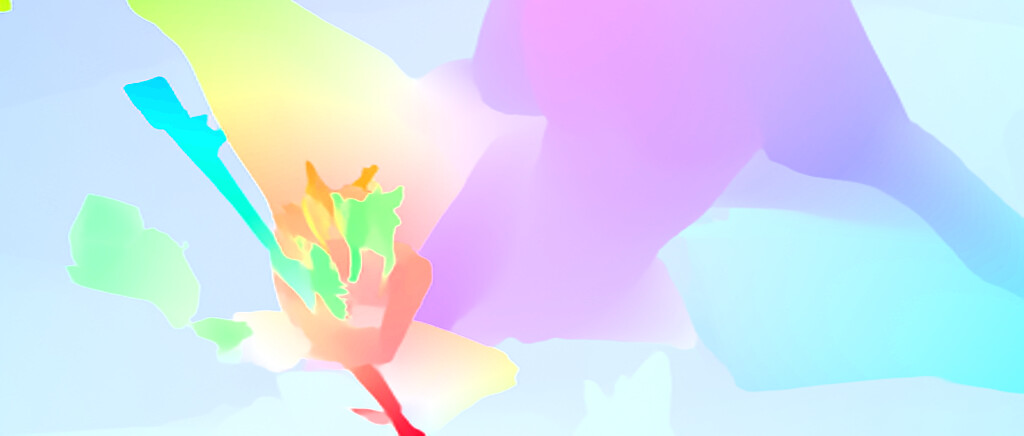}\\ [0.3em] \\
    
    \includegraphics[trim={0 0 0 0},clip,width=\Figwidth]{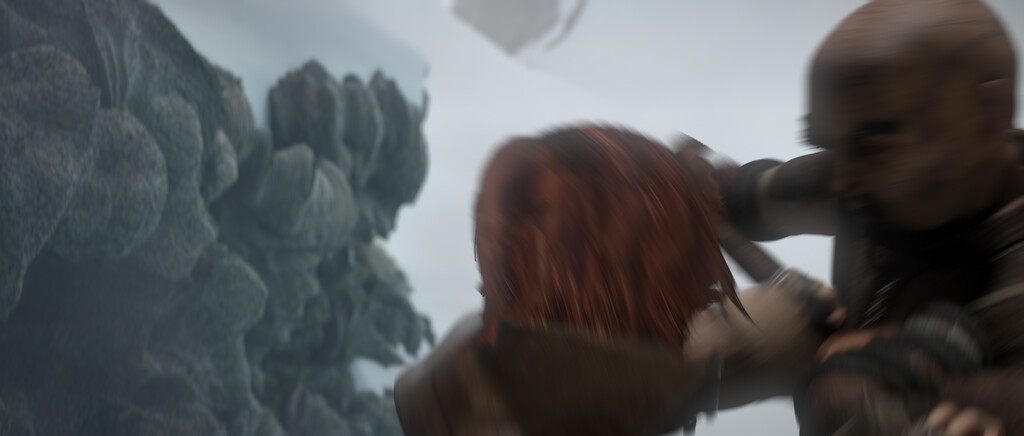} &
    \includegraphics[trim={0 0 0 0},clip,width=\Figwidth]{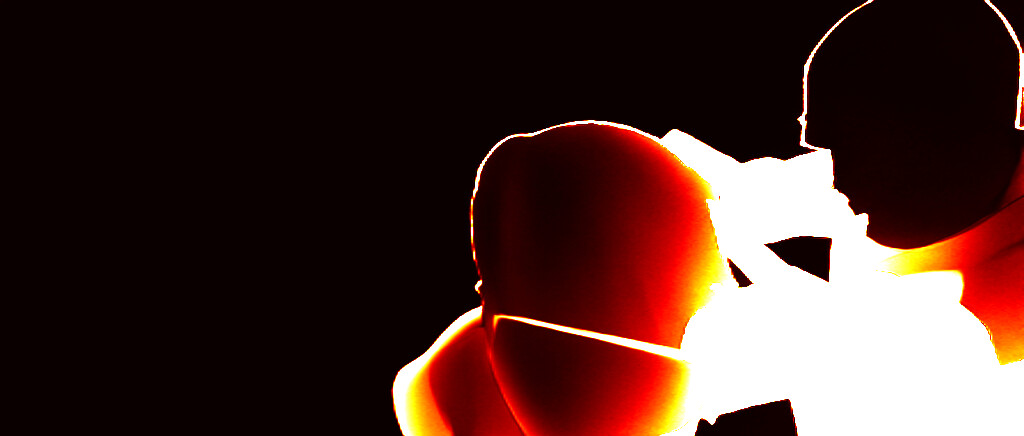} &
    \includegraphics[trim={0 0 0 0},clip,width=\Figwidth]{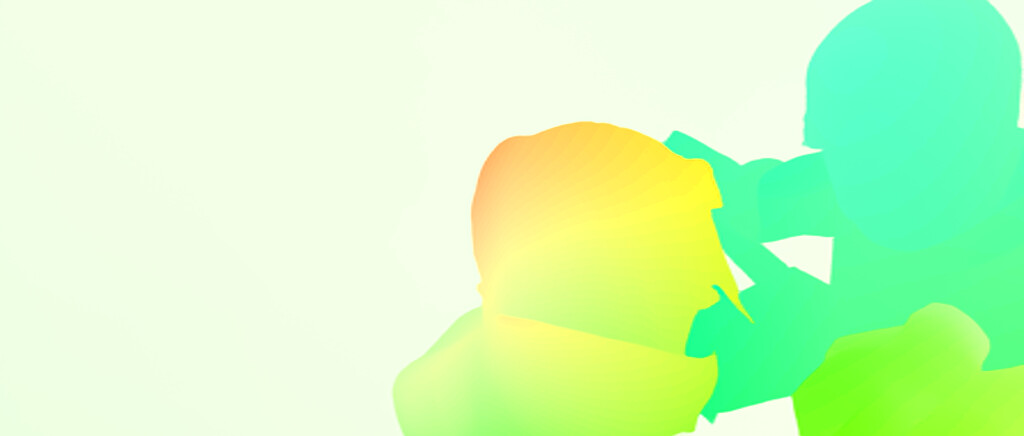} &
    \includegraphics[trim={0 0 0 0},clip,width=\Figwidth]{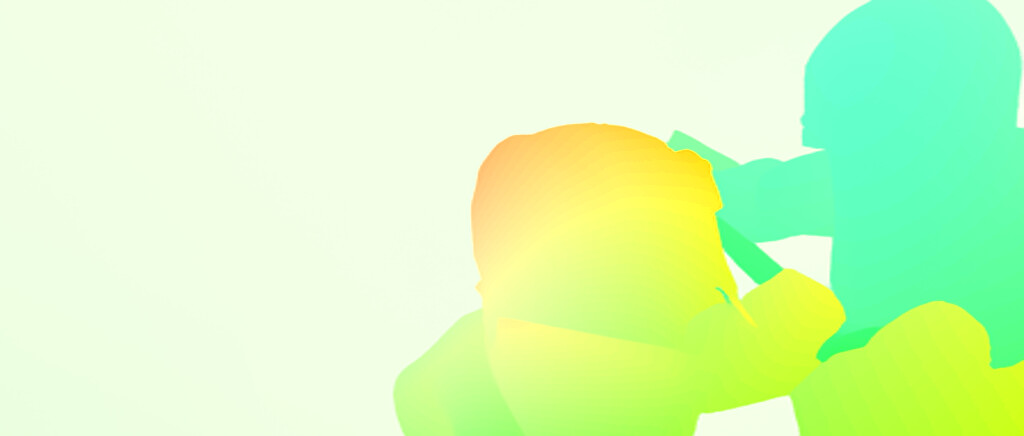} \\
    [0.1em] \\
    \includegraphics[trim={0 0 0 0},clip,width=\Figwidth]{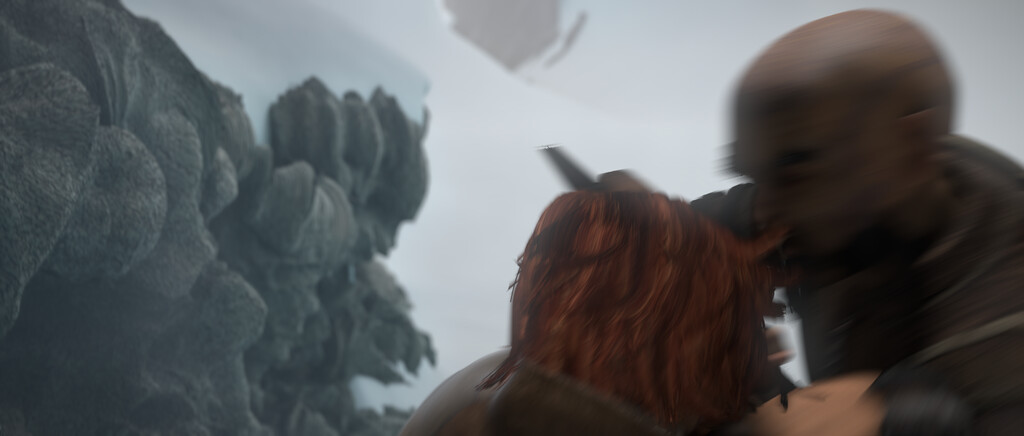} &
    \includegraphics[trim={0 0 0 0},clip,width=\Figwidth]{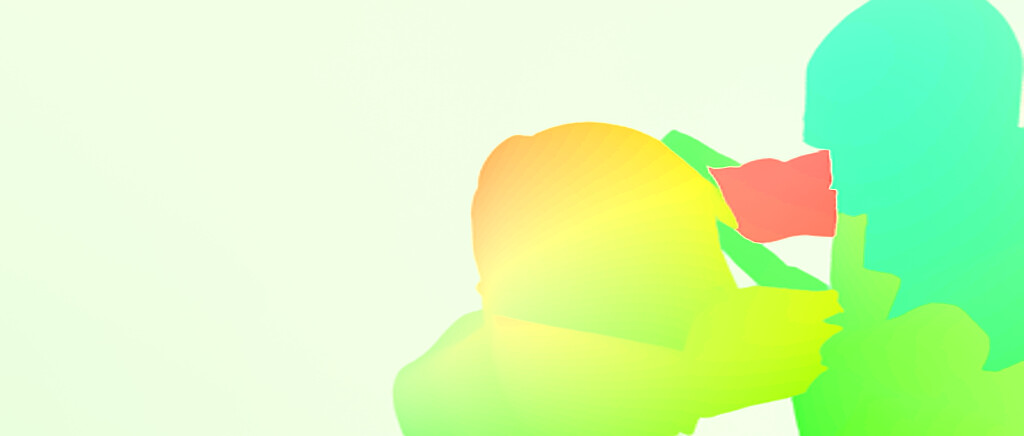} &
    \includegraphics[trim={0 0 0 0},clip,width=\Figwidth]{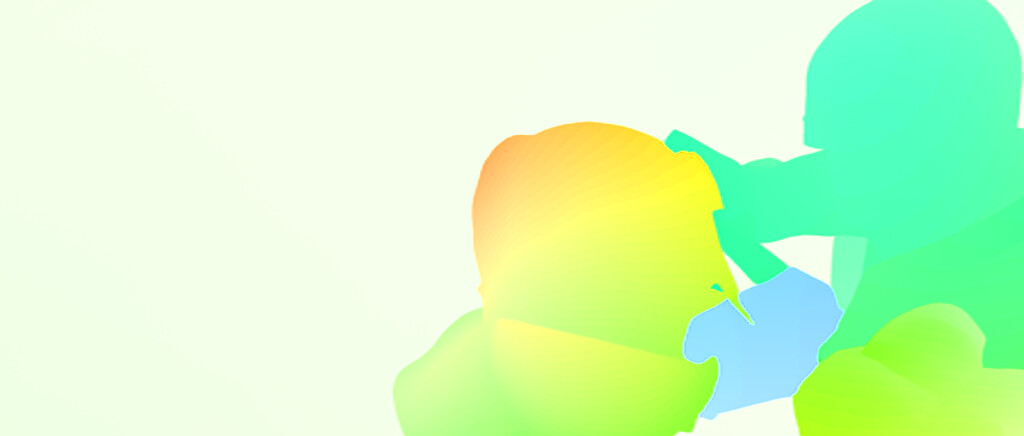} &
    \includegraphics[trim={0 0 0 0},clip,width=\Figwidth]{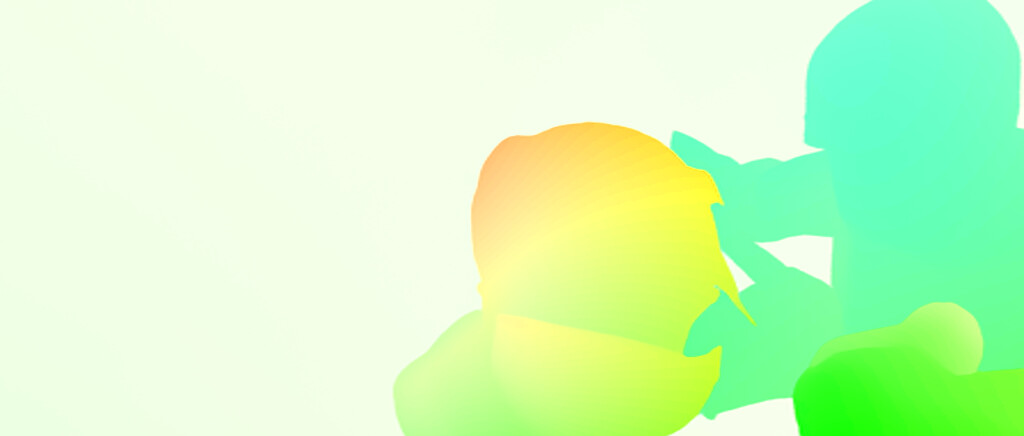}\\ [0.3em] \\
    
    \includegraphics[trim={0 0 0 0},clip,width=\Figwidth]{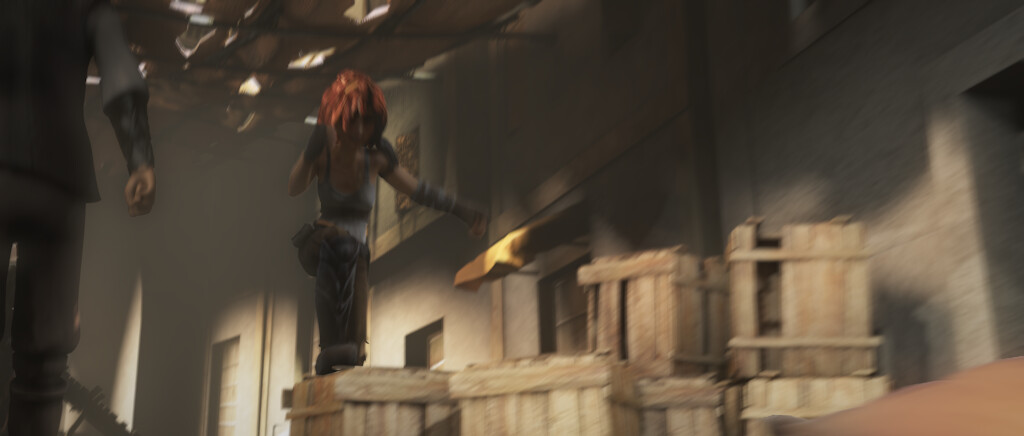} &
    \includegraphics[trim={0 0 0 0},clip,width=\Figwidth]{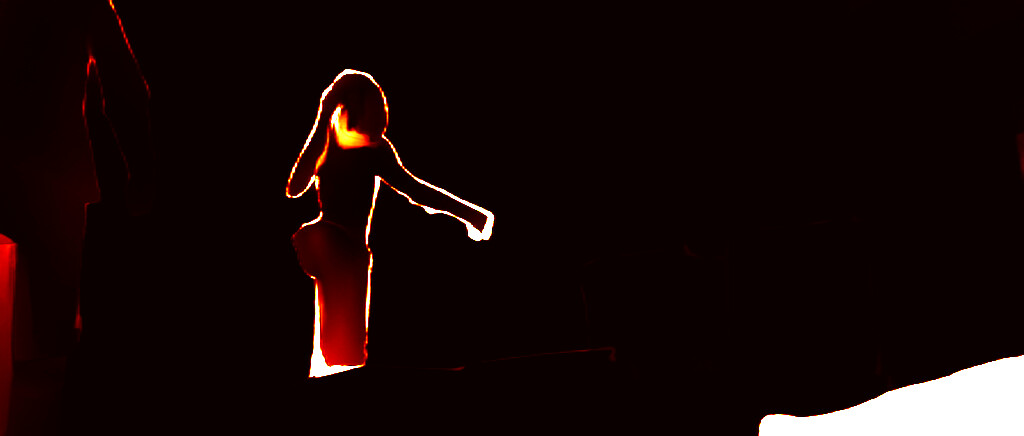} &
    \includegraphics[trim={0 0 0 0},clip,width=\Figwidth]{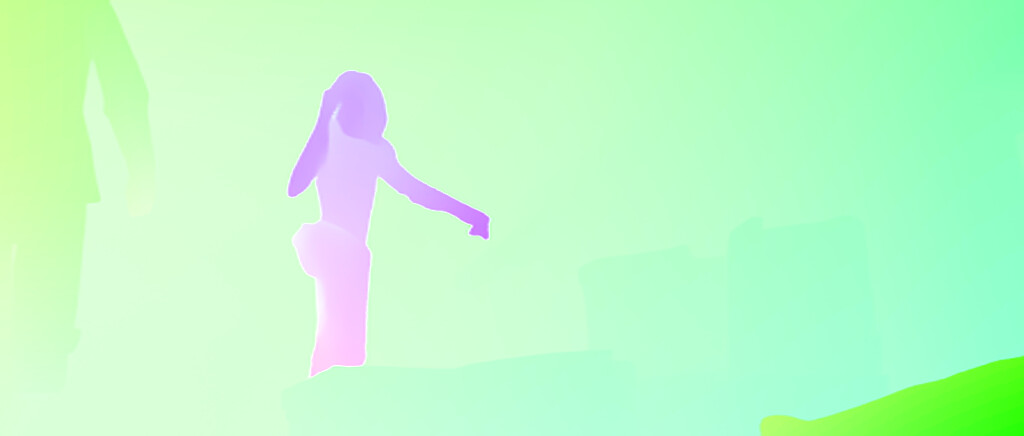} &
    \includegraphics[trim={0 0 0 0},clip,width=\Figwidth]{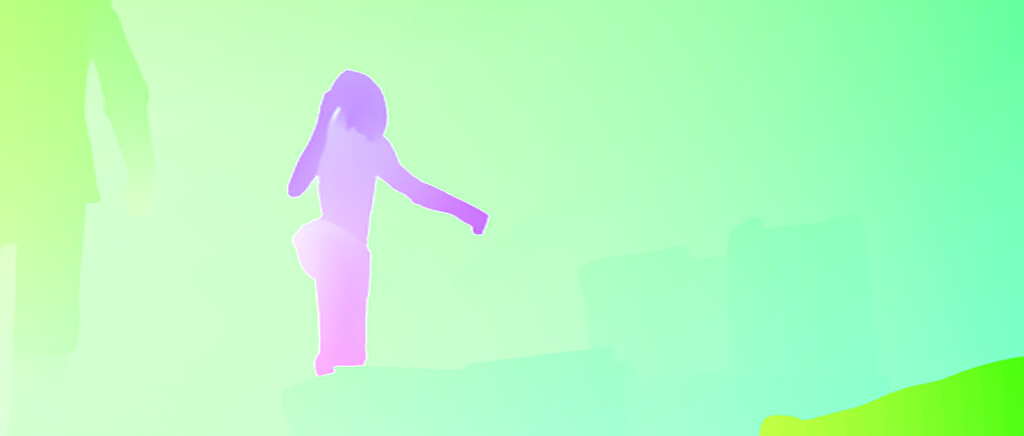} \\
    [0.1em] \\
    \includegraphics[trim={0 0 0 0},clip,width=\Figwidth]{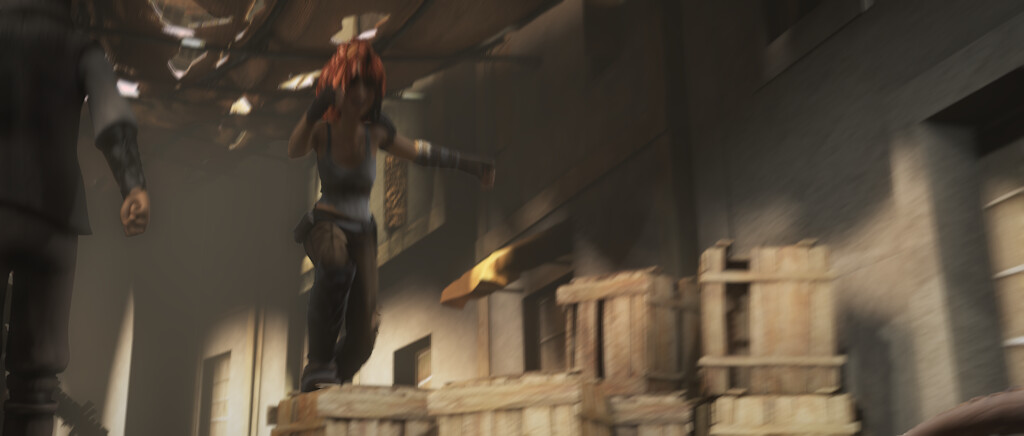} &
    \includegraphics[trim={0 0 0 0},clip,width=\Figwidth]{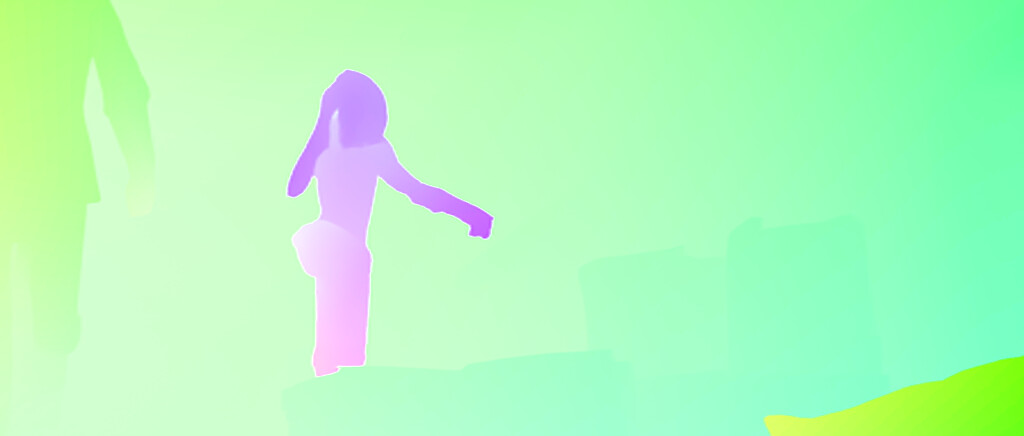} &
    \includegraphics[trim={0 0 0 0},clip,width=\Figwidth]{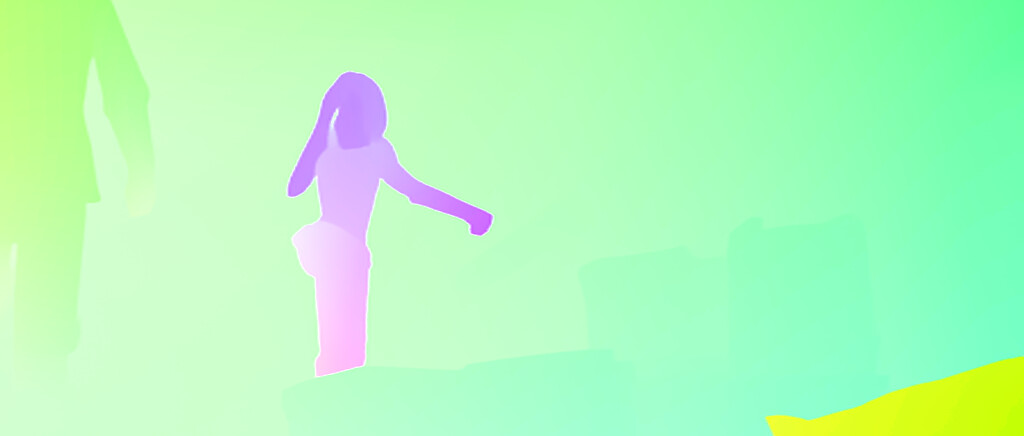} &
    \includegraphics[trim={0 0 0 0},clip,width=\Figwidth]{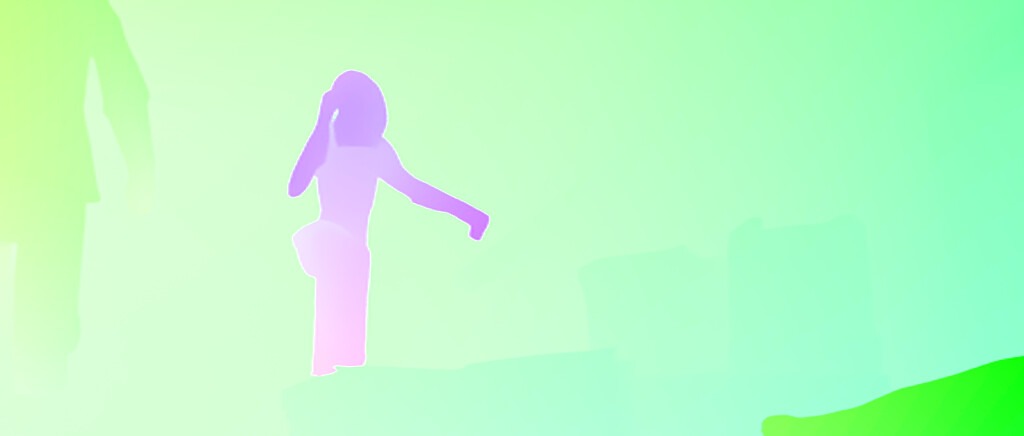}\\ [0.3em] \\

    \includegraphics[trim={0 0 0 0},clip,width=\Figwidth]{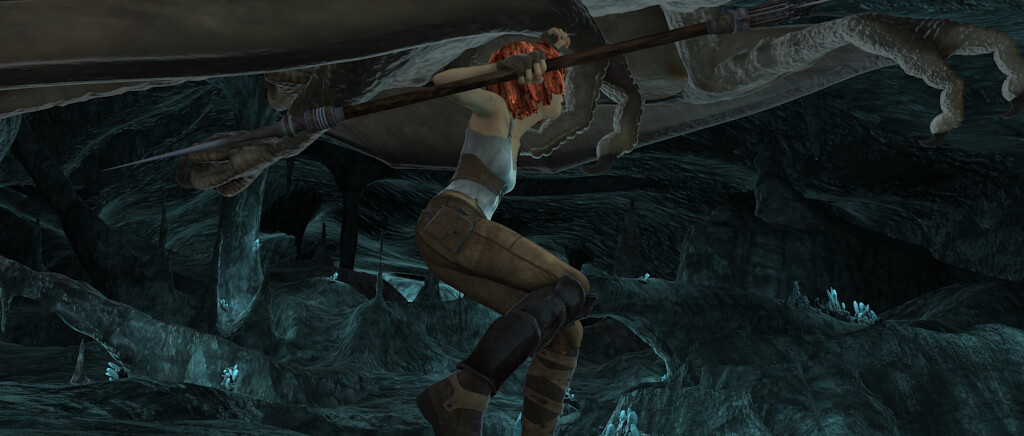} &
    \includegraphics[trim={0 0 0 0},clip,width=\Figwidth]{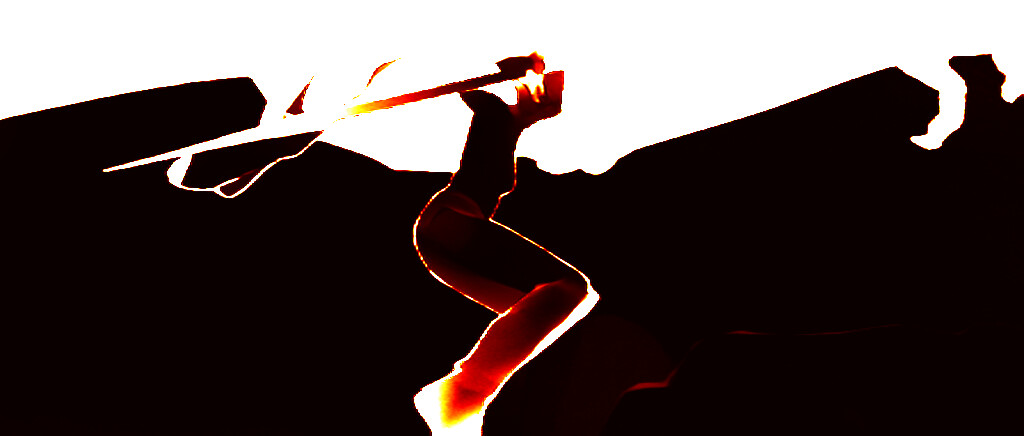} &
    \includegraphics[trim={0 0 0 0},clip,width=\Figwidth]{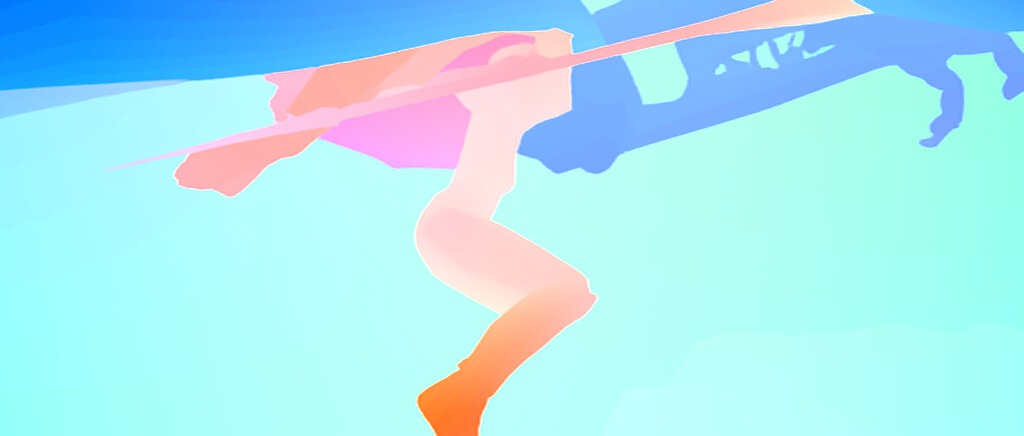} &
    \includegraphics[trim={0 0 0 0},clip,width=\Figwidth]{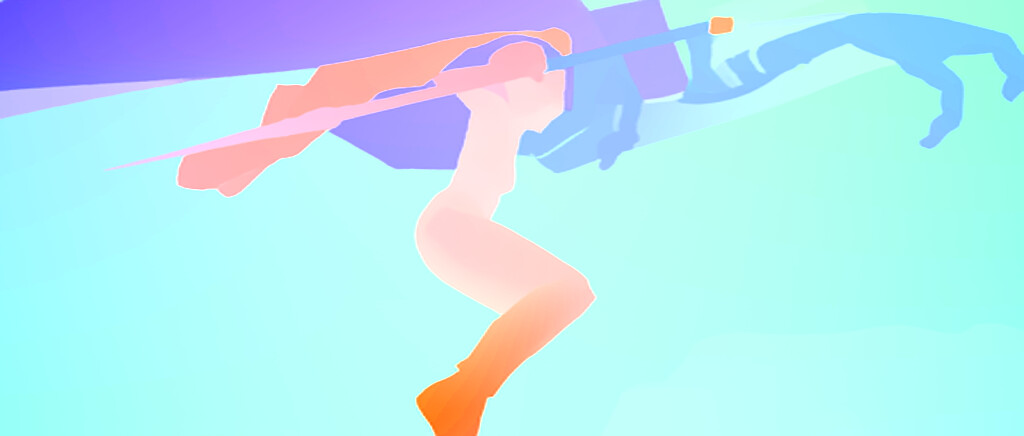} \\
    [0.1em] \\
    \includegraphics[trim={0 0 0 0},clip,width=\Figwidth]{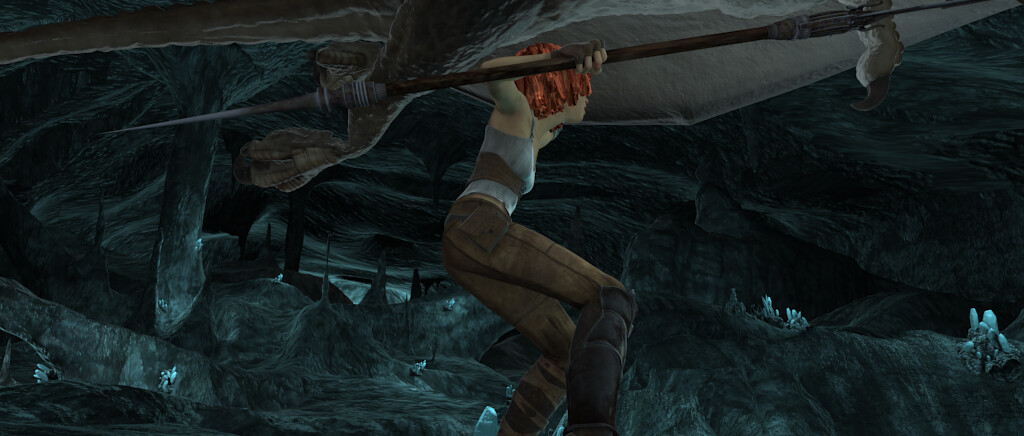} &
    \includegraphics[trim={0 0 0 0},clip,width=\Figwidth]{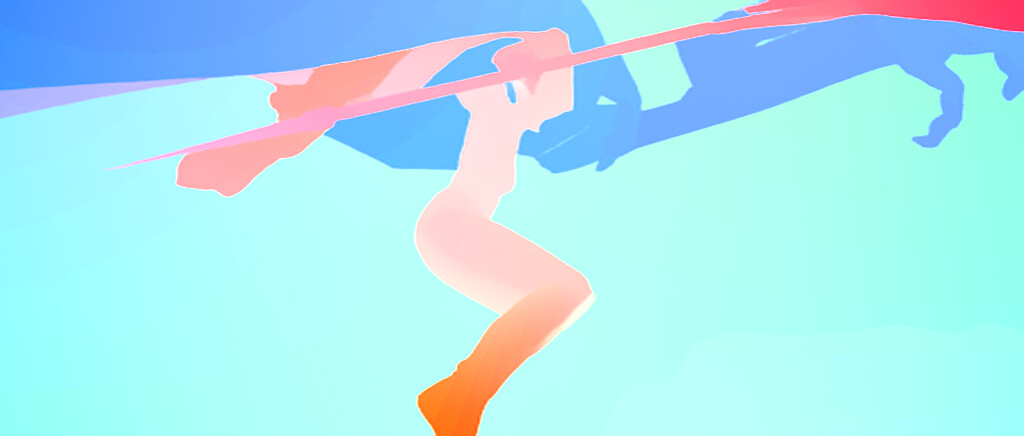} &
    \includegraphics[trim={0 0 0 0},clip,width=\Figwidth]{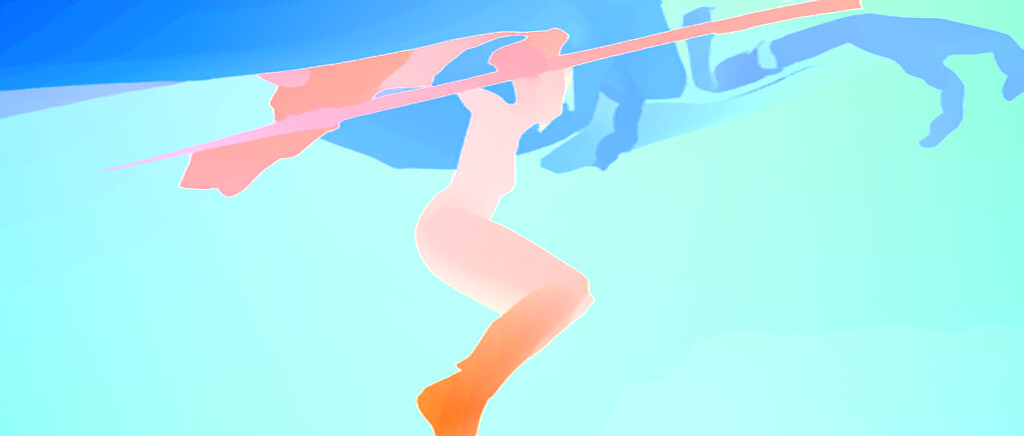} &
    \includegraphics[trim={0 0 0 0},clip,width=\Figwidth]{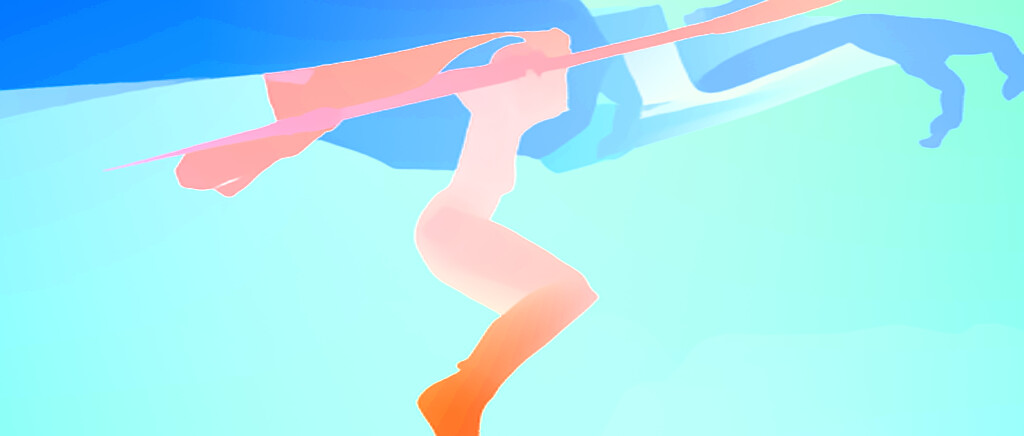}\\
    \end{tabular}

	\caption{\textbf{Qualitative examples of multimodal optical flow estimation on Sintel}. Multimodality also exists on examples with challenging occlusion or out-of-bound cases.}
    \label{fig:supp:multi_modal_samples_flow_sintel}
    %   \vspace*{0.2cm}
\end{figure*}

Figure \ref{fig:supp:multi_modal_samples_flow_KITTI} illustrates different samples on KITTI from the optical flow diffusion model, also capturing multiple modes of the predictive posterior.
Multimodality exists on transparent surfaces and near occlusions.
As shown in Figure \ref{fig:supp:multi_modal_samples_flow_sintel}, on Sintel, multimodality also exists on occluded or out-of-bounds pixels where multiple predictions are plausible.

\clearpage % to remove the spacing problem after tables

% \textbf{Examples of multimodal prediction} on depth (NYU) and optical flow (Sintel and KITTI). Each row shows an input image (or overlayed two images for optical flow), a variance heat map from 8 samples, and 3 individual samples. Our model captures multimodal samples on uncertain/ambiguous cases, such as reflective (\eg~mirror on NYU), transparent (\eg~vehicle window on KITTI), and translucent (\eg~fog on Sintel) regions. High variance also exists near object boundaries due to inaccurate estimates, which are often challenging cases for optical flow, and also partially originate from noisy ground truth measurements for depth.

\vspace*{0.5cm}

\section{Qualitative comparison of depth estimation with DPT}
\label{sec:supp-dpt}
\vspace*{-0.25cm}

\begin{figure*}[h!]

\setlength\tabcolsep{0.1pt} % default value: 6pt
\centering
\footnotesize
\renewcommand{\arraystretch}{0.05}
% \addtolength{\tabcolsep}{-5pt}

\newcommand{\image}[2]{
\begin{tikzpicture}
\node[anchor=south west,inner sep=0] at (0,0)
{\includegraphics[width=0.246\linewidth]{figures/dpt/#1.jpg}};
#2
\end{tikzpicture}
}

\newcommand{\row}[2]{
\image{image#1}{#2}
& 
\image{gt#1}{#2}
&
\image{dpt#1}{#2}
&
\image{ours#1}{#2}
}
\newcommand{\arrow}[2]{
\draw[red,-latex,thick]#1 -- #2;
}

\begin{tabular}{cccc}
    Image & Ground Truth & DPT & \textbf{Ours} \\
    
\row{0136}{
\arrow{(1.85,1.6)}{(1.55,1.6)};
\arrow{(1.75,1.3)}{(2.05,1.3)};
\arrow{(1.55,0.5)}{(1.25,0.5)};
} \\
% \row{0008}{} \\
\row{0011}{
\arrow{(2.6,1.4)}{(2.3,1.4)};
\arrow{(3.2,2.2)}{(3.2,1.9)};
} \\
% \row{0070}{} \\
\row{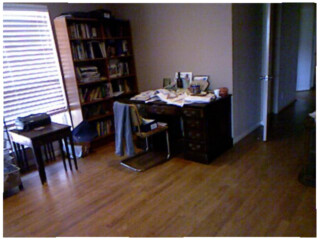}{
\arrow{(0.95,1.3)}{(1.25,1.3)};
\arrow{(2.55,2.0)}{(2.85,2.0)};
} \\
% \row{0134}{} \\
\row{0149}{
\arrow{(1.2,2)}{(1.5,2)};
\arrow{(2.6,2)}{(2.9,2)};
\arrow{(0.75,0.5)}{(0.5,0.4)};
} \\

\row{0005}{
\arrow{(1.7,1.7)}{(2.0,1.7)};
\arrow{(2.1,2.0)}{(2.4,2.0)};
} \\

\row{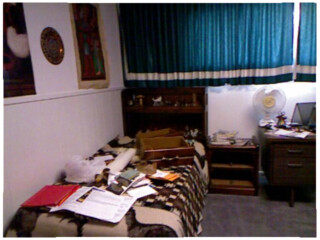}{
\arrow{(3.3,1.8)}{(3.1,1.65)};
\arrow{(2.55,1.4)}{(2.25,1.4)};
} \\
\row{0370}{
\arrow{(1.8,1)}{(1.8,1.25)};
\arrow{(1.9,1.7)}{(2.2,1.7)};
} \\
\row{0633}{
% \arrow{(0.7,1.0)}{(1.0,1.0)};
\arrow{(1.3,0.4)}{(1.6,0.4)};
} \\

\end{tabular}
\vspace*{-0.05cm}
\caption{\textbf{Qualitative comparison of our model with DPT-Hybrid} \cite{dpt} (fine-tuned on NYU) on the NYU depth v2 val set. Our method infers better depth for both scene structure (walls, floors, \etc.) and individual objects. Specific differences are highlighted with red arrows.}
\vspace*{-0.5cm}
\label{fig:nyu_dpt}
\end{figure*}
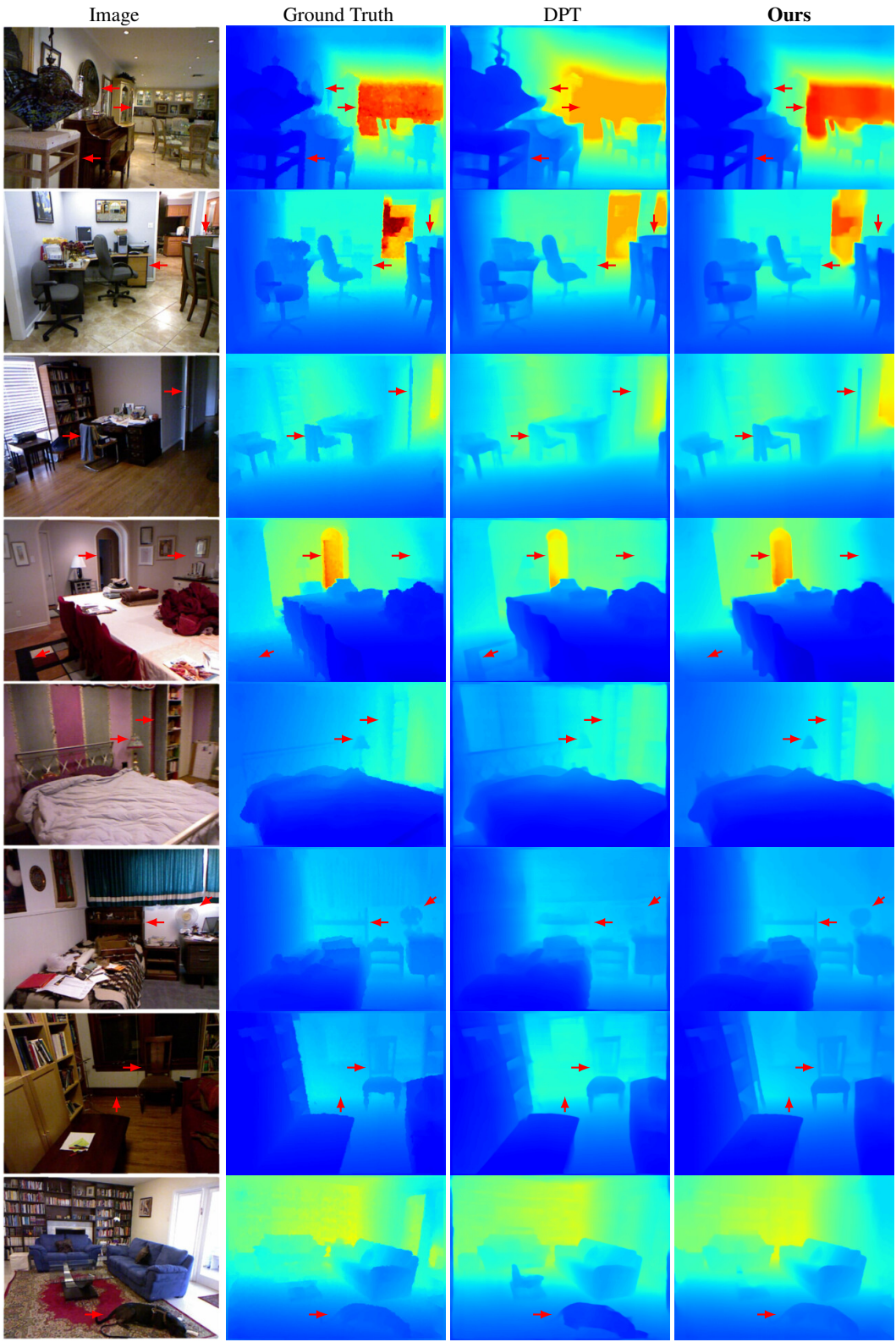

Figure \ref{fig:nyu_dpt} provides a qualitative comparison of our model with DPT-Hybrid \cite{dpt} finetuned on the NYU depth v2 \cite{nyu} dataset. 
The depth estimates of our diffusion model are more accurate both on coarse-scale scene structure (walls, floors, \etc) and on individual objects.

% \vspace*{-0.1cm}
\section{More samples for zero-shot imputation of depth}
\label{sec:supp-imputation}
\vspace*{-0.25cm}

Figure \ref{fig:text_to_3d_samples_extras} provides samples generated using our iterative text-to-3D pipeline. We note that such pipelines for iteratively generating 3D scenes have been previously proposed in literature \cite{infinite_nature_2020,3dphotos,wiles2020synsin}. However, these methods explicitly learn networks to refine the color \cite{infinite_nature_2020,3dphotos,wiles2020synsin} and the depth map \cite{infinite_nature_2020,3dphotos}. In contrast, we propose leveraging the text-conditioned image prior from existing large scale text-to-image \cite{sahariac-imagen} and text-conditional image completion \cite{ImagenEditor} models, and use our depth estimation model \textit{zero-shot} for depth completion. 
One caveat with our current approach of using the replacement method for conditional inference \cite{song2021scorebased} for imputing depth, is that it does not enable one to fix errors in the depth predicted in the previous step. One approach to fix artifacts would be by noising-denoising, like that used for coarse-to-fine refinement. We leave further exploration into this to future work.

\def\imwidth{0.095}
\def\imwidthlast{0.19}
\def\lastcoloffset{1.25cm}
\def\textpadding{3pt}
\def\ncols{9}

\begin{figure*}[h!]
\centering
\footnotesize
\renewcommand{\arraystretch}{0.2}
\addtolength{\tabcolsep}{-5pt}
\begin{tabular}{ccccccccc}

    \includegraphics[width=\imwidth\linewidth]{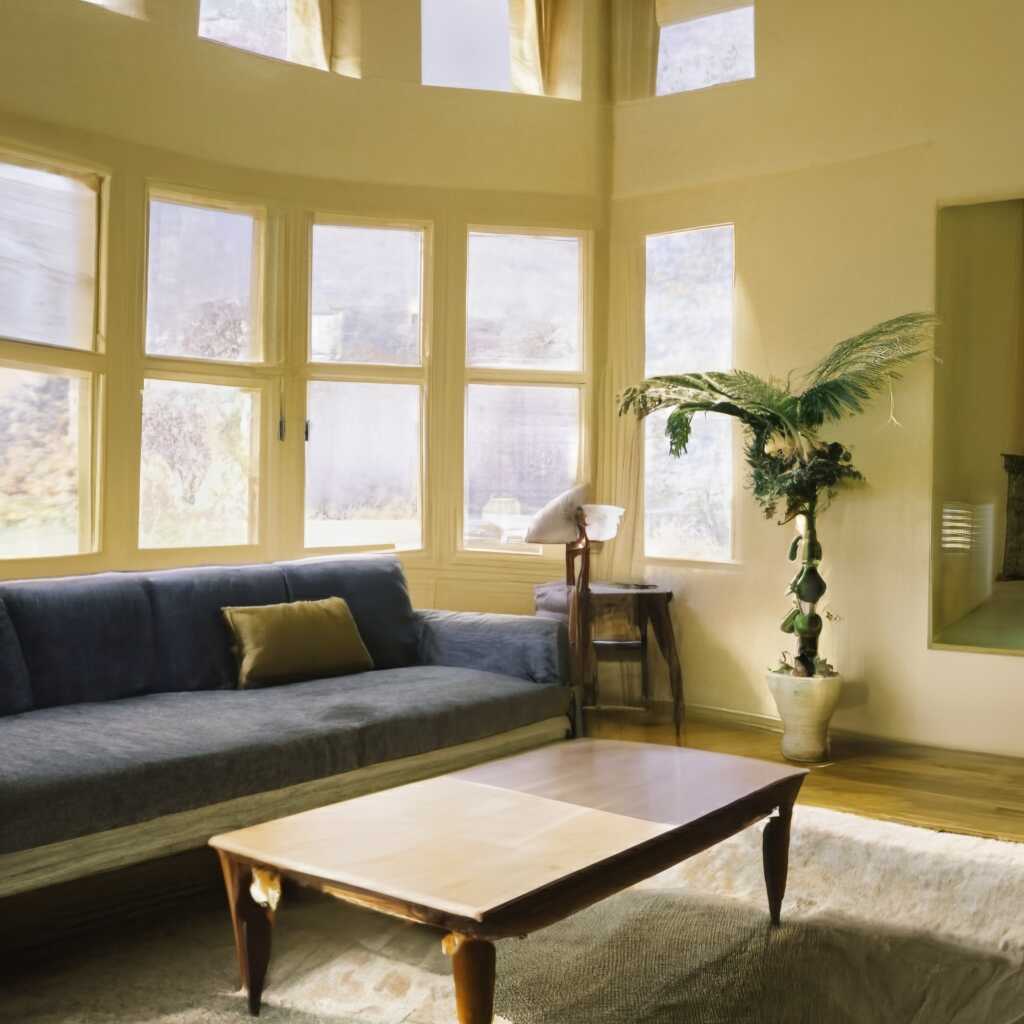} & \includegraphics[width=\imwidth\linewidth]{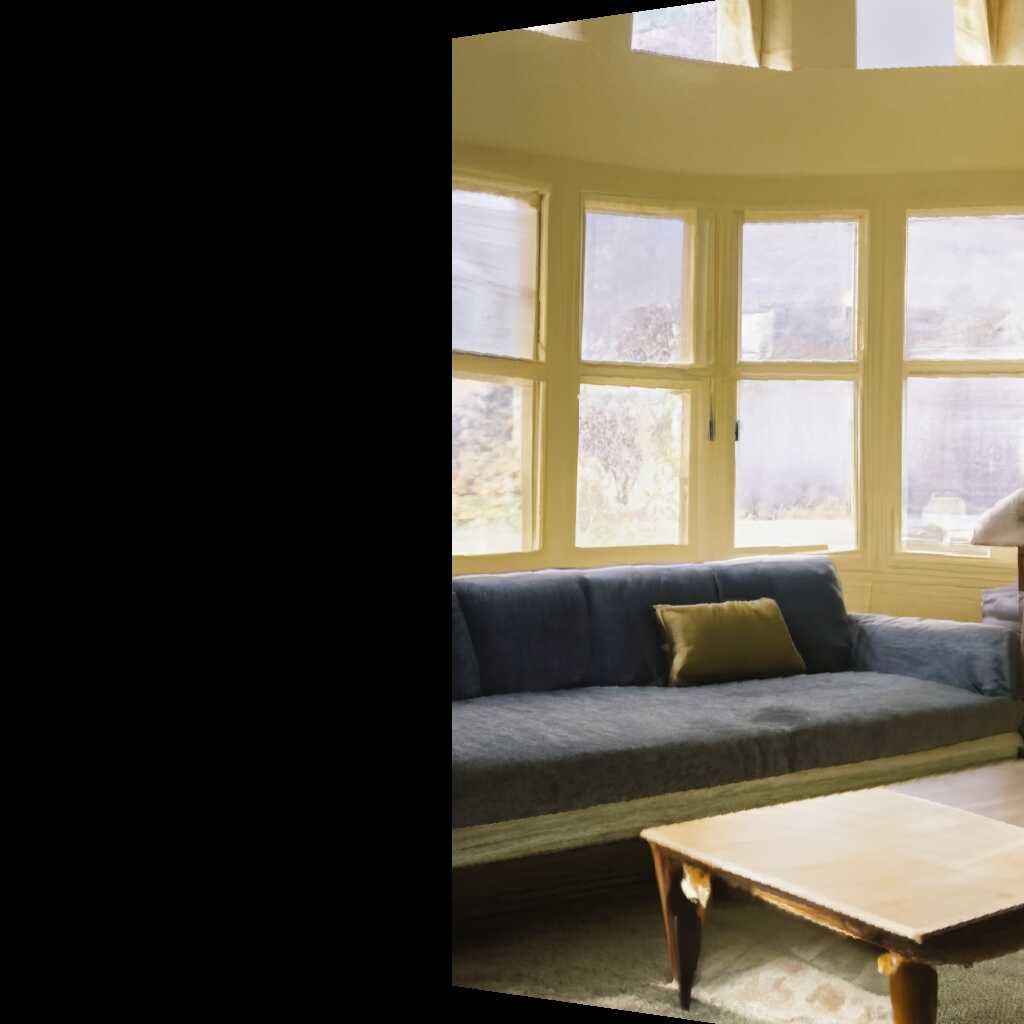} &\includegraphics[width=\imwidth\linewidth]{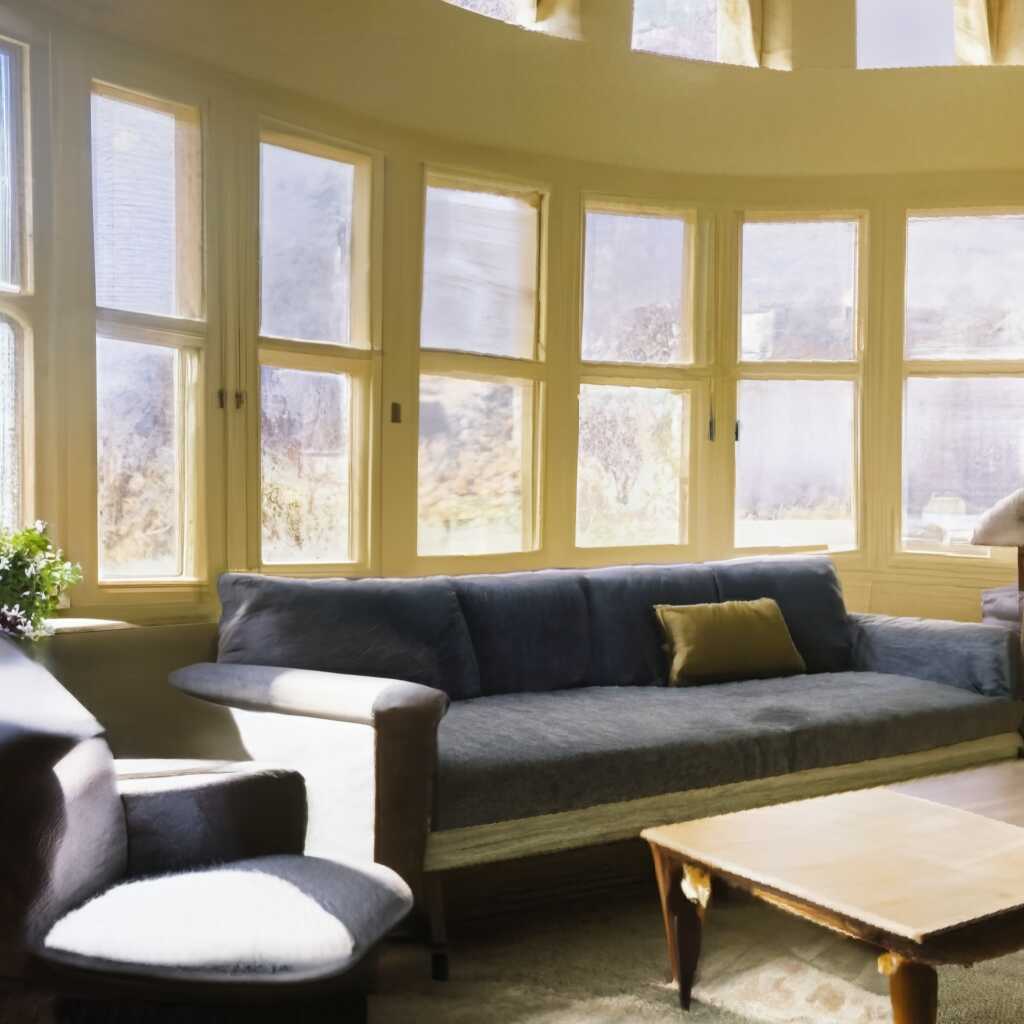} &\includegraphics[width=\imwidth\linewidth]{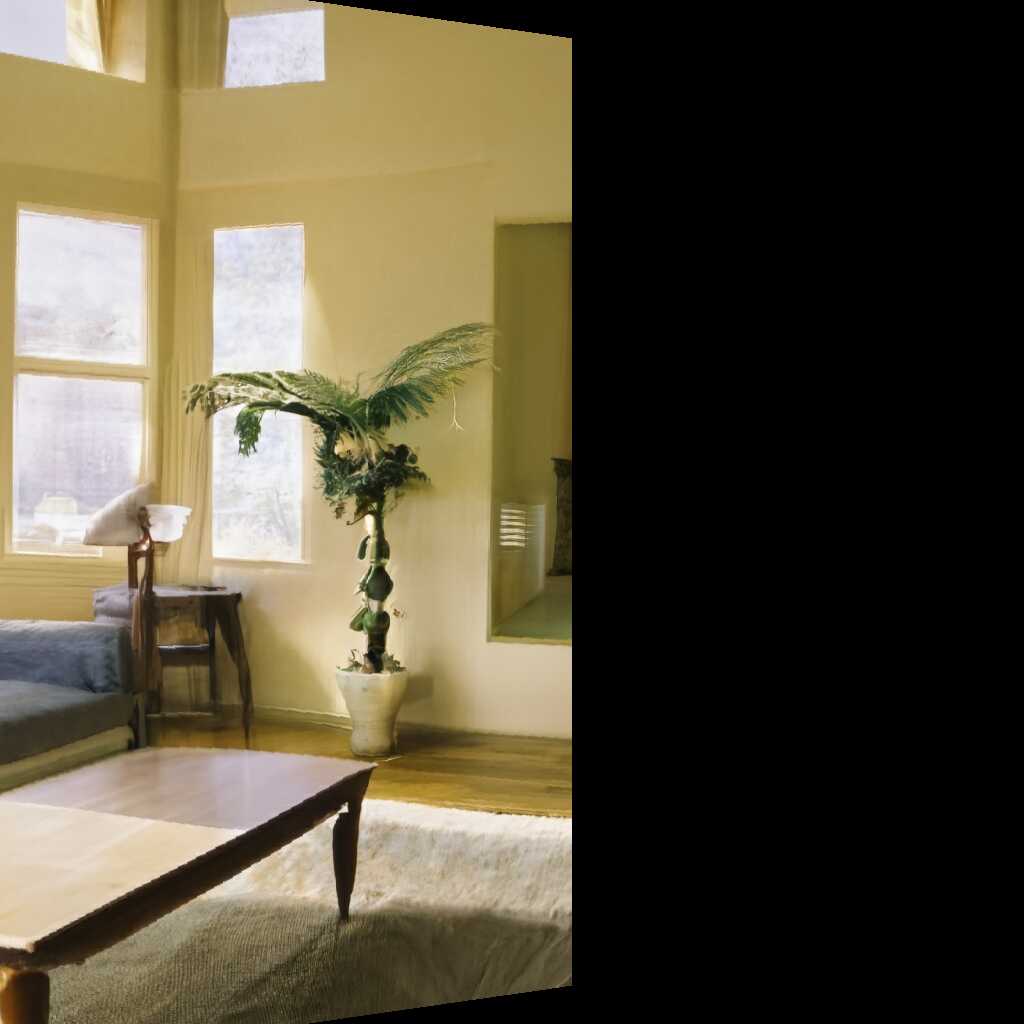} &\includegraphics[width=\imwidth\linewidth]{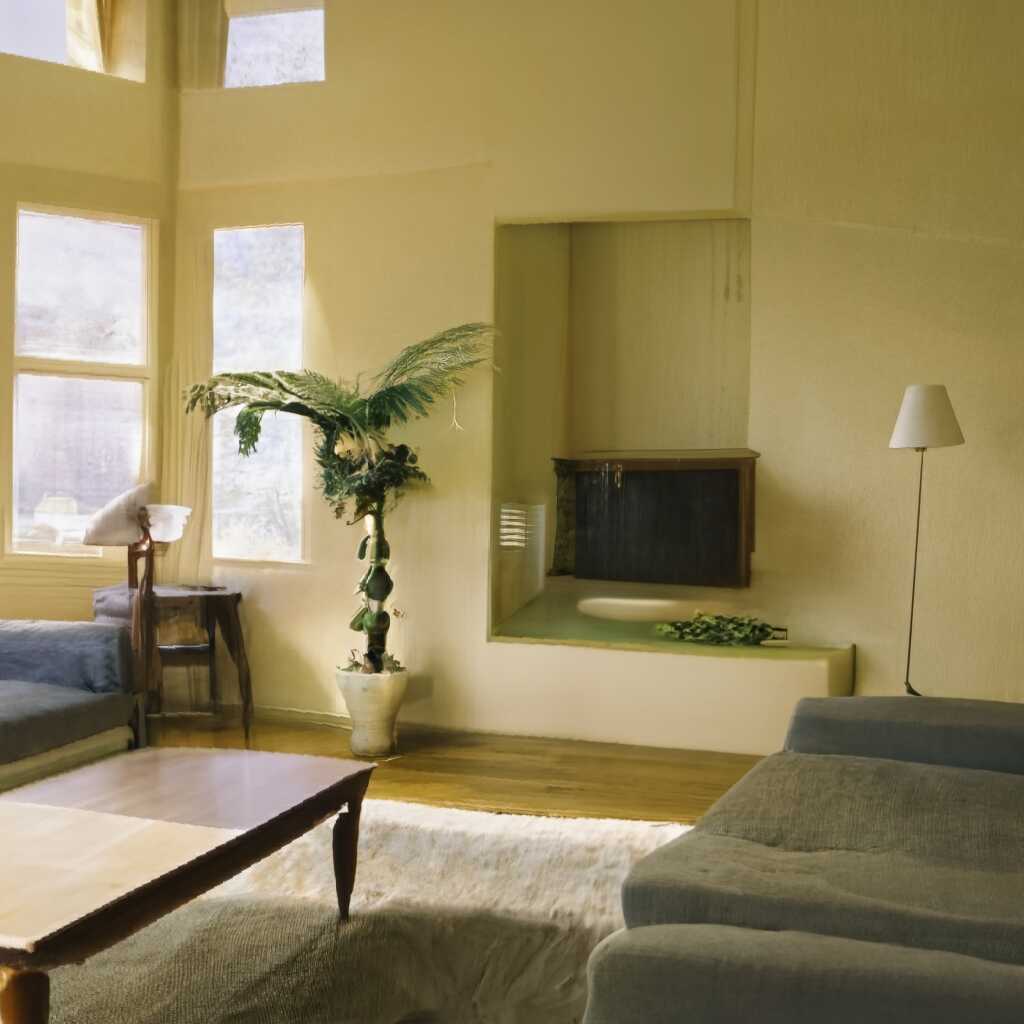} & \includegraphics[width=\imwidth\linewidth]{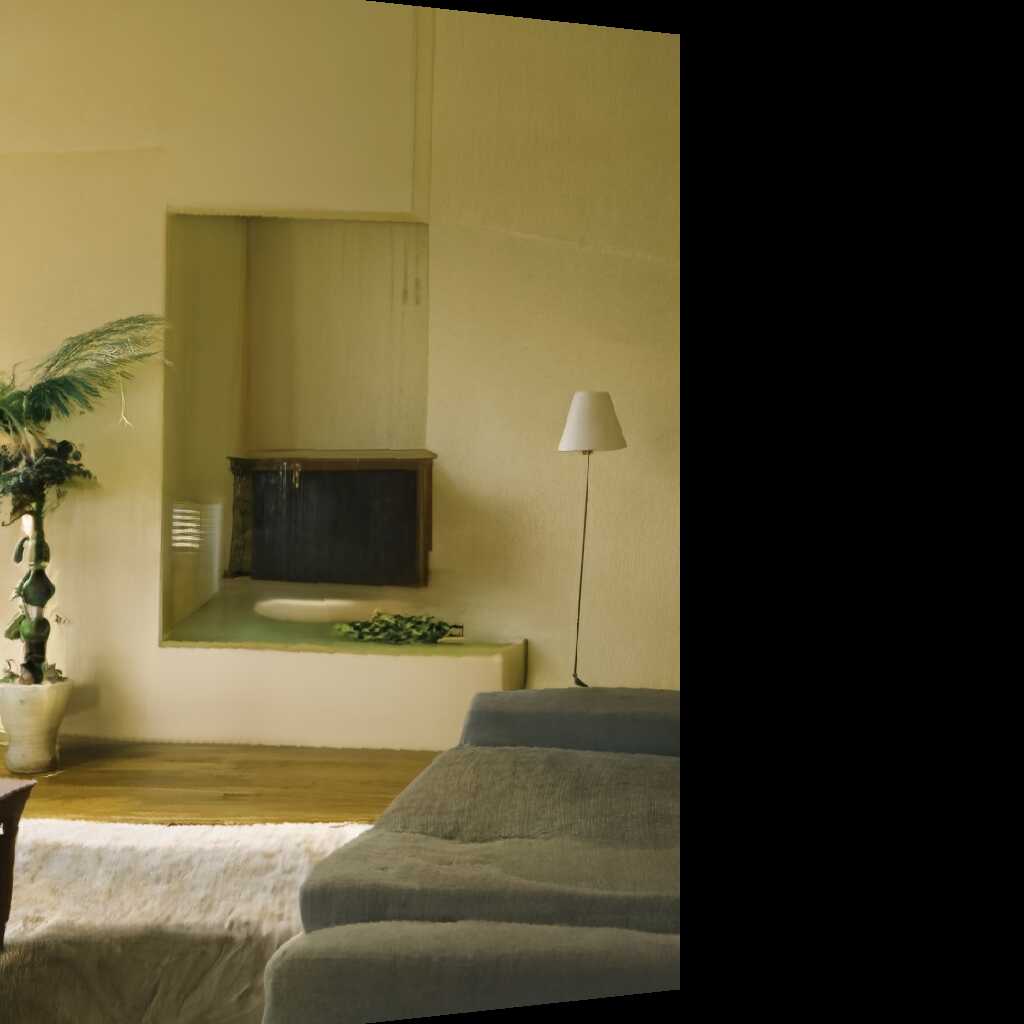} &\includegraphics[width=\imwidth\linewidth]{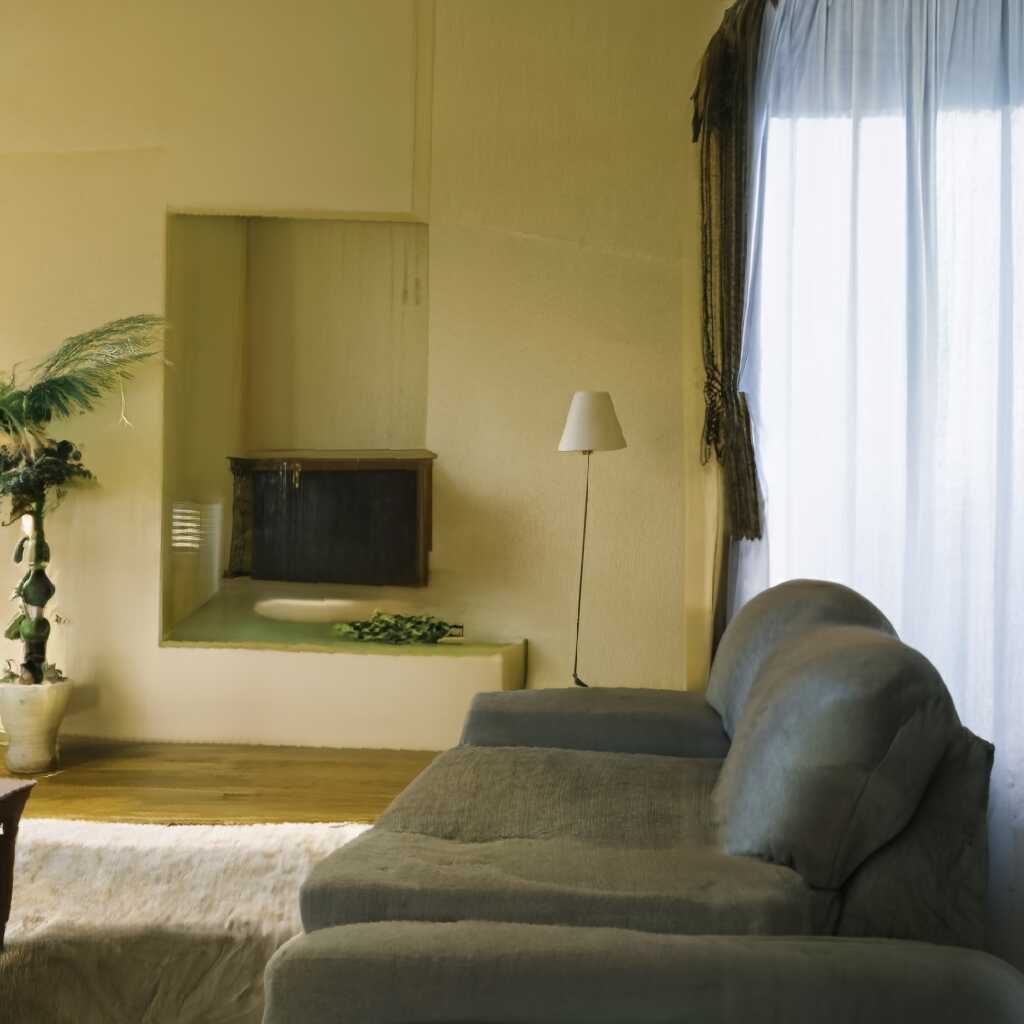} & \multicolumn{2}{c}{\multirow{2}{*}[\lastcoloffset]{\includegraphics[width=\imwidthlast\linewidth]{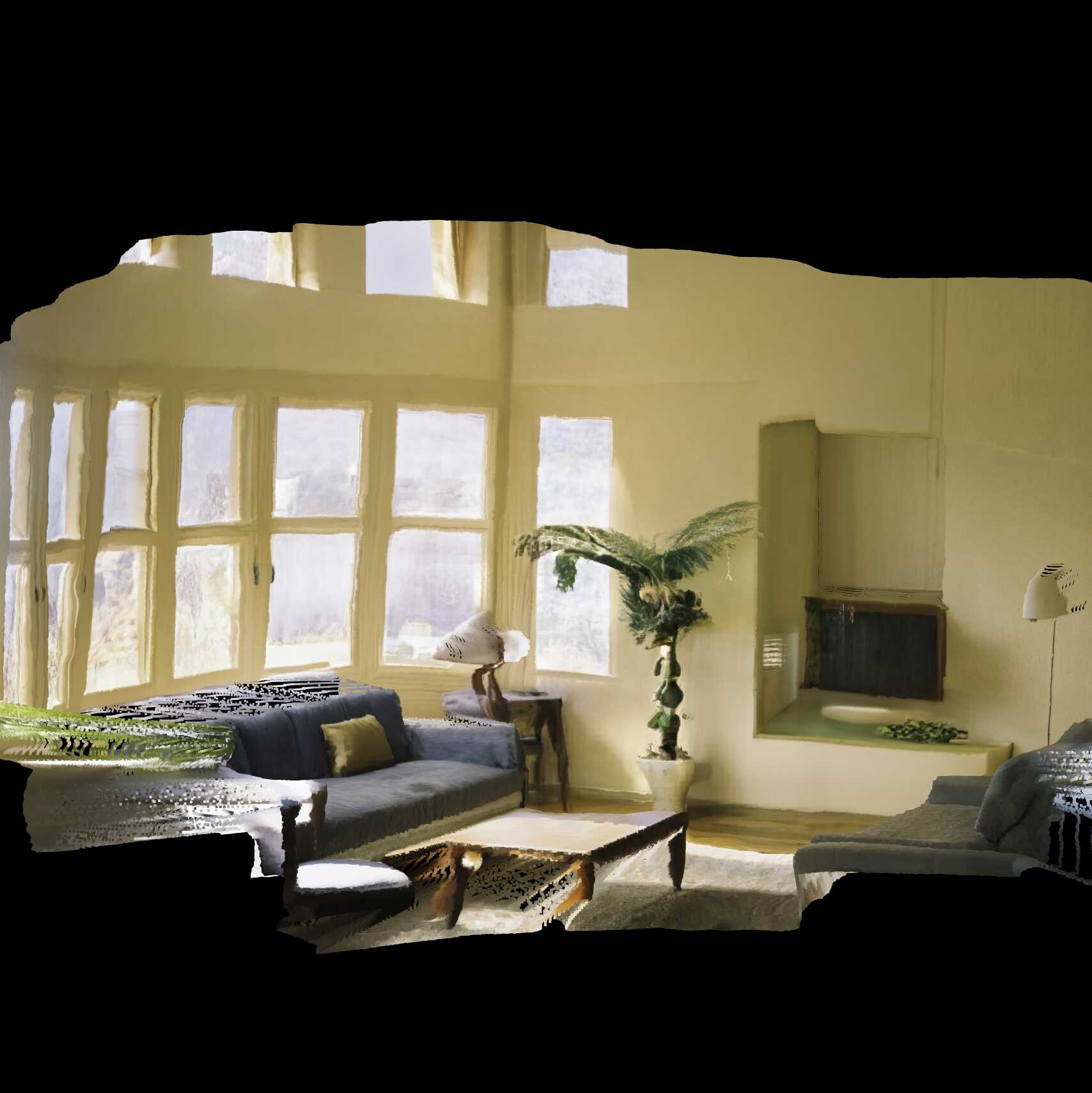}}} \\
    \includegraphics[width=\imwidth\linewidth]{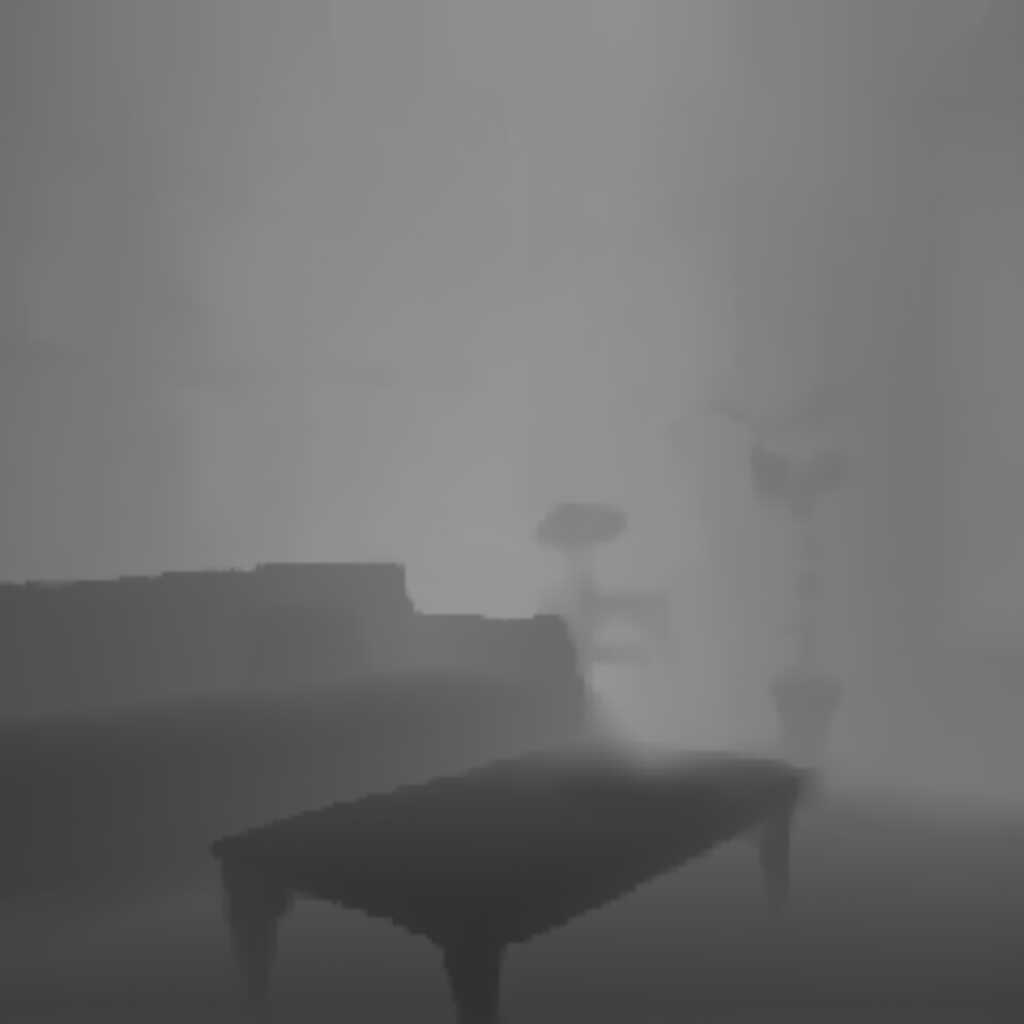} & \includegraphics[width=\imwidth\linewidth]{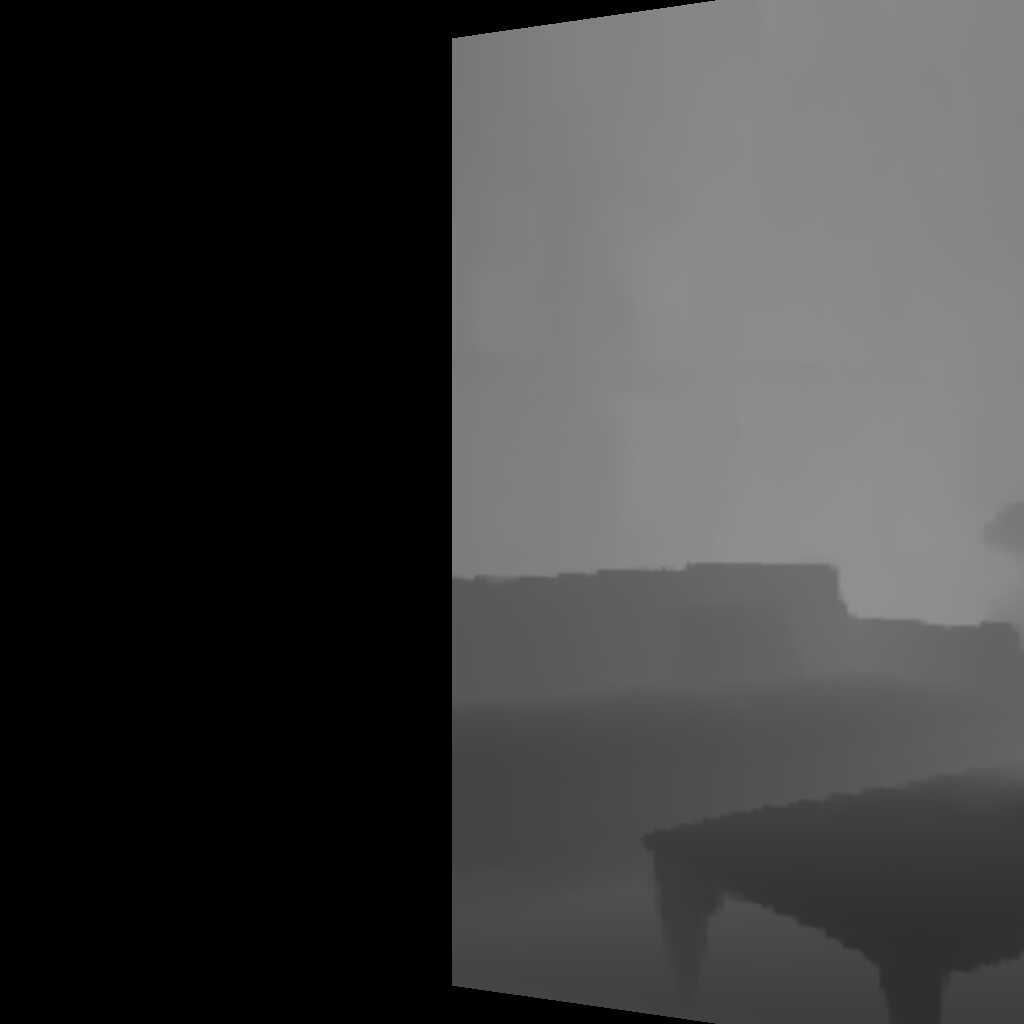} &\includegraphics[width=\imwidth\linewidth]{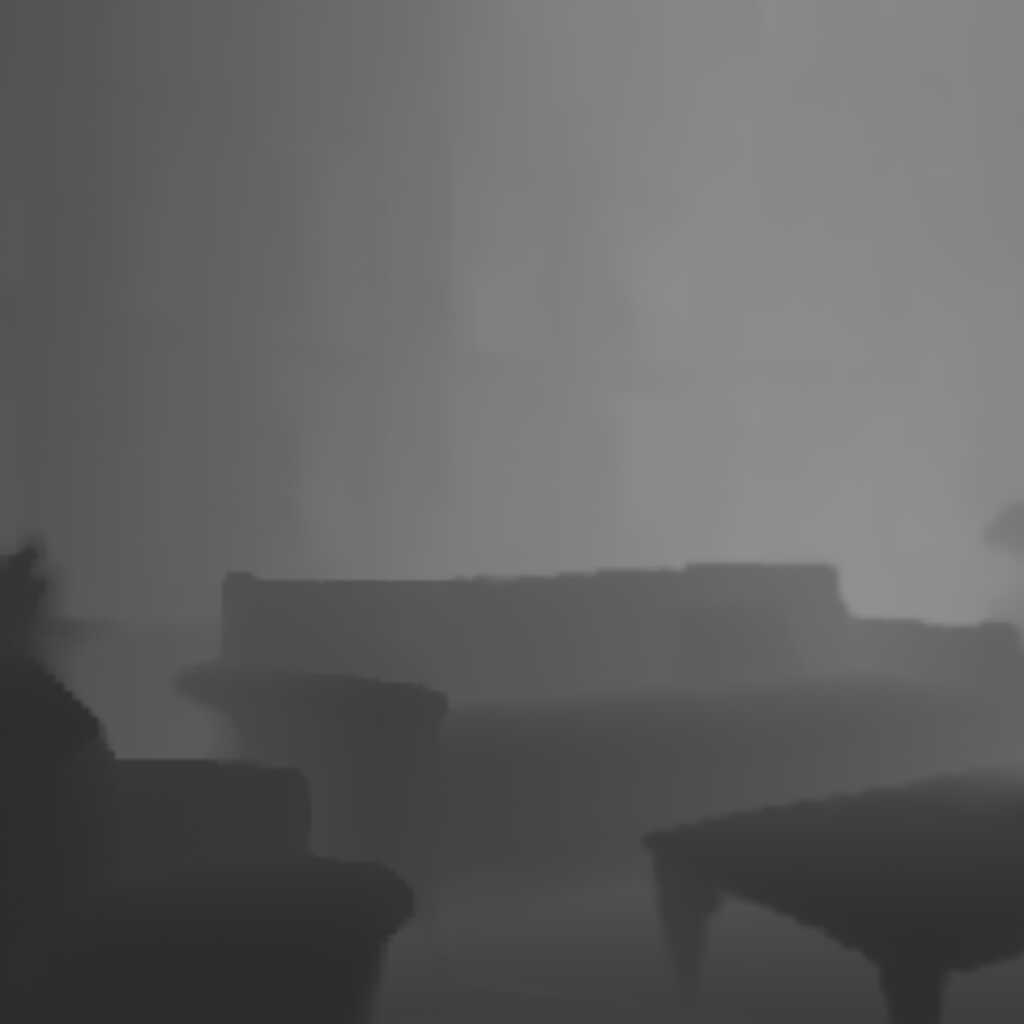} &  \includegraphics[width=\imwidth\linewidth]{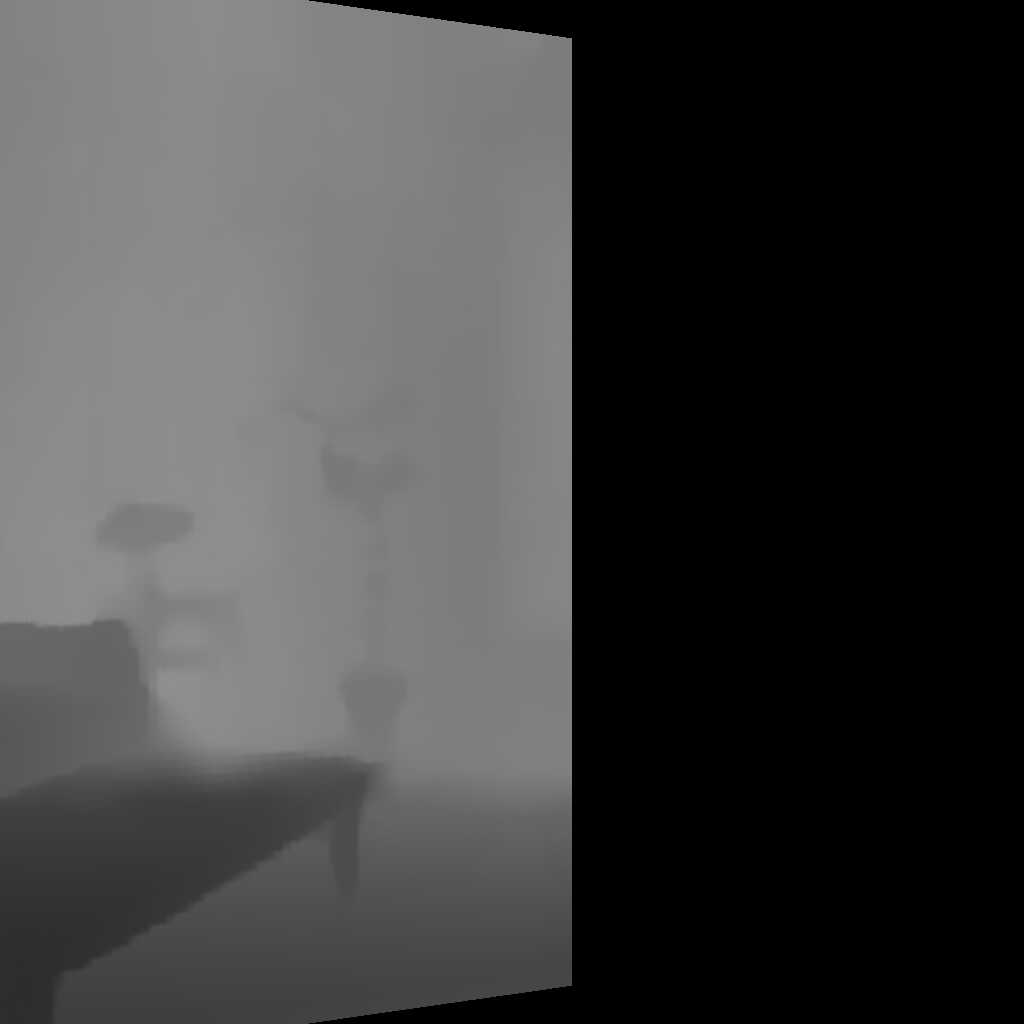} &\includegraphics[width=\imwidth\linewidth]{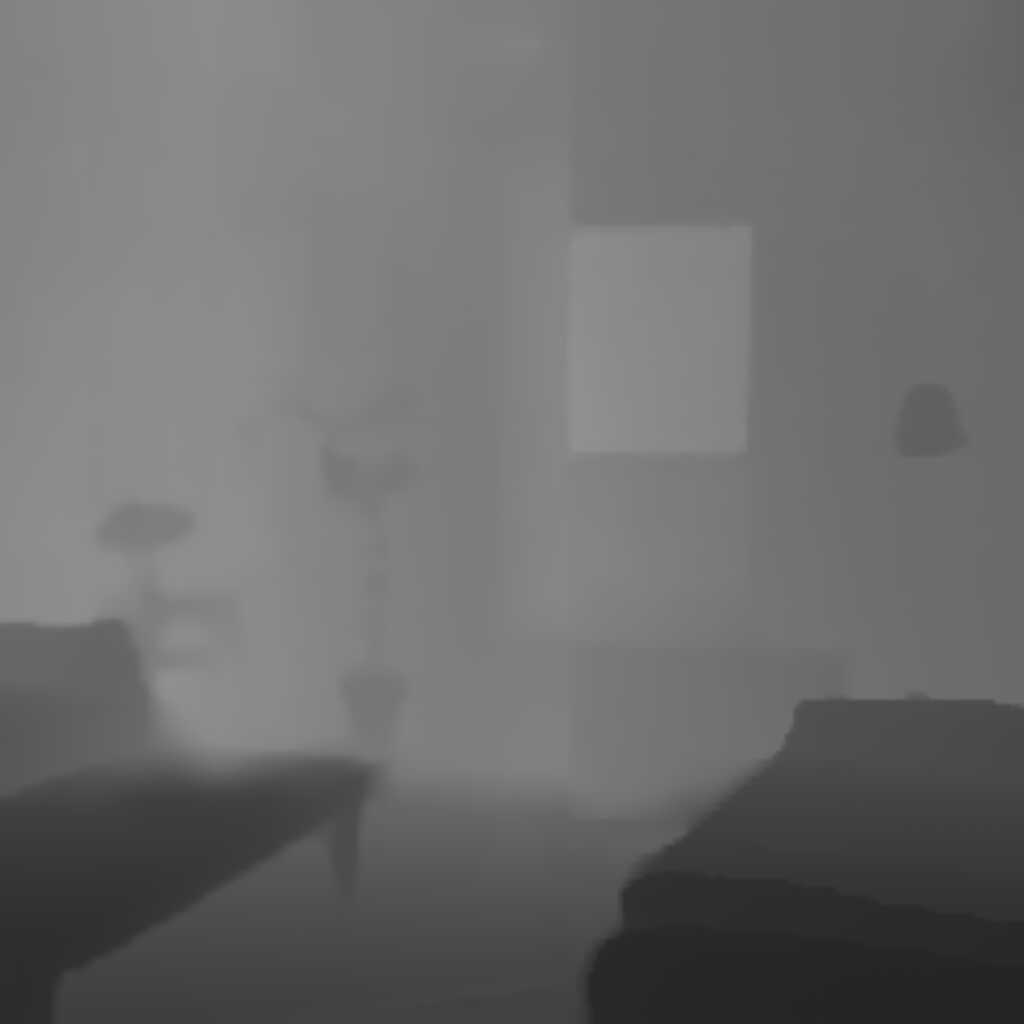} &\includegraphics[width=\imwidth\linewidth]{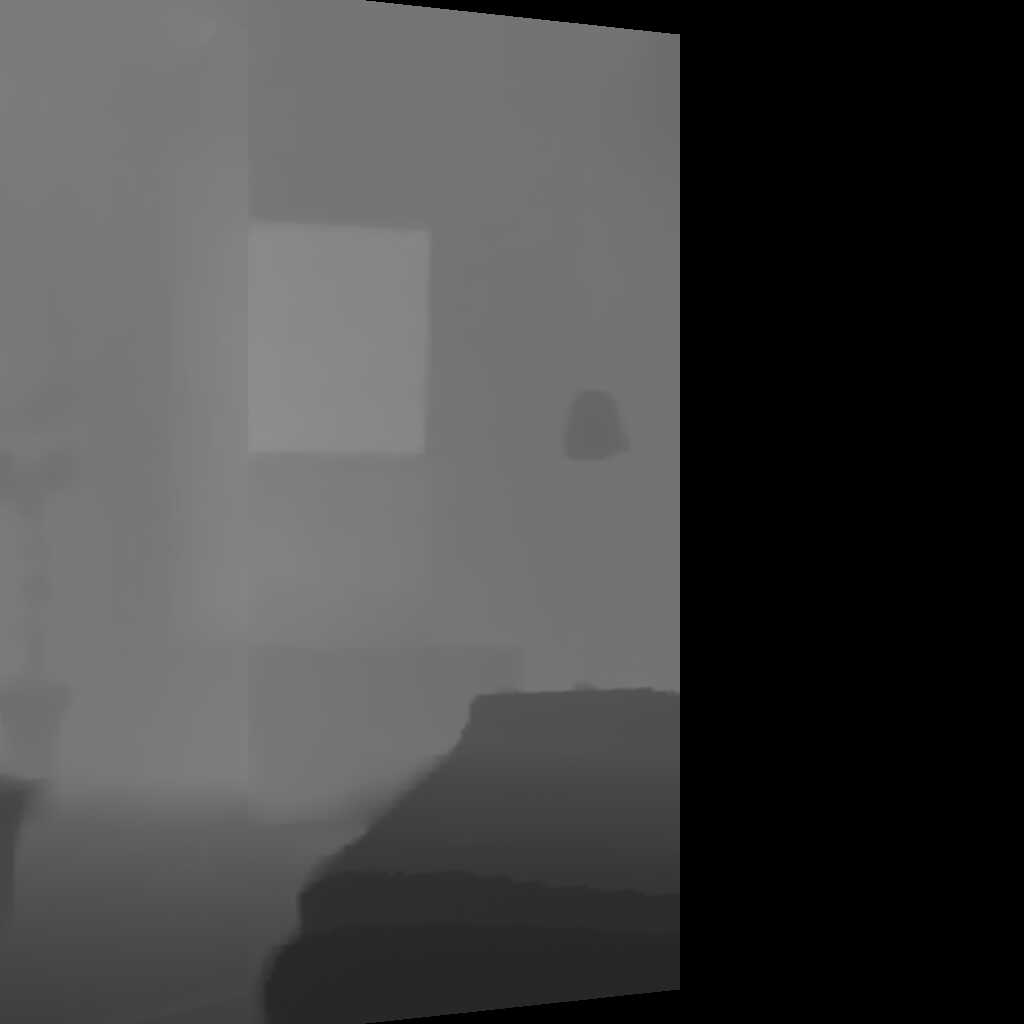} &\includegraphics[width=\imwidth\linewidth]{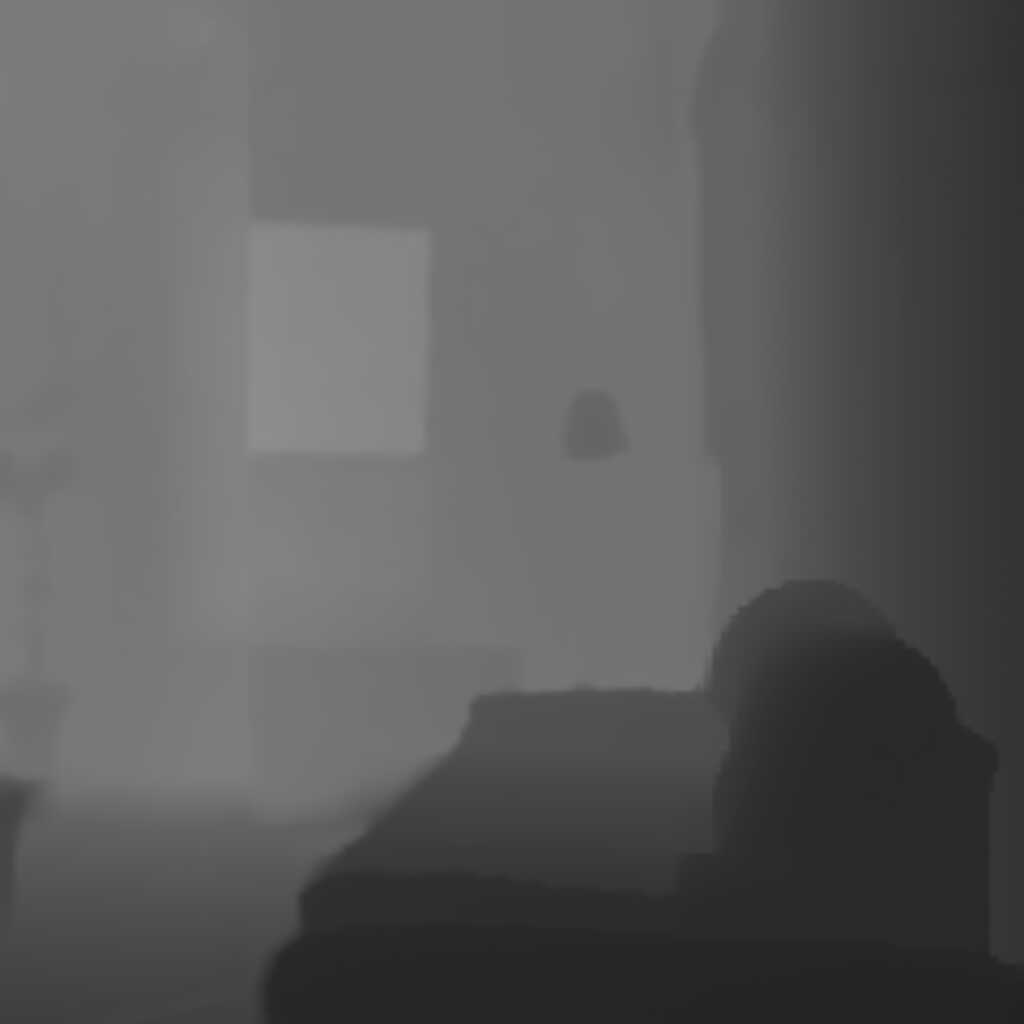} & \multicolumn{2}{c}{} \\
    \\[\textpadding]\multicolumn{\ncols}{c}{\footnotesize{Prompt: A living room.}}\\\\[\textpadding] \\

    \includegraphics[width=\imwidth\linewidth]{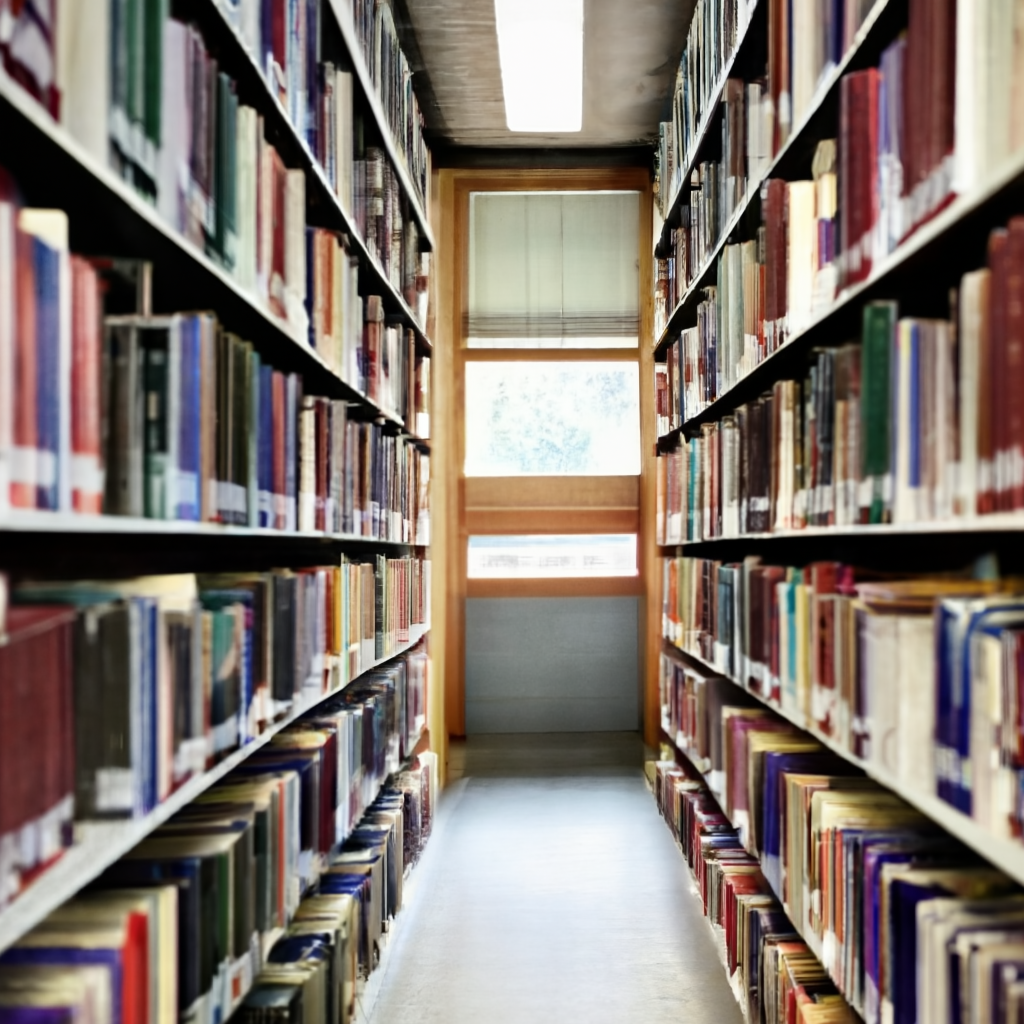} & \includegraphics[width=\imwidth\linewidth]{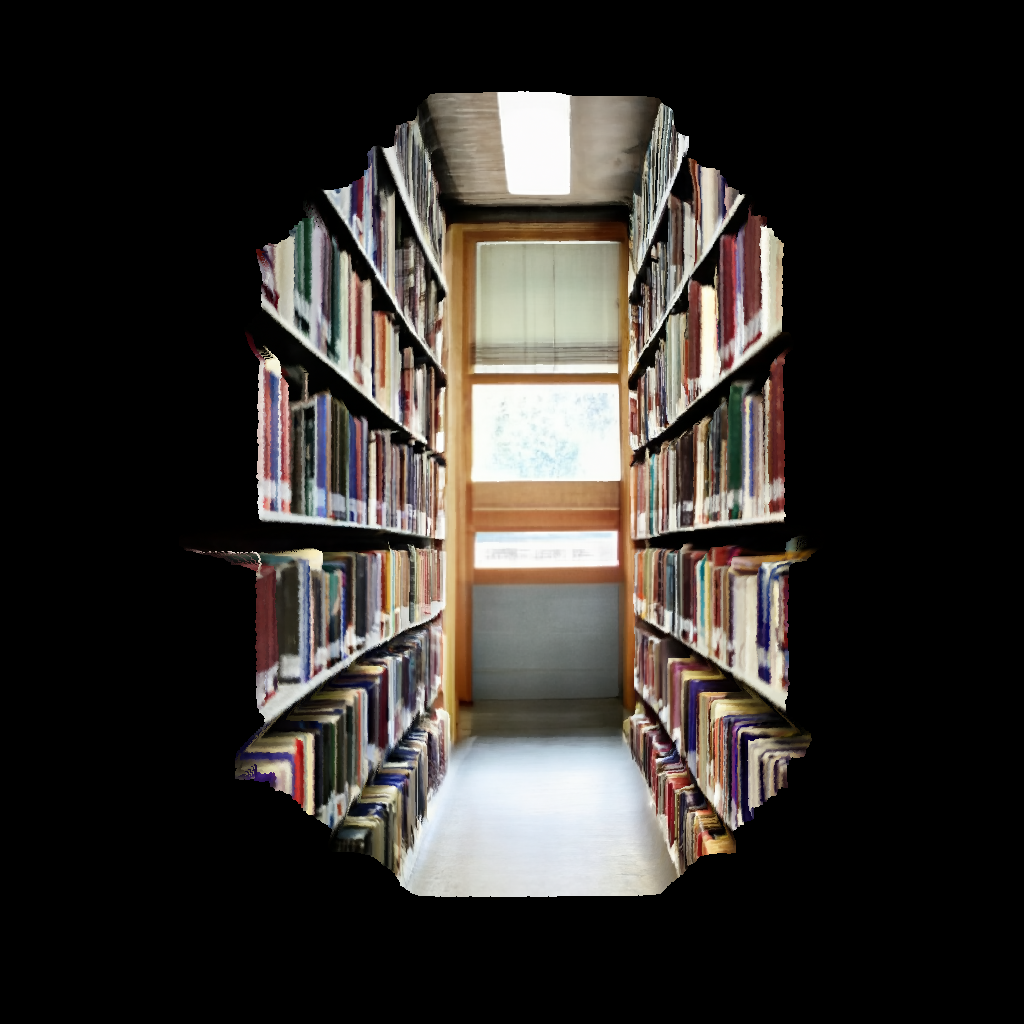} &\includegraphics[width=\imwidth\linewidth]{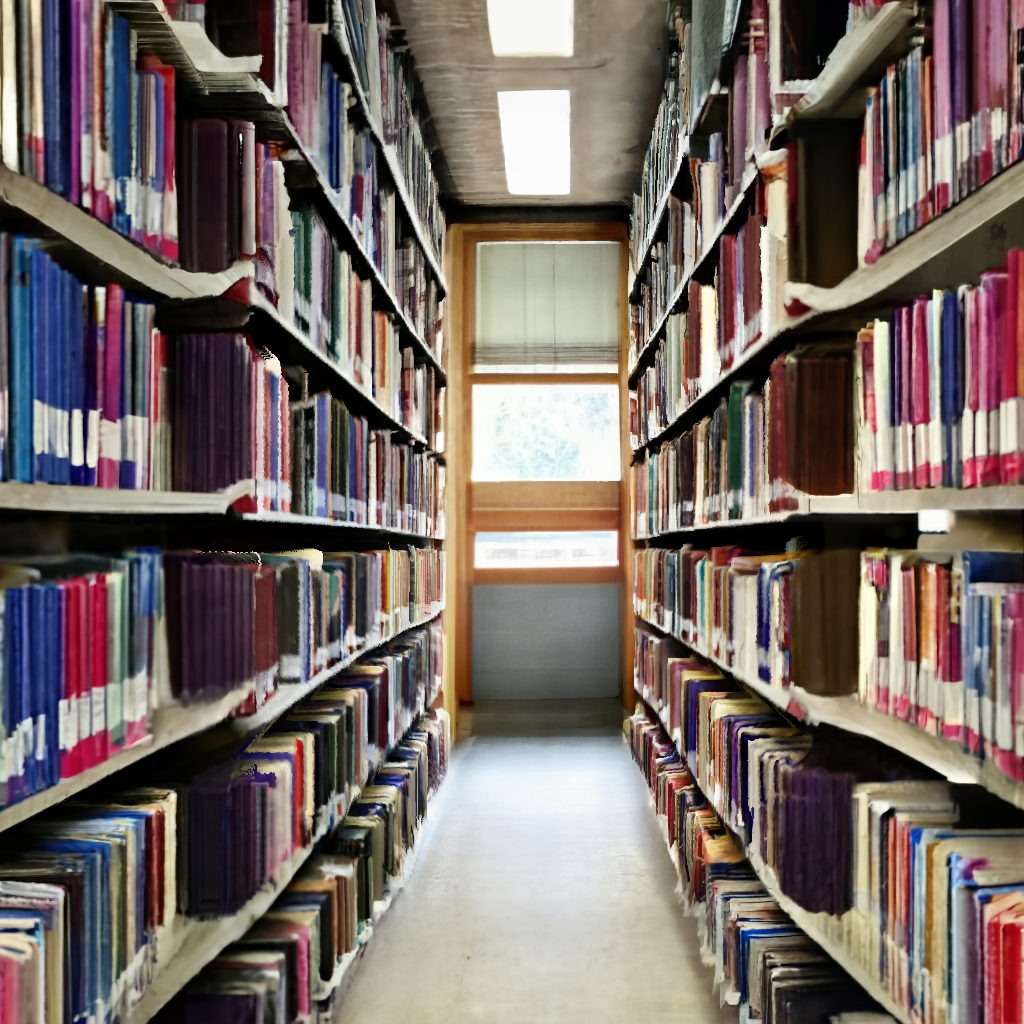} &\includegraphics[width=\imwidth\linewidth]{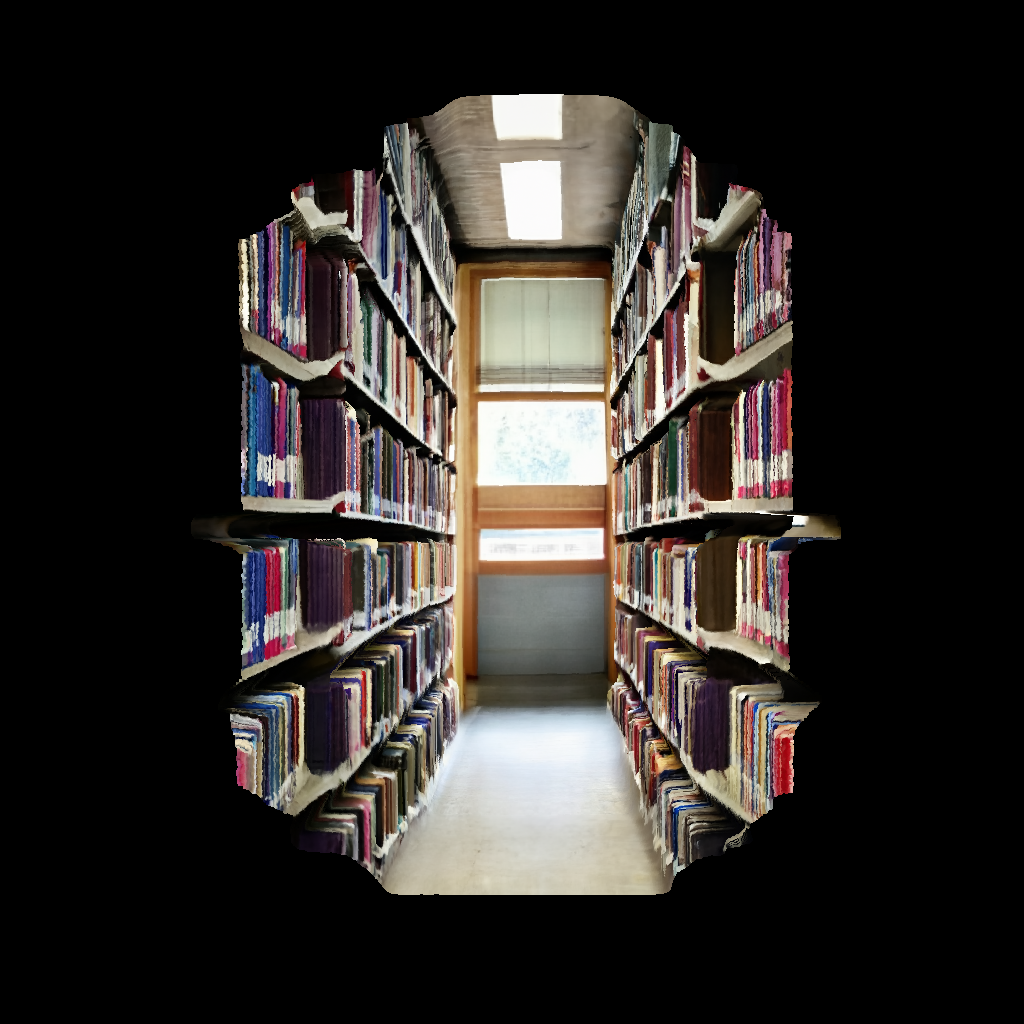} &\includegraphics[width=\imwidth\linewidth]{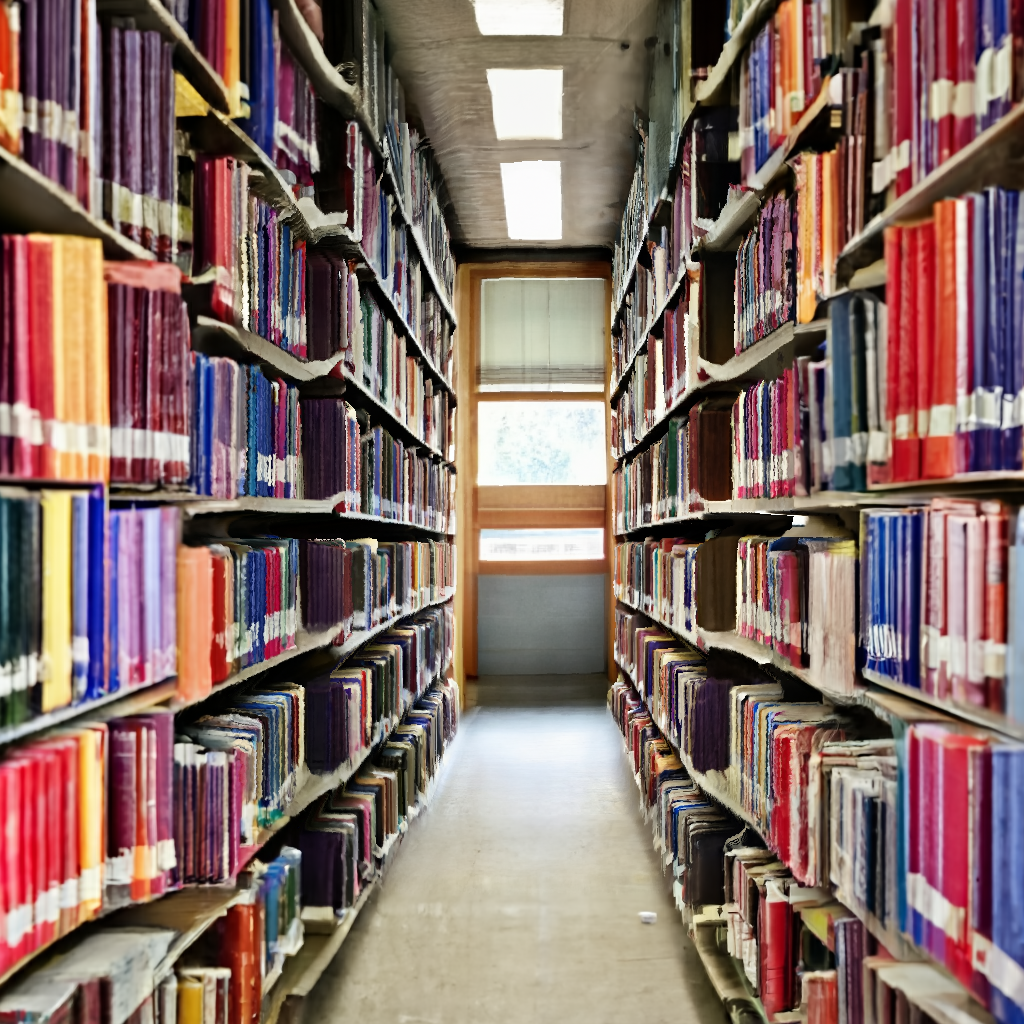} & \includegraphics[width=\imwidth\linewidth]{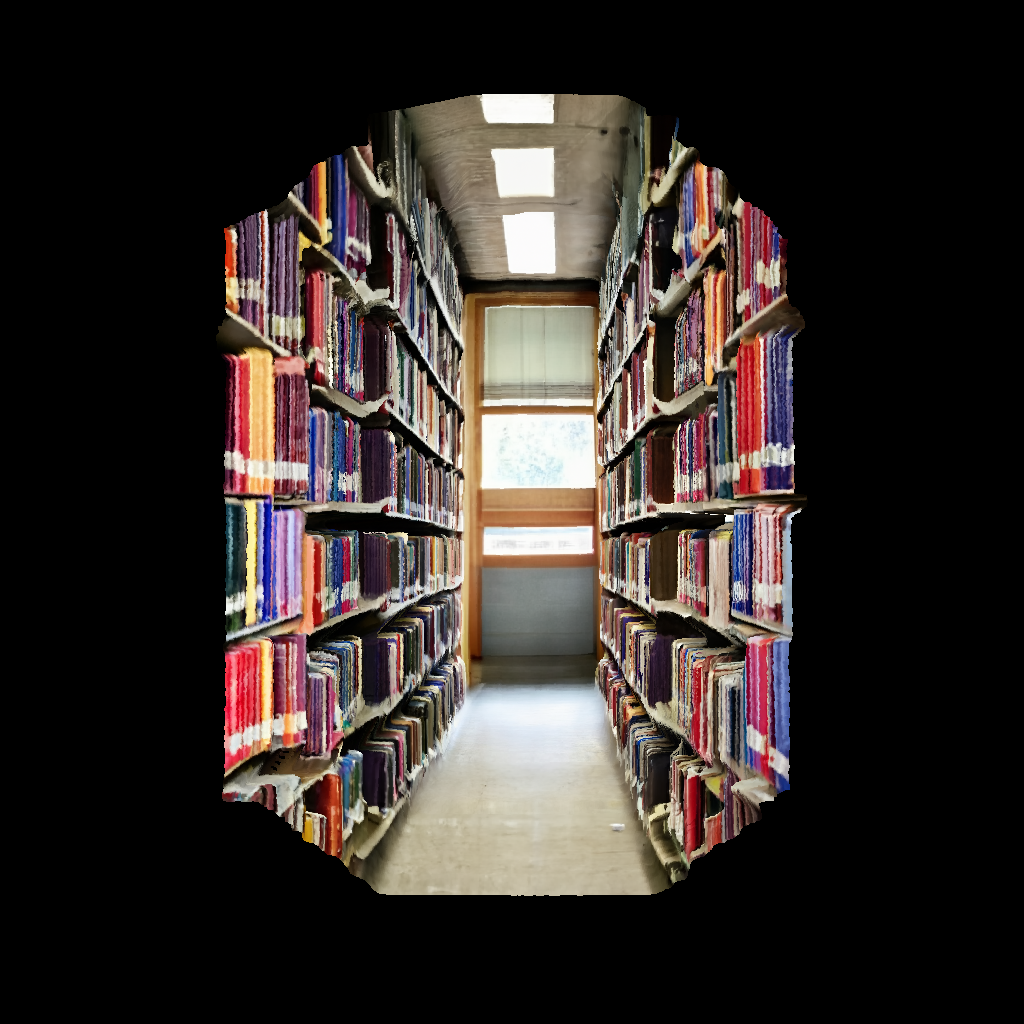} &\includegraphics[width=\imwidth\linewidth]{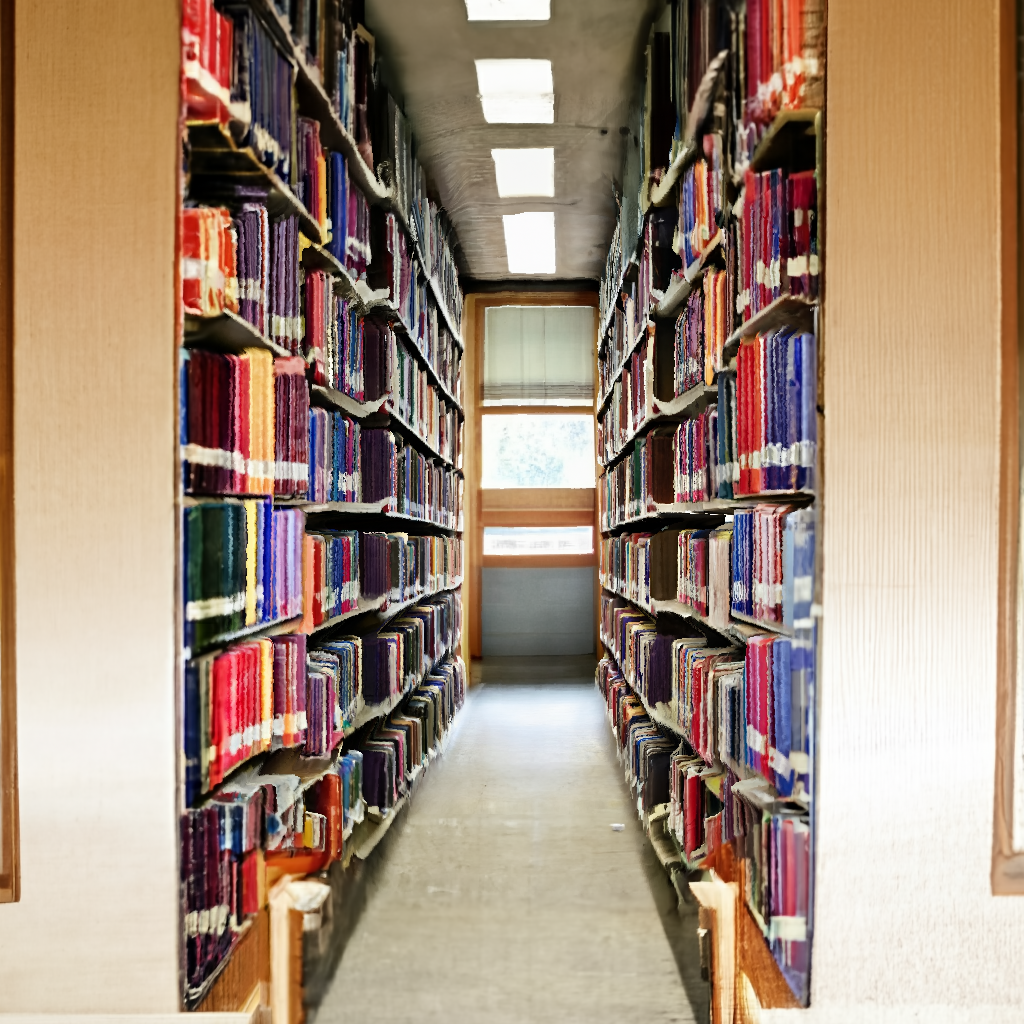} & \multicolumn{2}{c}{\multirow{2}{*}[\lastcoloffset]{\includegraphics[width=\imwidthlast\linewidth]{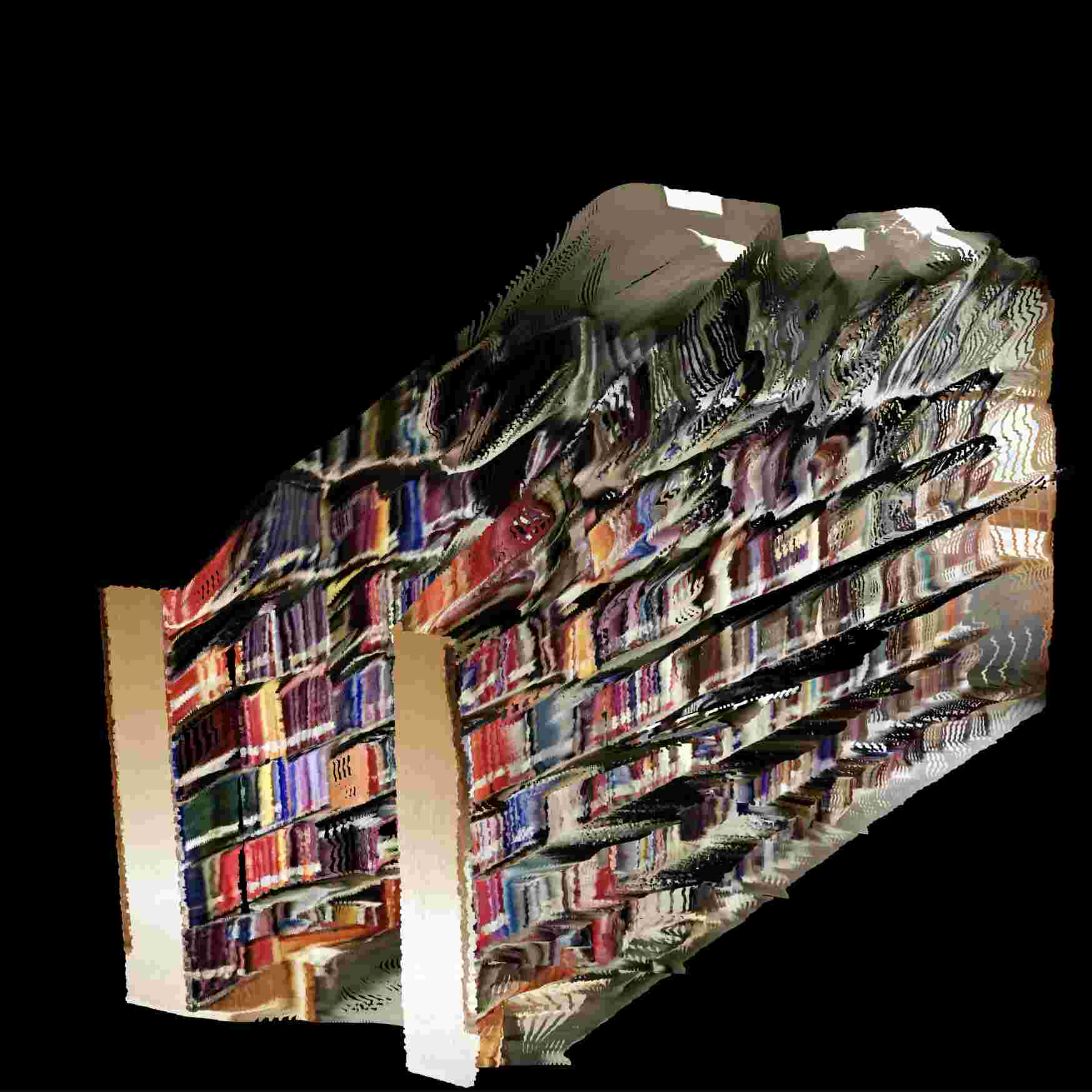}}} \\
    \includegraphics[width=\imwidth\linewidth]{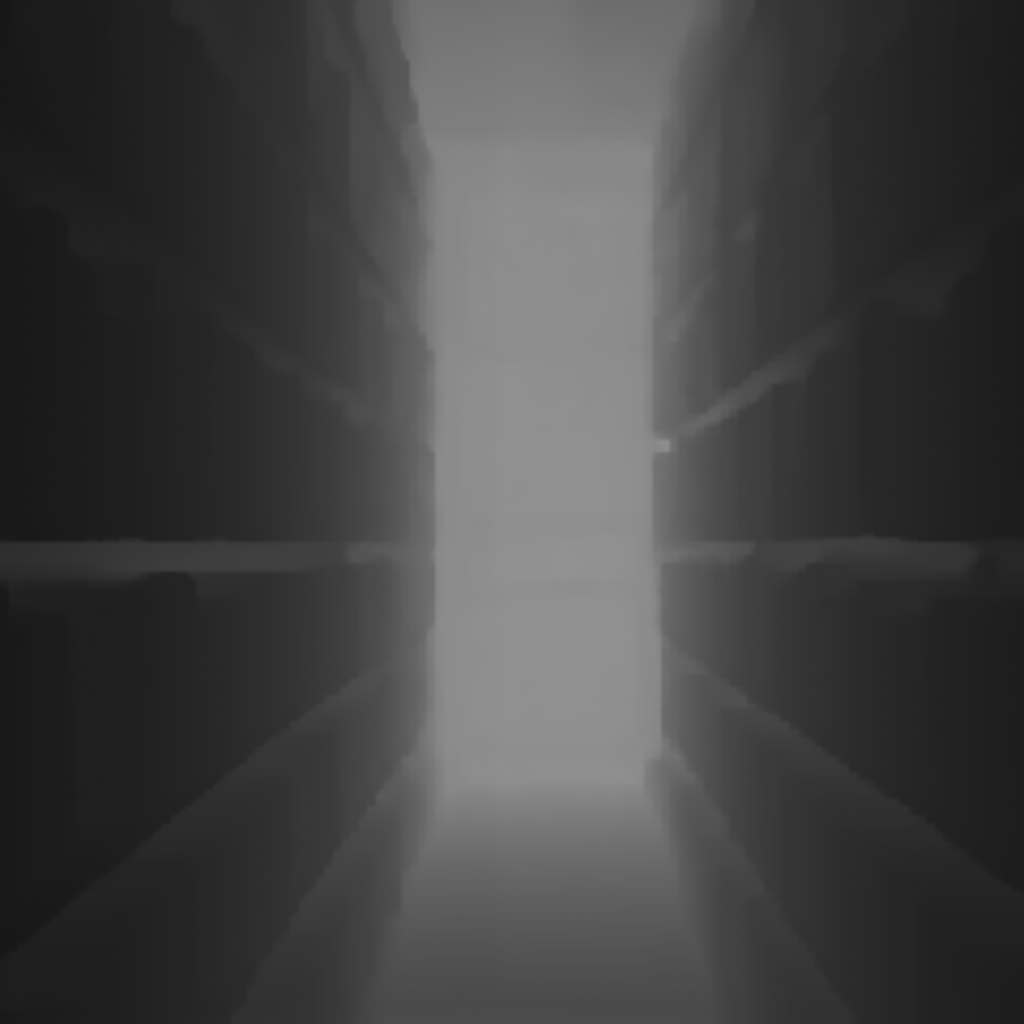} & \includegraphics[width=\imwidth\linewidth]{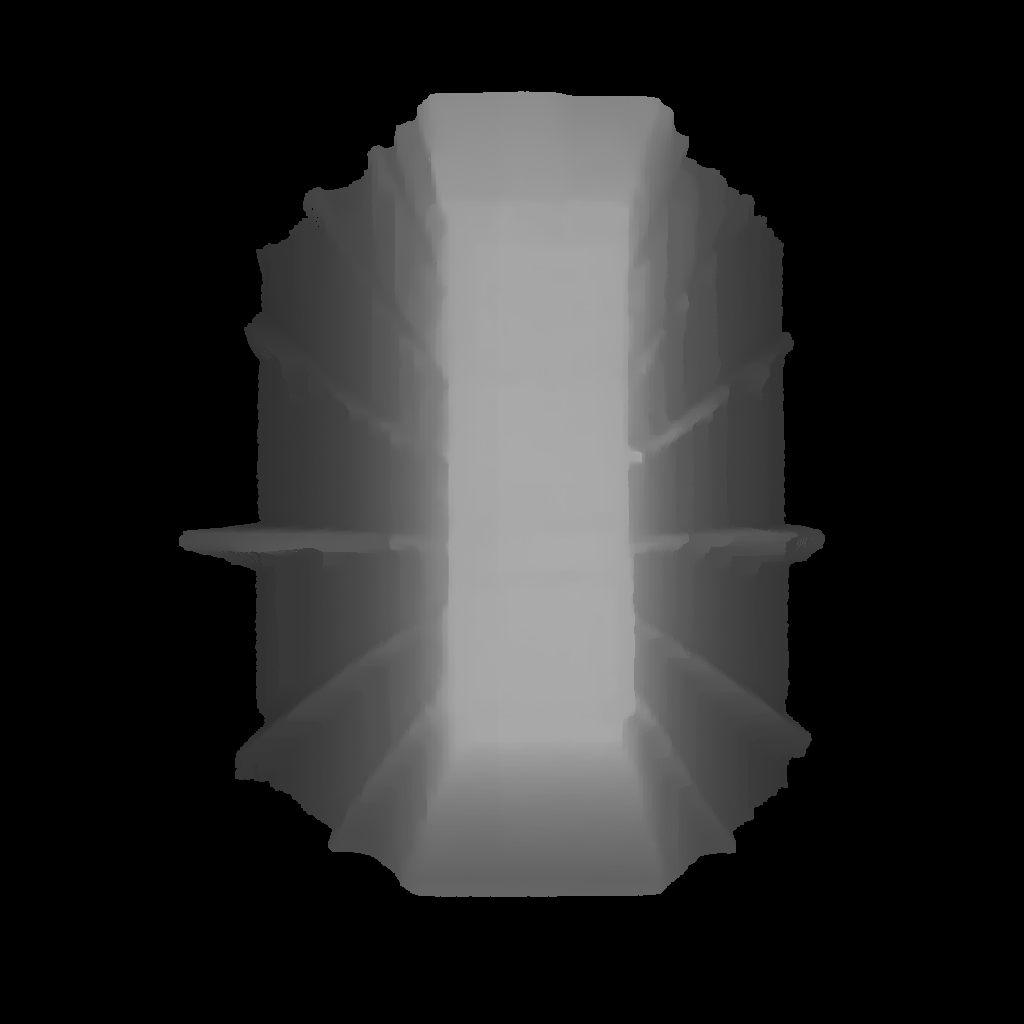} &\includegraphics[width=\imwidth\linewidth]{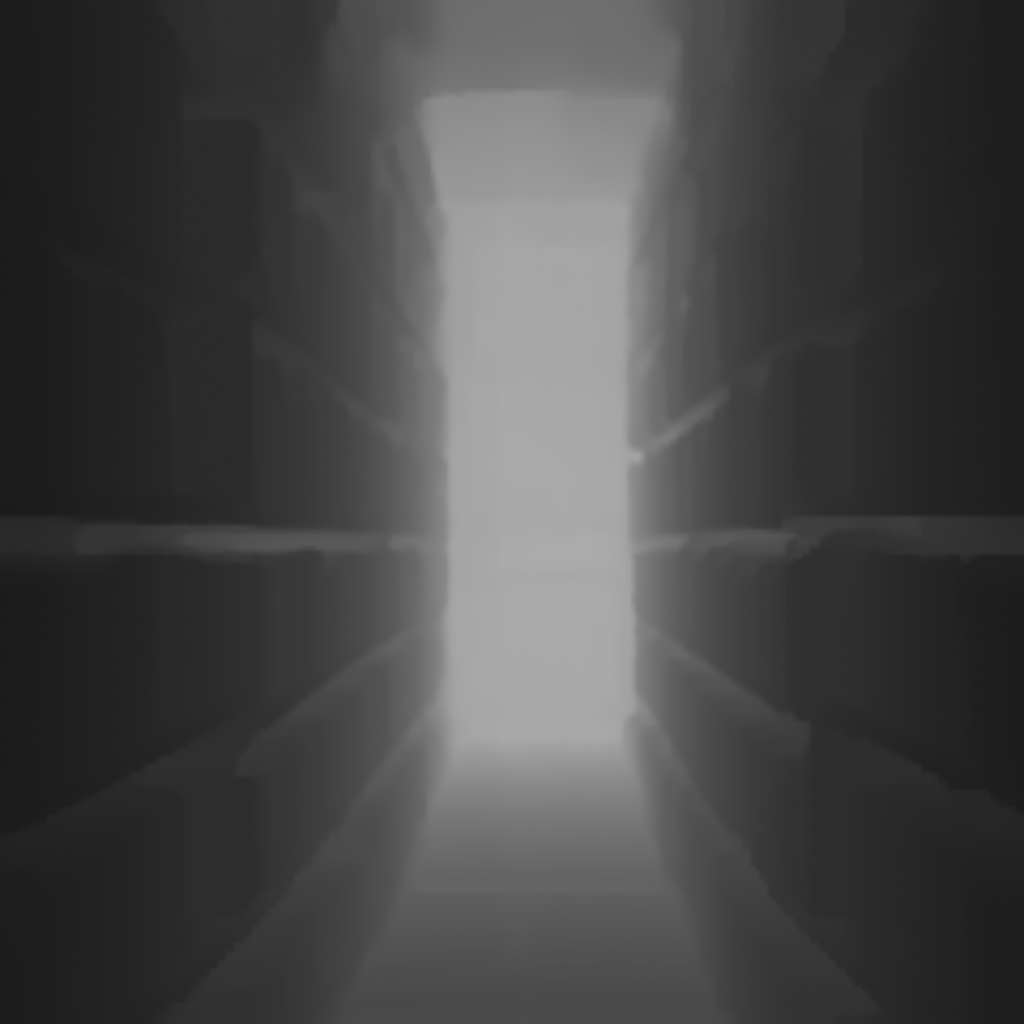} &  \includegraphics[width=\imwidth\linewidth]{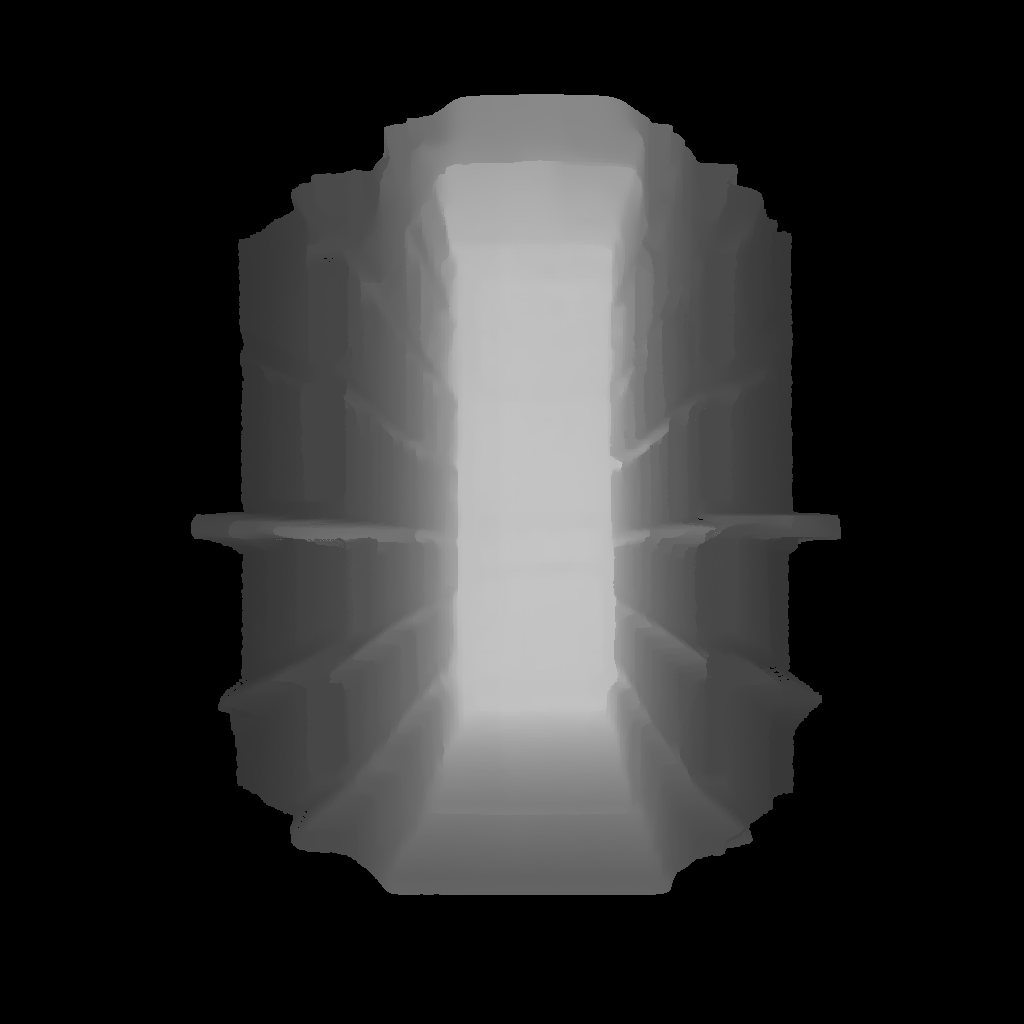} &\includegraphics[width=\imwidth\linewidth]{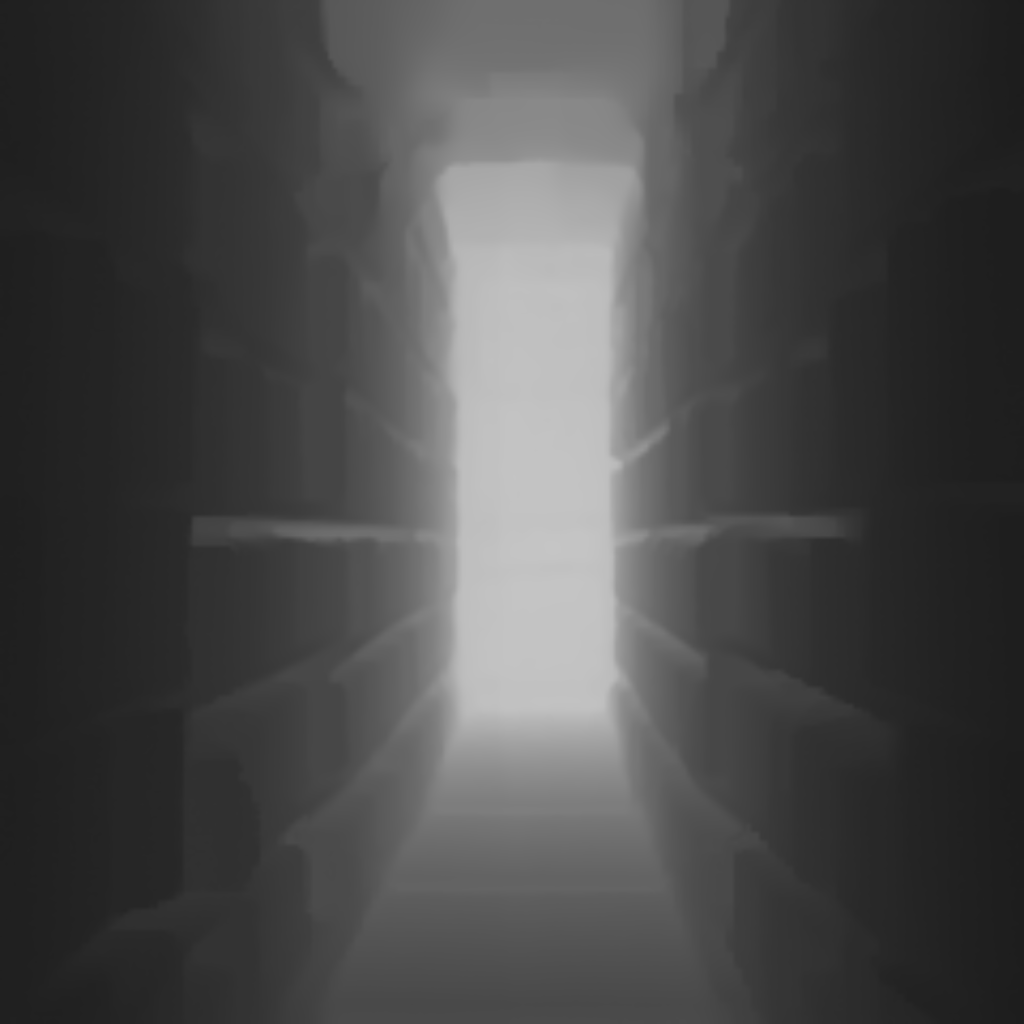} &\includegraphics[width=\imwidth\linewidth]{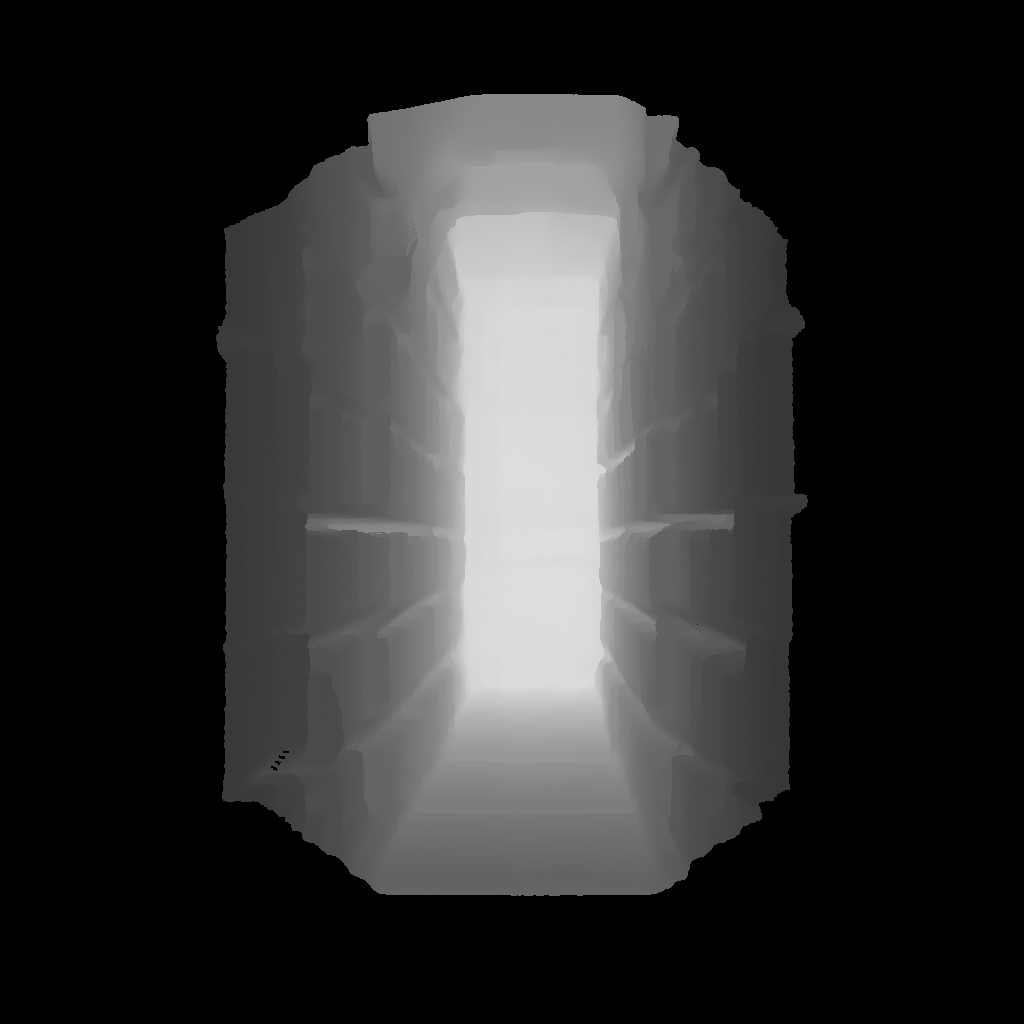} &\includegraphics[width=\imwidth\linewidth]{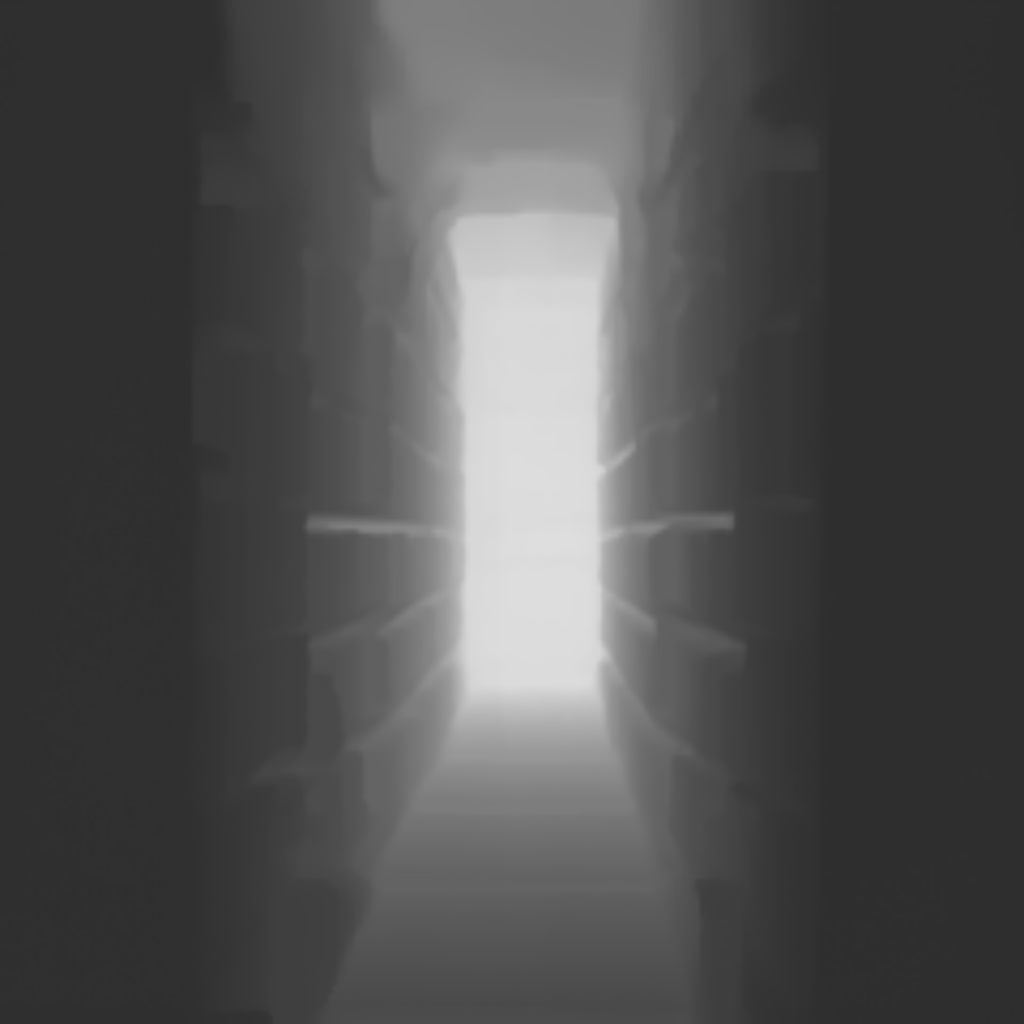} & \multicolumn{2}{c}{} \\
    \\[\textpadding]\multicolumn{\ncols}{c}{Prompt: A library.}\\\\[\textpadding]

    \includegraphics[width=\imwidth\linewidth]{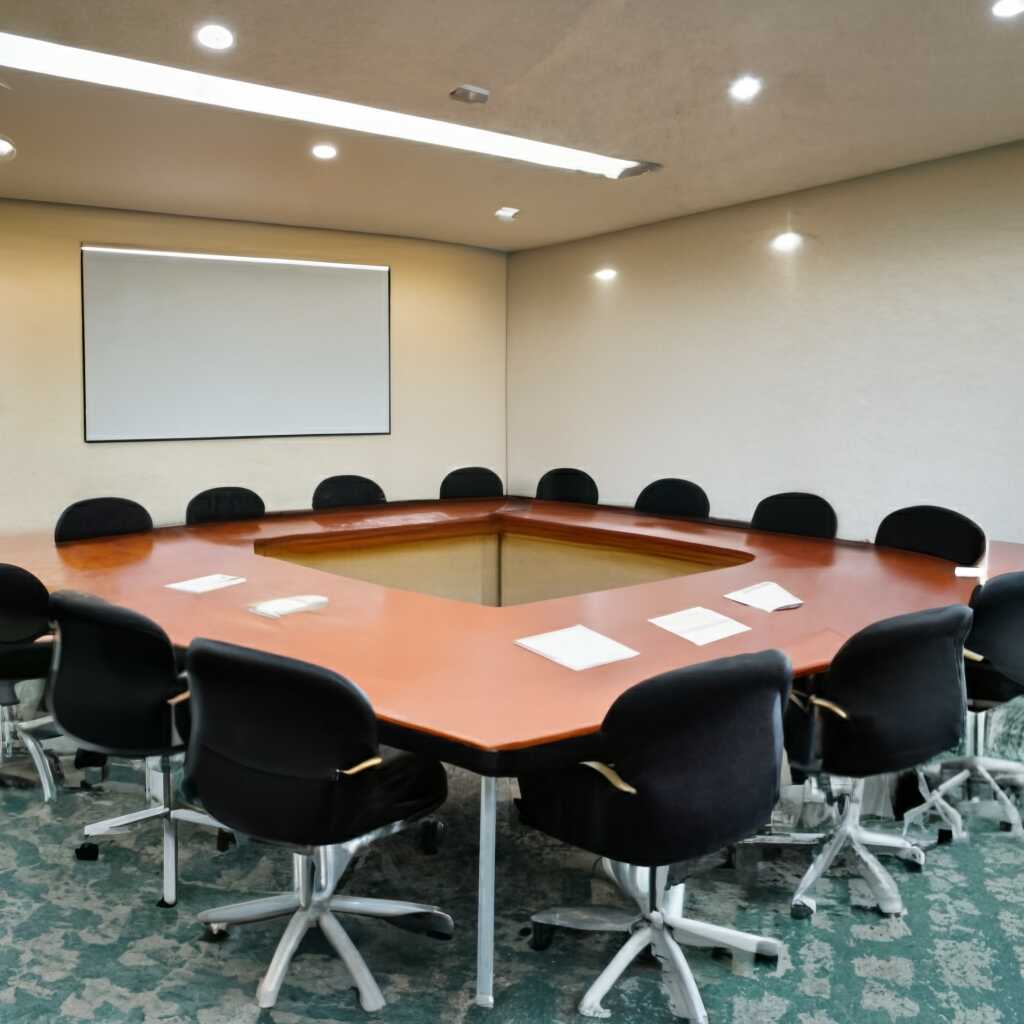} & \includegraphics[width=\imwidth\linewidth]{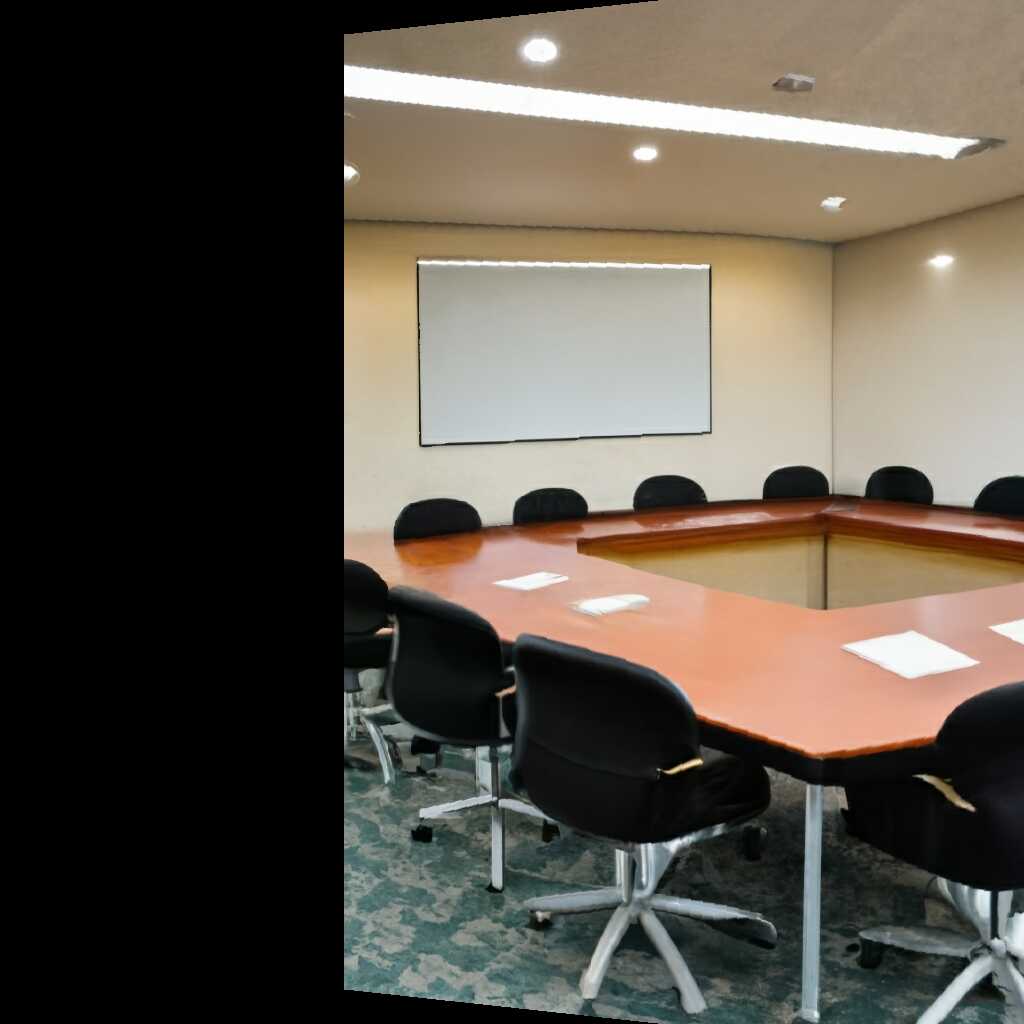} &\includegraphics[width=\imwidth\linewidth]{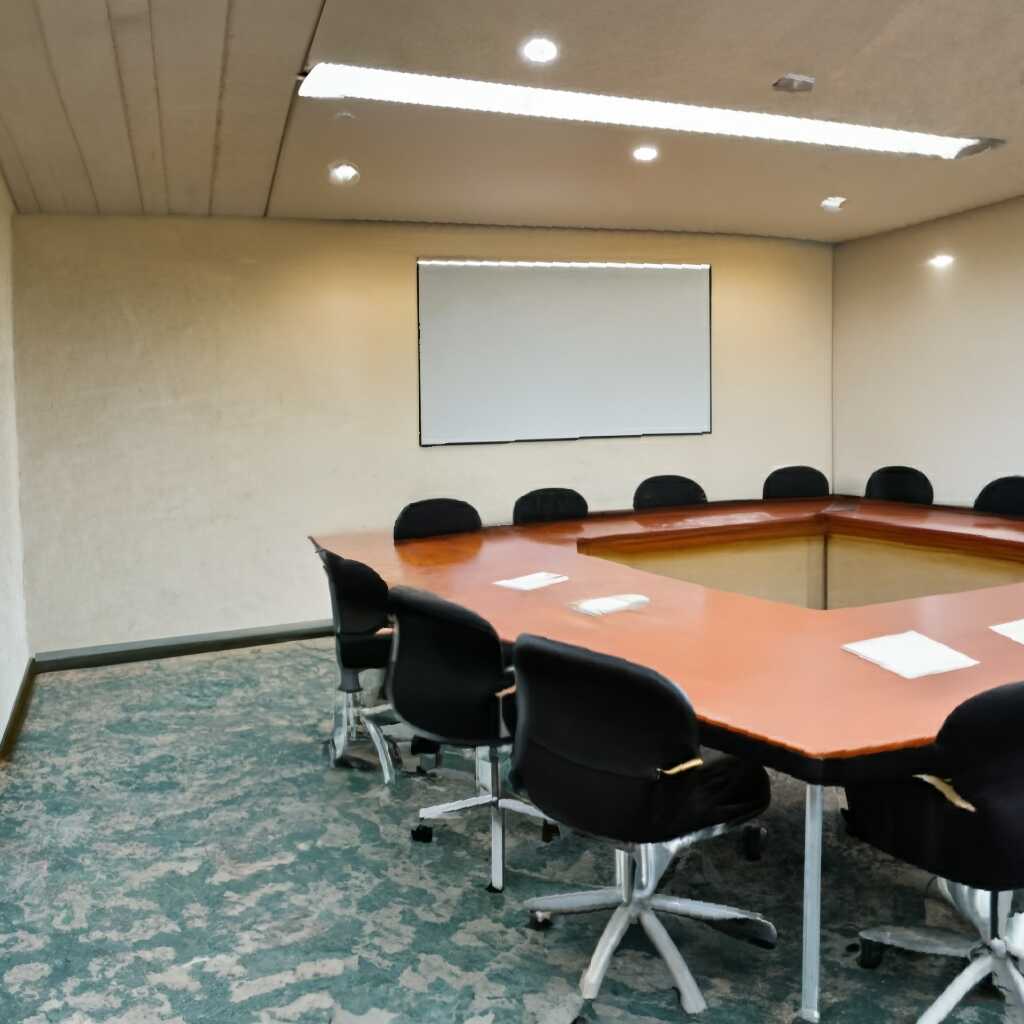} &\includegraphics[width=\imwidth\linewidth]{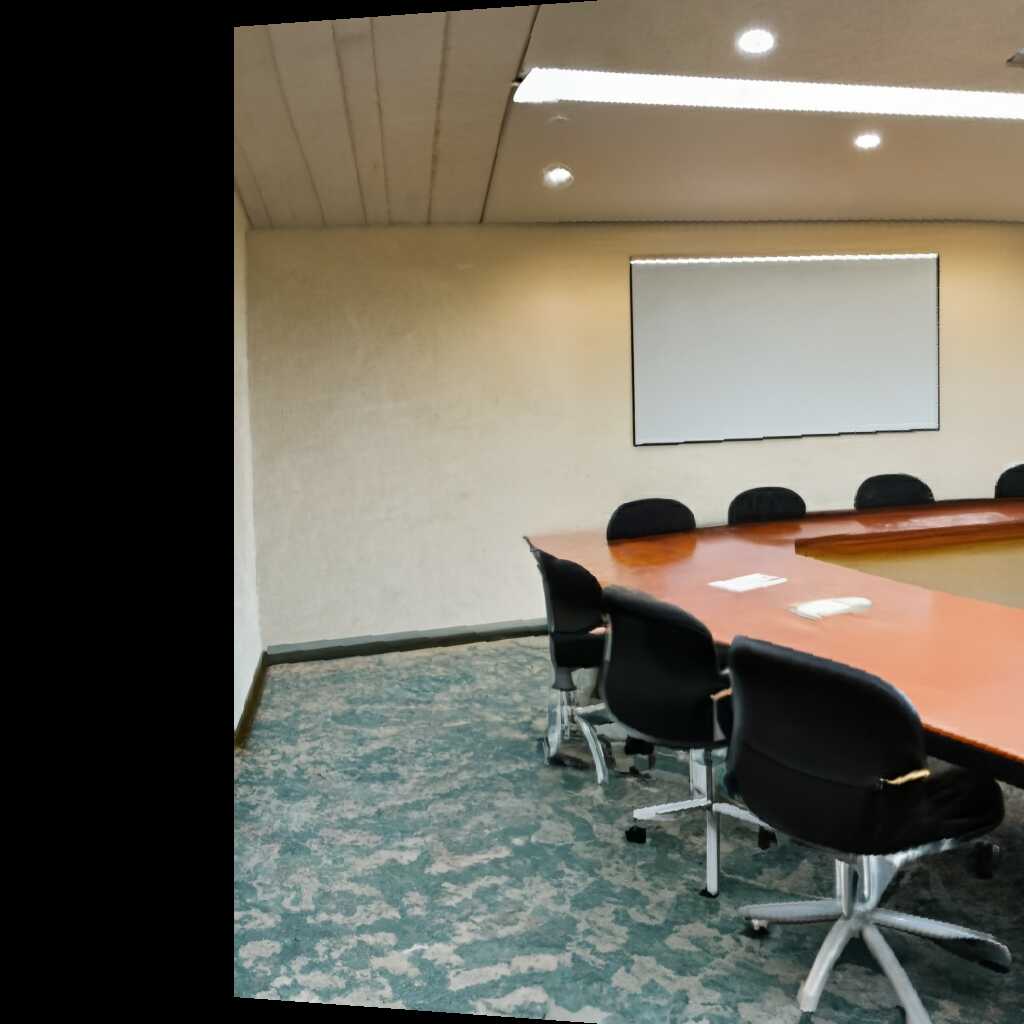} &\includegraphics[width=\imwidth\linewidth]{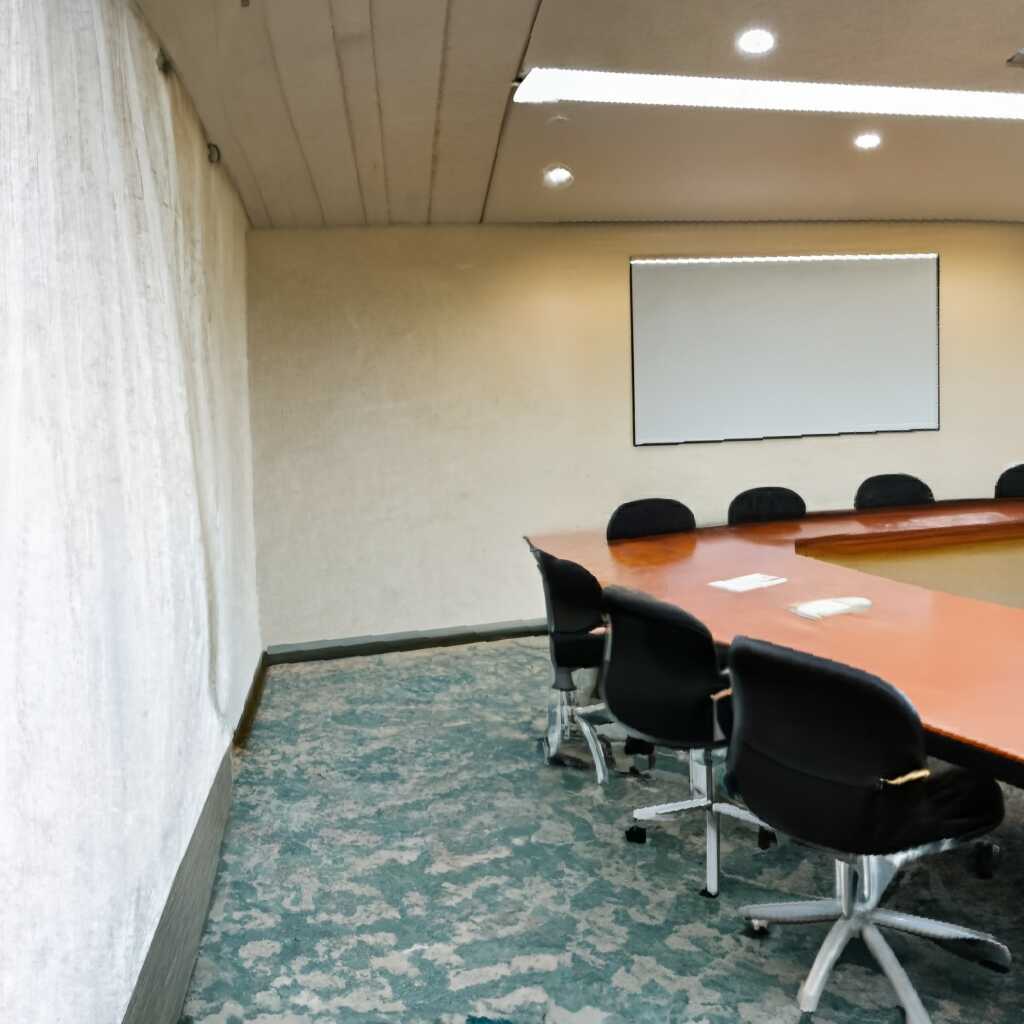} & \includegraphics[width=\imwidth\linewidth]{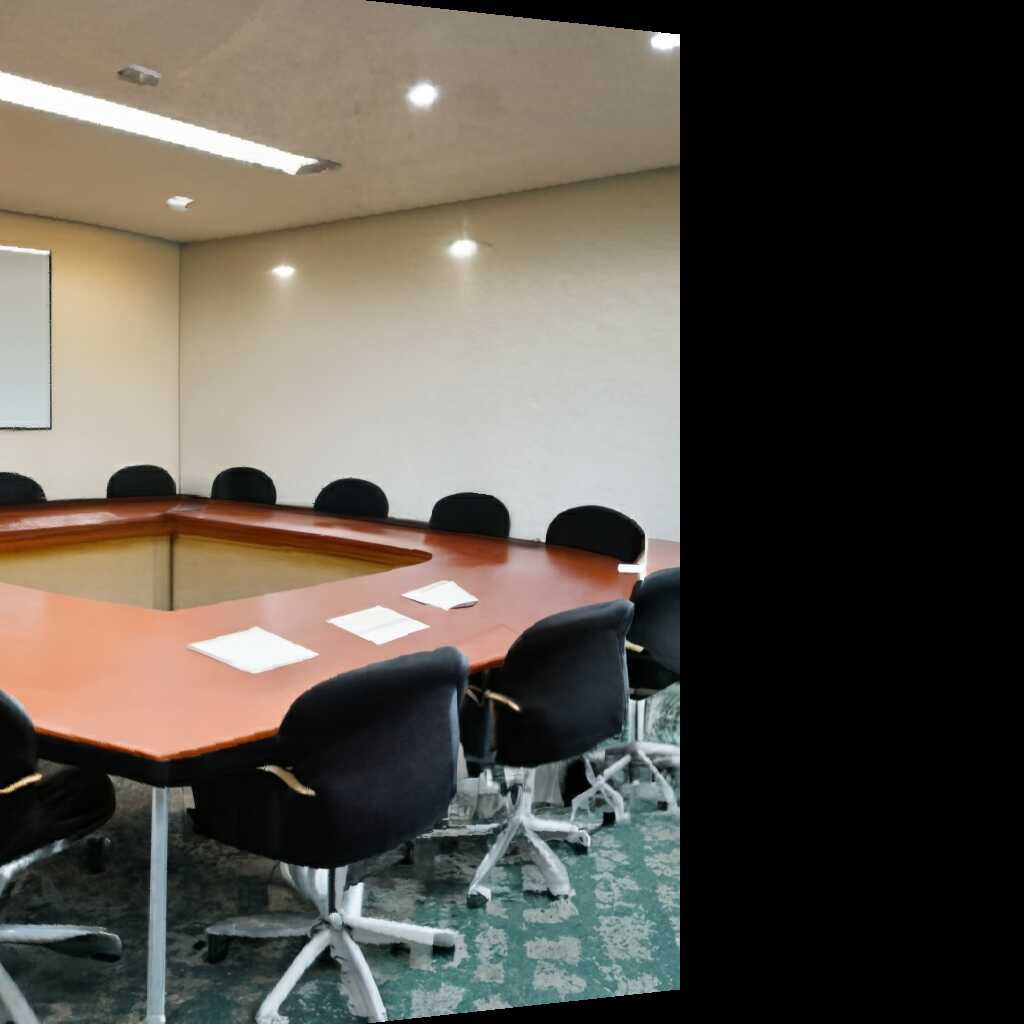} &\includegraphics[width=\imwidth\linewidth]{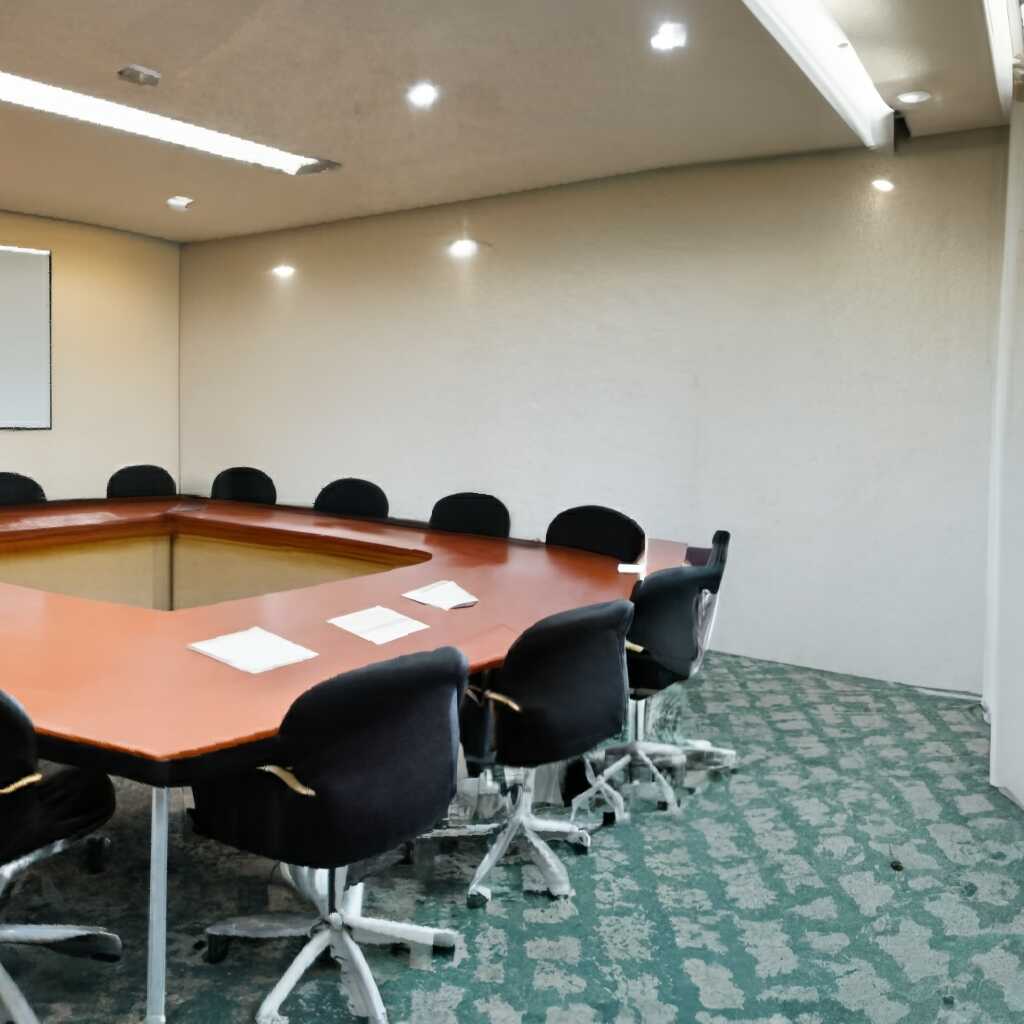} & \multicolumn{2}{c}{\multirow{2}{*}[\lastcoloffset]{\includegraphics[width=\imwidthlast\linewidth]{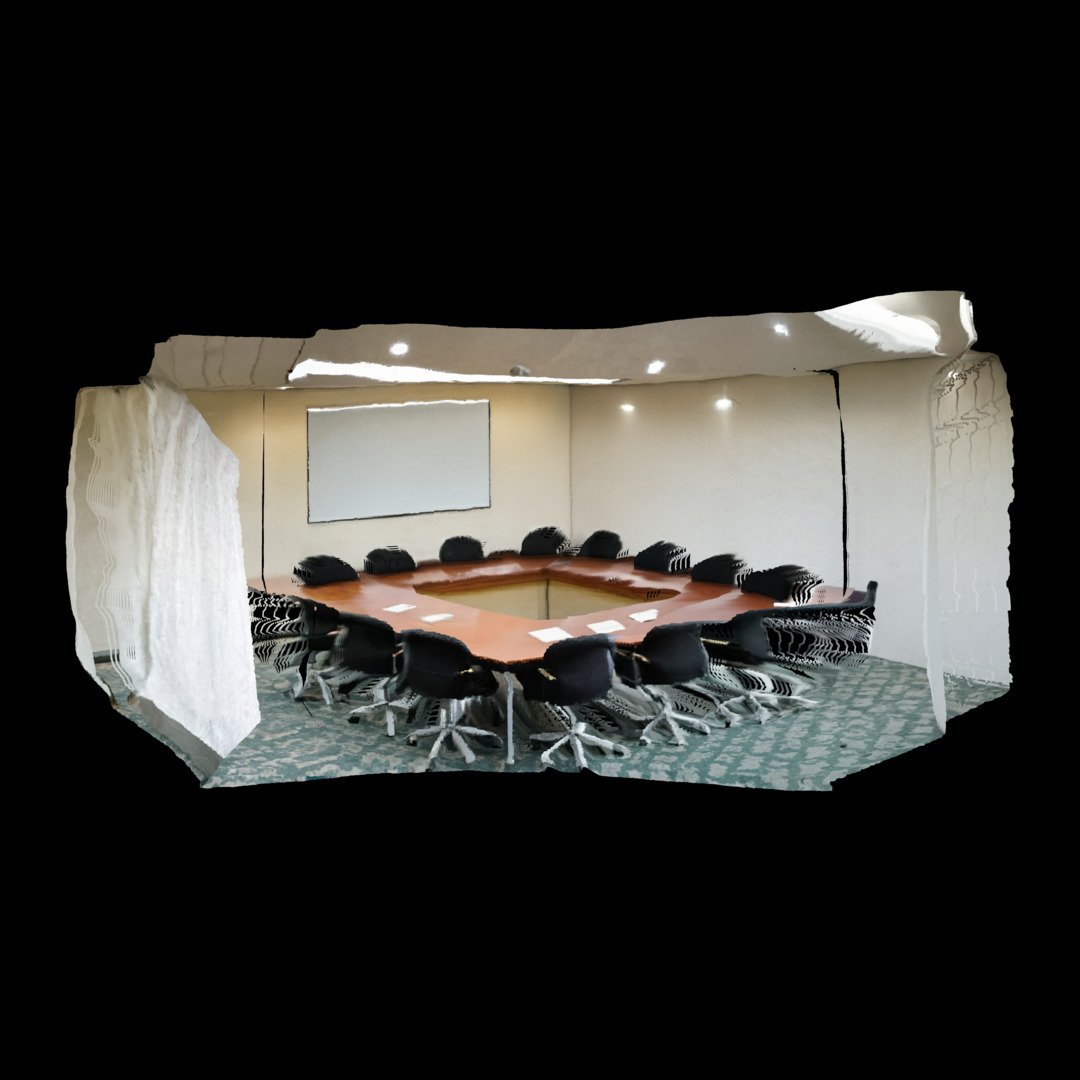}}} \\
    \includegraphics[width=\imwidth\linewidth]{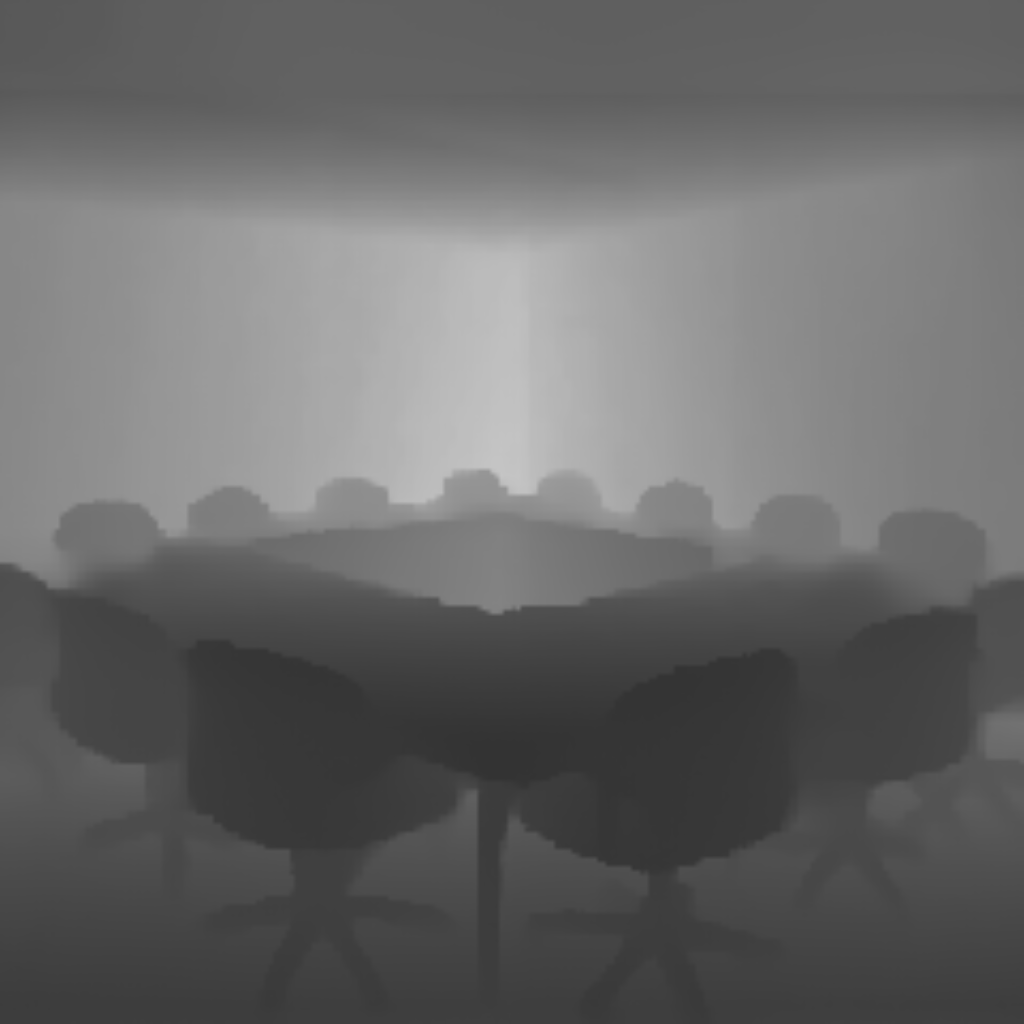} & \includegraphics[width=\imwidth\linewidth]{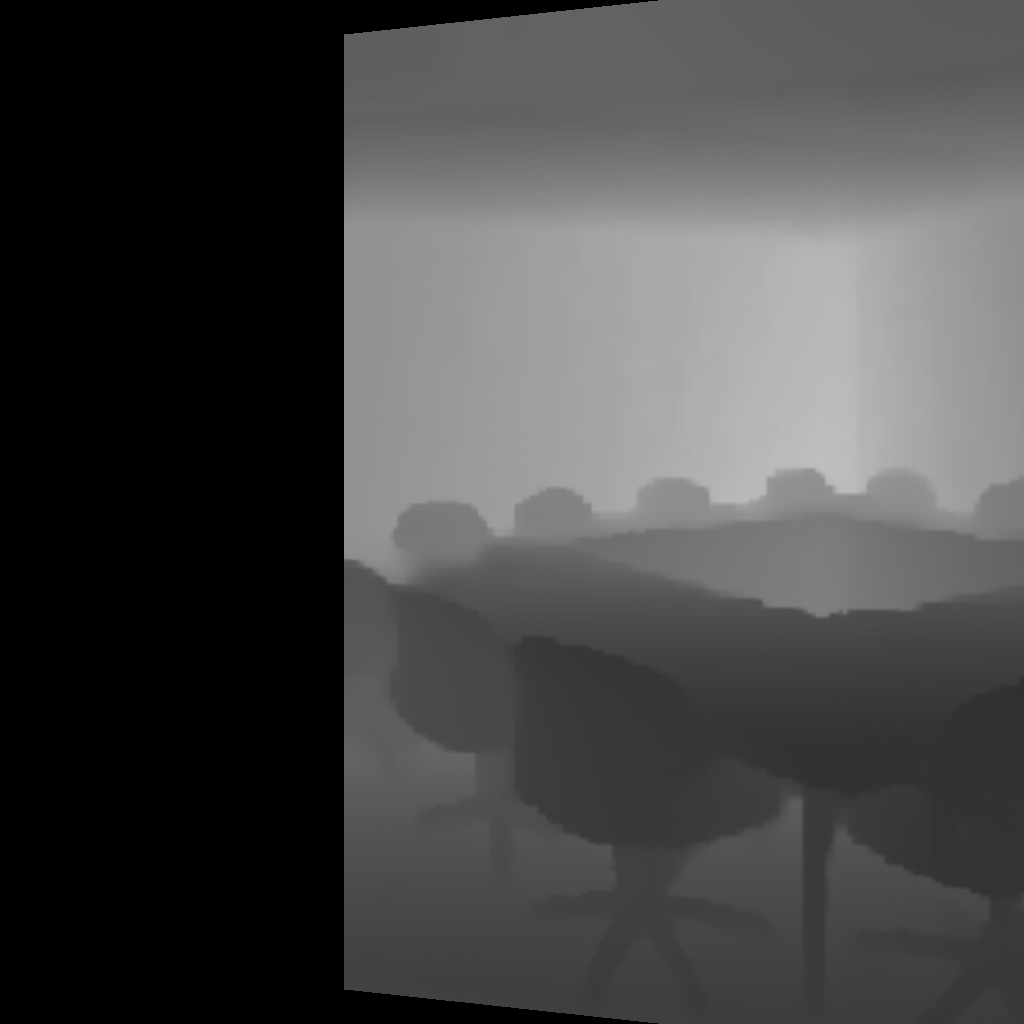} &\includegraphics[width=\imwidth\linewidth]{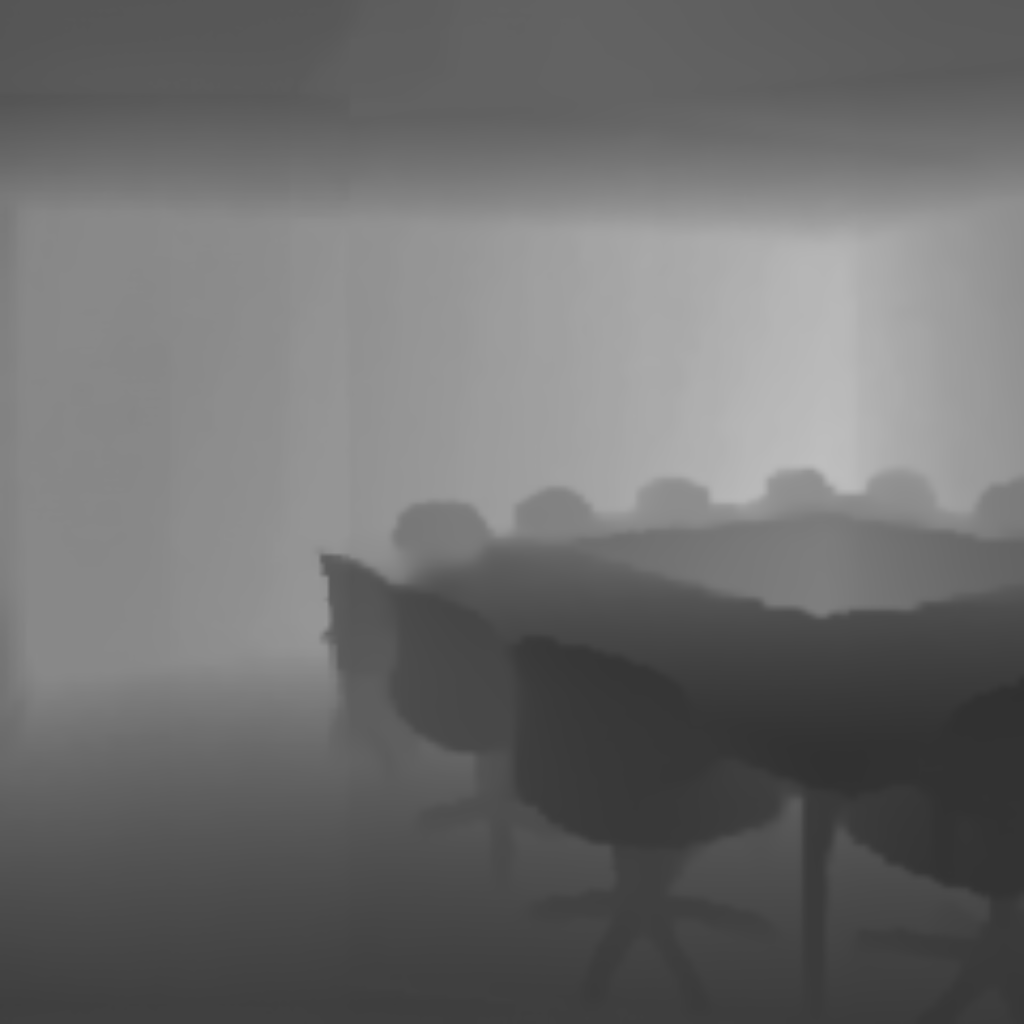} &  \includegraphics[width=\imwidth\linewidth]{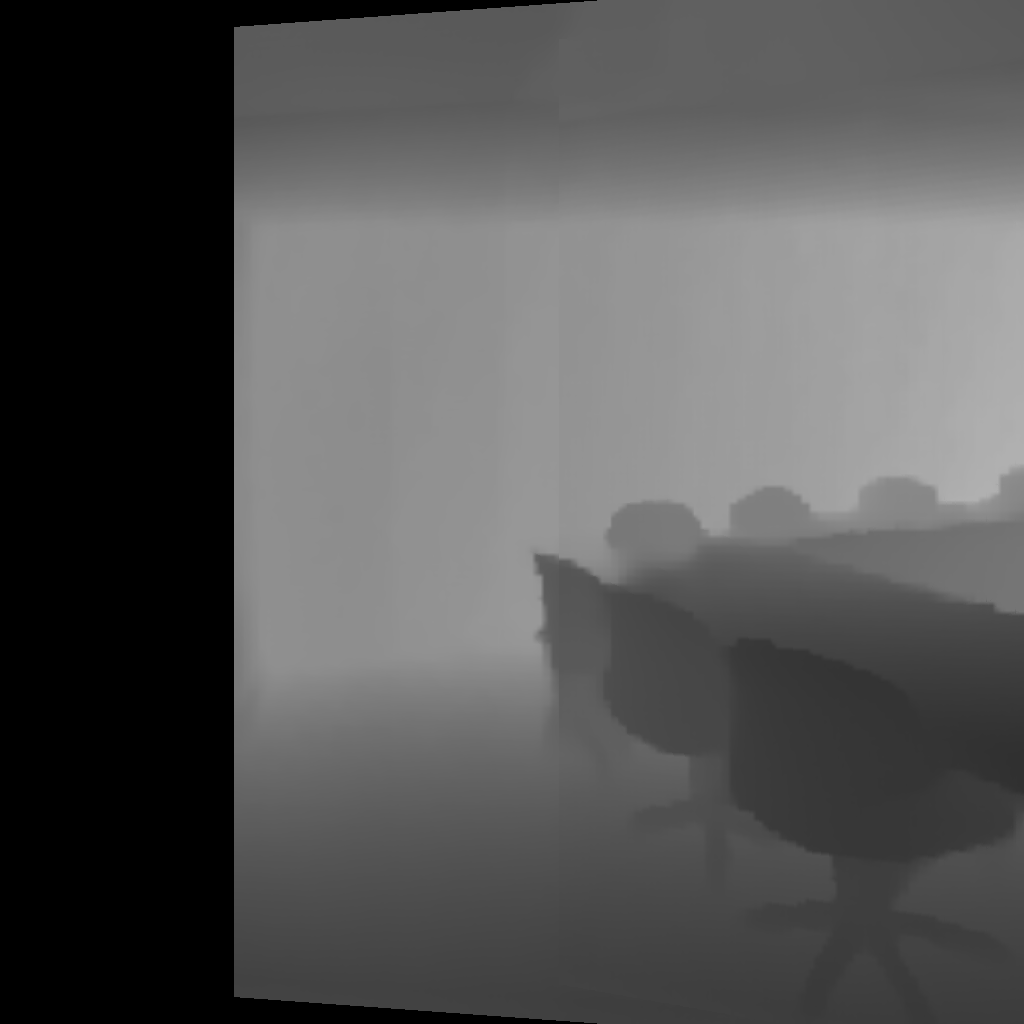} &\includegraphics[width=\imwidth\linewidth]{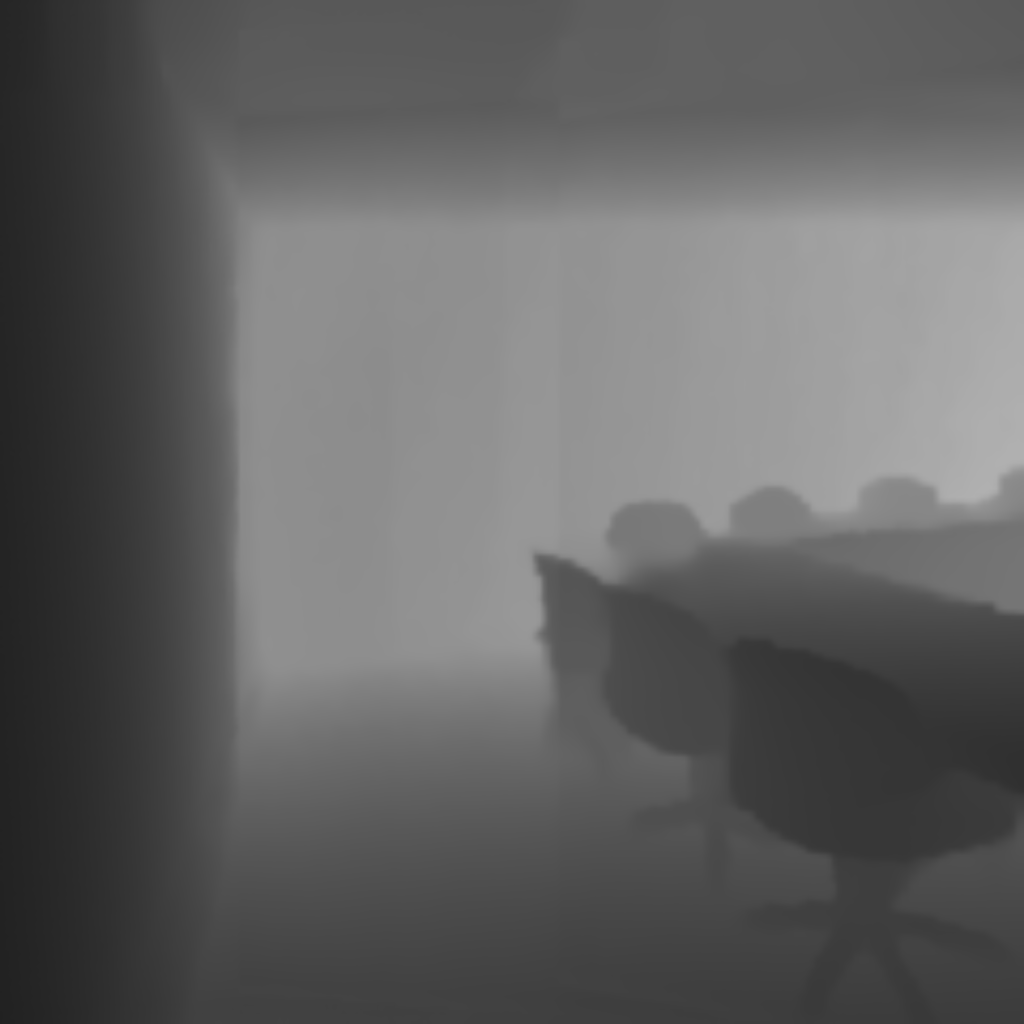} &\includegraphics[width=\imwidth\linewidth]{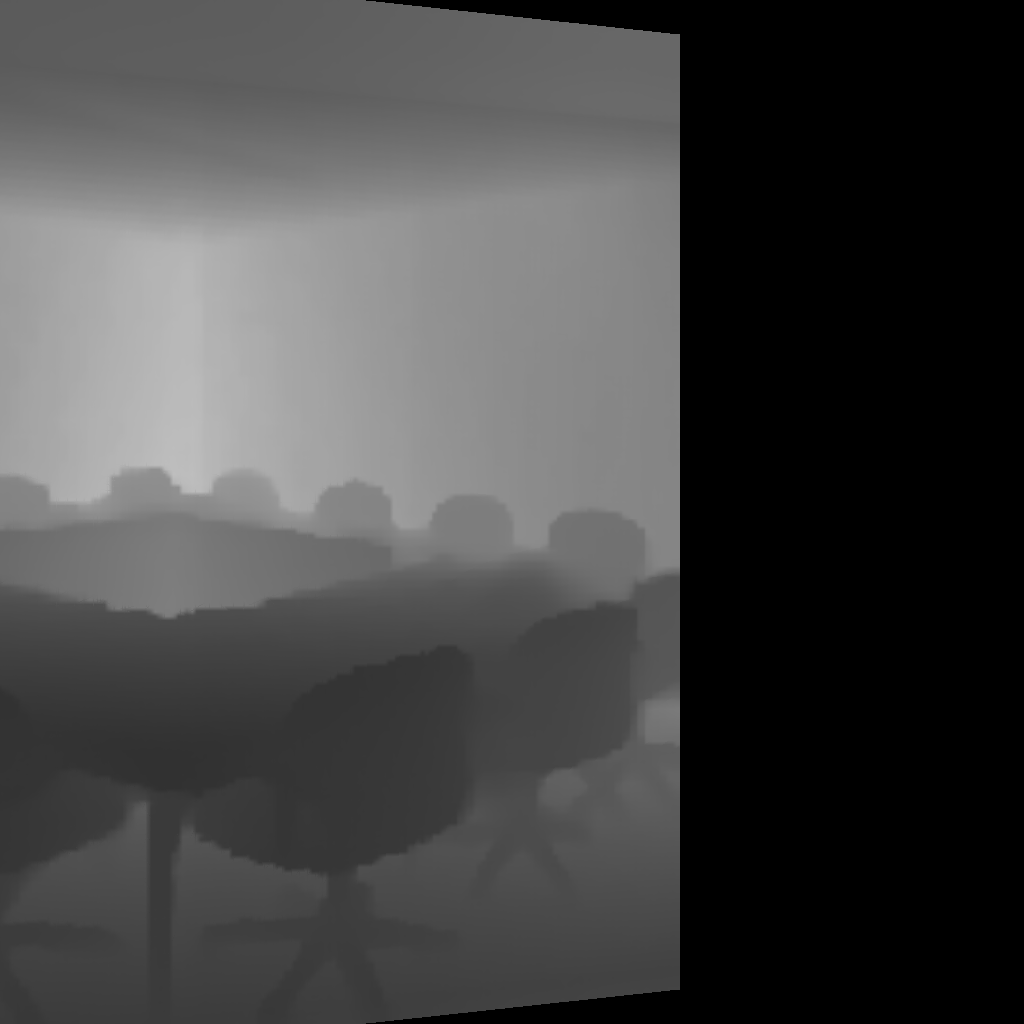} &\includegraphics[width=\imwidth\linewidth]{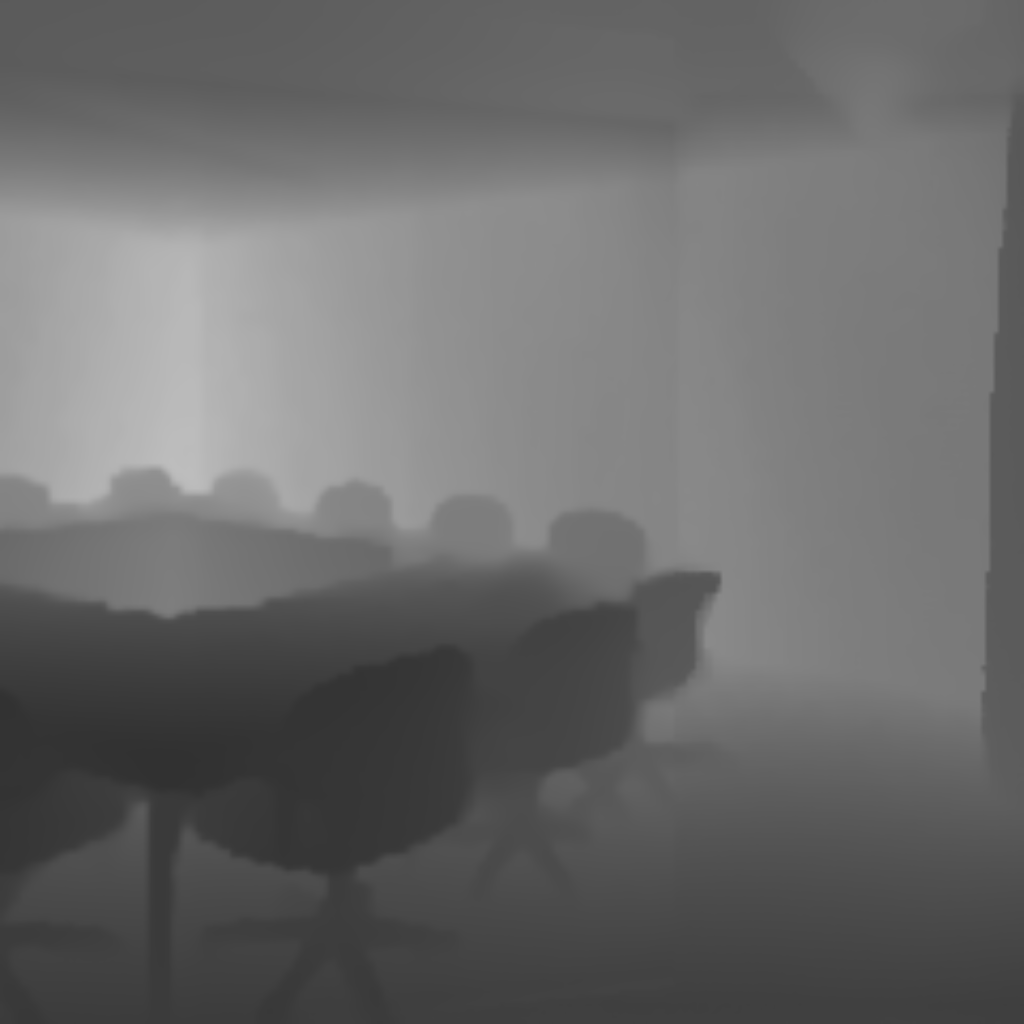} & \multicolumn{2}{c}{} \\
    \\[\textpadding]\multicolumn{\ncols}{c}{Prompt: A meeting room.}\\\\[\textpadding]

    \includegraphics[width=\imwidth\linewidth]{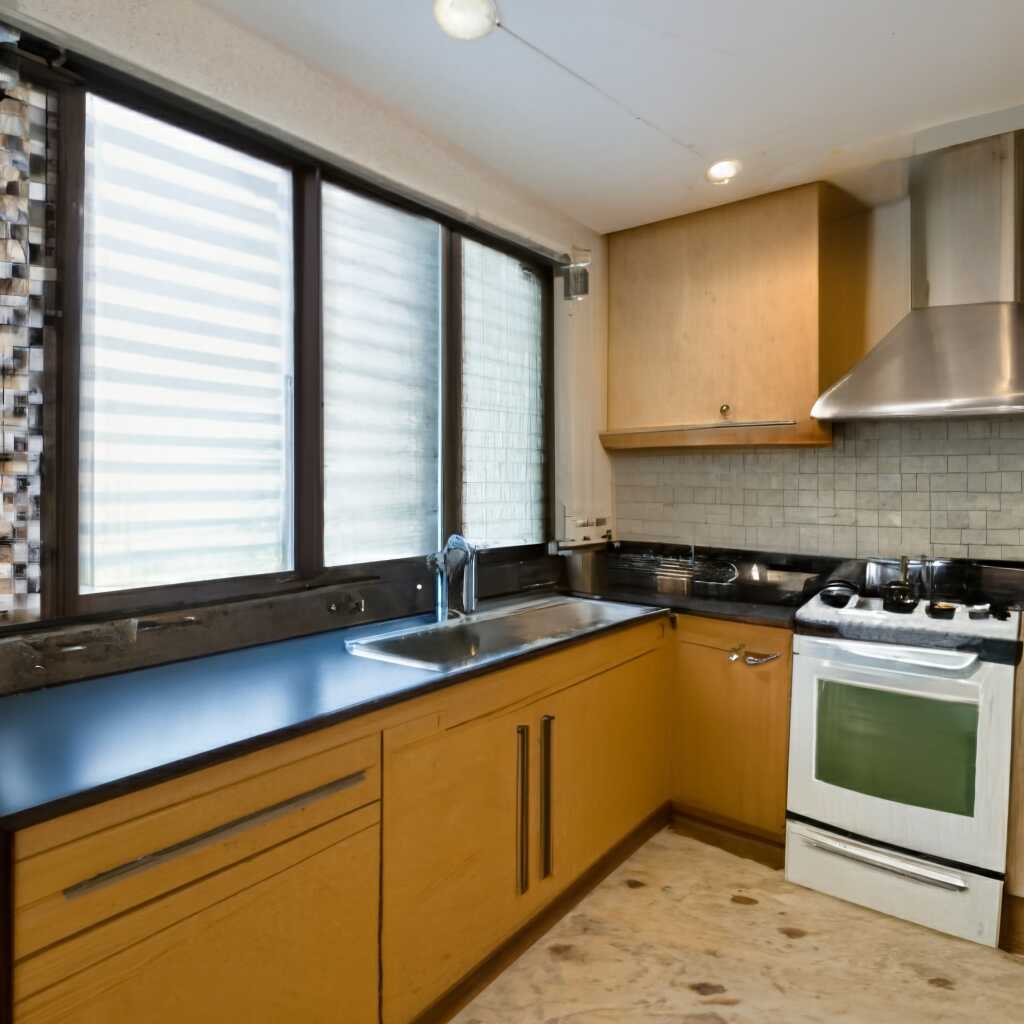} & \includegraphics[width=\imwidth\linewidth]{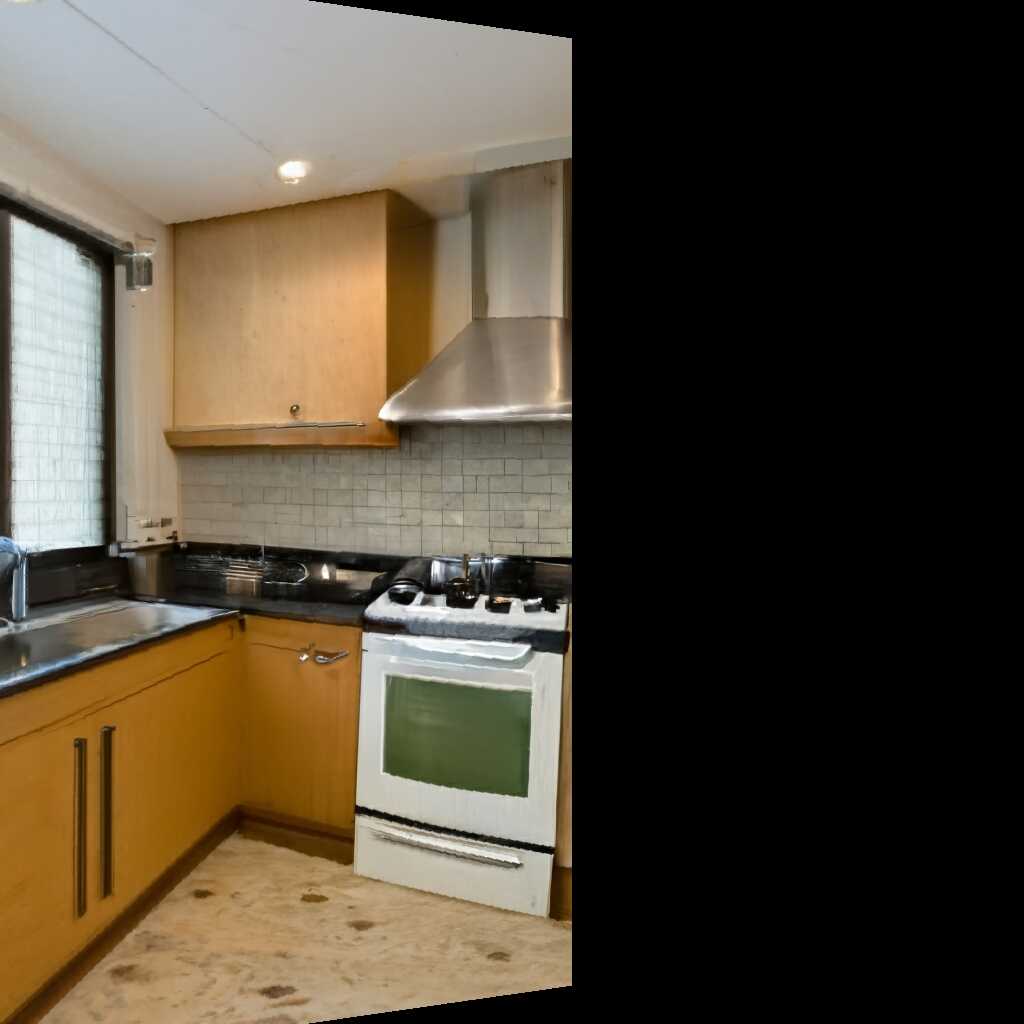} &\includegraphics[width=\imwidth\linewidth]{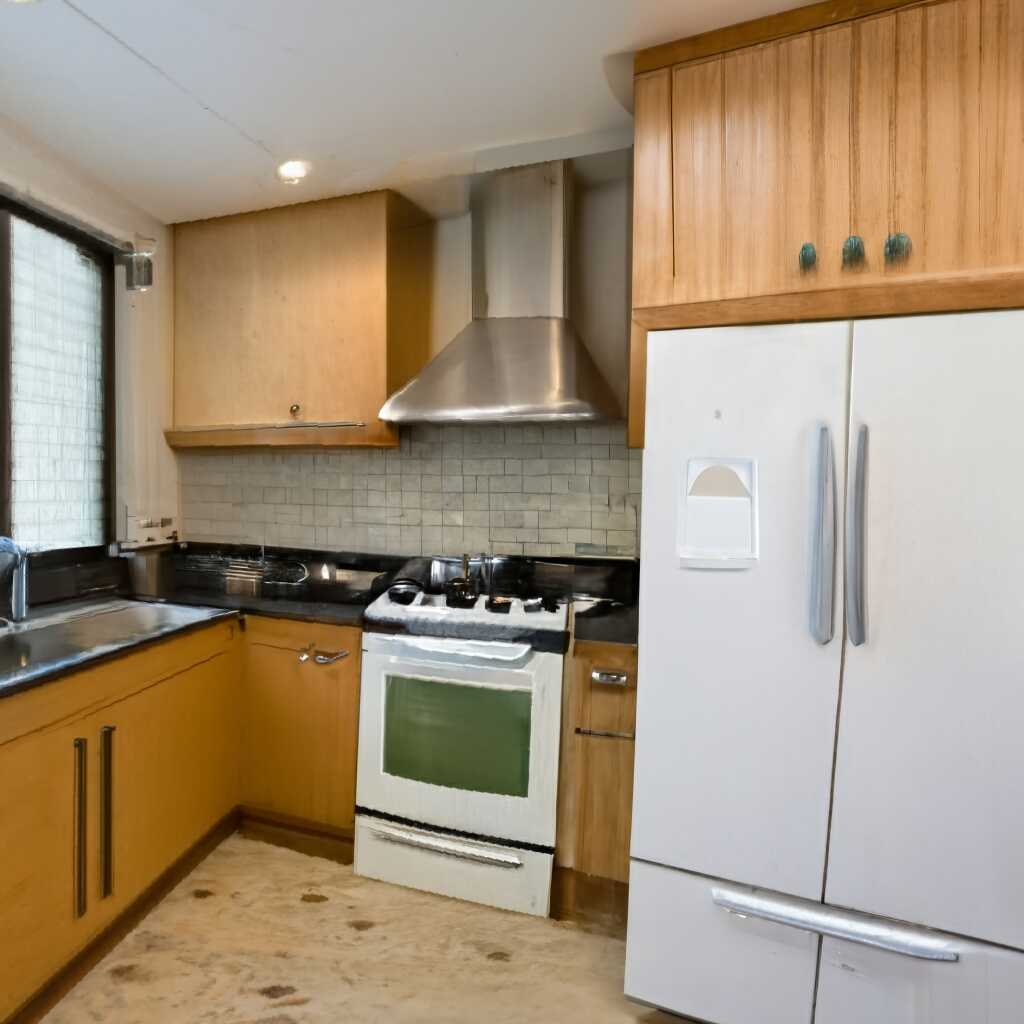} &\includegraphics[width=\imwidth\linewidth]{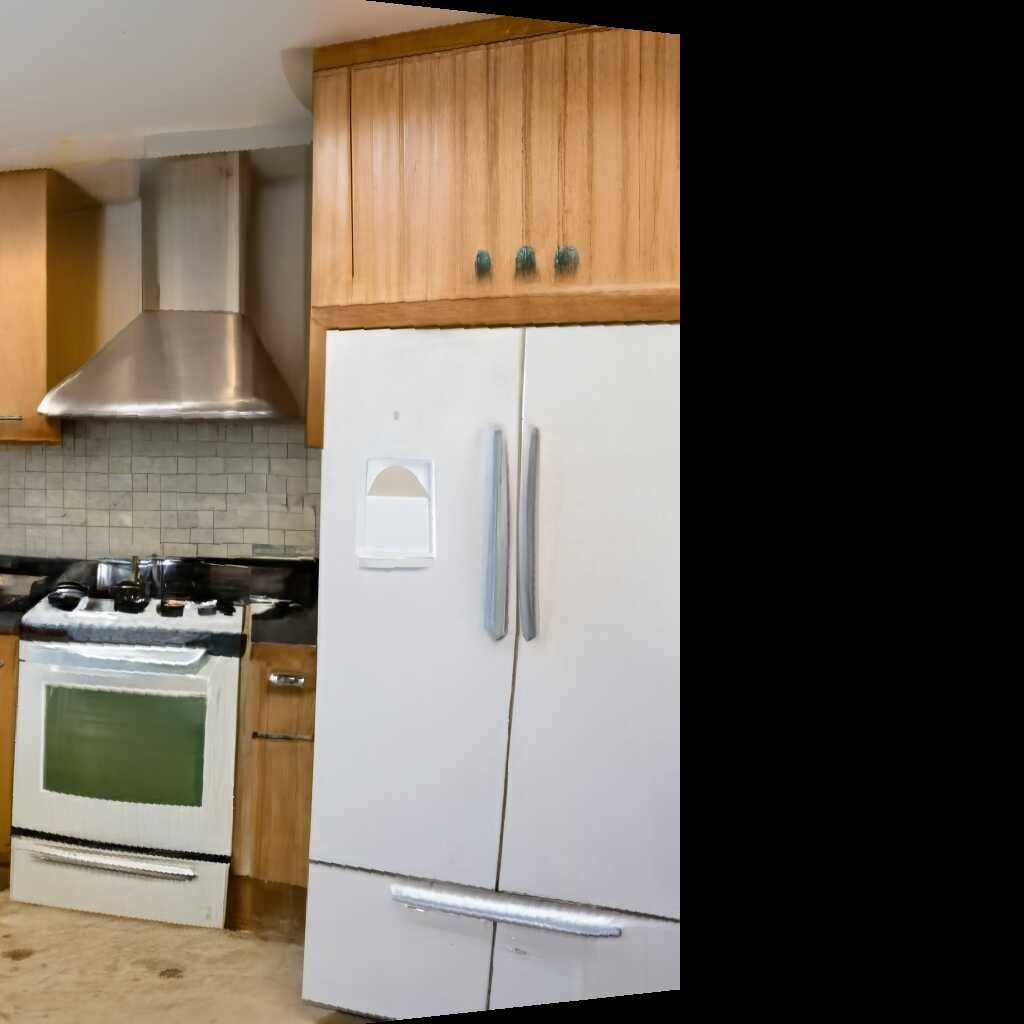} &\includegraphics[width=\imwidth\linewidth]{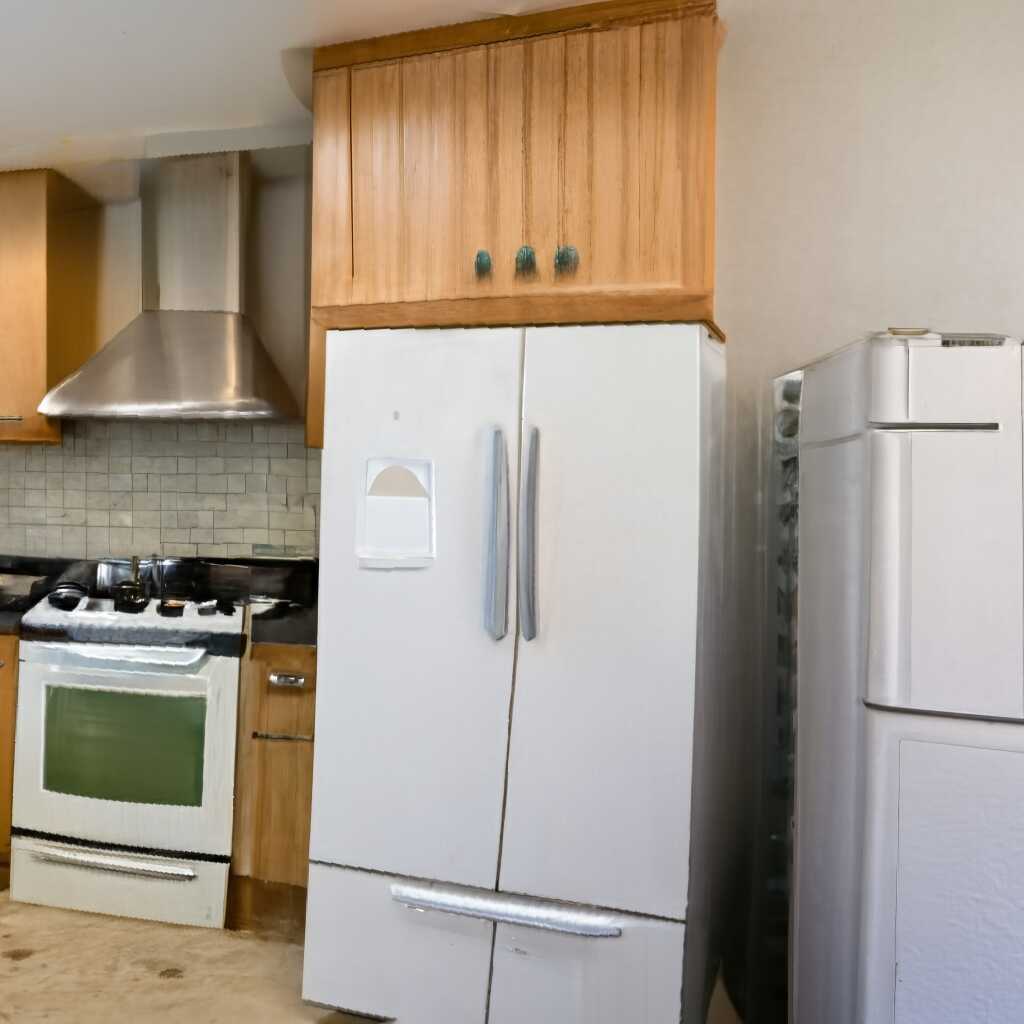} & \includegraphics[width=\imwidth\linewidth]{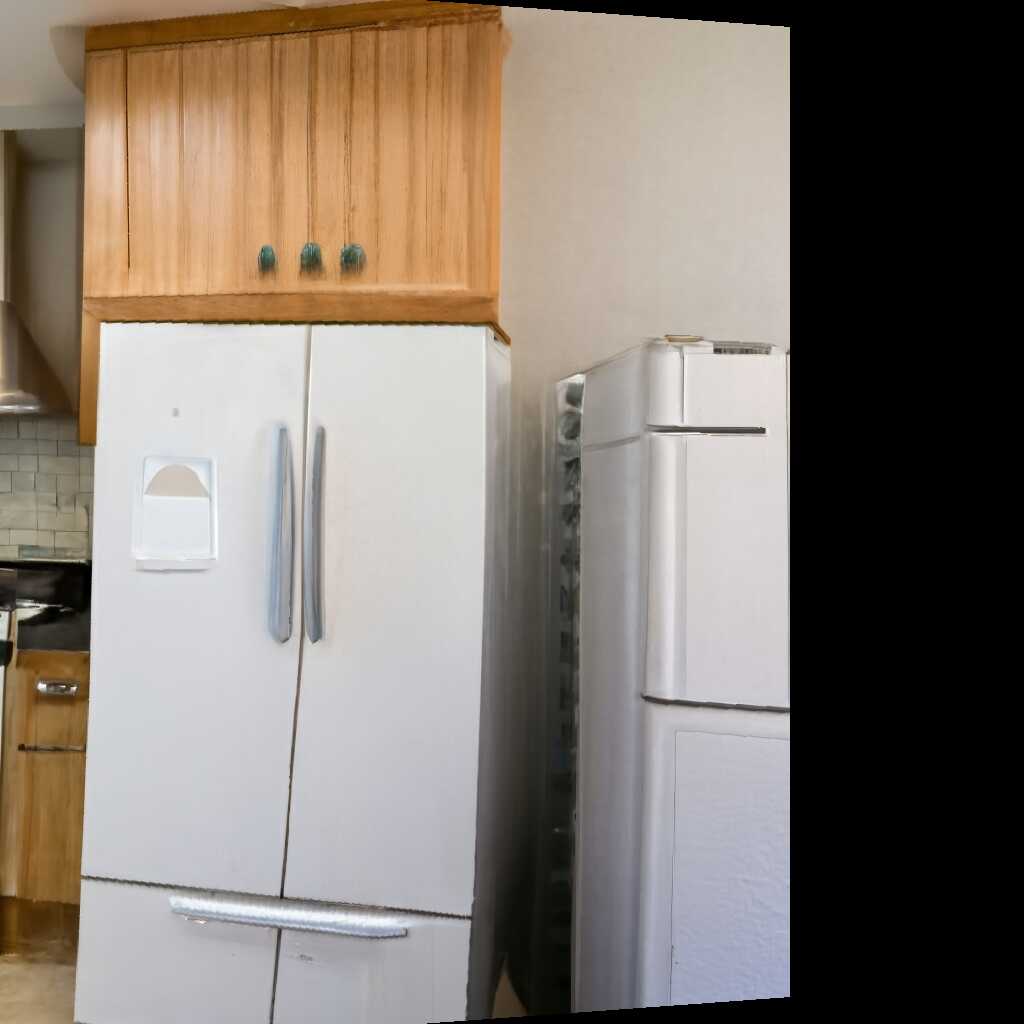} &\includegraphics[width=\imwidth\linewidth]{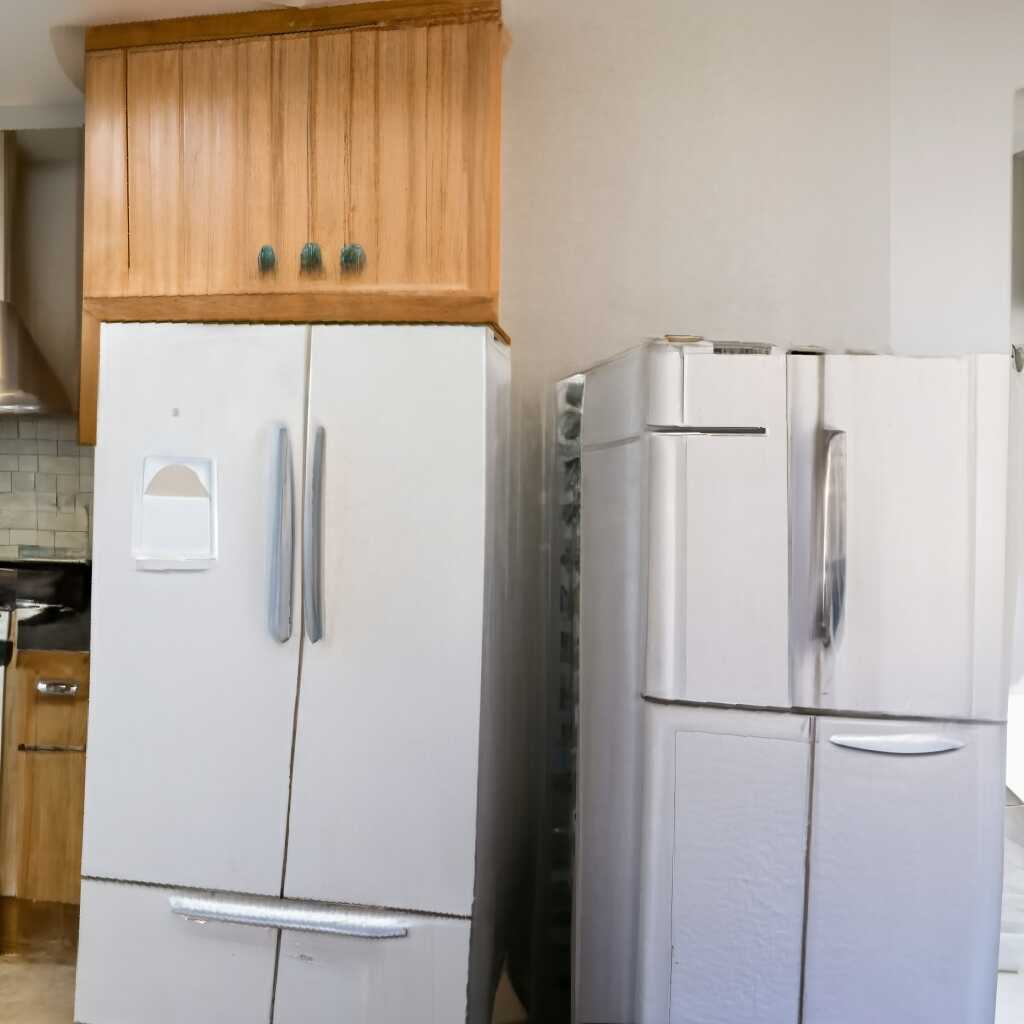} & \multicolumn{2}{c}{\multirow{2}{*}[\lastcoloffset]{\includegraphics[width=\imwidthlast\linewidth]{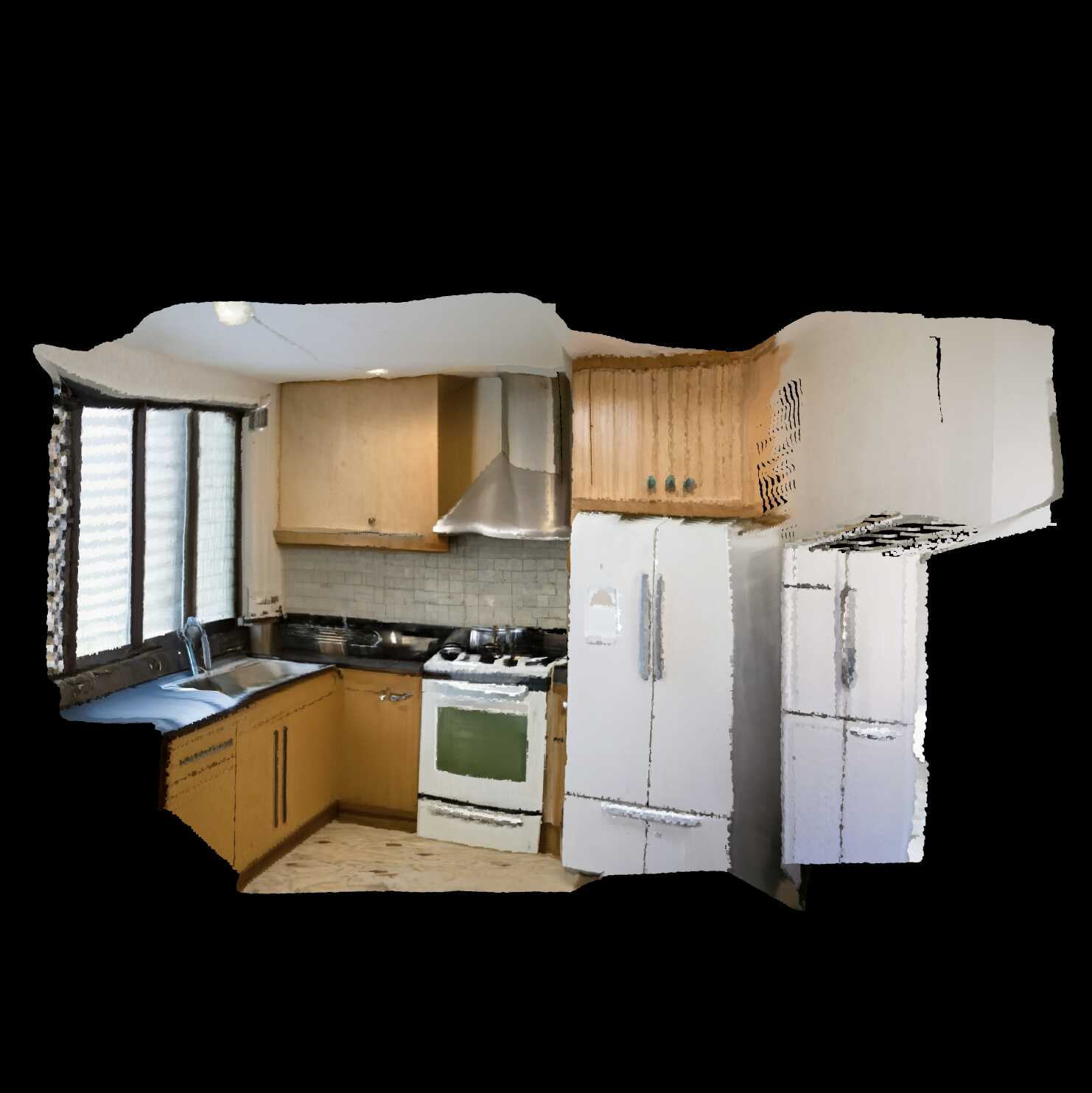}}} \\
    \includegraphics[width=\imwidth\linewidth]{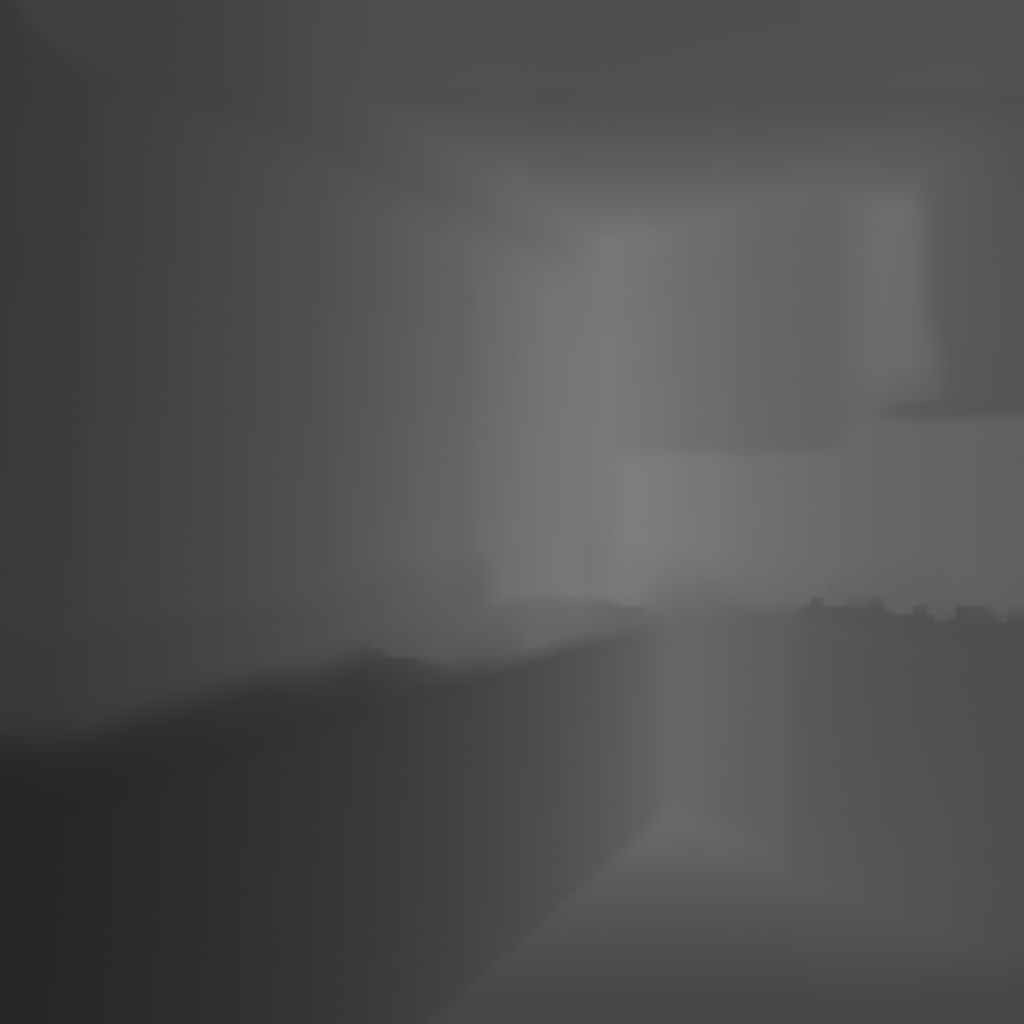} & \includegraphics[width=\imwidth\linewidth]{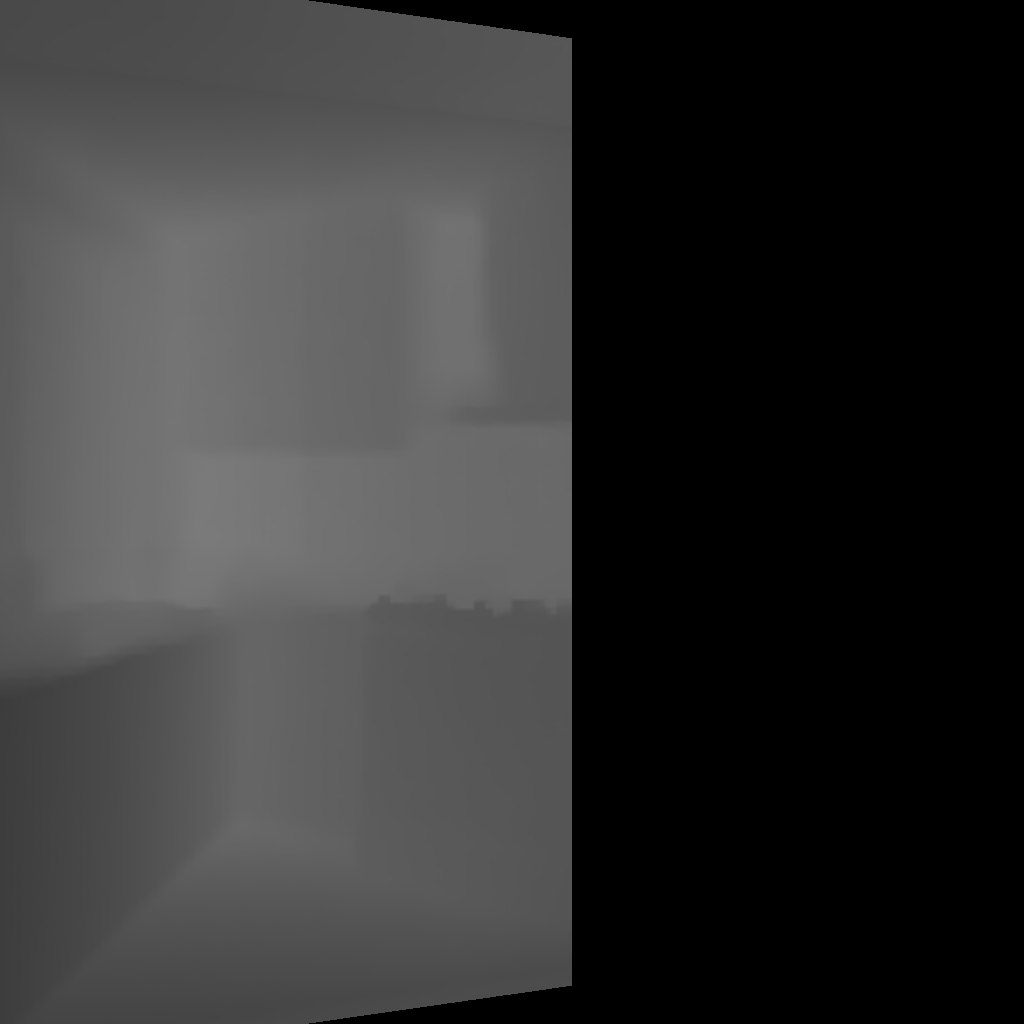} &\includegraphics[width=\imwidth\linewidth]{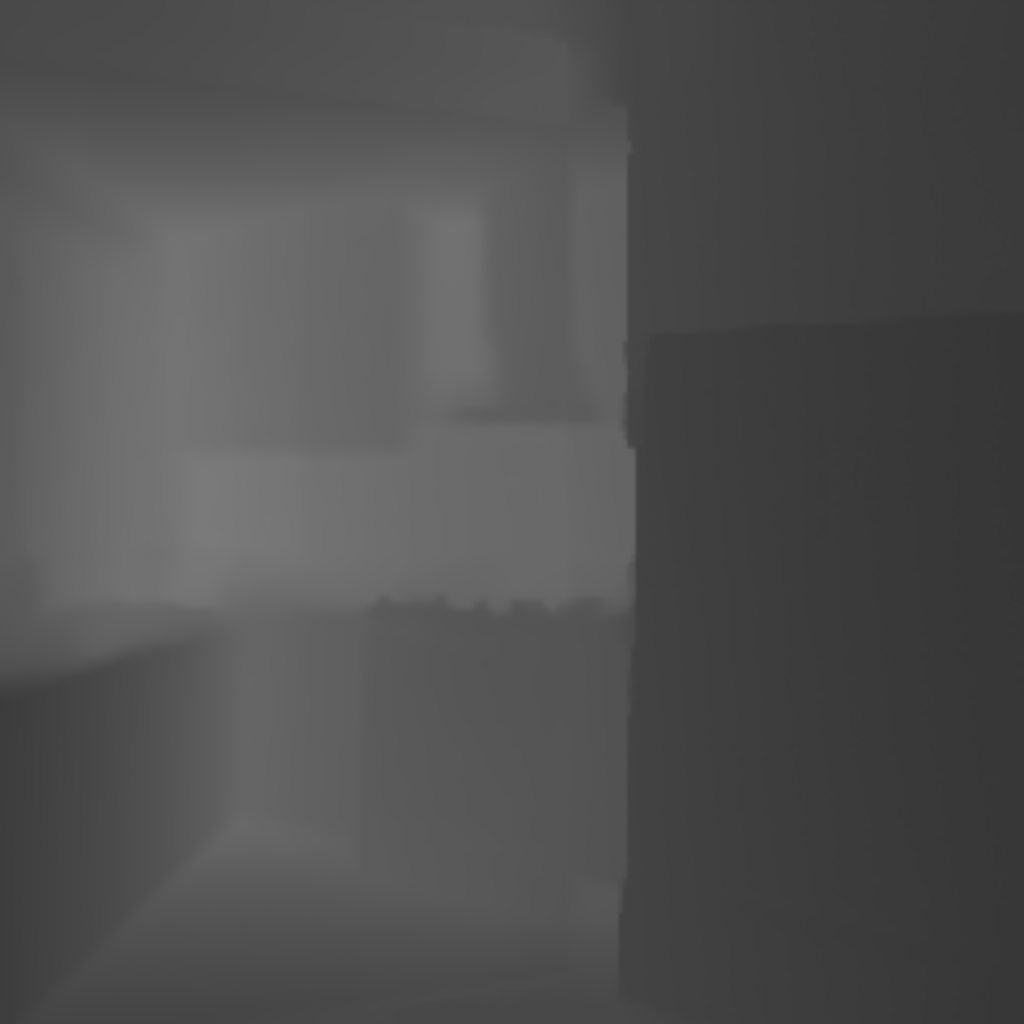} &  \includegraphics[width=\imwidth\linewidth]{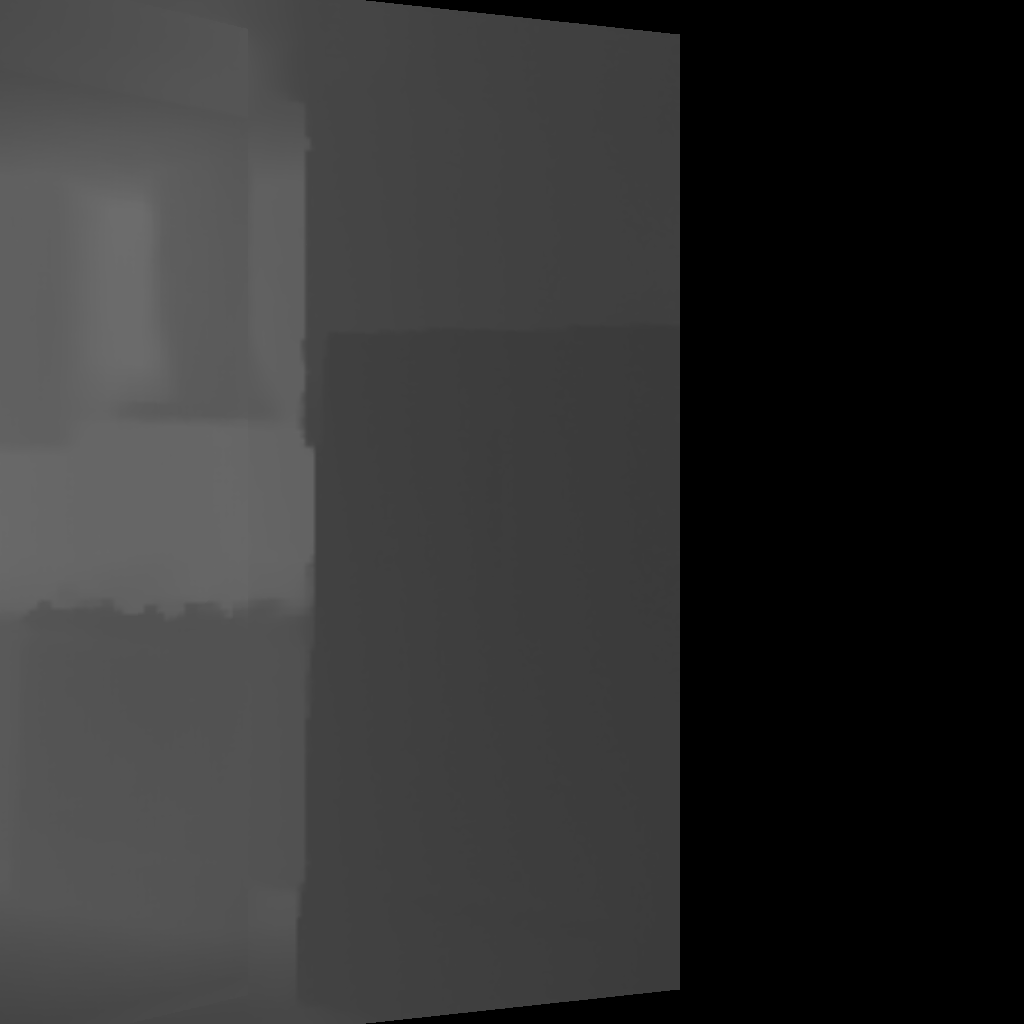} &\includegraphics[width=\imwidth\linewidth]{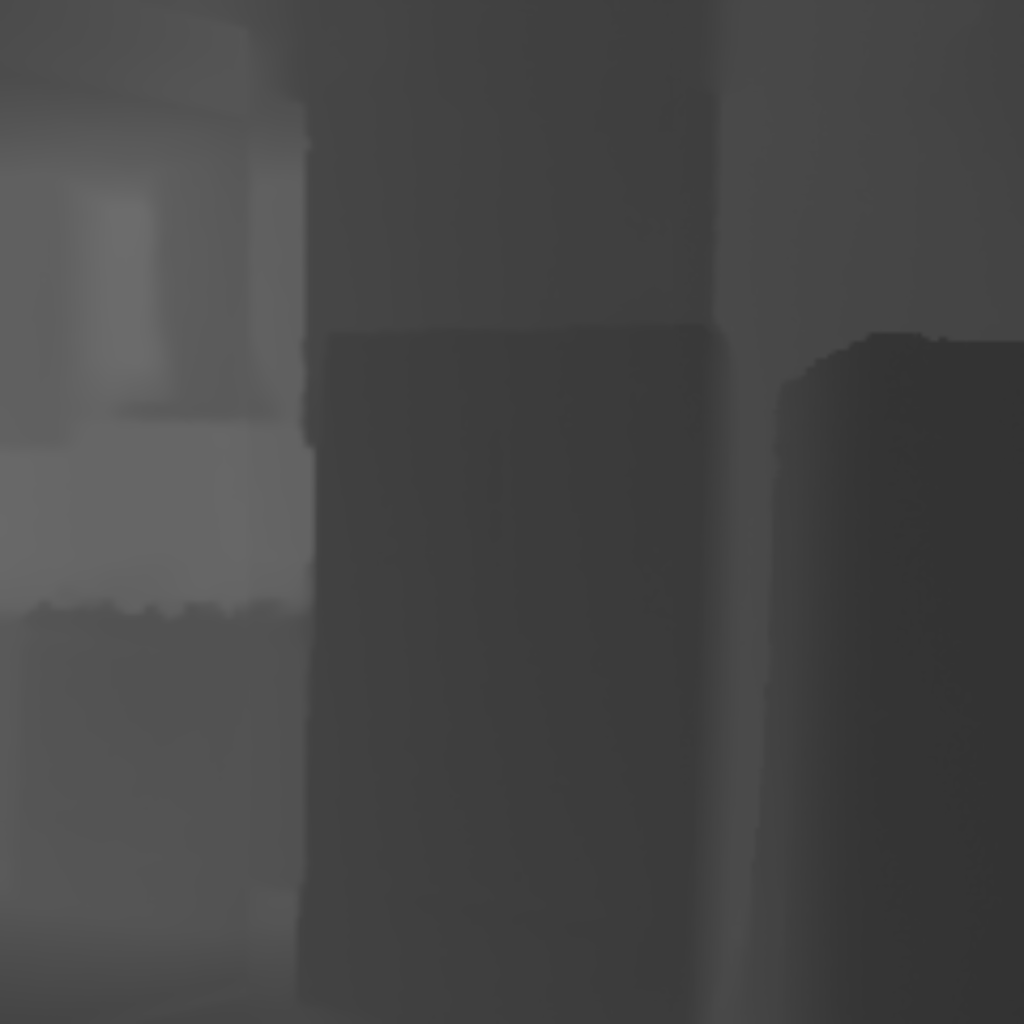} &\includegraphics[width=\imwidth\linewidth]{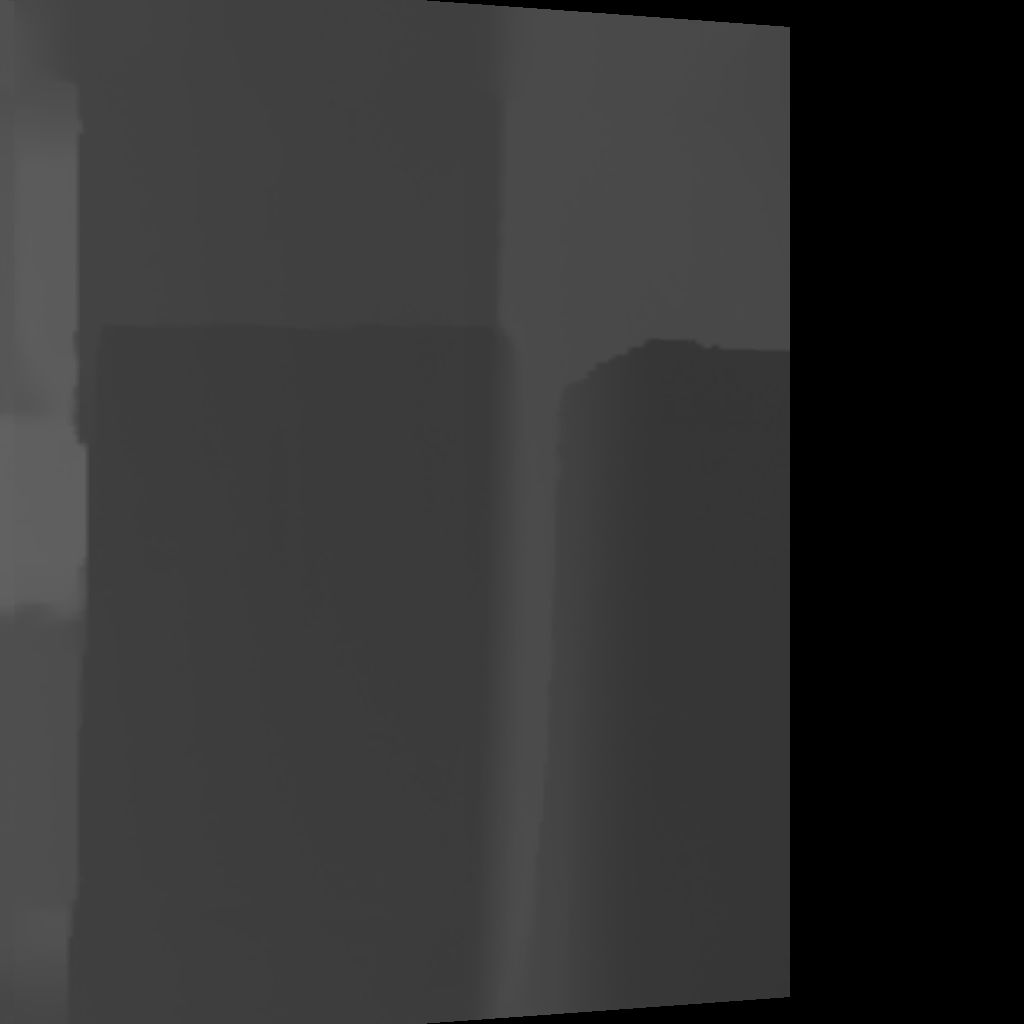} &\includegraphics[width=\imwidth\linewidth]{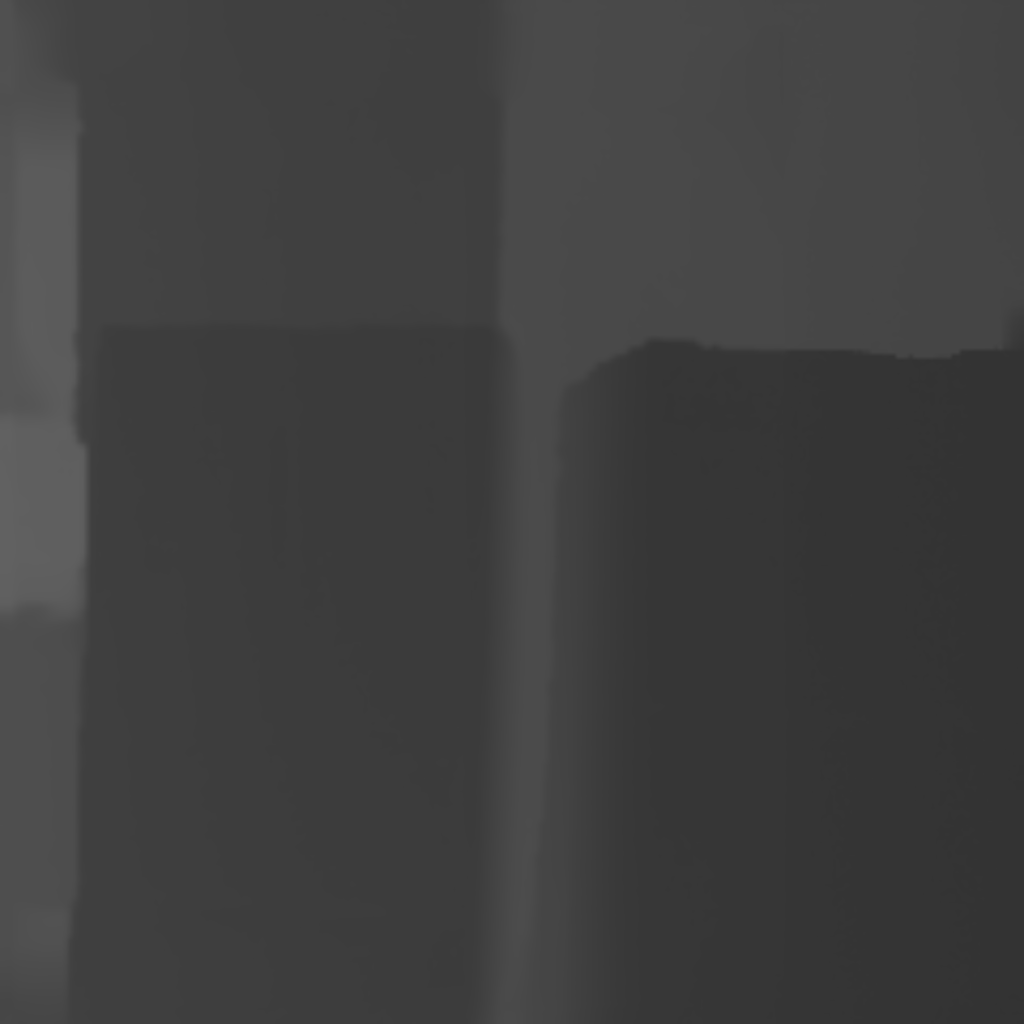} & \multicolumn{2}{c}{} \\
    \\[\textpadding]\multicolumn{\ncols}{c}{Prompt: A kitchen.}\\\\[\textpadding]

    \includegraphics[width=\imwidth\linewidth]{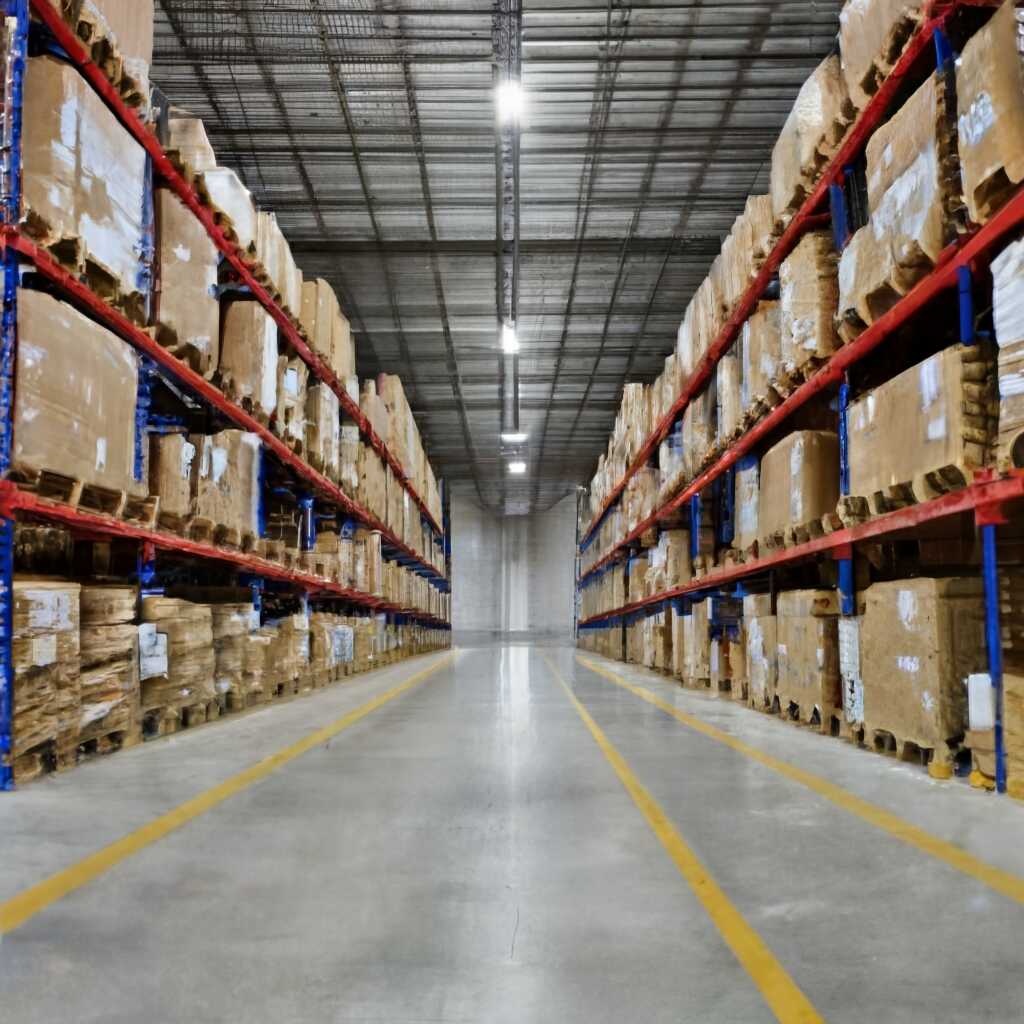} & \includegraphics[width=\imwidth\linewidth]{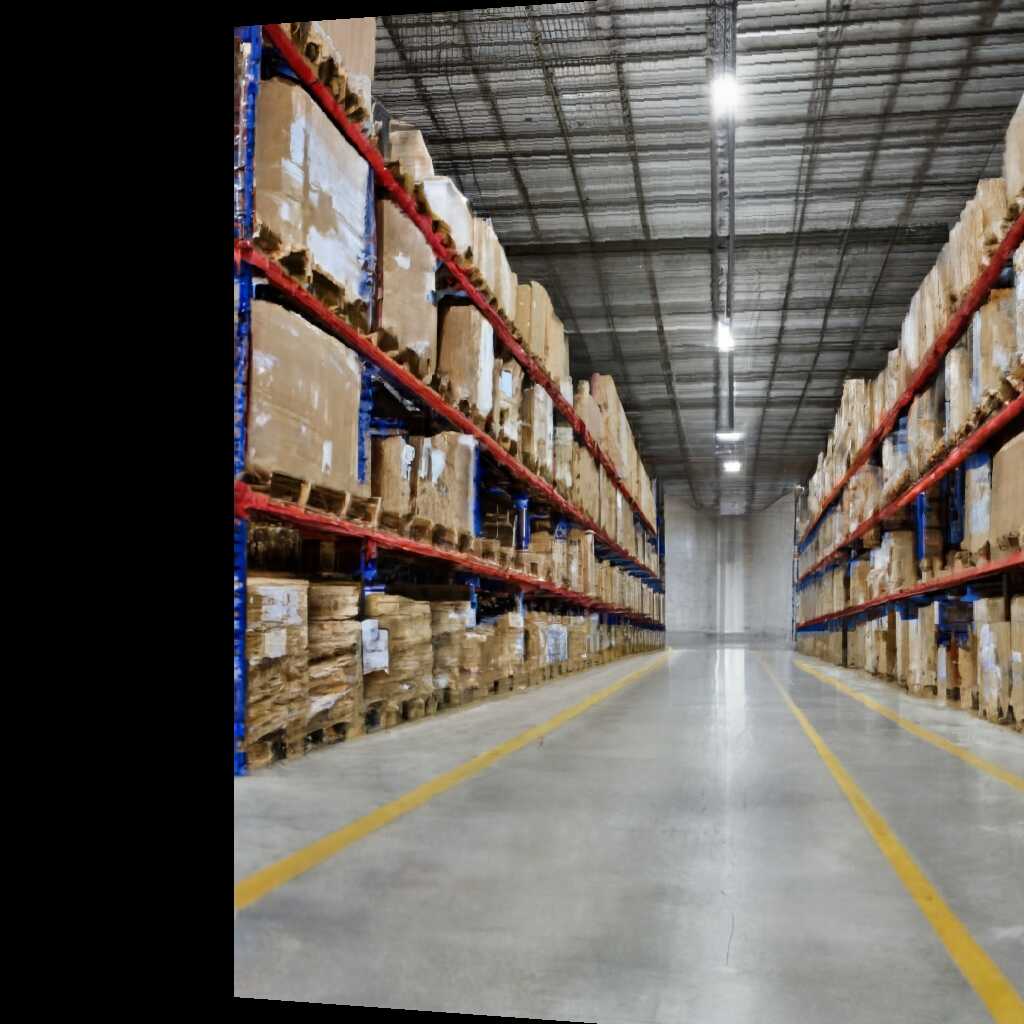} &\includegraphics[width=\imwidth\linewidth]{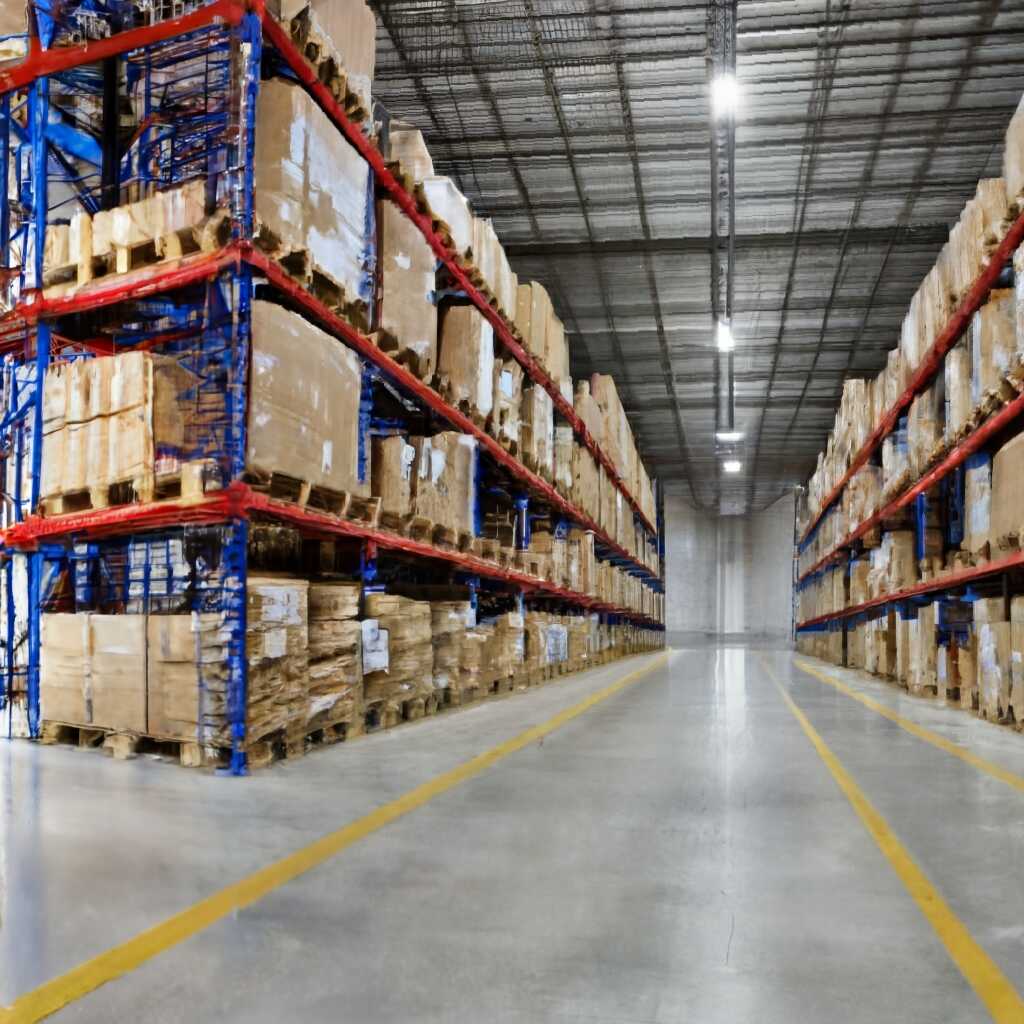} &\includegraphics[width=\imwidth\linewidth]{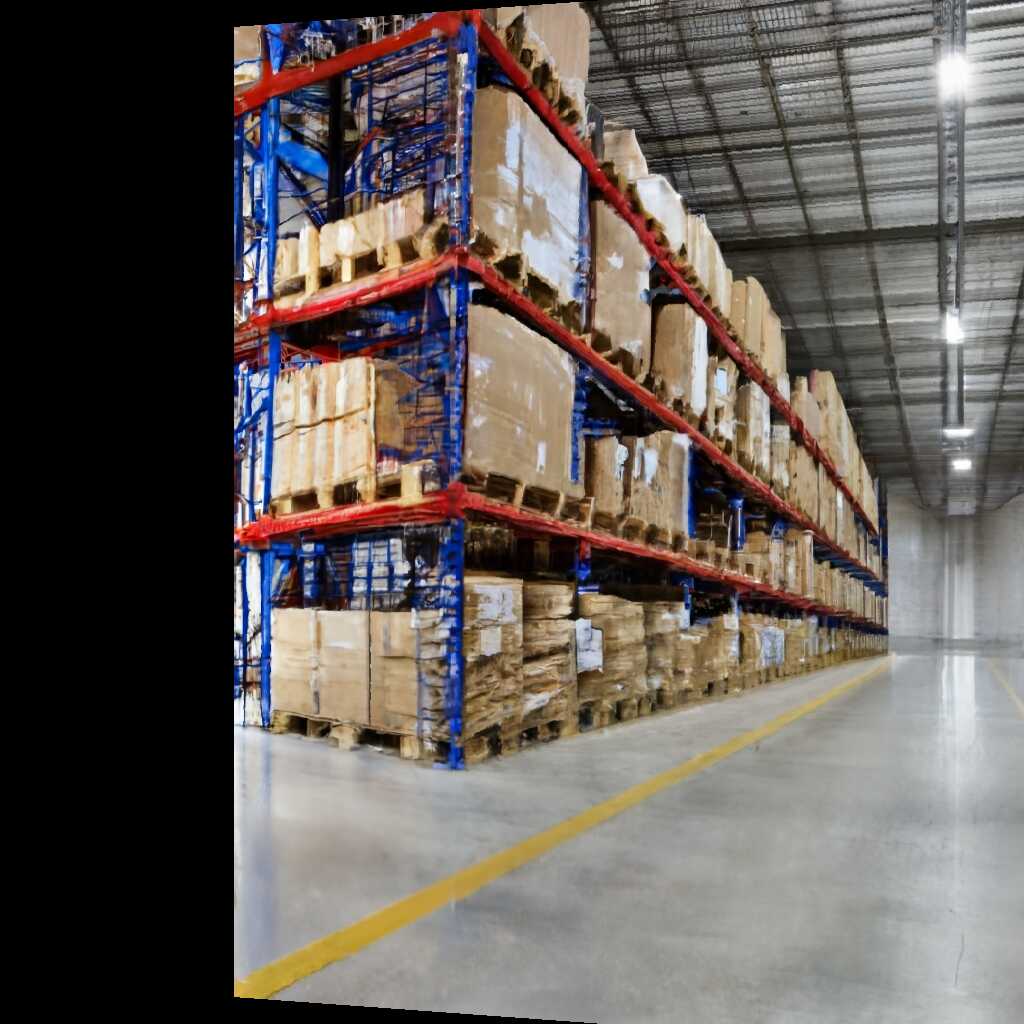} &\includegraphics[width=\imwidth\linewidth]{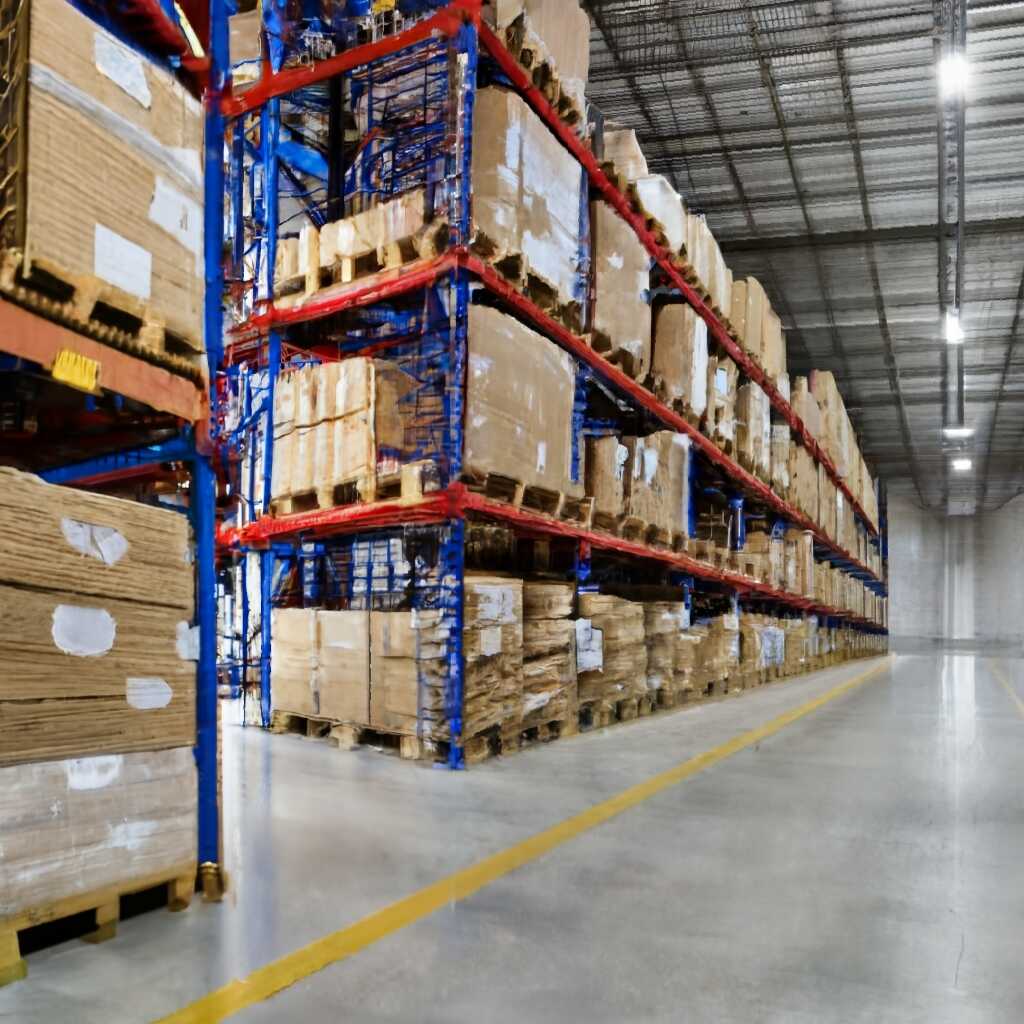} & \includegraphics[width=\imwidth\linewidth]{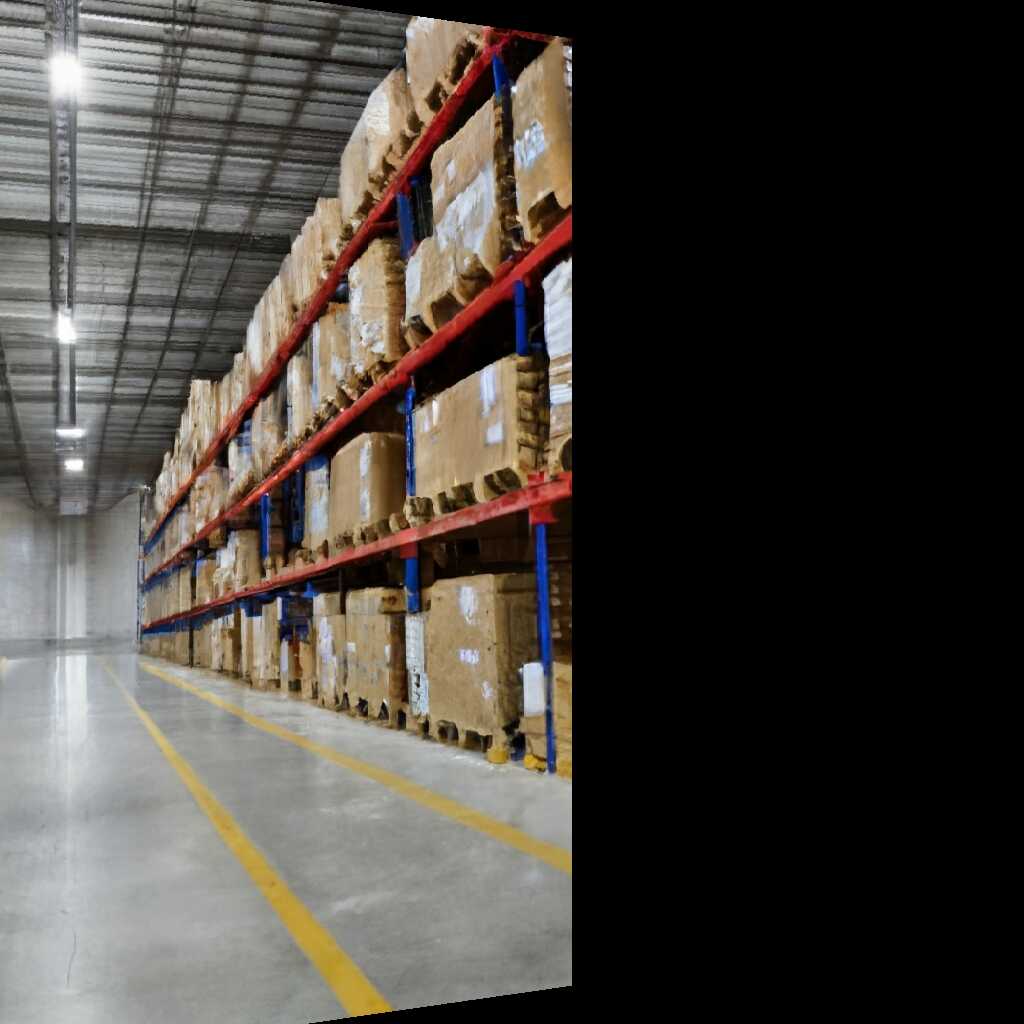} &\includegraphics[width=\imwidth\linewidth]{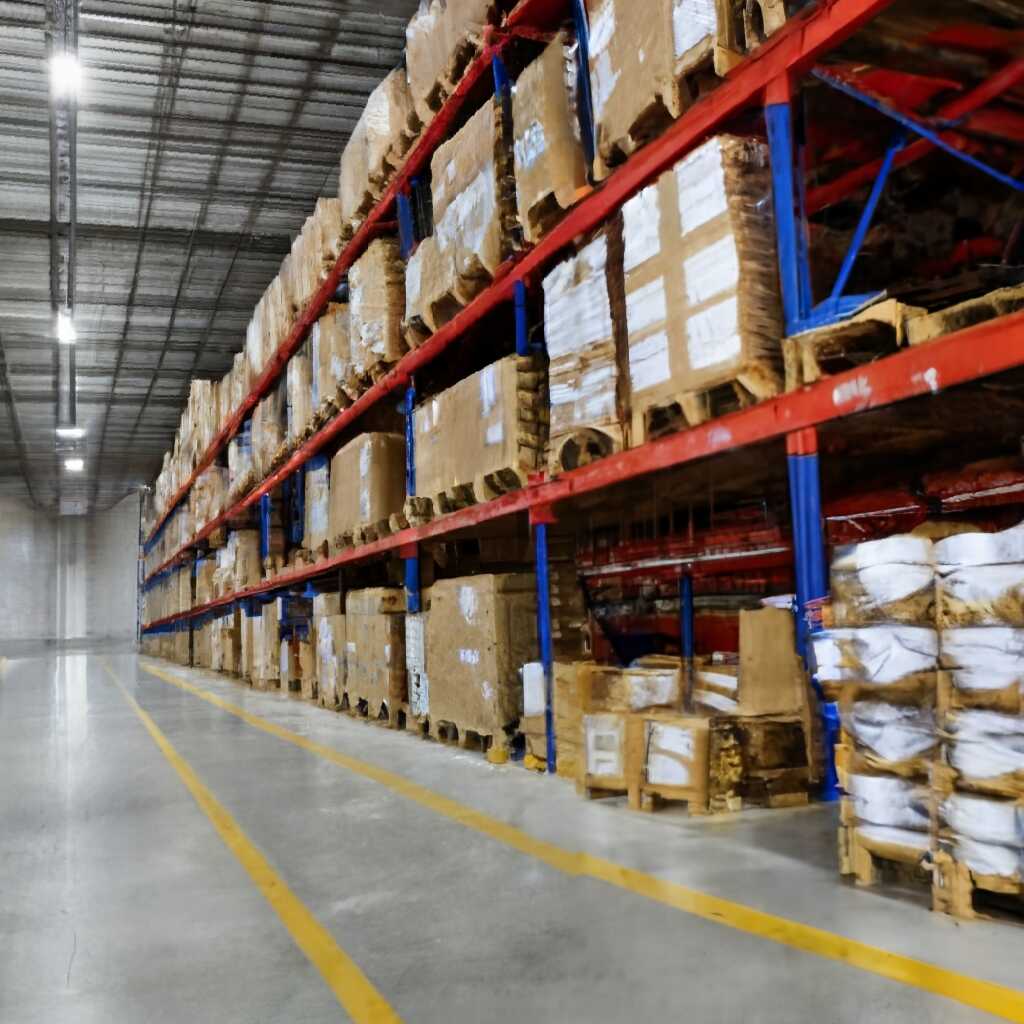} & \multicolumn{2}{c}{\multirow{2}{*}[\lastcoloffset]{\includegraphics[width=\imwidthlast\linewidth]{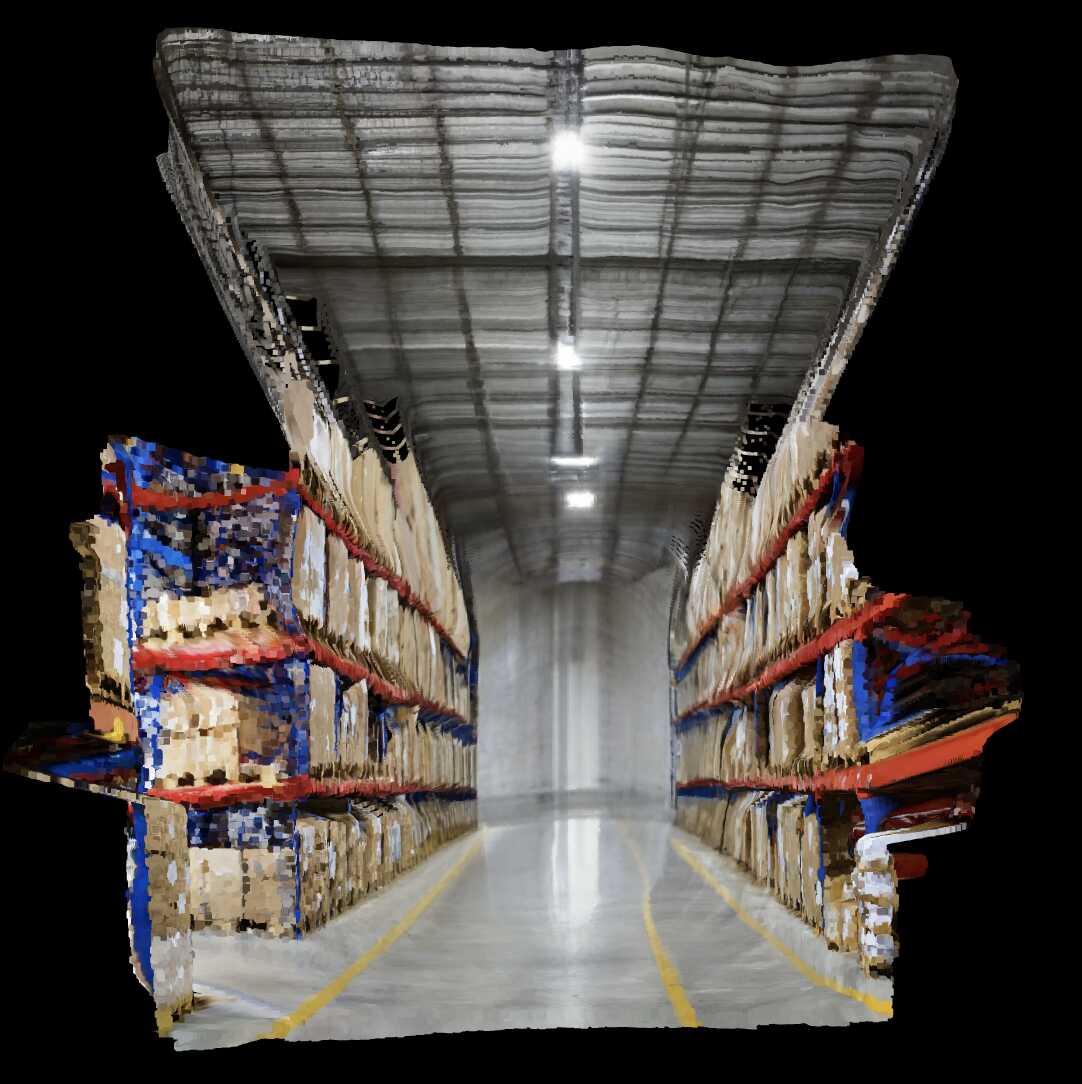}}} \\
    \includegraphics[width=\imwidth\linewidth]{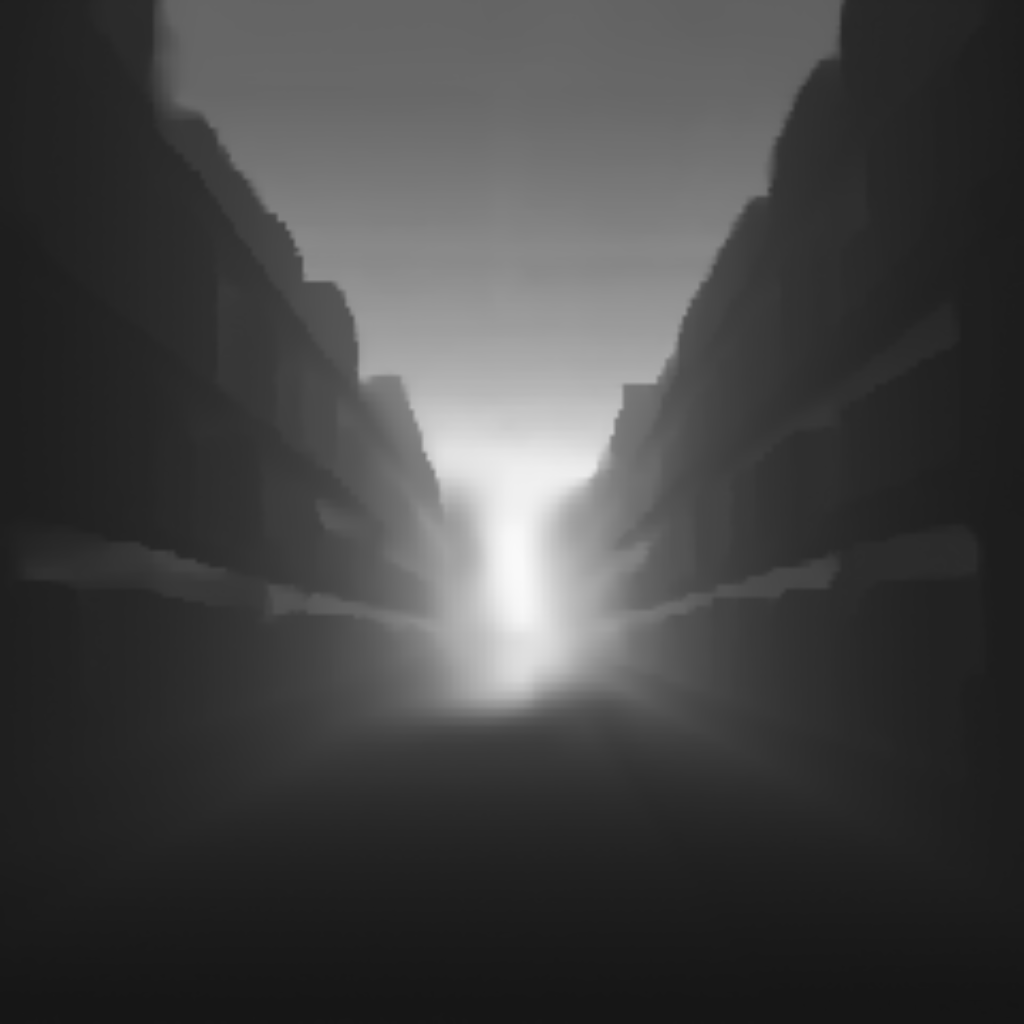} & \includegraphics[width=\imwidth\linewidth]{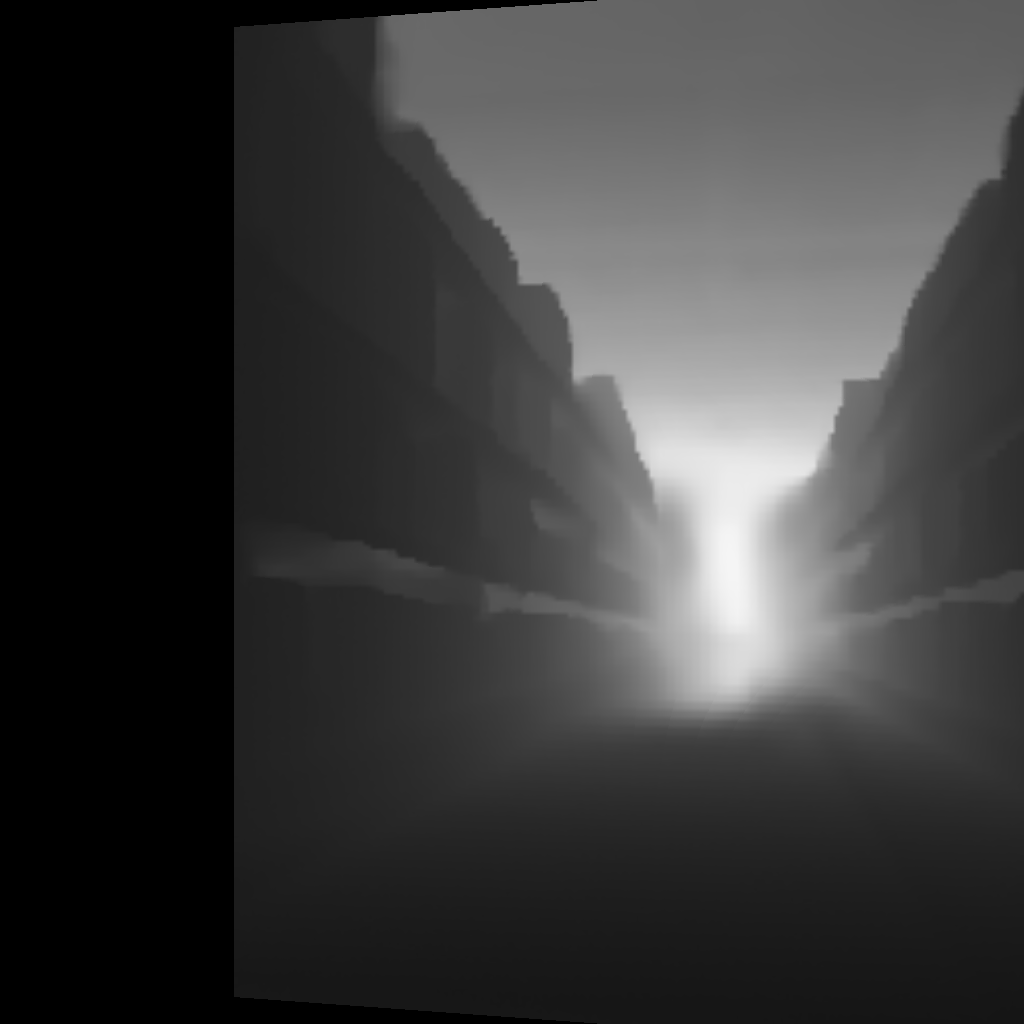} &\includegraphics[width=\imwidth\linewidth]{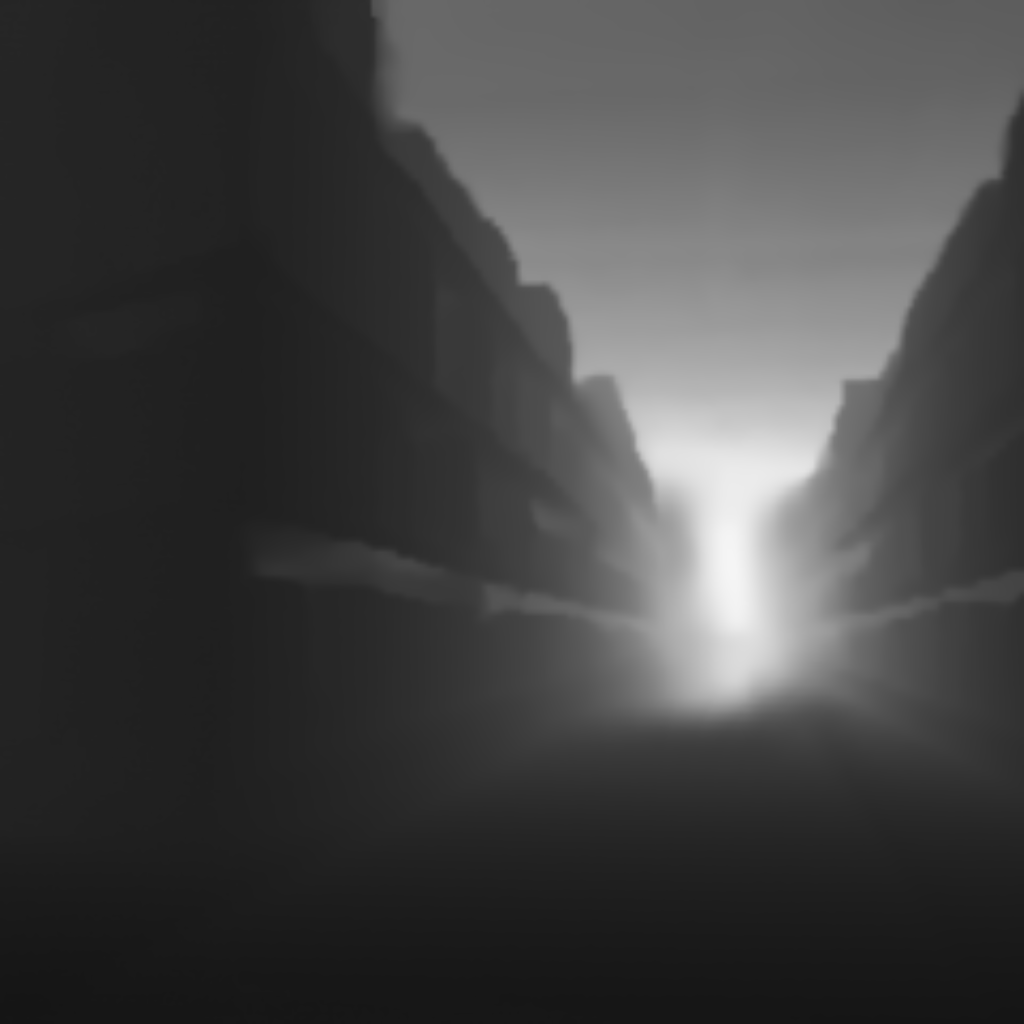} &  \includegraphics[width=\imwidth\linewidth]{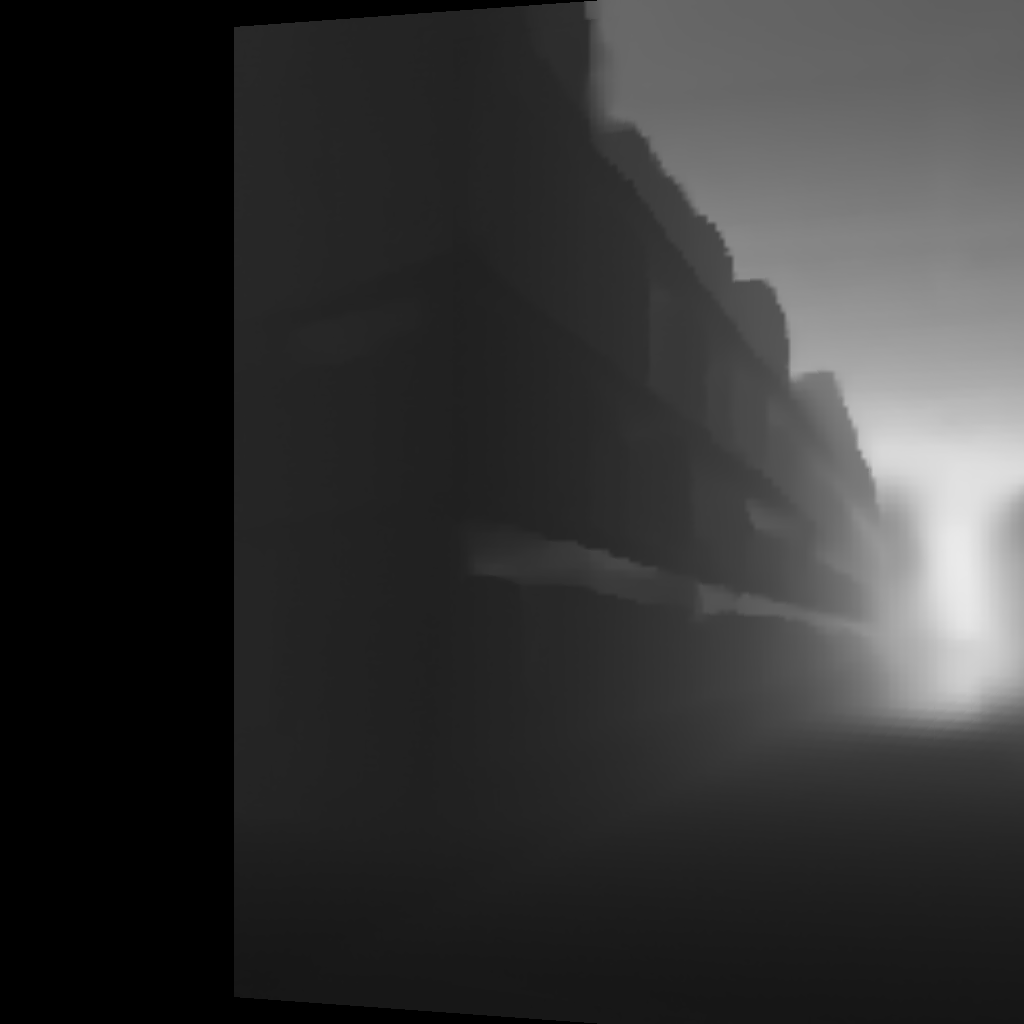} &\includegraphics[width=\imwidth\linewidth]{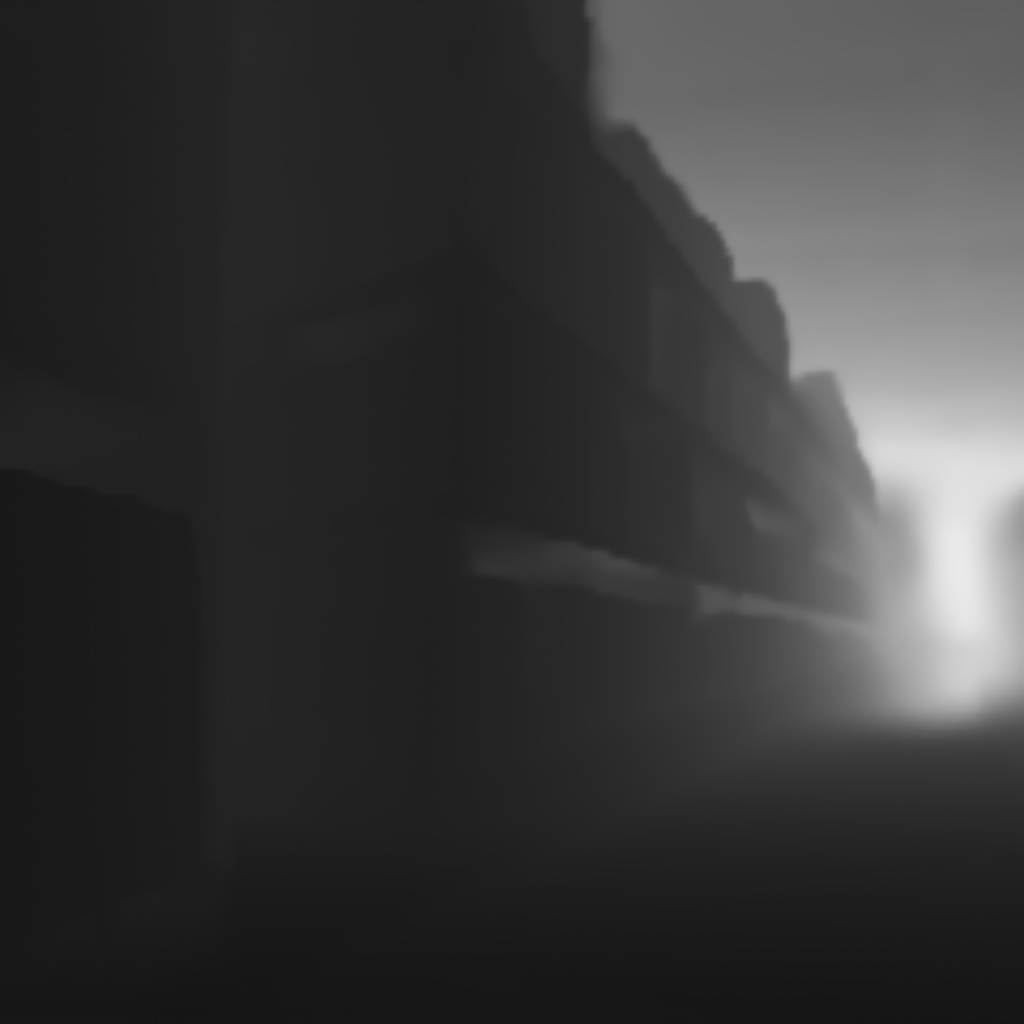} &\includegraphics[width=\imwidth\linewidth]{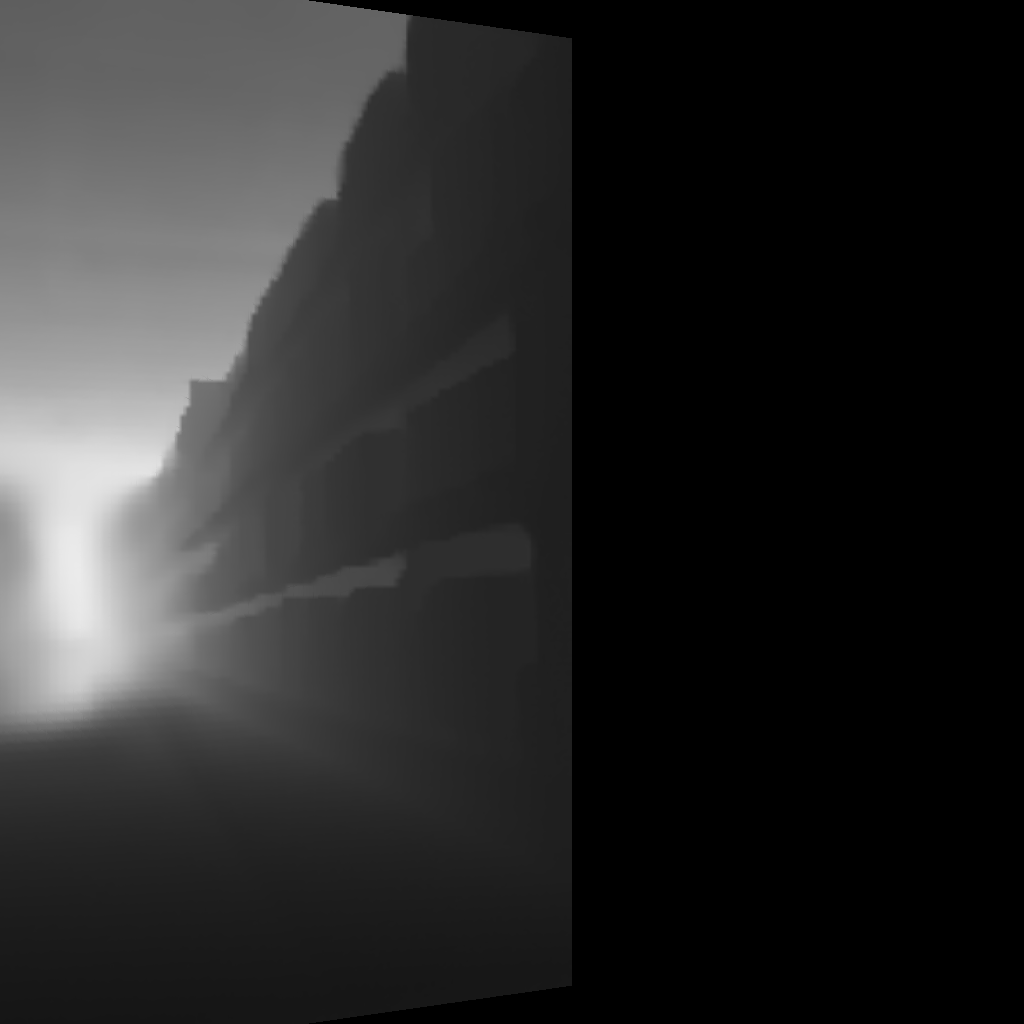} &\includegraphics[width=\imwidth\linewidth]{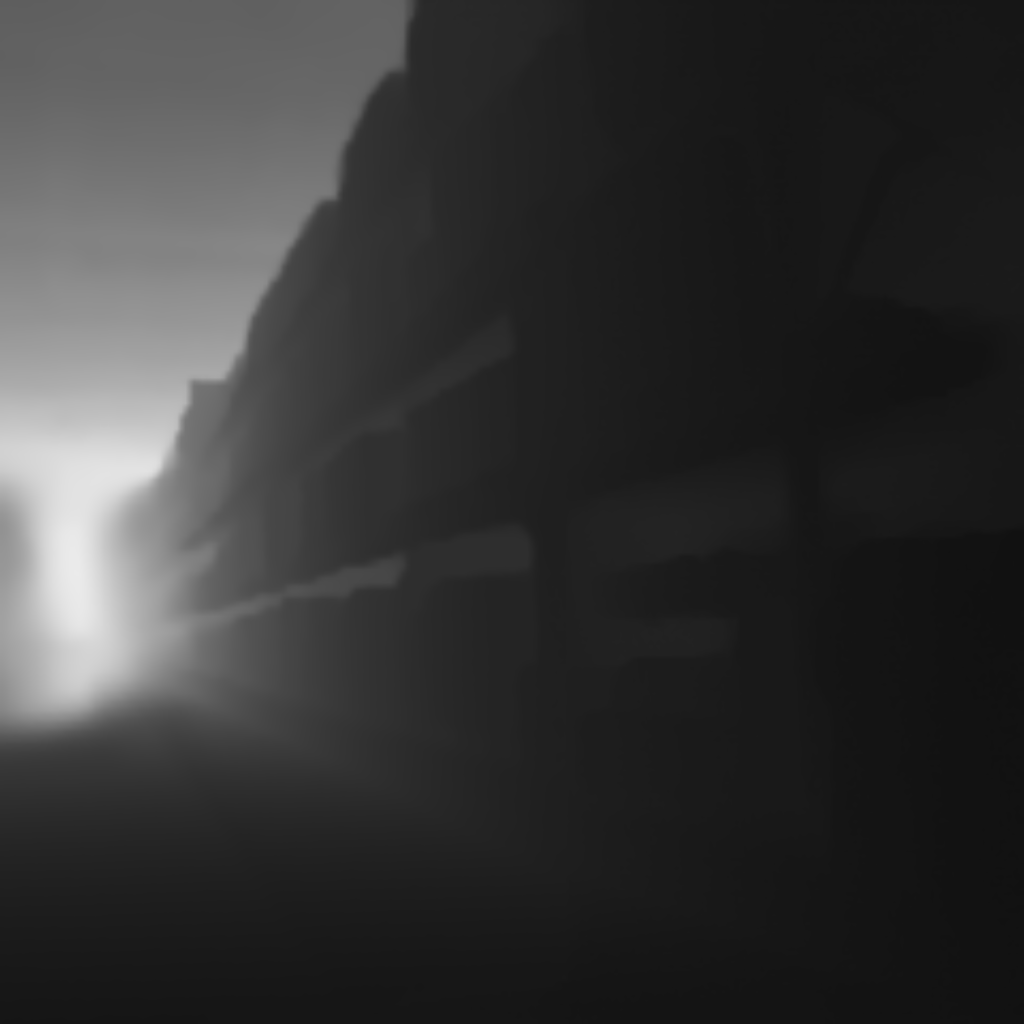} & \multicolumn{2}{c}{} \\
    \\[\textpadding]\multicolumn{\ncols}{c}{Prompt: A warehouse.}\\\\[\textpadding]

    \includegraphics[width=\imwidth\linewidth]{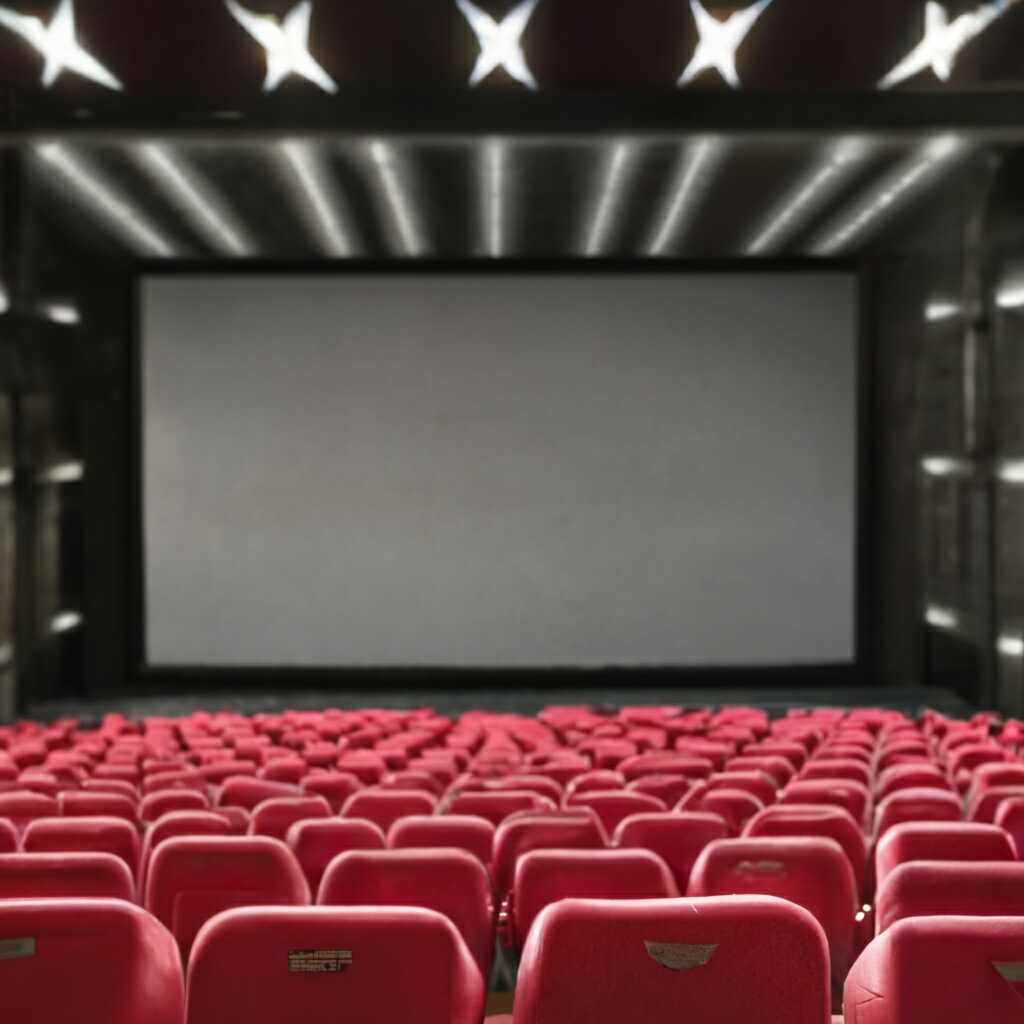} & \includegraphics[width=\imwidth\linewidth]{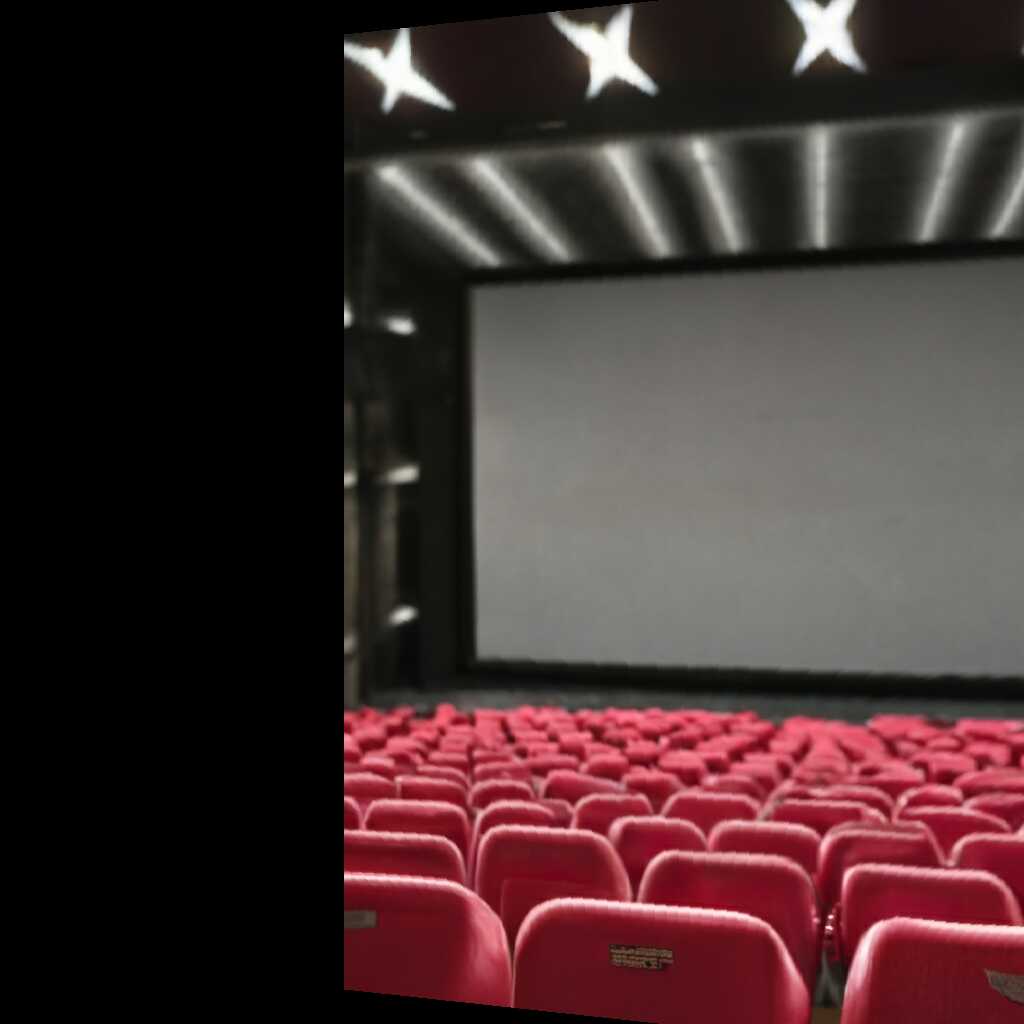} &\includegraphics[width=\imwidth\linewidth]{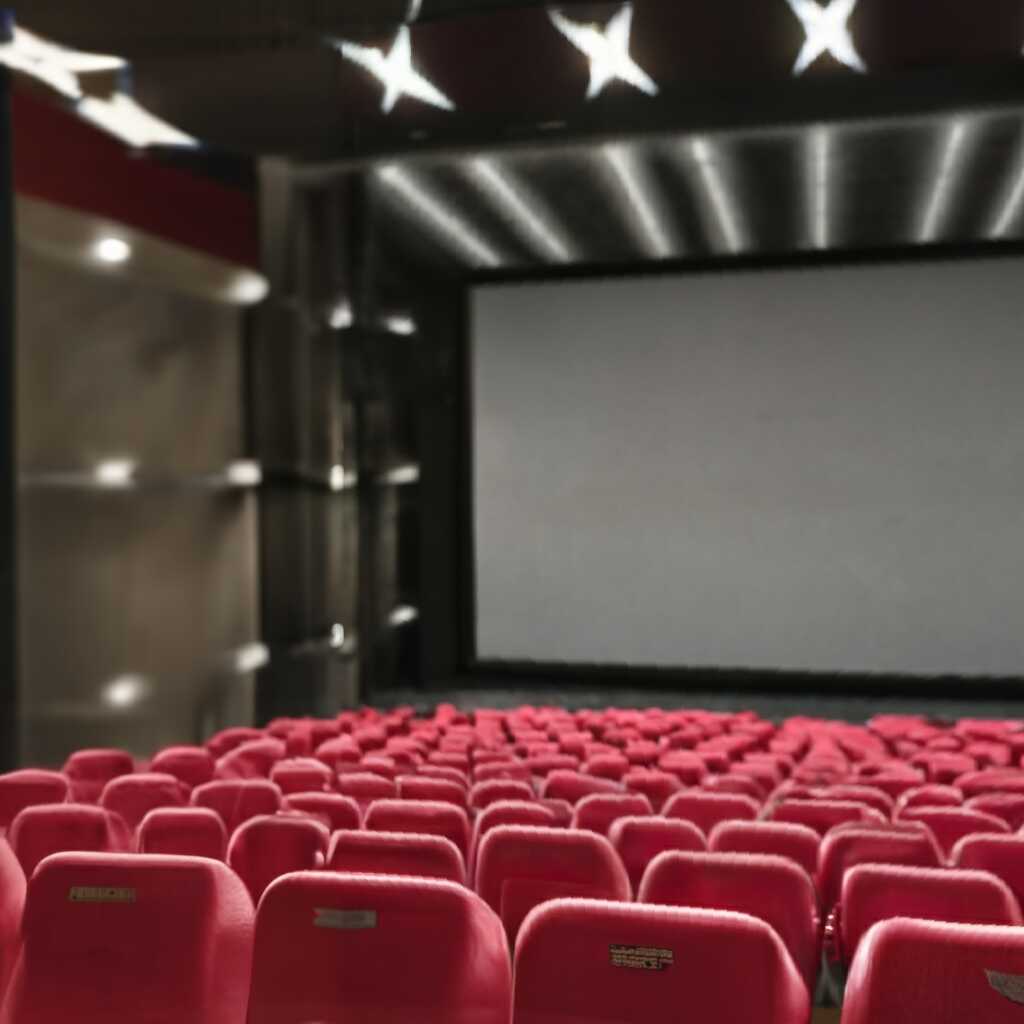} &\includegraphics[width=\imwidth\linewidth]{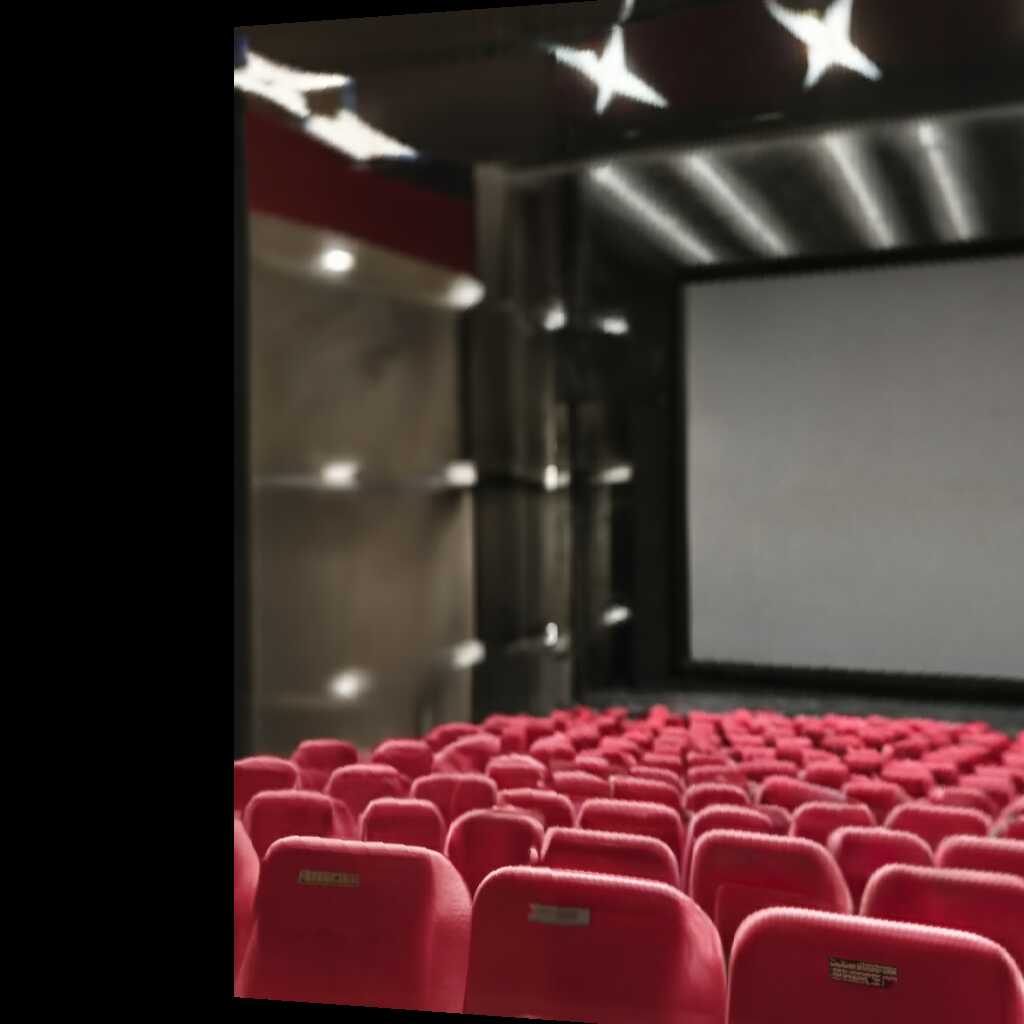} &\includegraphics[width=\imwidth\linewidth]{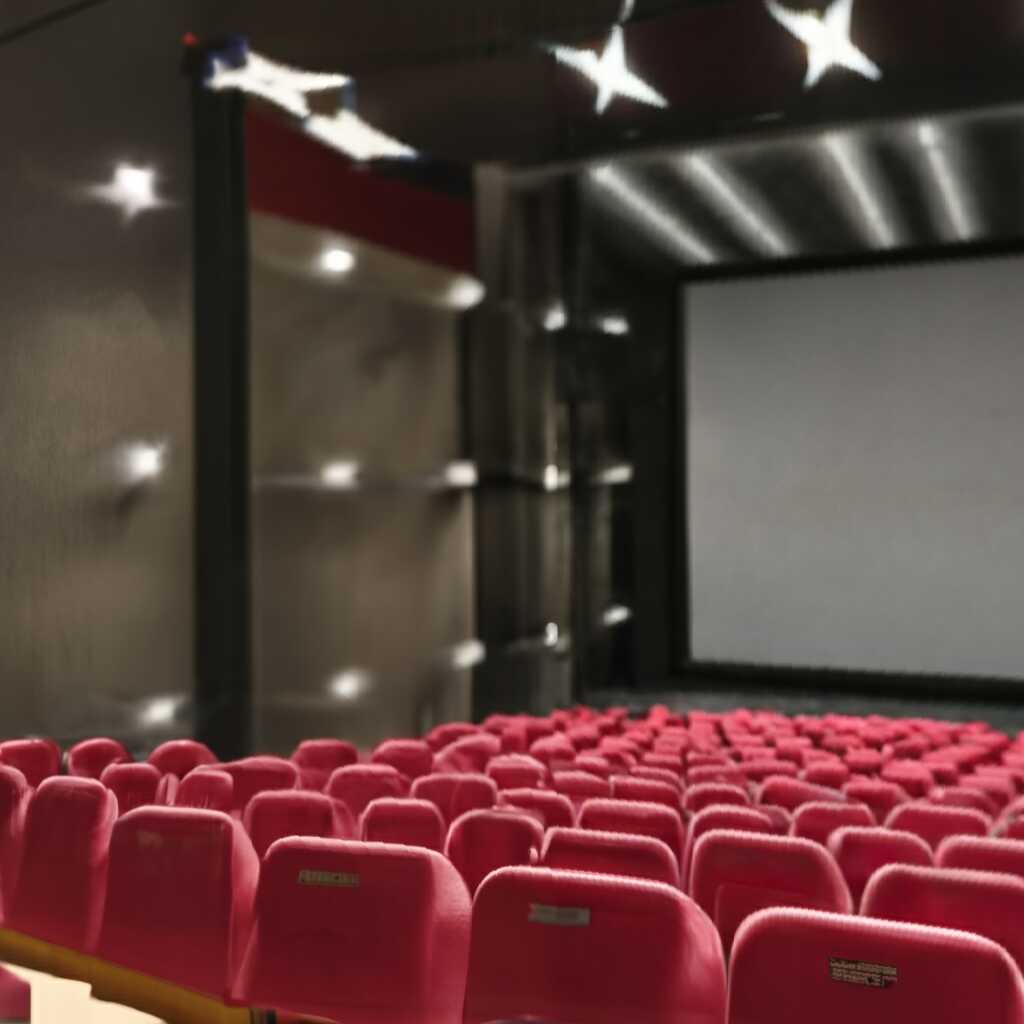} & \includegraphics[width=\imwidth\linewidth]{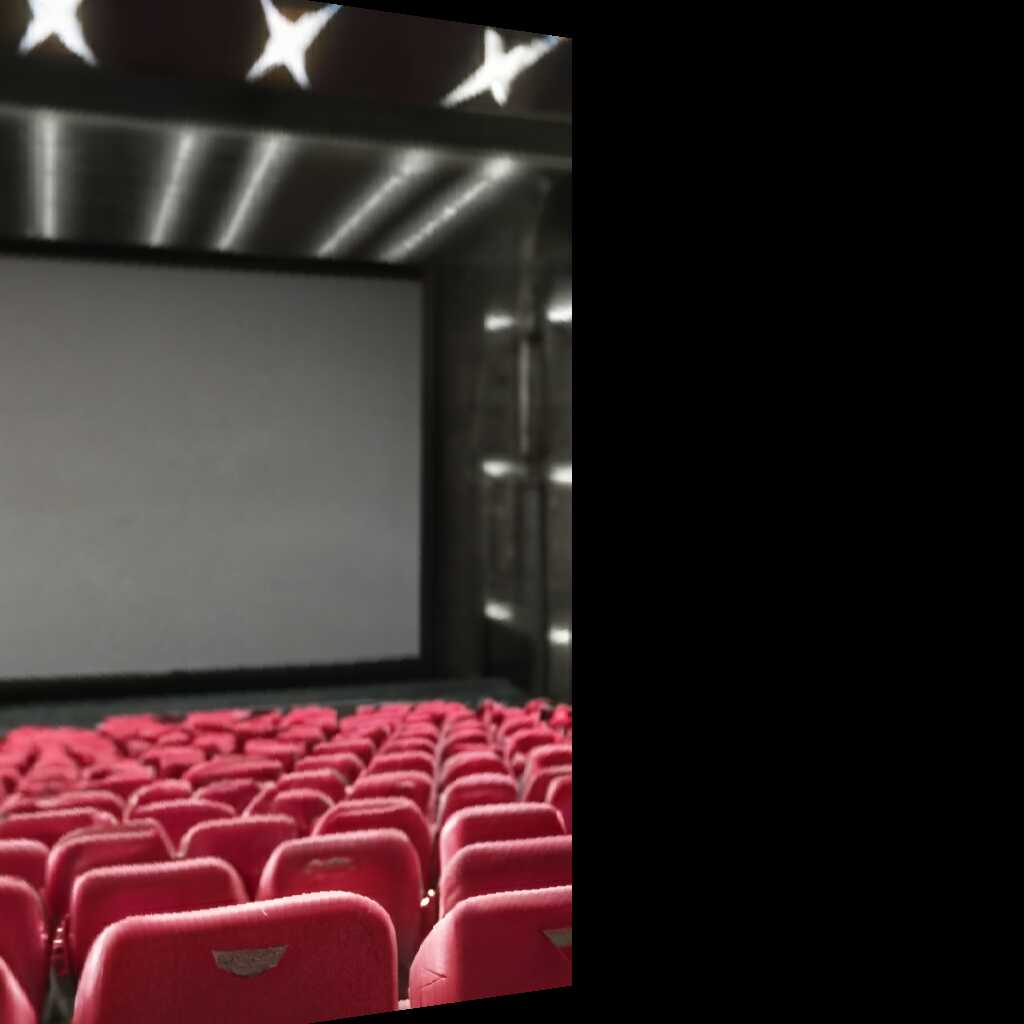} &\includegraphics[width=\imwidth\linewidth]{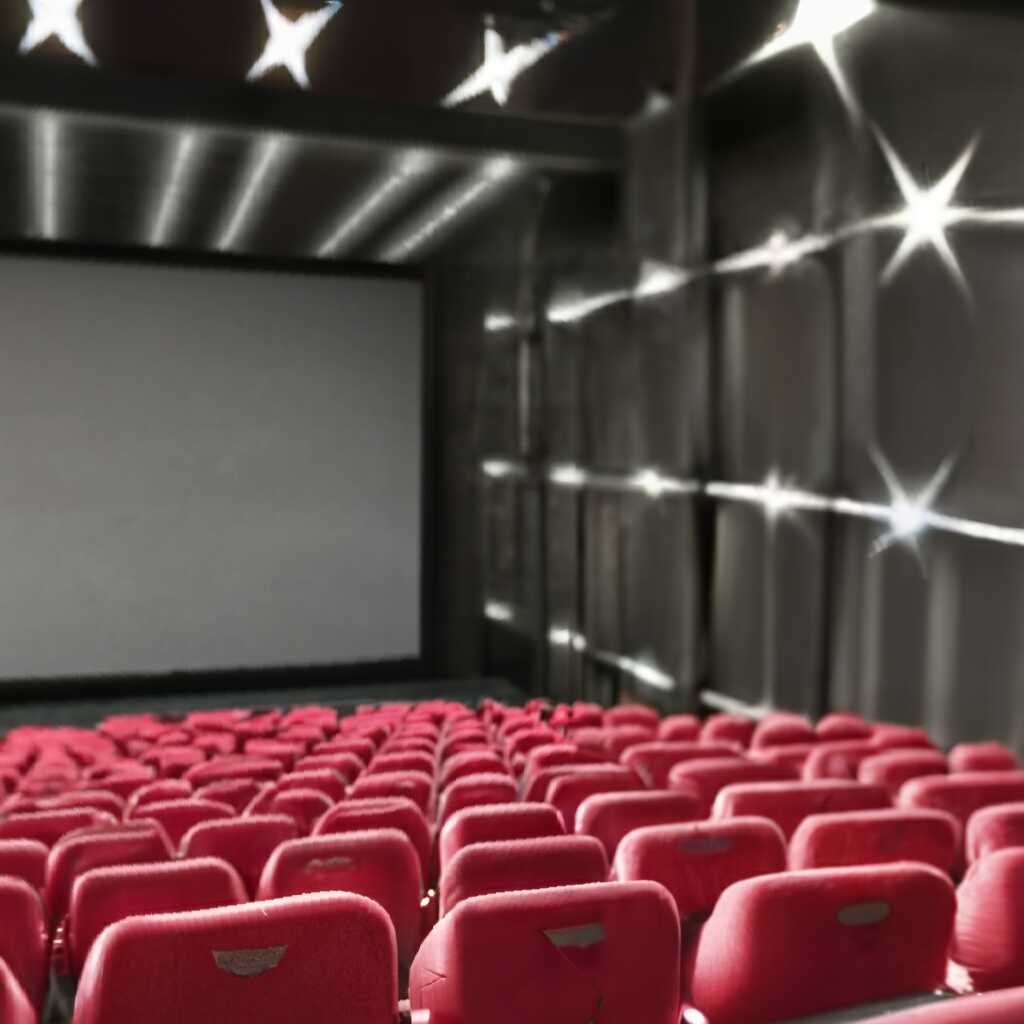} & \multicolumn{2}{c}{\multirow{2}{*}[\lastcoloffset]{\includegraphics[width=\imwidthlast\linewidth]{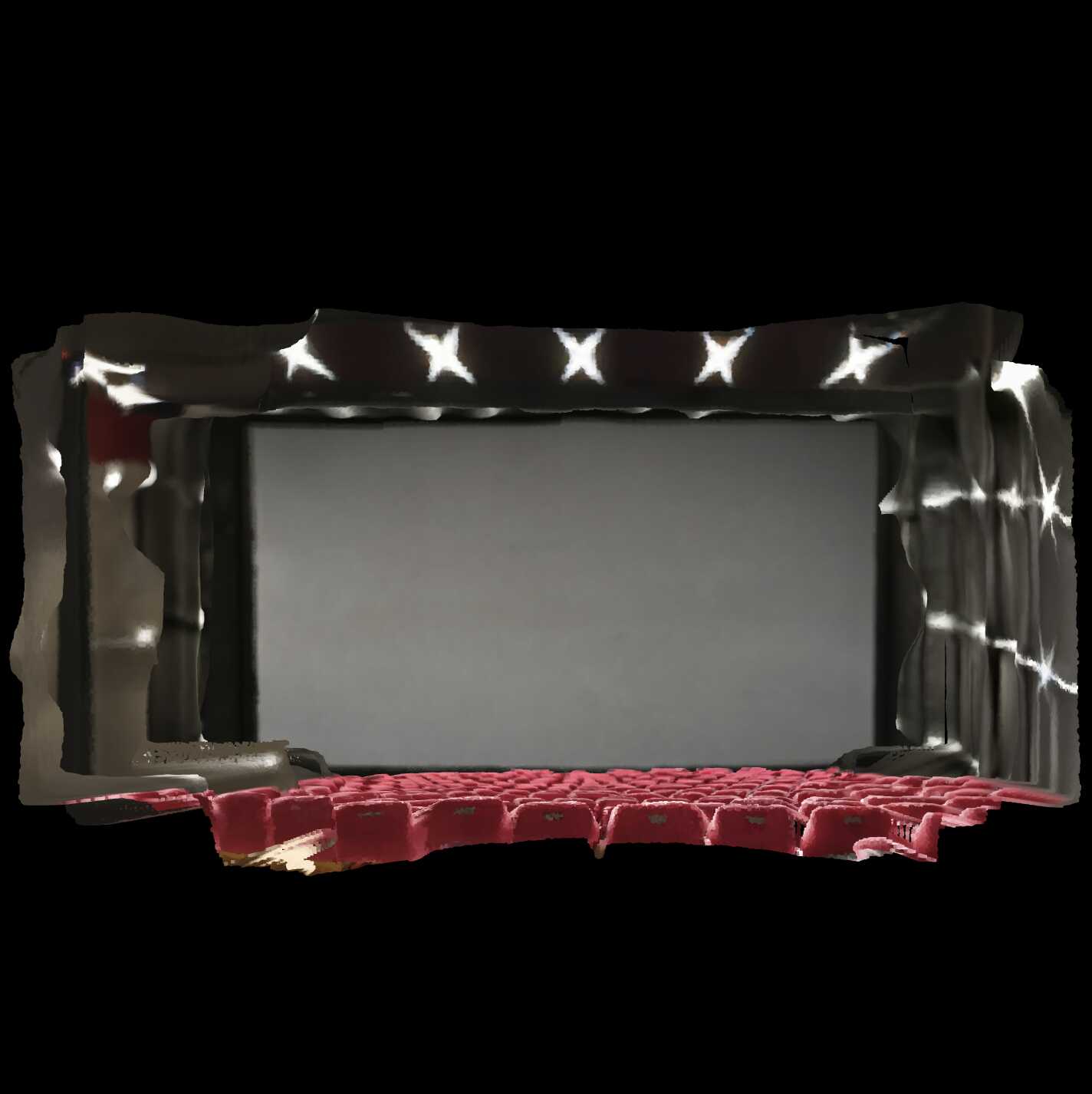}}} \\
    \includegraphics[width=\imwidth\linewidth]{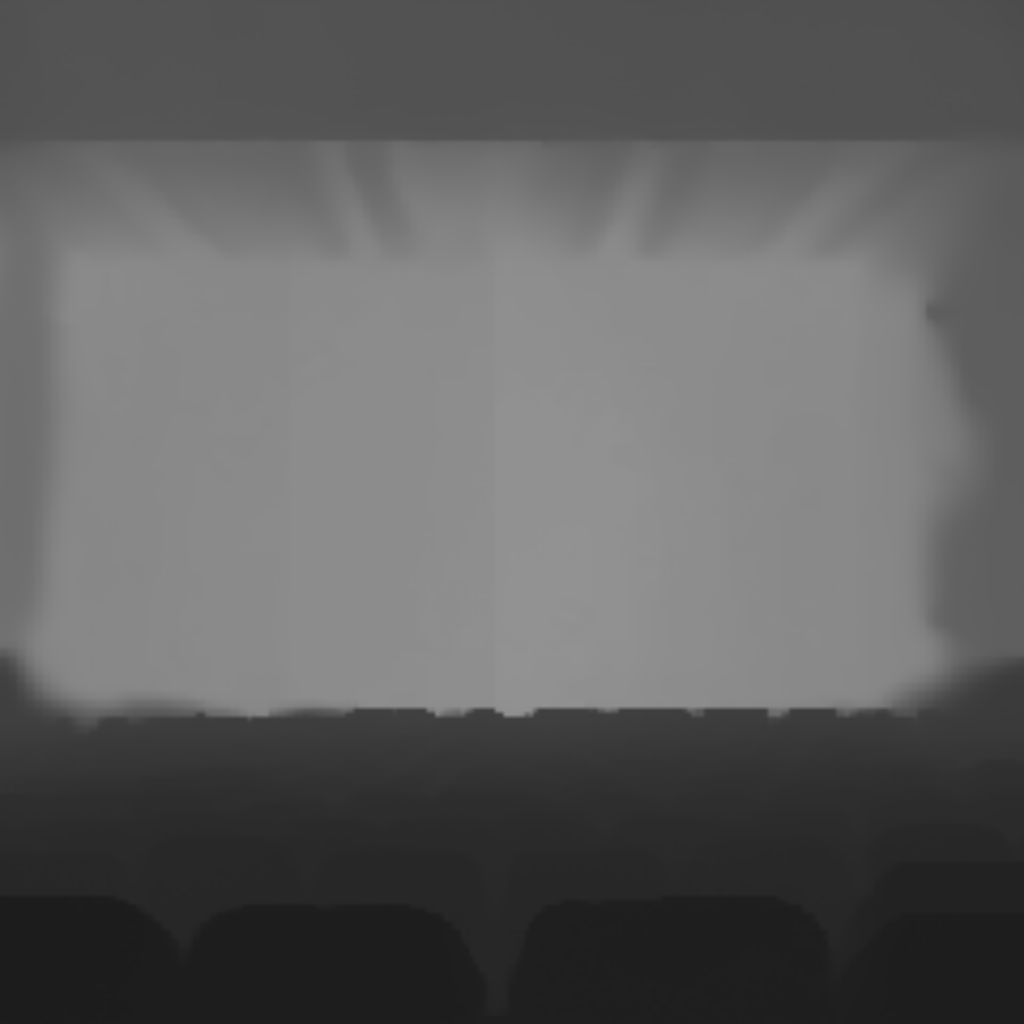} & \includegraphics[width=\imwidth\linewidth]{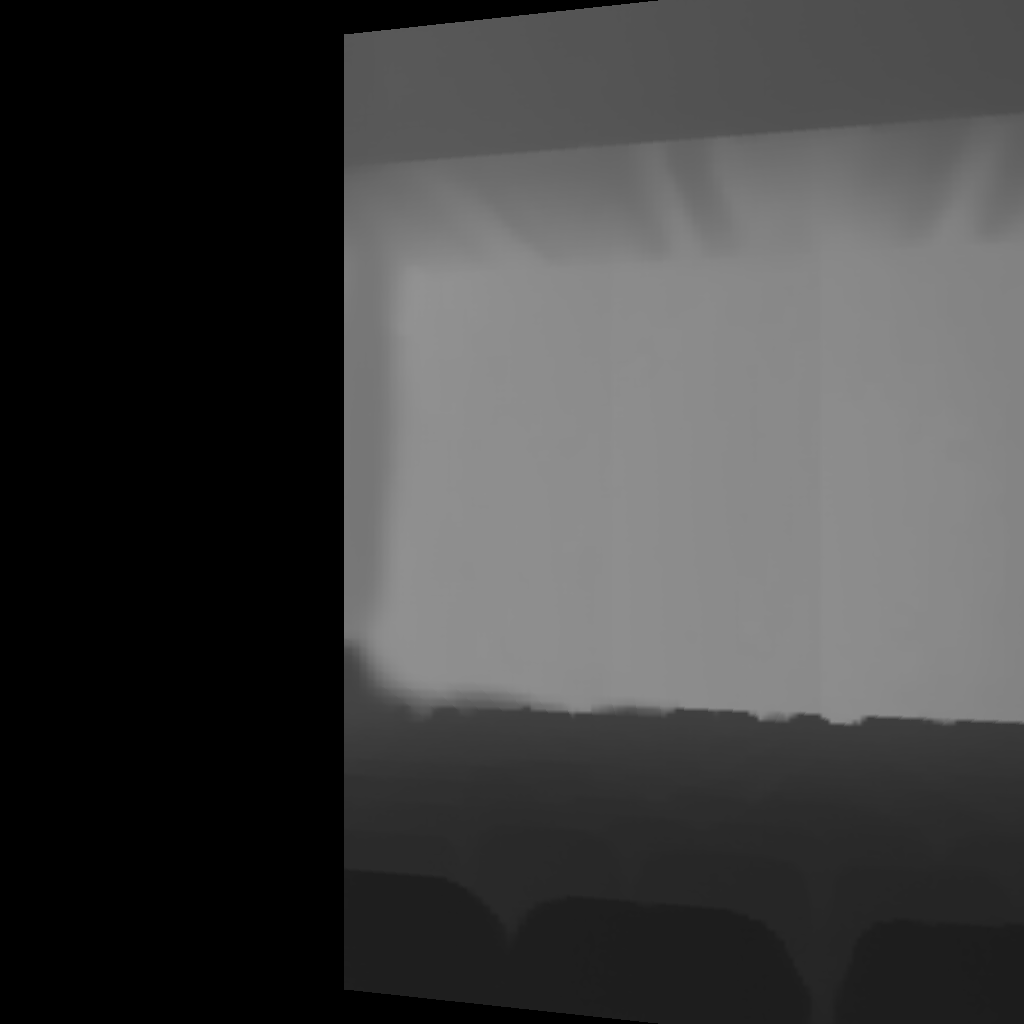} &\includegraphics[width=\imwidth\linewidth]{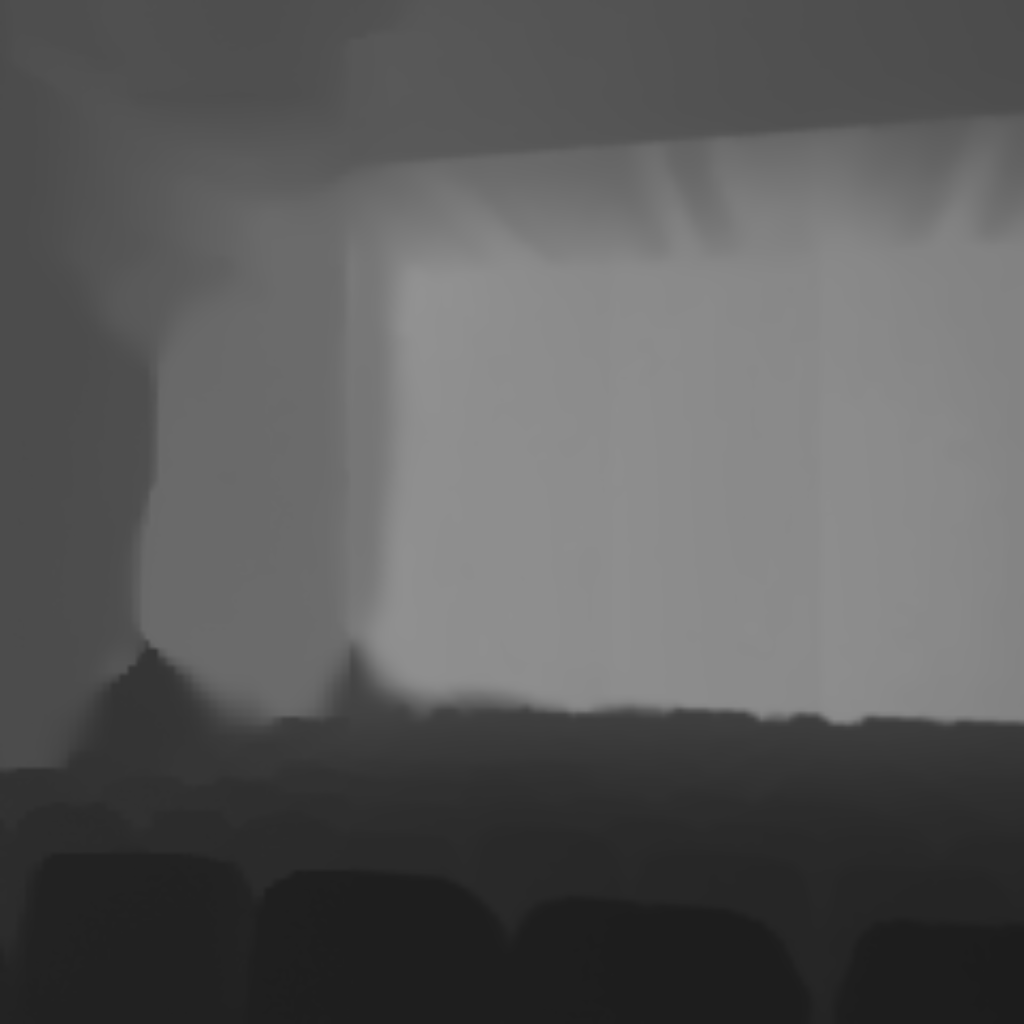} &  \includegraphics[width=\imwidth\linewidth]{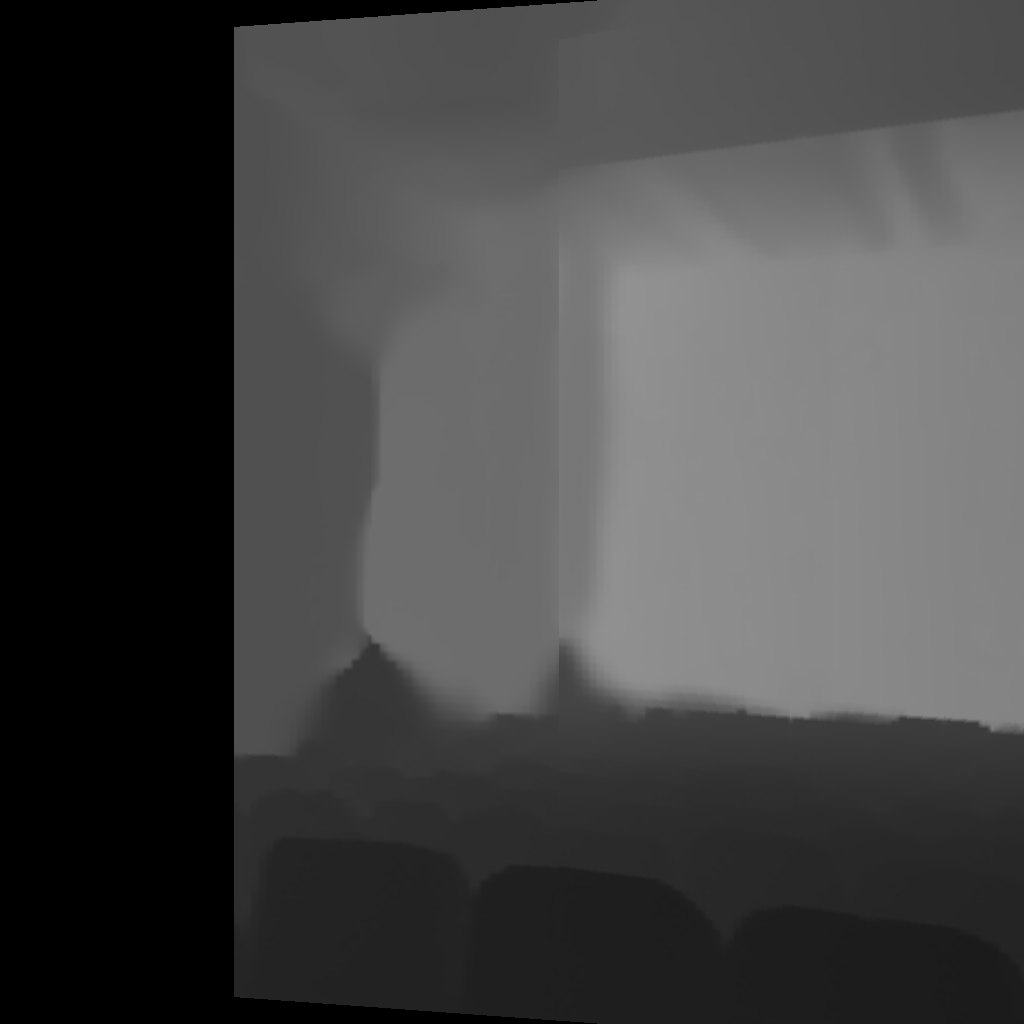} &\includegraphics[width=\imwidth\linewidth]{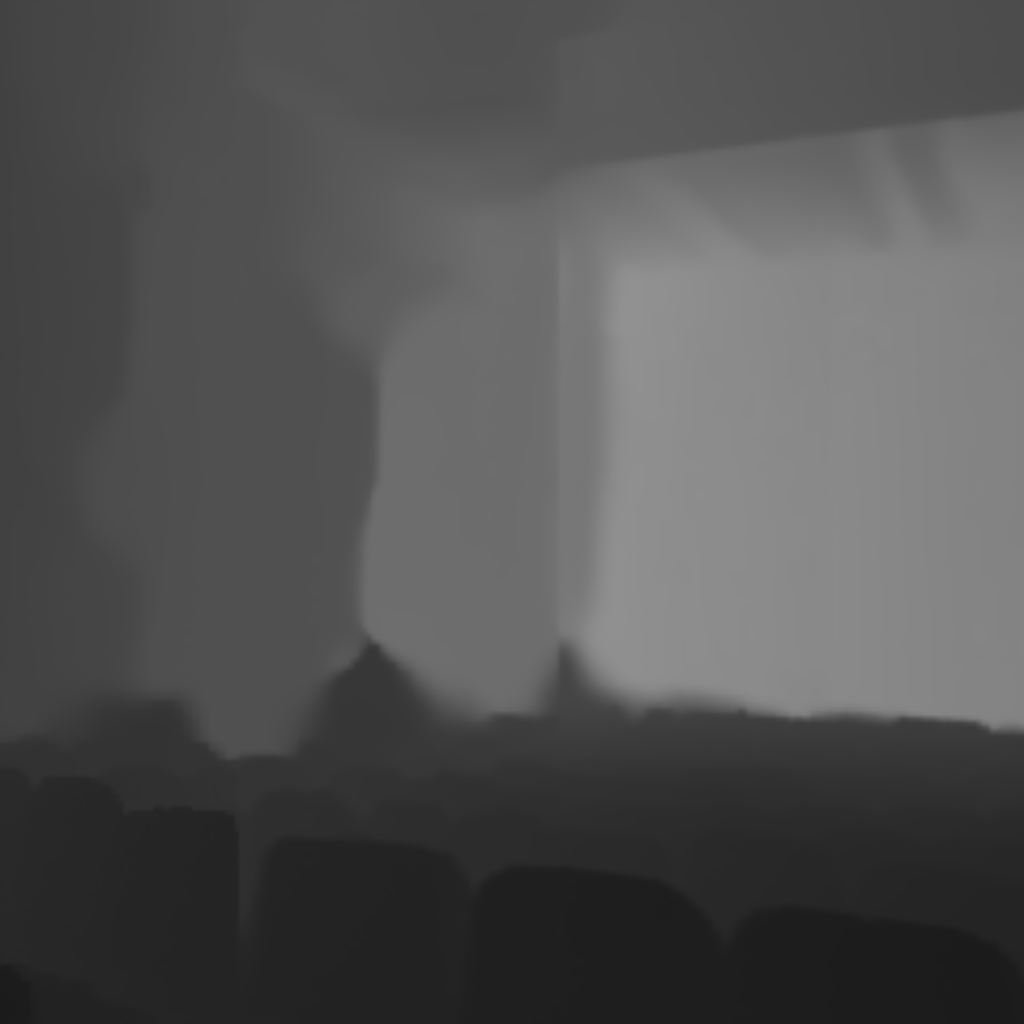} &\includegraphics[width=\imwidth\linewidth]{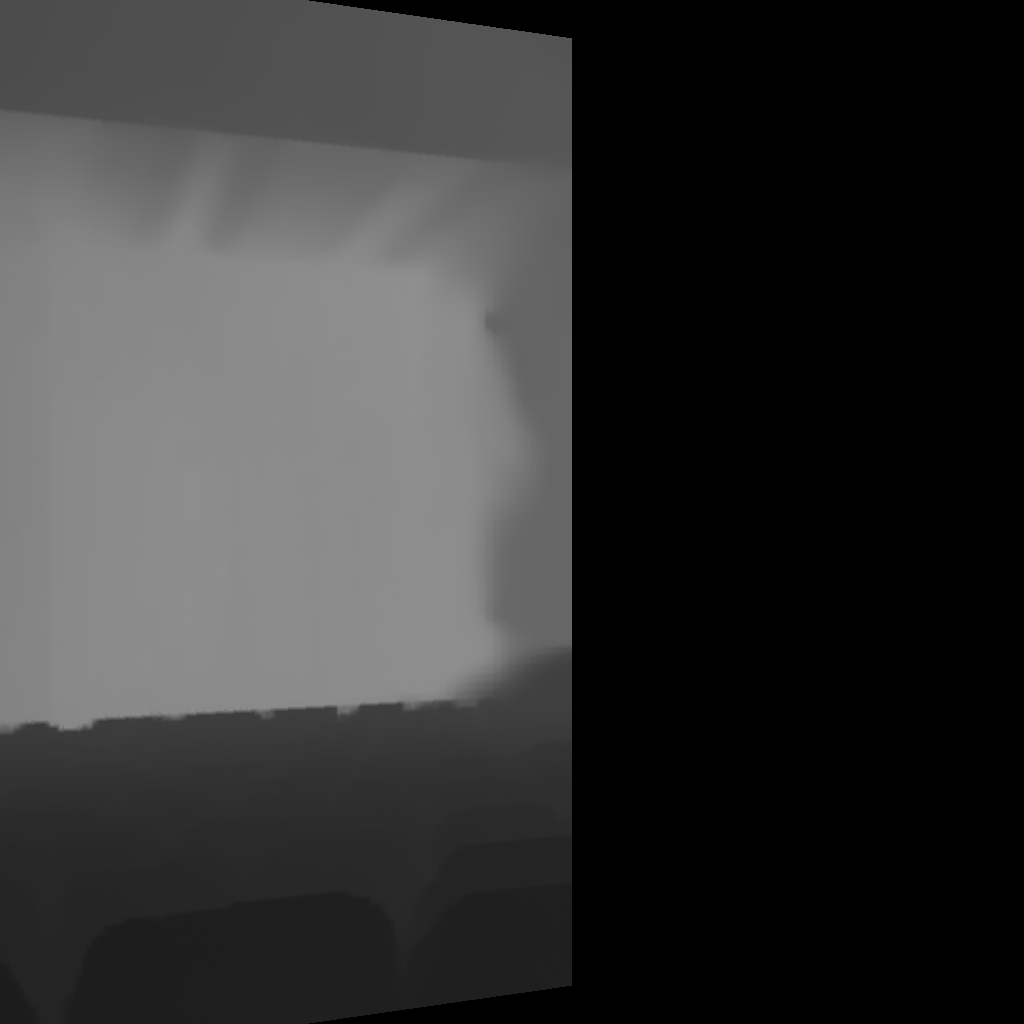} &\includegraphics[width=\imwidth\linewidth]{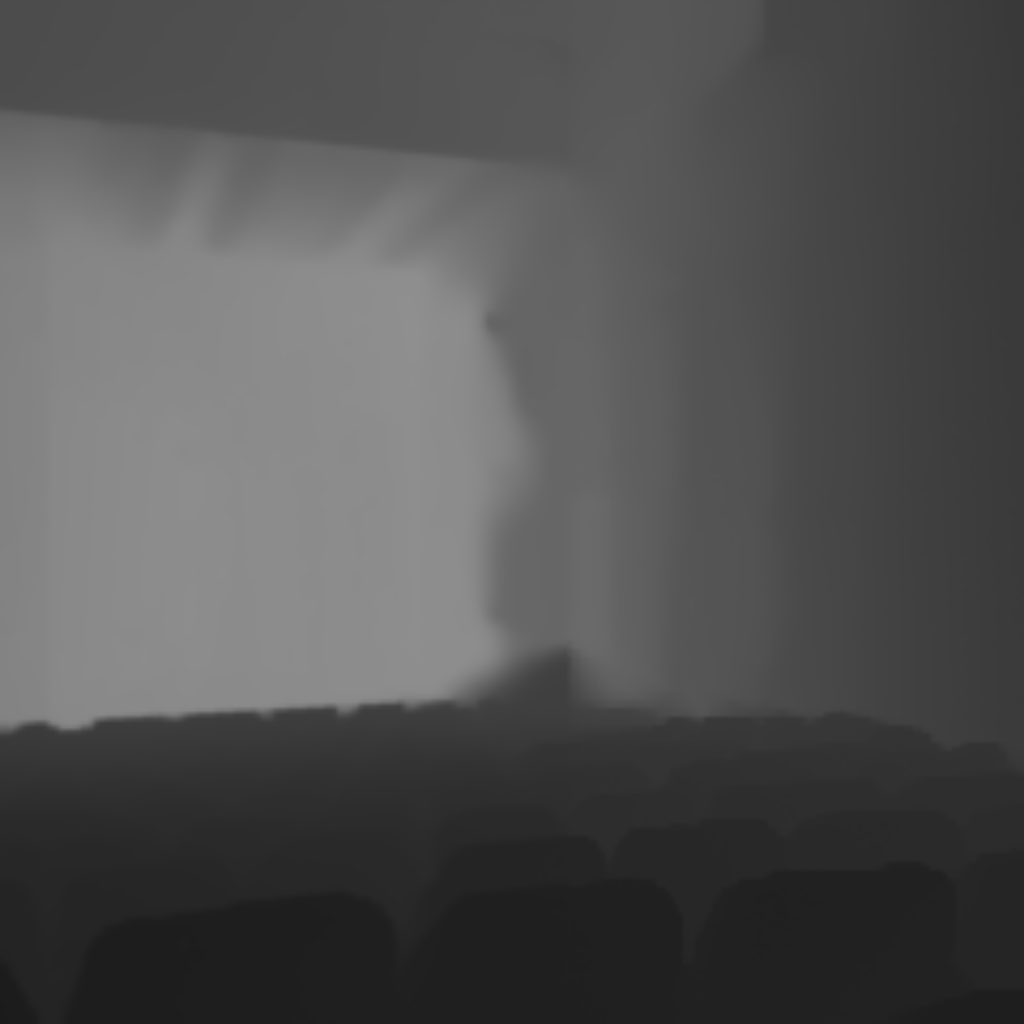} & \multicolumn{2}{c}{} \\
    \\[\textpadding]\multicolumn{\ncols}{c}{Prompt: A movie theatre.}\\\\[\textpadding]
\end{tabular}
\caption{\textbf{Text-to-3D samples}. Given a text prompt, an image is first generated using Imagen \cite{sahariac-imagen} (first row of first column), after which depth is estimated (second row of first column). Subsequently the camera is moved to reveal new parts of the scene which are infilled using an image completion model and our model (which conditions on both the incomplete depth map and the filled image). At each step, newly generated RGBD points are added to a global point cloud which is visualized in the rightmost column.}
\label{fig:text_to_3d_samples_extras}
\end{figure*}

%  

% \vspace*{13cm}
% \vspace*{-1cm}
\section{Complete depth results on NYU and KITTI}
\label{sec:supp-depth-results}
% \vspace*{-0.3cm}

\begin{table*}[h]
    \centering
    \scriptsize
    \caption{\textbf{Comparison of performance on the NYU-Depth-v2 dataset.} $\top$ indicates method uses unsupervised pretraining, \dag~indicates supervised pretraining and \ddag~indicates use of auxiliary supervised depth data. \textbf{Best} / \underline{second best} results are bolded / underlined respectively. $\downarrow$: lower is better and $\uparrow$: higher is better.
    }
    % \tableaftercaption
    % \begin{tabular}{llcccccccc}
    \setlength\tabcolsep{10pt}
    \begin{tabularx}{0.98\linewidth}{l @{\hskip5pt} l c @{\hskip 5em} ccccccc}
        \toprule
        Method & & Architecture & $\delta_1\uparrow$ & $\delta_2\uparrow$ & $\delta_3\uparrow$ & REL$\downarrow$ & RMS$\downarrow$ & $log_{10}\downarrow$ \\
        \midrule
        % \multicolumn{2}{l}{DORN \citep{dorn}} & ResNet-101$^\dag$ &0.828 & 0.965 & 0.992 & 0.115 & 0.509 & 0.051\\
        % \multicolumn{2}{l}{VNL \citep{yin2019enforcing}} & ResNeXt-101$^\dag$ & 0.875 & 0.976 & 0.994 & 0.108 & 0.416 & 0.048 \\
        % \multicolumn{2}{l}{BTS \citep{bts}} & DenseNet-161$^\dag$ &0.885 & 0.978 & 0.994 & 0.110 & 0.392 & 0.047\\
        % \multicolumn{2}{l}{DAV \citep{dav}} & DRN-D-22$^\dag$ &0.882 & 0.980 & 0.996 & 0.108 & 0.412 & --\\
        \multicolumn{2}{l}{TransDepth \citep{transdepth}} & Res-50+ViT-B$^\dag$ & 0.900 & 0.983 & 0.996 & 0.106 & 0.365 & 0.045\\
        \multicolumn{2}{l}{DPT \citep{dpt}} & Res-50+ViT-B$^{\dag\ddag}$ & 0.904 & 0.988 & \underline{0.998} & 0.110 & 0.357 & 0.045 \\
        \multicolumn{2}{l}{AdaBins \citep{adabins}} & E-B5+Mini-ViT$^{\dag}$ & 0.903 & 0.984 & \third{0.997} & 0.103 & 0.364 & 0.044 \\
        \multicolumn{2}{l}{BinsFormer \citep{binsformer}} & Swin-Large$^\dag$ & 0.925 & 0.989 & \third{0.997} & 0.094 & 0.330 & 0.040 \\
        \multicolumn{2}{l}{PixelFormer \citep{pixelformer}} & Swin-Large$^\dag$ & 0.929 & \third{0.991} & \underline{0.998} & 0.090 & 0.322 & 0.039 \\
        \multicolumn{2}{l}{MIM \citep{xie2022revealing}} & SwinV2-L$^\top$ & \underline{0.949} & \first{0.994} & \first{0.999} & 0.083 & \second{0.287} & \third{0.035} \\
        \multicolumn{2}{l}{AiT-P \citep{AllInTokens}} & SwinV2-L$^\top$ &  \first{0.953} &  \second{0.993} & \first{0.999} & \third{0.076} & \first{0.279} & \second{0.033} \\
         \midrule
%        # \ours &  &  &  &  &  & & \\
        \multirow{3}{*}{\textbf{\ours}} & samples=1 & \unet$^{\top\ddag}$ & 0.944 & 0.986 & 0.995 & \second{0.075} & 0.324 & \first{0.032} \\
        & samples=2 & \unet$^{\top\ddag}$ & 0.944 & 0.987 & 0.996 & \first{0.074} & 0.319 & \first{0.032} \\
        & samples=4 & \unet$^{\top\ddag}$ & \third{0.946} & 0.987 & 0.996 & \first{0.074} & 0.315 & \first{0.032} \\
        % & samples=8 & \unet$^{\top\ddag}$ & \third{0.946} & 0.987 & 0.996 & \first{0.074} & \third{0.314} & \first{0.032} \\
        \bottomrule
    \end{tabularx} 
    % \end{tabular} 
    \label{tab::results-nyu}
    % \vskip -0.1in
\end{table*}
\vspace*{-0.2cm}
\begin{table*}[h]
    \centering
    \scriptsize
    \caption{\textbf{Comparison of performance on the KITTI dataset.} $\top$ indicates method uses unsupervised pretraining, \dag~indicates supervised pretraining and \ddag~indicates use of auxiliary supervised depth data. \textbf{Best} / \underline{second best} results are bolded / underlined respectively. $\downarrow$: lower is better and $\uparrow$: higher is better. E-B5: EfficientNet-B5~\cite{tan2019efficientnet}.}
    \label{tab::results-kitti}
    % \tableaftercaption
    
    \setlength{\tabcolsep}{4.5pt}
    \begin{tabularx}{0.98\linewidth}{l @{\hskip 5pt} l @{\hskip 3em} c @{\hskip 4em} ccccccc}
    % \begin{tabular}{lcccccccc}
       \toprule
        Method & & Backbone & ~~~\textbf{$\delta_1$}$\uparrow$~~~ & ~~~\textbf{$\delta_2$}$\uparrow$~~~ & ~~~\textbf{$\delta_3$}$\uparrow$~~~ & ~REL~$\downarrow$~ & Sq-rel~$\downarrow$ & RMS~$\downarrow$ & RMS log~$\downarrow$ \\
        \midrule
       % Saxena~\etal~\cite{saxena2005learning} & - & 0.601 & 0.820  & 0.926 & 0.280   & 3.012  & 8.734 & 0.361    \\
       % Eigen~\etal~\cite{eigen2014depth} & Convolutions & 0.702 & 0.898 & 0.967 & 0.203   & 1.548  & 6.307 & 0.282    \\        % Liu~\etal~\cite{liu2015learning} & Convolutions & 0.680 & 0.898 & 0.967 & 0.201   & 1.584  & 6.471 & 0.273    \\
    %   Godard~\etal~\cite{godard2017unsupervised} & ResNet-50 & 0.861 & 0.949 & 0.976 & 0.114   & 0.898  & 4.935 & 0.206 \\
    %   Johnston~\etal~\cite{johnston2020self} & ResNet-101$^\dag$ & 0.889 & 0.962 & 0.982 & 0.106 & 0.861 & 4.699 & 0.185 \\
    %   Gan~\etal~\cite{gan2018monocular} & ResNet-101 & 0.890 & 0.964 & 0.985 & 0.098   & 0.666  & 3.933 & 0.173 \\
      % PLADE-Net~\cite{gonzalez2021plade} & FAL-Net & 0.900 & 0.967 & 0.985 & 0.089 & 0.590 & 4.008 & 0.172\\
    %   DORN \cite{dorn}  & ResNet-101$^\dag$ & 0.932 & 0.984 & 0.994 & 0.072   & 0.307  & 2.727 & 0.120\\
    %   VNL \cite{yin2019enforcing} & ResNext-101$^\dag$ & 0.938 & 0.990  & \third{0.998} & 0.072 & -- & 3.258 & 0.117 \\
    %   PGA-Net \cite{xu2020probabilistic} & ResNet-50$^\dag$ & 0.952 & 0.992 & \third{0.998} & 0.063 & 0.267 & 2.634 & 0.101 \\
    \multicolumn{2}{l}{BTS \cite{bts}} & DenseNet-161$^\dag$ & 0.956 & 0.993 & \third{0.998} & 0.059   & 0.245  & 2.756 & 0.096 \\
    \multicolumn{2}{l}{TransDepth \cite{transdepth}} & ResNet-50+ViT-B$^\dag$ & 0.956 & 0.994 & \second{0.999} & 0.064 & 0.252  & 2.755 & 0.098 \\
    \multicolumn{2}{l}{DPT \cite{dpt}} &ResNet-50+ViT-B$^{\dag\ddag}$ & 0.959 & \third{0.995} & \second{0.999} & 0.062 & --& 2.573 & 0.092 \\
    \multicolumn{2}{l}{AdaBins \cite{adabins}} & E-B5+mini-ViT$^{\dag}$ & 0.964 & \third{0.995} & \second{0.999} & 0.058  & 0.190  & 2.360 & 0.088\\
    \multicolumn{2}{l}{BinsFormer \cite{binsformer}} & Swin-Large$^{\dag}$ & \third{0.974} &  \second{0.997} & \second{0.999} & \third{0.052} & \third{0.151} & \second{2.098} & \third{0.079} \\
    \multicolumn{2}{l}{PixelFormer \cite{pixelformer}} & Swin-Large$^{\dag}$ & \second{0.976} & \second{0.997}  & \second{0.999} & \second{0.051} & \second{0.149} & \second{2.081} & \second{0.077} \\
    \multicolumn{2}{l}{MIM \cite{xie2022revealing}} & SwinV2-L$^\top$ & \first{0.977} & \first{0.998} & \first{1.000} & \first{0.050} & \first{0.139} & \first{1.966} & \first{0.075} \\

    \midrule
    \multirow{3}{*}{\textbf{\ours}} & samples=1 & \unet$^{\top\ddag}$ & 0.964 &   0.994 &   0.998 &   0.056 &   0.339 &  2.700 &   0.091 \\
    & samples=2 & \unet$^{\top\ddag}$ & 0.965&   0.994&   \third{0.998}&   0.055&   0.325 &   2.660 & 0.090 \\
    & samples=4 & \unet$^{\top\ddag}$ &  0.965 &   0.994 &   \third{0.998}&   0.055 &   0.292 &   2.613 &   0.089 \\
    % & samples=8 & \unet$^{\top\ddag}$ &  0.965&   0.994 &   \third{0.998}&   0.055 &   0.295 &   2.613 &   0.089 \\
    \bottomrule
\end{tabularx}
\end{table*} 
\vspace*{-0.2cm}

Tables \ref{tab::results-nyu} and \ref{tab::results-kitti} provide detailed results on the val set of NYU depth v2 and KITTI depth datasets. We follow the standard evaluation protocol used in prior work \citep{binsformer}. For both the NYU depth v2 and KITTI datasets we report the absolute relative error (REL), root mean squared error (RMS) and accuracy metrics ($\delta_i < 1.25^i$ for $i \in 1, 2, 3$). For NYU we also report absolute error of log depths ($log_{10}$). For KITTI we additionally report the squared relative error (Sq-rel) and root mean squared error of log depths (RMS log). The predicted depth is up-sampled to the full resolution using bilinear interpolation before evaluation. For the indoor model we evaluate on the cropped region proposed by \cite{eigen2014depth} and for the outdoor model the cropped region proposed by \cite{garg2016unsupervised} as is standard in prior work.

\section{Ablations}
\vspace*{-0.2cm}

Tables \ref{tab::loss-ablation-nyu} and \ref{tab::loss-ablation-kitti} show that an $L_1$ loss in training the diffusion model performs much better than an $L_2$ loss for monocular depth estimation on NYU and KITTI. Tables \ref{tab::pretraining-ablation-nyu} and \ref{tab::pretraining-ablation-kitti} show the effectiveness of Palette-style \cite{sahariac-palette} self-supervised pretraining for monocular depth estimation on NYU and KITTI respectively. All results use a single sample. Because  these findings are reasonable and expected to generalize to other dense vision tasks, we do not further ablate them for optical flow estimation for compute efficiency.
% \srbs{Consider placing the indoor/outdoor ablation tables side-by-side}

\begin{table*}[!htbp]
    \centering
    \footnotesize
    \caption{Ablation for the choice of loss function on the NYU depth v2 dataset.}
    \vspace*{-0.1cm}
    \begin{tabular}{lccccccc}
        \toprule
         & $\delta_1\uparrow$ & $\delta_2\uparrow$ & $\delta_3\uparrow$ & REL$\downarrow$ & RMS$\downarrow$ & $log_{10}\downarrow$ \\
        \midrule
        $L_2$ & 0.932 & 0.981 & 0.994 & 0.085 & 0.349 & 0.037  \\
        $L_1$ & \textbf{0.944} & \textbf{0.986} & \textbf{0.995} & \textbf{0.075} & \textbf{0.324} & \textbf{0.032} \\
        \bottomrule
    \end{tabular} 
    \label{tab::loss-ablation-nyu}
    % \vskip -0.1in
\end{table*}
\begin{table*}[!htbp]
    \centering
    \footnotesize
    \caption{Ablation for the choice of loss function on the KITTI dataset.}
    \label{tab::loss-ablation-kitti}
    \vspace*{-0.1cm}
    \begin{tabular}{lcccccccc}
       \toprule
         & ~~~\textbf{$\delta_1$}$\uparrow$~~~ & ~~~\textbf{$\delta_2$}$\uparrow$~~~ & ~~~\textbf{$\delta_3$}$\uparrow$~~~ & ~REL~$\downarrow$~ & Sq-rel~$\downarrow$ & RMS~$\downarrow$ & RMS log~$\downarrow$ \\
        \midrule
        $L_2$ & 0.954 & 0.993 & \textbf{0.998} & 0.065 & \textbf{0.321} & 2.773 & 0.099   \\
        $L_1$ & \textbf{0.964} & \textbf{0.994}  &  \textbf{0.998} & \textbf{0.056}  & 0.339  & \textbf{2.700}  &  \textbf{0.091} \\
    \bottomrule
\end{tabular}
\end{table*} 
\begin{table*}[!htbp]
    \centering
    \footnotesize
    \caption{Ablation for self-supervised pretraining on the NYU depth v2 dataset.}
    \vspace*{-0.1cm}
    \setlength\tabcolsep{1.5pt}
    \begin{tabular}{l @{\hskip 4em} *{7}{@{\hskip 1.2em}c}}
        \toprule
         & $\delta_1\uparrow$ & $\delta_2\uparrow$ & $\delta_3\uparrow$ & REL$\downarrow$ & RMS$\downarrow$ & $log_{10}\downarrow$ \\
        \midrule
        % From scratch & 0.777 & 0.920 & 0.969 & 0.155 & 0.601 & 0.069  \\
        % Unsupervised pretraining only & 0.877 & 0.964 & 0.988 & 0.116 & 0.450 & 0.049 \\
        No self-supervised pre-training & 0.936 & 0.980 & 0.992 & 0.081 & 0.352 & 0.035  \\
        With self-supervised pre-training & \textbf{0.944} & \textbf{0.986} & \textbf{0.995} & \textbf{0.075} & \textbf{0.324} & \textbf{0.032} \\
        \bottomrule
    \end{tabular} 
    \label{tab::pretraining-ablation-nyu}
    % \vskip -0.1in  
\end{table*}
\begin{table*}[!htbp]
    \centering
    \footnotesize
    \caption{Ablation for self-supervised pretraining on the KITTI depth dataset.}
    \vspace*{-0.1cm}
    \setlength\tabcolsep{1.5pt}
    \begin{tabular}{l @{\hskip 2em} cccccccc}
       \toprule
         & ~~~\textbf{$\delta_1$}$\uparrow$~~~ & ~~~\textbf{$\delta_2$}$\uparrow$~~~ & ~~~\textbf{$\delta_3$}$\uparrow$~~~ & ~REL~$\downarrow$~ & Sq-rel~$\downarrow$ & RMS~$\downarrow$ & RMS log~$\downarrow$ \\
        \midrule
        % From scratch & 0.886 & 0.962 & 0.987 & 0.102 & 0.776 & 4.477 & 0.166  \\
        % Unsupervised pretraining only & 0.912 & 0.976  & 0.993 & 0.090 &  0.892 & 4.094  & 0.145  \\
        No self-supervised pre-training & 0.952 & 0.990 & 0.997 & 0.064  & 0.389 & 2.998 &  0.104 \\
        With self-supervised pre-training & \textbf{0.965} &  \textbf{0.994}  & \textbf{0.998}  &  \textbf{0.055}  & \textbf{0.332}   & \textbf{2.696}   &  \textbf{0.091}  \\
    \bottomrule
\end{tabular}
\label{tab::pretraining-ablation-kitti}
\end{table*}

\vspace*{0.1cm}
\section{Coarse-to-fine refinement for depth}
\label{sec:supp-coarse-to-fine-depth}
\vspace*{-0.2cm}

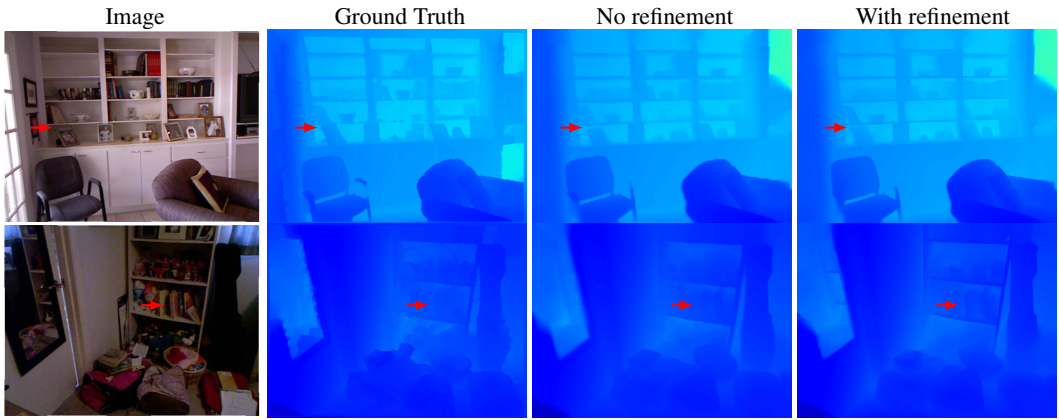
\begin{figure*}[h!]
\vspace*{0.1cm}
\setlength\tabcolsep{0.1pt} % default value: 6pt
\centering
\footnotesize
\renewcommand{\arraystretch}{0.05}
% \addtolength{\tabcolsep}{-5pt}

\newcommand{\image}[2]{
\begin{tikzpicture}
\node[anchor=south west,inner sep=0] at (0,0)
{\includegraphics[width=0.246\linewidth]{figures/tiled_refinement_nyu/#1.jpg}};
#2
\end{tikzpicture}
}

\newcommand{\row}[2]{
\image{image#1}{#2}
& 
\image{gt#1}{#2}
&
\image{pred#1}{#2}
&
\image{tiled#1}{#2}
}
\newcommand{\arrow}[2]{
\draw[red,-latex,thick]#1 -- #2;
}

\begin{tabular}{cccc}
    Image & Ground Truth & No refinement & With refinement \\
    
\row{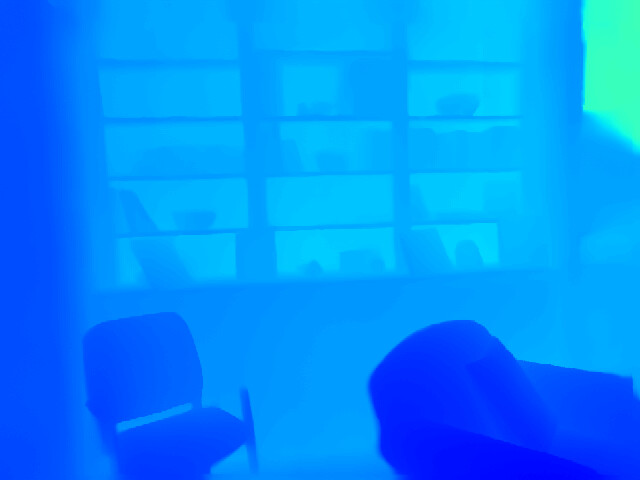}{
\arrow{(0.375,1.28)}{(0.675,1.28)};
} \\
% \row{0059}{} \\
\row{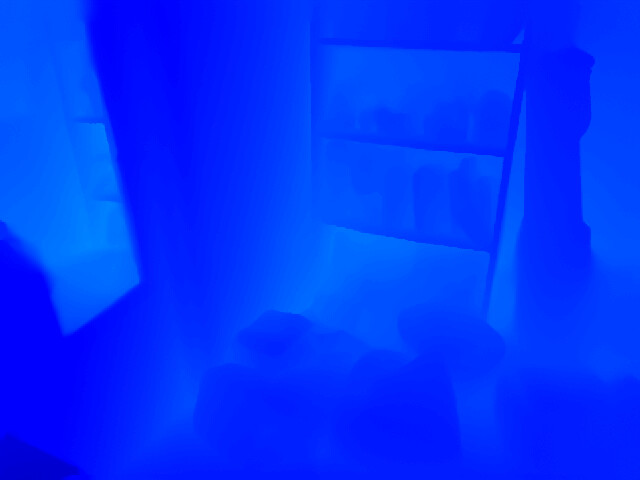}{
\arrow{(1.85,1.5)}{(2.15,1.5)};
} \\
% \arrow{(1.85,1.6)}{(1.55,1.6)};
% \arrow{(1.75,1.3)}{(2.05,1.3)};
% \arrow{(1.55,0.5)}{(1.25,0.5)};
\end{tabular}

\caption{\textbf{Samples with coarse-to-fine refinement on the NYU depth v2 dataset.} We find that refinement adds sharpness and detail to the depth estimation but does not provide quantitative improvements.}
\label{fig:nyu_tiled_refinement}
\vspace*{0.5cm}
\end{figure*}

Figure \ref{fig:nyu_tiled_refinement} demonstrates performance of coarse-to-fine refinement on the NYU depth v2 dataset.
While refinement improves fine-scale details in the estimated depth maps, the qualitative improvements are small and we do not find significant quantitative improvements. Hence the results reported in this work do not use coarse-to-fine refinement for depth estimation.
Further work is needed to develop a coarse-to-fine algorithm capable of more robust gains in depth estimation.

\vspace*{0.25cm}

\vspace*{-1cm}
\section{Coarse-to-fine optical flow refinement for RAFT}
\label{sec:supp-coarse-to-fine-raft}
\vspace*{-0.2cm}

For a fair comparison with optical flow estimation, we also apply our coarse-to-fine refinement scheme to RAFT \cite{teed2020raft}, to determine whether our performance gains translate to RAFT as well.
We first estimate flow at a low resolution, $320\times448$, upsample the low-resolution flow to the original resolution, divide original-resolution input images into $2\times5$ 
% \srbs{should we mention that 2x5 (the one we use) does worse for RAFT?} 
overlapping patches
% \footnote{We also tried  $2\times5$ patches but this degraded performance interestingly.} 
of size $320\times448$, then estimate flow on the cropped patches using the upsampled flow field as the initial guess for the recurrent refinement (12 steps in total) of RAFT \cite{teed2020raft}.
After estimating flow of each patch, we merge them using weighted masks \cite{jaegleperceiver}.
Table \ref{tab::raft_coarse2fine_refinement} reports the result.
Unlike our diffusion-based method, the coarse-to-fine scheme actually hurts the accuracy of RAFT on Sintel Clean and KITTI and only marginally improves the accuracy on Sintel Final.
Further exploration into better approaches for coarse-to-fine refinement for RAFT is warranted. We leave that to future work.

\begin{table*}[t]
    \centering
    \footnotesize
    \caption{\textbf{Our coarse-to-fine refinement} scheme marginally improves the performance of RAFT on Sintel Final while hurting performance on Sintel Clean and KITTI. We report the EPE on the Sintel and KITTI datasets.}
    \label{tab::raft_coarse2fine_refinement}
    \begin{tabular}{l @{\hskip 3em} ccc}
    \toprule
    & Sintel Clean & Sintel Final & KITTI \\
    \midrule
    RAFT baseline & 1.27 & 2.28 & 2.71 \\
    RAFT with the coarse-to-fine refinement & 1.35 & 2.26 & 2.85 \\
    \bottomrule
\end{tabular}
\end{table*}

\vspace*{0.25cm}
\vspace*{-0.25cm}
\section{Training and inference details}
\label{sec:supp-training-inference-details}
\vspace*{-0.2cm}

\subsection{Architecture}
\vspace*{-0.25cm}
\begin{figure}[h!]
    \centering
    %   \vspace*{0.1cm}
    \includegraphics[width=\textwidth]{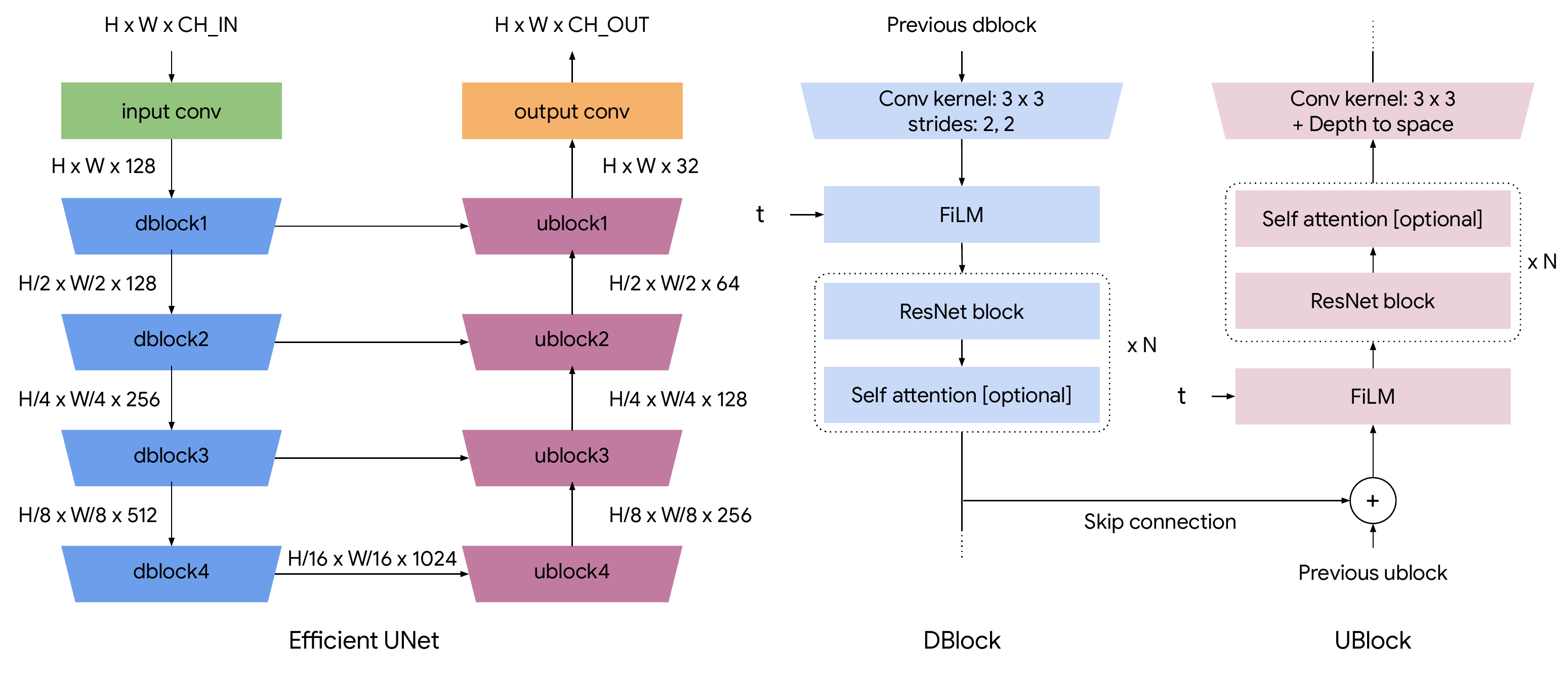}
    \vspace*{-0.5cm}
    \caption{\textbf{Overview of the Efficient UNet architecture} proposed in \cite{sahariac-imagen}. \textit{CH\_IN} and \textit{CH\_OUT} refer to the number of input and output channels respectively. \textit{t} refers to the time embedding. \textit{FiLM} refers to the modulation layers proposed in \cite{perez2018film}. \textit{N} is the number of ResNet + self-attention blocks.}
    \label{fig:efficient_unet}
    \vspace*{-0.1cm}
\end{figure}

\textbf{UNet.} The predominant architecture for diffusion models is the U-Net developed for the DDPM model \cite{ho2020denoising}, and 
later improved in several respects  \cite{dhariwal2021diffusion,nichol2021improved,song2021scorebased}. 
Here we adapt the {\em \unet}~architecture that was developed for Imagen \cite{sahariac-imagen}. It is more efficient that the U-Nets used in prior work
owing to the use of fewer self-attention layers, fewer parameters and less computation at higher resolutions, along with other adjustments that make it well suited to training medium resolution diffusion models. 

Specifically we adopt the configuration for the 64$\times$64 $\rightarrow$ 256$\times$256 super-resolution model (see Figure \ref{fig:efficient_unet} for an overview) with several changes. We drop the text cross-attention layers but preserve the self-attention in the lowest resolution layers \textit{dblock4} and \textit{ublock4} (see Figure \ref{fig:efficient_unet}). For supervised training for the flow model, we find it beneficial to additionally enable self-attention for the last-but-one layers \textit{dblock3} and \textit{ublock3}.
The number of input and output channels differ across self-supervised pre-training and supervised pre-training and are also different for flow and depth models.
For self-supervised pre-training \textit{CH\_IN=6} and \textit{CH\_OUT=3} (see Figure \ref{fig:efficient_unet}) since the input consists of a 3-channel source RGB image and a 3-channel noisy target image concatenated along the channel dimension and the output is a RGB image.
The supervised depth model has \textit{CH\_IN=4} (RGB image + noisy depth) and \textit{CH\_OUT=1}.
The supervised optical flow model has \textit{CH\_IN=8} (2 RGB images + noisy flow along $x$ and $y$) and \textit{CH\_OUT=2}. Note that this means we need to reinitialize the input and output convolutional kernels and biases before the supervised pretraining stage. All other weights are re-used.

\textbf{Resolution.} Our self-supervised model was trained at a resolution of $256\times256$. The indoor depth model is trained at 240$\times$320. For Waymo we use 256$\times$384 and for KITTI depth 256$\times$832.
Flow pretraining is done at a resolution of 320$\times$448, and finetuning at 320$\times$768.

\subsection{Datasets and augmentation}
\vspace*{-0.1cm}

For unsupervised pre-training, we use the ImageNet-1K \cite{imagenet_cvpr09} and Places365 \cite{zhou2017places} datasets and train on the self-supervised tasks of colorization, inpainting, uncropping, and JPEG decompression, following \cite{sahariac-palette}. Throughout, we mix datasets at the batch level.

\textbf{Flow.} For supervised flow pretraining we use a mix of AutoFlow (native resolution 448$\times$576), FlyingThings (540$\times$960), Kubric (512$\times$512) and TartanAir (480$\times$640) synthetic datasets.
We finetune on the standard mixture consisting of AutoFlow, FlyingThings, Viper (540$\times$960), HD1K (540$\times$1280), Sintel (436$\times$1024), and KITTI (375$\times$1242).

We follow the same photometric and geometric augmentation schemes from \cite{sun2022disentangling}, comprising random affine transformation, flipping, and cropping.

\textbf{Depth.} For supervised pre-training of the indoor model we mix the following datasets. \textit{ScanNet} \cite{scannet} is a dataset of 2.5M examples captured using a Kinect v1-like sensor. It provides depth maps at 480$\times$640 and RGB images at 968$\times$1296. \textit{SceneNet RGB-D} \citep{scenenet-rgbd} is a synthetic dataset of 5M images generated by rendering ShapeNet \citep{shapenet} objects in scenes from SceneNet \citep{https://doi.org/10.48550/arxiv.1511.07041} at a resolution of 240$\times$320.

For the outdoor model training we use the \textit{Waymo Open Dataset} \citep{waymoOpenDataset}, a large-scale driving dataset consisting of about 200k frames. Each frame provides RGB images from 5 cameras and LiDAR maps. We use the RGB images from the FRONT, FRONT\_LEFT and FRONT\_RIGHT cameras and the TOP LiDAR only to build about 600k aligned RGB depth maps.

For indoor fine-tuning and evaluation we use 
\textit{NYU depth v2} \citep{nyu}, a commonly used dataset for evaluating indoor depth prediction models. It provides aligned image and depth maps at 480$\times$640 resolution. We use the official split comprising ~50k images for training and 654 for evaluation.

For outdoor fine-tuning and evaluation, we use 
\textit{KITTI}~\citep{kitti}, an outdoor driving dataset which provides RGB images and LiDAR scans at resolutions close to 370$\times$1226. We use the training/test split proposed by \cite{eigen2014depth}, comprising  26k training images and 652 test images.

We use random horizontal flip data augmentation which is common in prior work. Where needed, images and dense depth maps are resized using bilinear interpolation to the model's resolution for training and nearest neighbor interpolation is used for sparse maps.

\subsection{Step-unrolling and interpolation of missing depth and flow}
\label{sec:supp-step-unroll-infill}
\vspace*{-0.1cm}

As discussed in Section 3.2 of the main paper, infilling and step-unrolling are used to mitigate distribution shift between training and inference with diffusion models.
The problem arises due to the missing data in the training depth maps and flow fields.

\textbf{Infilling.} For indoor depth maps, we use nearest neighbor interpolation during training (see Section 3.2 in the main paper).  For the outdoor depth 
data we use nearest neighbor interpolation except for sky regions, as they are often large and are much further from the camera than adjacent objects in the image.
We use an off-the-shelf sky segmenter \cite{Liba_2020_CVPR_Workshops}, and then set all sky pixels to be the maximum modeled depth (here, 80m). 
For missing optical flow ground truth we employ a simple sequence of 1D nearest neighbor interpolations first along rows, and then along columns.

\begin{table*}[h]
\centering
\small
% \begin{minipage}[t]{0.56\textwidth}
    \centering
    \small
    \caption{\textbf{Ablation of the number of unroll steps for monocular depth estimation}. Performs improves up to four steps of unrolling and plateaus thereafter. The models trained without infilling missing depth benefit more from a larger number of unroll steps, which is to be expected. \textbf{Best} and \underline{second best} results are bolded and underlined respectively.
    }
    \label{tab::num_unroll_steps_ablation}
    \setlength\tabcolsep{4.6pt}
    \begin{tabular}{c @{\hskip 2em} cccccccc}
        \toprule
        & \multicolumn{4}{c}{NYU} & \multicolumn{4}{c}{KITTI} \\
        \cmidrule(lr){2-5} \cmidrule(lr){6-9}
        & \multicolumn{2}{c}{No infill} & \multicolumn{2}{c}{Infill} & \multicolumn{2}{c}{No infill} & \multicolumn{2}{c}{Infill} \\
        \cmidrule(lr){2-3} \cmidrule(lr){4-5} \cmidrule(lr){6-7} \cmidrule(lr){8-9}
        Unroll steps & REL & RMS & REL & RMS & REL & RMS & REL & RMS \\
        \midrule
        0 &  0.079 &  0.331 & 0.077 & 0.338 &  0.222 & 3.770 & 0.057 & 2.744 \\
        1 & \underline{0.076} &  0.324 & \underline{0.075} &  0.324 & 0.085 & 2.844 &  0.056 &  2.700 \\
        2 & \underline{0.076} & \underline{0.317} & \underline{0.075} & \textbf{0.315} & 0.068 & 2.799 & \underline{0.054}  & 2.629 \\
        3 & \textbf{0.075} & \textbf{0.316} & \textbf{0.074} & \underline{0.317} & \underline{0.061} & \underline{2.789} & \underline{0.054}  & \underline{2.591} \\
        4 & \textbf{0.075} & \textbf{0.316} & \textbf{0.074} & \textbf{0.315} & \textbf{0.059} & \textbf{2.739} & \textbf{0.053}  & \textbf{2.568} \\

        \bottomrule
    \end{tabular} 
% \end{minipage} \hspace{1.5em}
\end{table*}

\textbf{Step-unrolling.} By default we use a single unroll step in all results where step-unrolling is enabled. In Table \ref{tab::num_unroll_steps_ablation}, we show that using multiple unroll steps can further improve performance on the task of monocular depth estimation.

Finally, while we use infilling and step-unrolling, there are other ways in which one might try to mitigate the problem.
One such approach was taken by \cite{AllInTokens}, which faced a similar problem when training a vector-quantizer on depth data. Their approach was to  synthetically add more holes following a carefully chosen masking ratio. We prefer our approach since nearest neighbor infilling is hyper-parameter free and \stepunrolled~could be more generally applicable to other tasks with sparse data.

\begin{table*}[h]
\centering
\small
% \begin{minipage}[t]{0.56\textwidth}
    \centering
    \small
    \caption{\textbf{Comparison of step-unrolling and self-conditioning} \cite{chen2023analog}
    }
    \label{tab::ablation_self_cond}
    \setlength\tabcolsep{4.6pt}
    \begin{tabular}{l @{\hskip 3em} cccccccc}
        \toprule
        & \multicolumn{4}{c}{NYU} & \multicolumn{4}{c}{KITTI} \\
        \cmidrule(lr){2-5} \cmidrule(lr){6-9}
        & \multicolumn{2}{c}{No infill} & \multicolumn{2}{c}{Infill} & \multicolumn{2}{c}{No infill} & \multicolumn{2}{c}{Infill} \\
        \cmidrule(lr){2-3} \cmidrule(lr){4-5} \cmidrule(lr){6-7} \cmidrule(lr){8-9}
         & REL & RMS & REL & RMS & REL & RMS & REL & RMS \\
        \midrule
        Baseline &  0.079 &  0.331 & 0.077 & 0.338 &  0.222 & 3.770 & 0.057 & 2.744 \\
        Self conditioning & 0.082 & 0.335 & 0.081 & 0.333 & 0.242 & 3.940 & 0.057 & 2.761 \\
        Step unrolling & \textbf{0.076} &  \textbf{0.324} & \textbf{0.075} &  \textbf{0.324} & \textbf{0.085} & \textbf{2.844} &  \textbf{0.056} &  \textbf{2.700} \\
        \bottomrule
    \end{tabular} 
% \end{minipage} \hspace{1.5em}
\end{table*}

We also considered the approach of \textit{self-conditioning} \cite{chen2023analog} as an alternative to step-unrolling. However, as we show in Table \ref{tab::ablation_self_cond}, we find that self-conditioning is unable to bridge the train-inference distribution shift of the noisy latent for the task of monocular depth estimation. This is specially apparent in the results for KITTI without infilling where self conditioning leads to no improvement whereas step-unrolling substantially improves performance.

\subsection{Hyper-parameters}
\vspace*{-0.1cm}

\label{sec:hparams}
\textbf{Self-supervised.} The self-supervised model is trained for 2.8M steps with an $L_2$ loss and a mini-batch size of 512.  Other hyper-parameters are same as those in the original Palette paper \cite{sahariac-palette}. 

\textbf{Supervised.} The supervised flow and depth models are trained with $L_1$ loss. Usually a constant learning rate of $1\times 10^{-4}$ with a warm-up over 10k steps is used. However, for depth fine-tuning we find that a lower learning rate of $3\times 10^{-5}$ achieves slightly better results. All models are trained with a mini-batch size of 64. The indoor depth model is pre-trained for 2M steps and then fine-tuned on NYU for 40k steps. The outdoor depth model is pre-trained for 0.9M steps and fine-tuned on KITTI for 40k steps. For flow, we pretrain for 3.7M steps, followed by finetuning for 50k steps. Other details, like the optimizer and the use of EMA are the same as \cite{sahariac-palette}.

\subsection{Inference}
\label{sec:supp-inference}
\vspace*{-0.1cm}

\textbf{Sampler.} We use the DDPM ancestral sampler \citep{ho2020denoising} with 128 denoising steps for monocular depth models and 64 steps for optical flow models. Increasing the number of denoising steps further did not greatly improve performance.

\myparagraph{Coarse-to-fine refinement.}
We use 2$\times$5 overlapping patches (\{top, bottom\} $\times$ \{left, center-left, center, center-right, right\}) for coarse-to-fine refinement. For Sintel we use $t'=32/64$ and for KITTI $t'=8/64$.
% Note that our noising-denoising process for coarse-to-fine refinement assumes that, given a coarse estimation $z$ $\exists~t, t'$ such that $y_{t'} \sim \mathcal{N}(y_{t'};\sqrt{\gamma_{t'}}\,z, (1-\gamma_{t'})I)$ is a fair sample from $\mathcal{N}(y_t;\sqrt{\gamma_t}y_0, (1-\gamma_t)I)$ and further that $t = t'$. However, this may not be the case due to interpolation artifacts in $z$ and one may want to use $t' > t$. Another way to deal with this would be to train a supervised model for super-resolution like in~\cite{saharia2021image}. This implies that the coarse-to-fine refinement improvements with out simple setup are a lower bound for the possible improvements that can be achieved. Notwithstanding, we find this approach, despite its simplicity, to be remarkably effective at improving fine-grained detail and we leave further exploration to future work.

\section{Limitations}
\label{sec:supp-limitations}
\vspace*{-0.2cm}

\begin{table*}[h!]
    \centering
    \footnotesize
    \caption{\textbf{Inference speed comparison} of our method with DPT \cite{dpt} on the indoor depth model finetuned on NYU. Diffusion model inference is bottlenecked by the large number of denoising steps. We show that some efficiency gains can be achieved by simply reducing the number of denoising steps. Our model with 24 denoising steps is comparable in performance to DPT while being $\sim$5x slower (modulo differences in hardware). * we use the step-time reported in the DPT paper at a resolution of $384\times384$, however, the DPT performance metrics on NYU are with a model trained at a resolution of $480\times640$, for which the step time will be higher.}
    \vspace*{0.05cm}
    \begin{tabular}{lccccccccc}
        \toprule
         Method & Architecture & Resolution & Total Time [ms] & Inference steps & REL~$\downarrow$ & RMS~$\downarrow$ \\
        \midrule
        DPT-Hybrid & Nvidia RTX 2080  & $384\times384$* & 38* & - & 0.110 & 0.357 \\
        \midrule
        \multirow{4}{*}{\textbf{\ours}} & \multirow{4}{*}{TPU v4}  & \multirow{4}{*}{$240\times320$} & 204 & 24 & 0.104 & 0.378  \\
        &&& 272 & 32 & 0.086 & 0.342 \\
        &&& 544 & 64 & 0.077 & 0.324 \\
        &&& 1089 & 128 & 0.075 & 0.324 \\
        \bottomrule
    \end{tabular} 
    \label{tab::latency}
    % \vskip -0.1in
\end{table*}

\textbf{Efficiency.} Inference speed with diffusion models is a well-known issue, as multiple denoising steps are used to transform noise to a target signal. This can be prohibitive for vision tasks where near real-time latency is often desired. Table \ref{tab::latency} compares the inference speed of our diffusion model for depth against DPT \cite{dpt}. 
Despite having an efficient denoiser backbone ($\sim$8.5 ms per denoising step on a TPU v4), the diffusion model is considerably slower than DPT in total wall time.
The most obvious way to reduce inference latency is to reduce the number of denoising steps. 
This can be done with only moderate reduction in performance.
As shown in Table \ref{tab::latency}, we perform comparably with DPT with as few as 24 denoising steps. However, a more thorough study into optimizing the inference speed of these models while preserving the generation quality is warranted. 
With the use of  progressive distillation \cite{meng2022on,salimans2022progressive} it is likely possible to reduce latency even further, as this approach has been shown to successfully distill generative image models with over 1000 denoising steps into those with just 2-4 steps.

\begin{figure*}[!h]
    \centering
    \small
    \setlength\tabcolsep{0.4pt}
    \newcommand{\Figwidth}{0.3\linewidth}
    \begin{tabular}{c @{\hskip 0.5em} c c c}
    & Temporally consecutive input frames & \textbf{Ours} & FlowFormer \cite{huang2022flowformer} \\
    \multirow{2}{*}{\rotatebox{90}{Ambush 1}} &
    \includegraphics[trim={0 0 0 0},clip,width=\Figwidth]{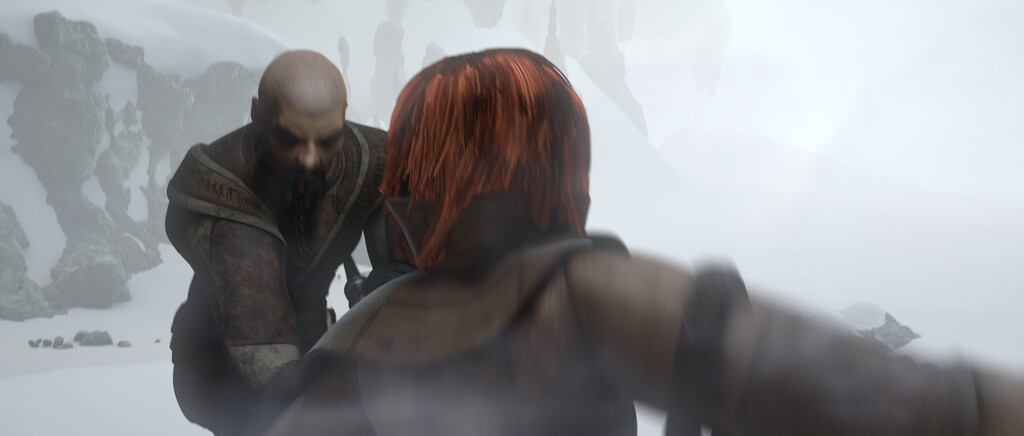} &
    \includegraphics[trim={0 0 0 0},clip,width=\Figwidth]{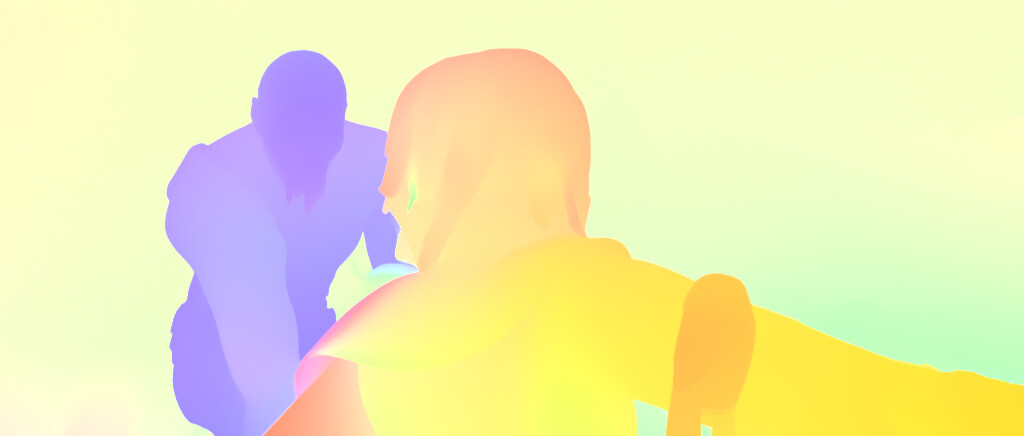} &
    \includegraphics[trim={0 0 0 0},clip,width=\Figwidth]{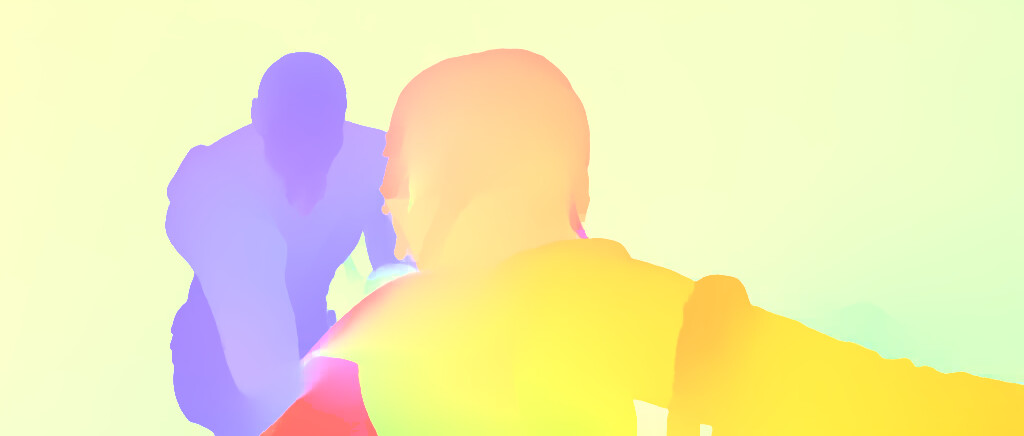} \\
    & \includegraphics[trim={0 0 0 0},clip,width=\Figwidth]{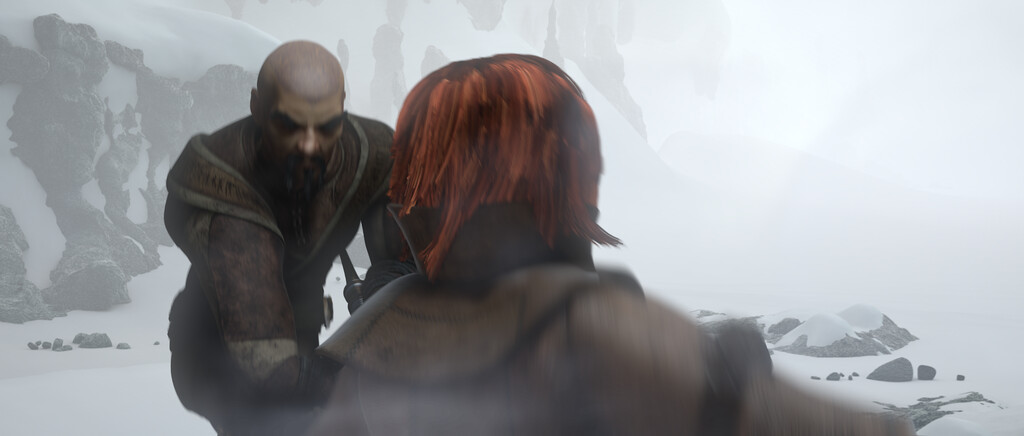} & 
    \includegraphics[trim={0 0 0 0},clip,width=\Figwidth]{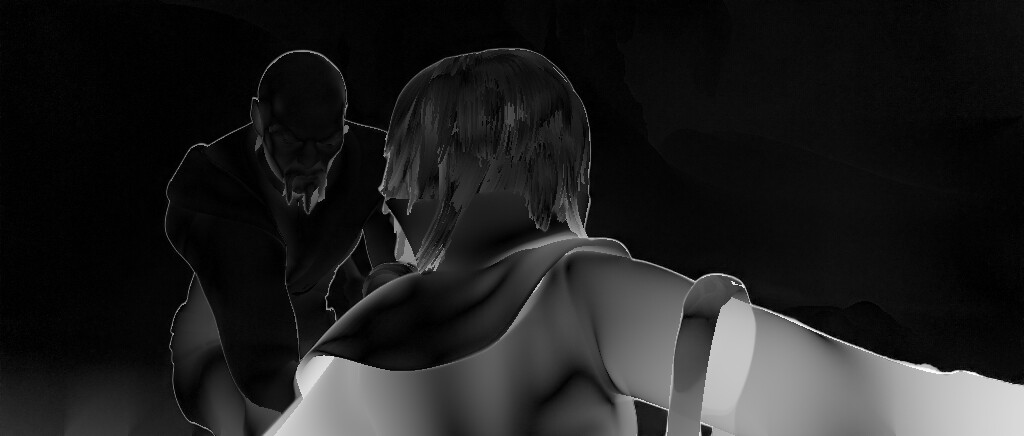} & 
    \includegraphics[trim={0 0 0 0},clip,width=\Figwidth]{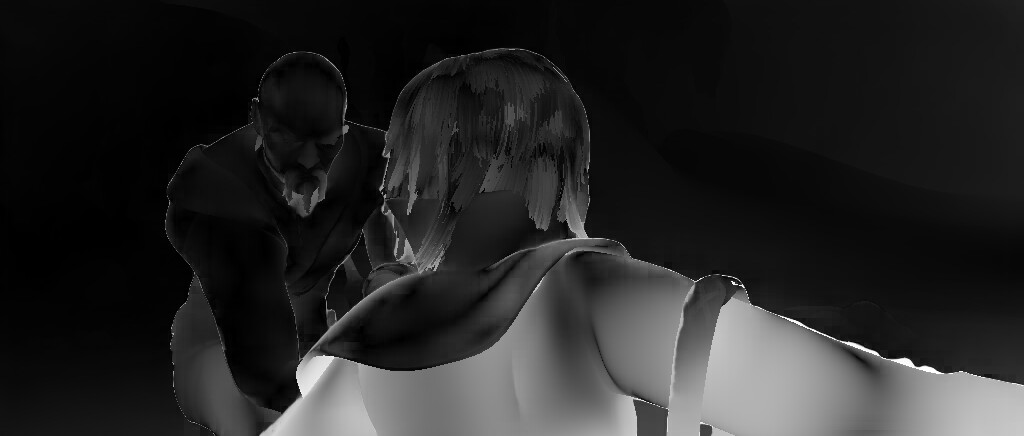} \\
    &  & EPE = 29.330 & EPE = 8.141 \\ 
    
    \multirow{2}{*}{\rotatebox{90}{Cave 3}} &
    \includegraphics[trim={0 0 0 0},clip,width=\Figwidth]{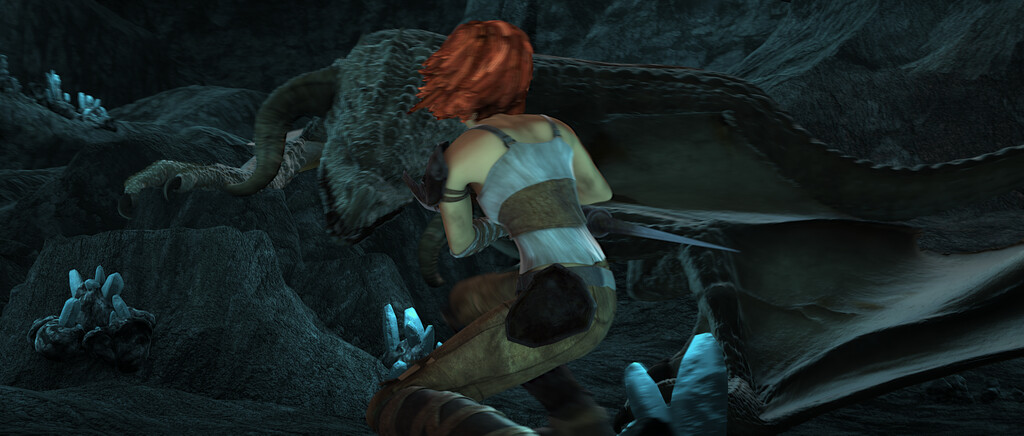} &
    \includegraphics[trim={0 0 0 0},clip,width=\Figwidth]{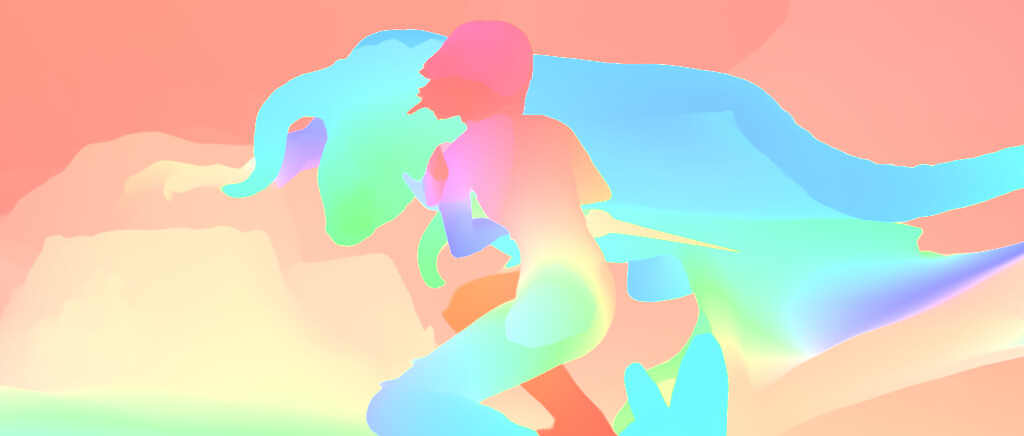} &
    \includegraphics[trim={0 0 0 0},clip,width=\Figwidth]{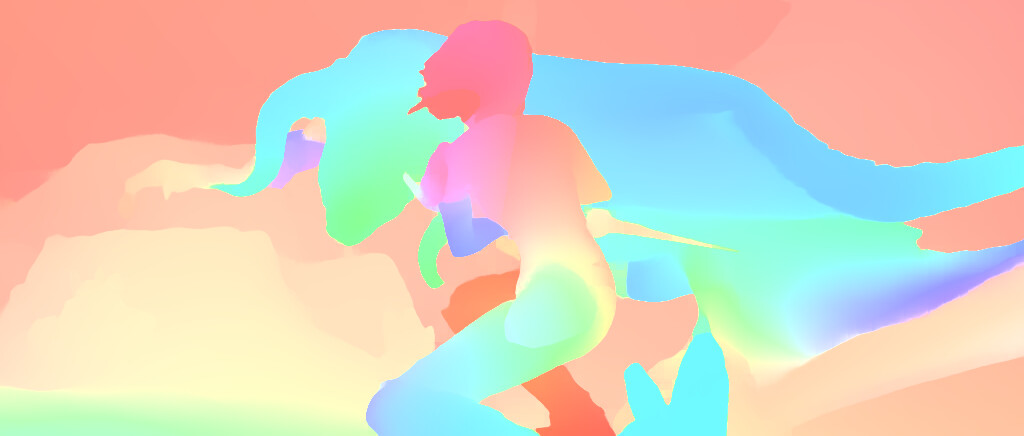} \\
    & \includegraphics[trim={0 0 0 0},clip,width=\Figwidth]{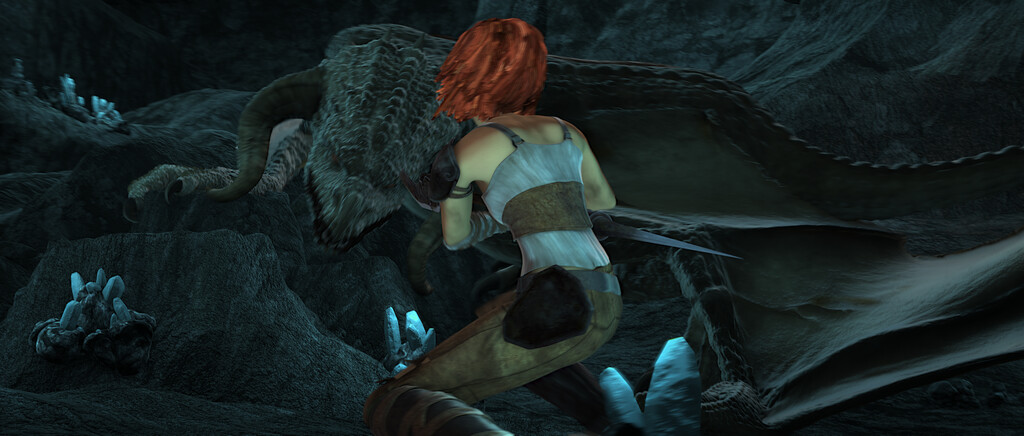} & 
    \includegraphics[trim={0 0 0 0},clip,width=\Figwidth]{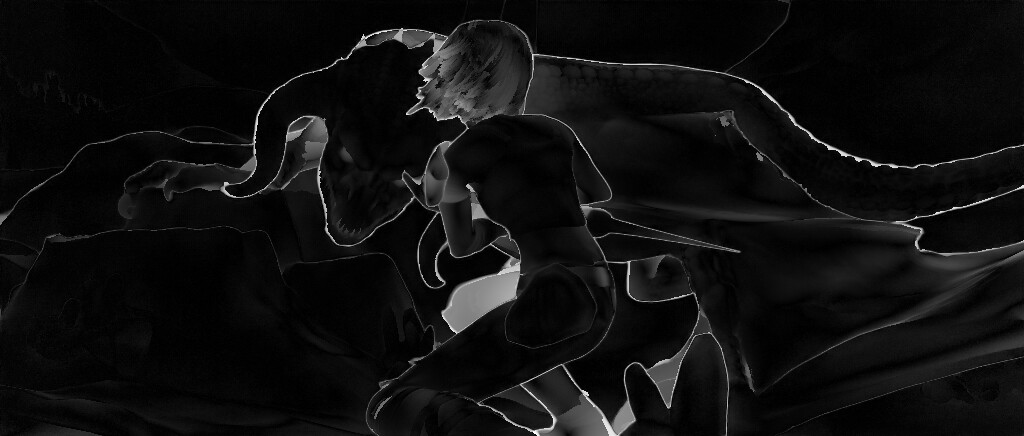} & 
    \includegraphics[trim={0 0 0 0},clip,width=\Figwidth]{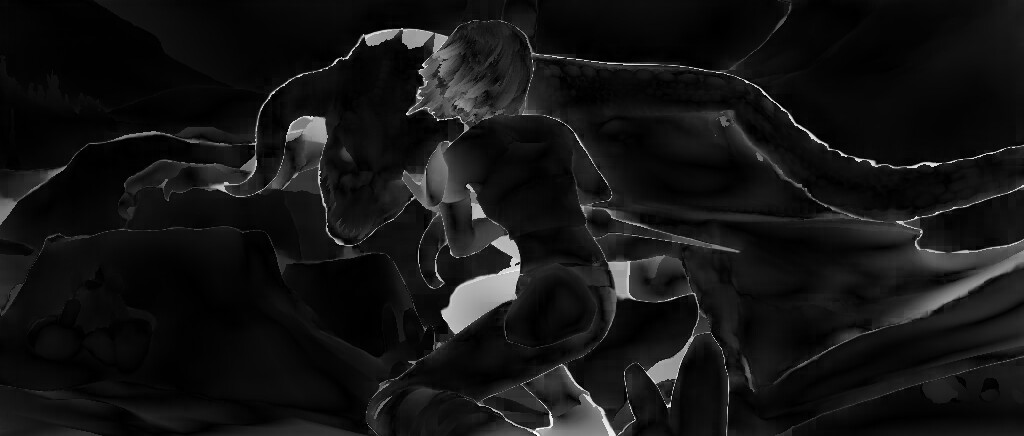} \\
    &  & EPE = 2.042 & EPE = 2.352 \\ 
    
    \multirow{2}{*}{\rotatebox{90}{Market 1}} &
    \includegraphics[trim={0 0 0 0},clip,width=\Figwidth]{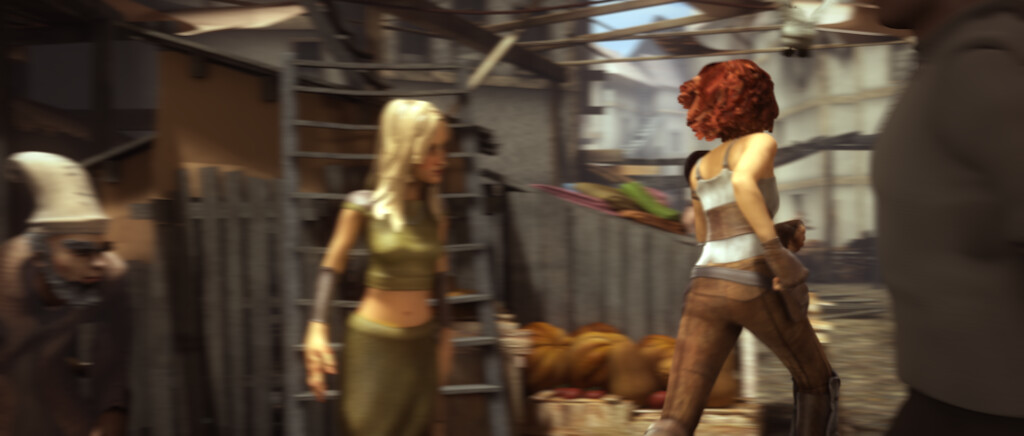} &
    \includegraphics[trim={0 0 0 0},clip,width=\Figwidth]{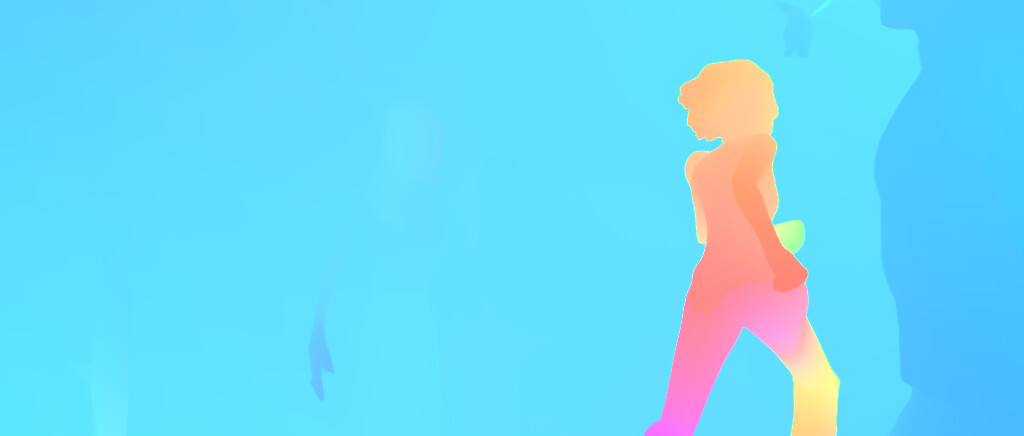} &
    \includegraphics[trim={0 0 0 0},clip,width=\Figwidth]{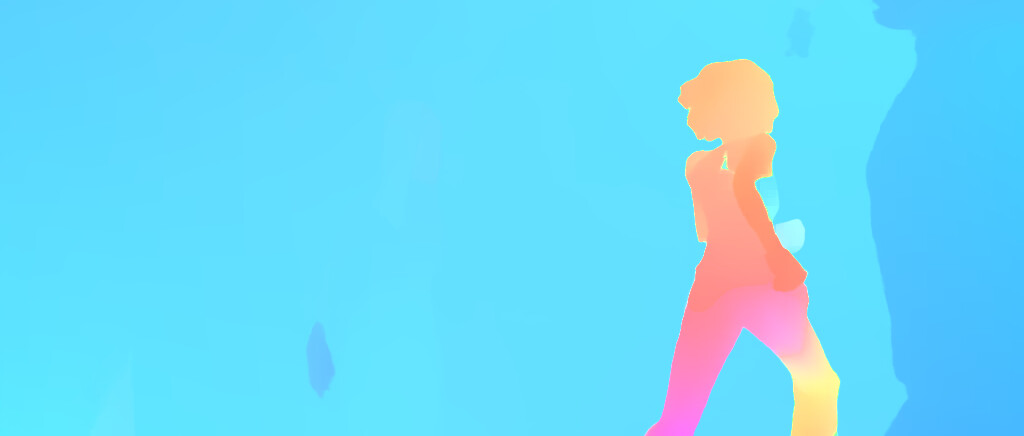} \\
    & \includegraphics[trim={0 0 0 0},clip,width=\Figwidth]{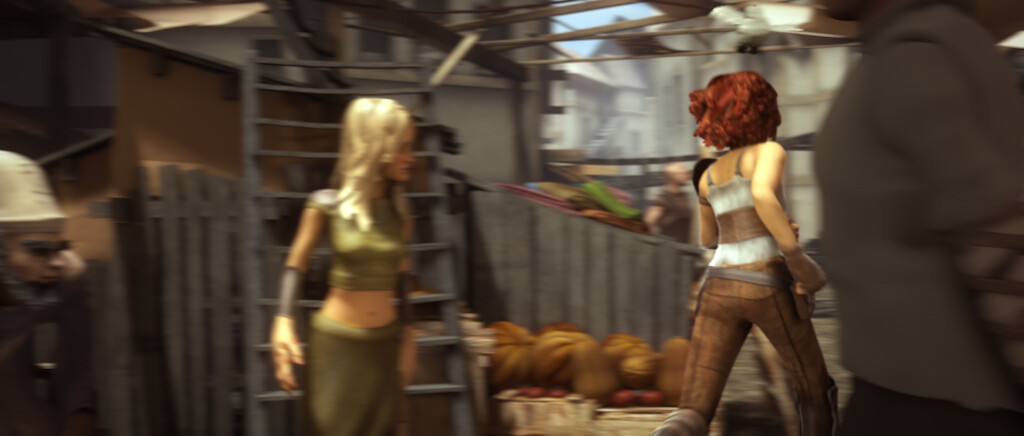} & 
    \includegraphics[trim={0 0 0 0},clip,width=\Figwidth]{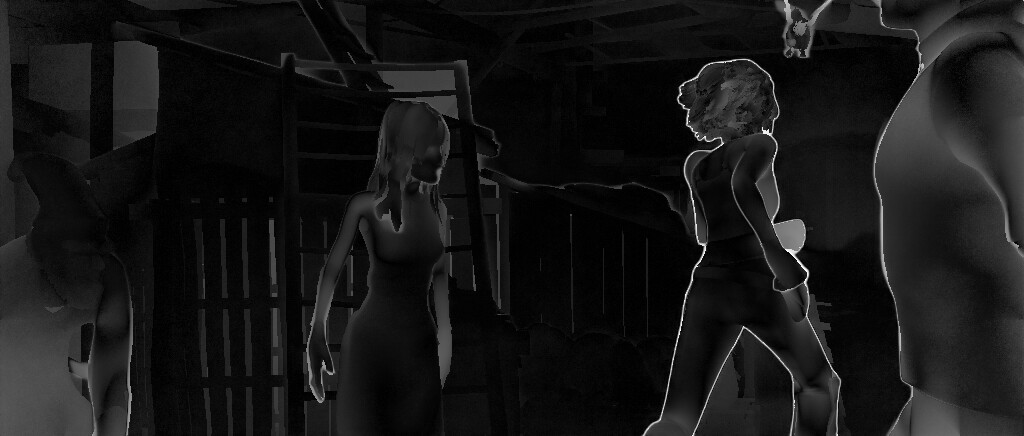} & 
    \includegraphics[trim={0 0 0 0},clip,width=\Figwidth]{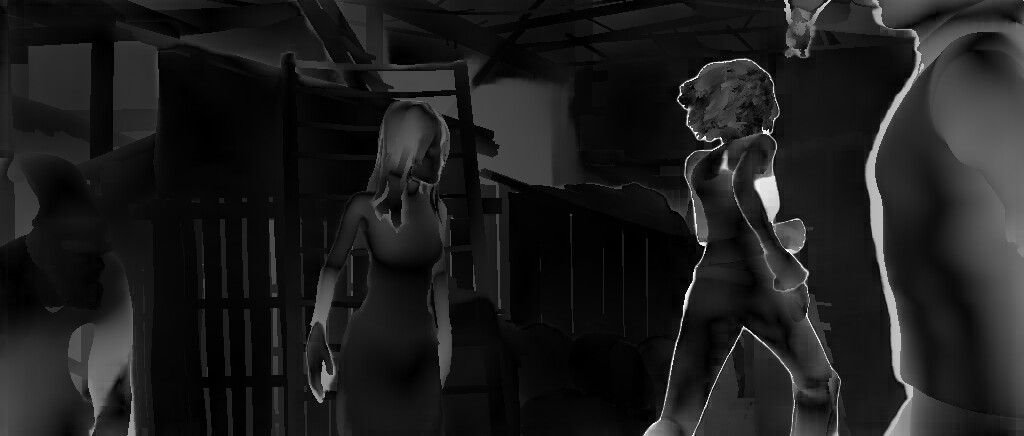} \\
    &  & EPE = 0.719 & EPE = 1.174 \\ 
    
    \end{tabular}
	\caption{
	\textbf{Visual results on Sintel test}. We compare with Flowformer~\cite{huang2022flowformer} and provide flow visualization and an error map on each scene. On Ambush 1, FlowFormer can better predict the motion of the girl's right arm that moves out of the image boundary, likely due to the global reasoning capability of attention. On Cave 3 and Market 1, our method provides much finer details on motion boundaries with lower end-point error (EPE).}
    \label{fig:supp:sintel_finetuning}
\end{figure*}

\textbf{Fine-tuning on Sintel.} In Section \ref{sec:limitations} we discuss possible reasons for why our model's superior zero-shot performance compared to FlowFormer \cite{huang2022flowformer} does not transfer to fine-tuning on Sintel. Figure \ref{fig:supp:sintel_finetuning} provides qualitative examples to further support the claims.

\clearpage
 
  \begin{figure*}[t]
    \centering
    \footnotesize
    \setlength\tabcolsep{0.8pt}
    \renewcommand{\arraystretch}{0.0}
    
    \newcommand{\Figwidth}{0.195\linewidth}
    \begin{tabular}{c @{\hskip 0.3em} c @{\hskip 0.3em} ccc}
    Input image & Variance heat map & \multicolumn{3}{c}{Multimodal prediction samples} \\
        \includegraphics[trim={0 0 0 0},clip,width=\Figwidth]{images/multi_modal_samples/depth_nyu/nyu_3_img.jpg} &
    \includegraphics[trim={0 0 0 0},clip,width=\Figwidth]{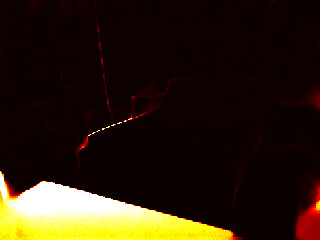} &
    \includegraphics[trim={0 0 0 0},clip,width=\Figwidth]{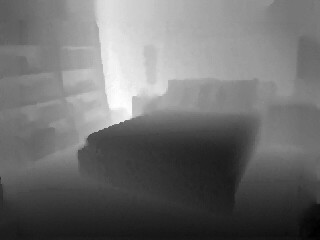} & 
    \includegraphics[trim={0 0 0 0},clip,width=\Figwidth]{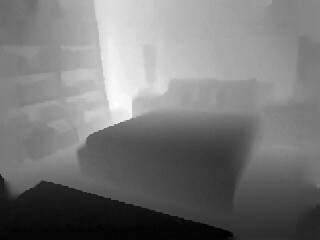} & 
    \includegraphics[trim={0 0 0 0},clip,width=\Figwidth]{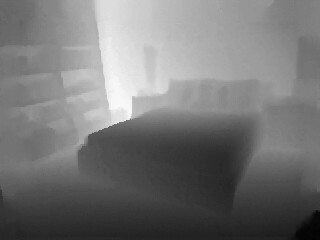}\\
    [0.2em] \\
    
    \includegraphics[trim={0 0 0 0},clip,width=\Figwidth]{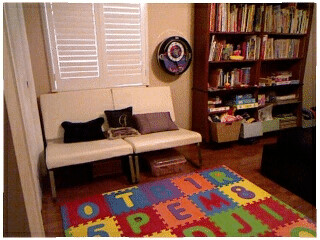} &
    \includegraphics[trim={0 0 0 0},clip,width=\Figwidth]{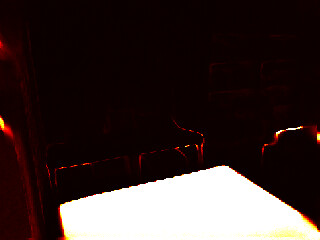} &
    \includegraphics[trim={0 0 0 0},clip,width=\Figwidth]{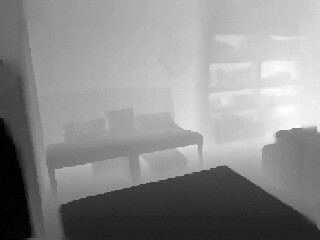} & 
    \includegraphics[trim={0 0 0 0},clip,width=\Figwidth]{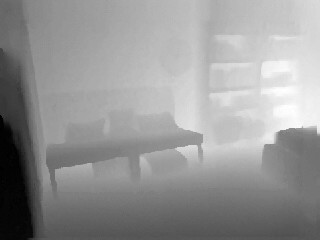} & 
    \includegraphics[trim={0 0 0 0},clip,width=\Figwidth]{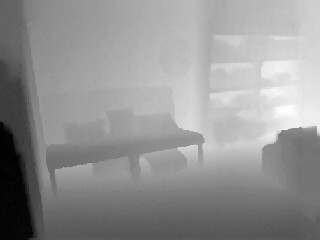}\\
    [0.2em] \\
    
    \end{tabular}
    \vspace*{-0.1cm}
	\caption{\textbf{Qualitative examples of multimodal estimation on the NYU depth dataset} showing examples where our model's \textit{uncertainty} gets captured in the multimodal posterior. In the examples above, the model confuses the play mat (farther away from the viewpoint) for a table (closer to the viewpoint).}
	   \vspace*{0.25cm}
    \label{fig:supp:multi_modal_samples_depth_NYU_limitation}
\end{figure*}
 \paragraph{Uncertainty in depth estimation.} We observe certain cases where the model is uncertain about the depth estimates. Interestingly, this uncertainty appears to be well captured in the predictive posterior, as illustrated in Figure \ref{fig:supp:multi_modal_samples_depth_NYU_limitation}.
 
 \vfill

% \ifdrafting
%     \input{supplementary}
% \fi
\end{document}